\documentclass[10pt]{article} 
\usepackage[accepted]{tmlr}

\usepackage{amsmath,amsfonts,bm}

\def\eqref#1{equation~\ref{#1}}

\def\1{\bm{1}}

\DeclareMathAlphabet{\mathsfit}{\encodingdefault}{\sfdefault}{m}{sl}
\SetMathAlphabet{\mathsfit}{bold}{\encodingdefault}{\sfdefault}{bx}{n}

\usepackage{hyperref}
\usepackage{url}
\usepackage{algorithm}
\usepackage{algpseudocode}
 \usepackage{amssymb}
  \usepackage{booktabs}
  \usepackage{graphicx}
\usepackage{subcaption}
  \usepackage{multirow}
\usepackage{booktabs}
\usepackage{tabularx}
\usepackage{array}
\usepackage{makecell}

\title{PairAlign: A Framework for Autoregressive Tokenization via Self-Alignment with Applications to Audio Tokenization}

\author{\name Adhiraj Banerjee \email adhirajbanerjee35@gmail.com \\
      \addr Department of Electrical Engineering\\
      Indian Institute of Technology, Kanpur 
      \AND
      \name Vipul Arora \email vipular@iitk.ac.in \\
      \addr Department of Electrical Engineering \\
            Indian Institute of Technology, Kanpur }

\def\month{11}  
\def\year{2026} 
\def\openreview{\url{https://openreview.net/forum?id=gd0IwqmO71}} 

\begin{document}

\maketitle

\begin{abstract}
Modern learning systems represent perceptual signals with continuous vectors,
but many operations over sensory data---comparison, retrieval, memory,
alignment, and reasoning---are naturally expressed over discrete symbolic
sequences. In language, this interface is given a priori through tokens; for
continuous sensory signals such as speech and audio, it must be learned.
Existing audio tokenizers typically induce symbols through frame- or
short-window-level quantization, clustering, or reconstruction, leaving
sequence-level properties such as cross-realization consistency, learned
length, termination, compactness, and edit geometry only indirectly controlled.
We introduce \textbf{PairAlign}, a framework for learning compact audio token
strings through sequence-level autoregressive self-alignment. PairAlign
formulates tokenization as conditional sequence generation: an encoder maps
speech to a continuous condition, and an autoregressive decoder emits a
symbolic sequence from $\mathrm{BOS}$ to $\mathrm{EOS}$, jointly learning token
identity, order, length, and termination. Given two content-preserving views,
PairAlign derives a canonical symbolic target from the anchor and trains both
the anchor and positive representations to predict the same sequence, while
unrelated in-batch anchor targets provide competing sequences. This yields a
scalable likelihood-based surrogate for learning relational sequence geometry
while discouraging many-to-one collapse. Starting from a VQ-style geometric
tokenizer, PairAlign first learns an autoregressive bridge from deterministic
anchor-derived VQ targets and then transitions to adaptive EMA-teacher
self-alignment, where an anchor-derived teacher sequence serves as the shared
target for both views, together with grounding and anti-collapse controls.
On $3$ s speech, PairAlign produces substantially shorter, broad-vocabulary,
non-degenerate sequences while retaining strong ordered paired-view consistency,
operating at only $8.28$ tokens/s on the TIMIT retrieval archive. Relative to the Stage~I geometric baseline,
PairAlign reduces archive token count by $70.65\%$ and the edit-distance work
proxy by $91.87\%$. Direct positive--negative probes further show the strongest
relative edit-distance separation across phone-disjoint, trigram-disjoint, and
temporally rearranged negatives, while rate-controlled post-hoc BPE does not
recover the same compactness--consistency operating point. These results
characterize PairAlign not as a uniformly stronger high-rate tokenizer, but as
a learned lower-rate symbolic interface that retains meaningful ordered and
relational structure while reducing sequence length and repeated comparison
cost. More broadly, PairAlign is motivated by the same high-level philosophy as JEPA-style predictive learning: predicting an abstract target representation rather than reconstructing the input directly. Here, the target is a learned, variable-length symbolic sequence. 

\end{abstract}
\section{Introduction}

\paragraph{Toward symbolic interfaces for audio.}

Discrete token sequences provide useful interfaces for comparison, retrieval,
alignment, memory, editing, generation, and structured reasoning. In language,
this interface is given by text; in audio, it must be learned from continuous
signals. Audio tokenization is therefore central to neural codecs, speech and
music generation, text-conditioned audio synthesis, multimodal modeling, and
retrieval-oriented speech processing
\citep{mousavi2025discrete}. Existing tokenizers are effective, but most induce
symbols through local frame- or short-window-level assignment: codec tokenizers
optimize reconstruction-oriented codes, semantic tokenizers discretize
self-supervised speech representations, and hierarchical or factorized
tokenizers distribute information across codebooks
\citep{zeghidour2021soundstream,defossez2022high,kumar2023high,
baevski2020vq,hsu2021hubert,chung2021w2v,zhang2023speechtokenizer,
ju2024naturalspeech}. PairAlign addresses a complementary question: how to
learn the emitted token string itself, including its ordering, compactness,
cross-view stability, relational edit geometry, sequence length, and
termination.

\paragraph{Why framewise assignment is not enough.}

Geometric discretization is a strong inductive bias: assigning each encoder
frame to a nearby centroid or quantizer entry preserves timing and local detail,
and works well for compression, reconstruction, and frame-synchronous modeling.
However, the resulting token stream is not usually optimized as a string. This
matters for retrieval, matching, indexing, alignment, and edit-based comparison,
where related views $(x,x^{+})$ should induce closer token strings than
unrelated examples $x^{-}$. Edit distance is natural for this requirement
because it jointly accounts for token identity, ordering, substitutions,
insertions, deletions, and sequence length. Retrieval-oriented systems such as
wav2tok and BEST-STD motivate this symbolic view
\citep{banerjee2022wav2tok,singh2025best}, but edit distance is combinatorial
and non-differentiable, making it difficult to optimize directly inside neural
tokenizers.

A second limitation is that symbolic rate is normally inherited from encoder
stride and local assignment. This couples sequence length to temporal
resolution, even when the downstream interface may not require a dense
frame-level trace. PairAlign instead asks whether sequence geometry and
symbolic rate can be learned jointly: can a substantially shorter symbolic
sequence remain stable across content-preserving views and discriminative
across different inputs?

\paragraph{Learning the string, not only the frame labels.}

We introduce \textbf{PairAlign}, a framework for learning compact discrete
audio token sequences through sequence-level autoregressive self-alignment.
Rather than assigning one discrete label to each encoder frame and optionally
compressing the resulting stream afterward, PairAlign formulates tokenization
itself as conditional sequence generation. Given an input segment $x$, an
encoder produces
\begin{equation}
    Z = Enc(x),
\end{equation}
and an autoregressive decoder defines
\begin{equation}
    p(\mathcal{T}\mid Z)
    =
    \prod_{l=1}^{|\mathcal{T}|}
    p(\tau_l\mid \tau_{<l},Z).
\end{equation}
Decoding begins from $\mathrm{BOS}$ and terminates when the model emits
$\mathrm{EOS}$. Token identity, ordering, sequence length, and termination are
therefore learned properties of the symbolic interface rather than direct
consequences of encoder stride, nearest-centroid assignment, or
de-duplication.

\paragraph{Shared-target self-alignment.}

The central learning signal is shared-target prediction across
content-preserving views. For an anchor--positive pair
$(x_i,x_i^{+})$, with representations $Z_i$ and $Z_i^{+}$, PairAlign defines
one anchor-derived symbolic target $\mathcal{T}_i$ and trains both conditions
to predict that same sequence:
\begin{equation}
    \log p(\mathcal{T}_i\mid Z_i),
    \qquad
    \log p(\mathcal{T}_i\mid Z_i^{+}).
\end{equation}
The anchor branch grounds the target in the input, while the positive branch
requires a nuisance-perturbed realization of the same content to recover the
same ordered symbolic description. The objective therefore encourages the two
views to share one symbolic realization rather than merely requiring two
independently generated sequences to predict one another.

Shared-target prediction alone is not sufficient: unrelated inputs could still
map to generic strings. PairAlign therefore contrasts the paired target against
mismatched anchor-derived sequences from other minibatch examples using
conditional sequence likelihood. The two terms have complementary roles.
Shared-target fitting encourages positive contraction around a common symbolic
reference, while likelihood contrast supplies explicit cross-example
discrimination.

\paragraph{From geometric initialization to adaptive symbols.}

PairAlign follows a three-stage training path. Stage~I learns a contextual
encoder and nearest-centroid VQ tokenizer, providing a stable geometric
initialization and the primary controlled baseline. Stage~II freezes this
tokenizer and trains an autoregressive decoder so that both the anchor and its
content-preserving positive view predict the deterministic Stage~I anchor
sequence. Stage~III replaces this fixed target with an adaptive anchor sequence
generated by a slowly updated EMA teacher over the complete
encoder--decoder tokenizer, while jointly refining the student encoder and
decoder.

The final representation is therefore not a post-hoc shortening of the
Stage~I token stream. The conditioning representation, symbolic target,
sequence length, EOS position, and autoregressive trajectory are allowed to
co-evolve under shared-target prediction and likelihood contrast. PairAlign
learns the lower-rate symbolic sequence itself rather than compressing a
symbolic organization fixed beforehand.

\paragraph{Keeping autoregressive tokenization grounded and non-degenerate.}

Autoregressive tokenization introduces failure modes that do not arise in the
same form for framewise tokenizers. Under teacher forcing, the decoder can rely
too strongly on token history, underuse the acoustic condition, or converge to
generic strings that produce deceptively high paired-view agreement. We
therefore treat acoustic grounding and collapse prevention as explicit design
constraints.

Prefix corruption weakens intact target-side history, structured
self-attention dropout reduces dependence on the decoder continuation pathway,
and encoder-summary conditioning supplies persistent global acoustic context.
In-batch likelihood contrast discriminates the paired target from competing
sequences, while entropy control, repetition-aware teacher generation, and rate
constraints provide complementary regularization. The ablations make these
failure modes visible: removing decoder-history regularization or likelihood
contrast can yield nominally perfect paired agreement through generic-string
collapse, whereas removing encoder-summary conditioning substantially degrades
sequence stability and compactness.

\paragraph{What should a learned symbolic interface preserve?}

A lower token rate is useful only if the resulting sequence remains
input-dependent, non-degenerate, and structurally meaningful. We therefore
evaluate PairAlign as a symbolic interface rather than merely as a source of
discrete codes. The central questions are whether substantial compression
preserves ordered paired-view structure, whether different inputs remain
separated, whether variable-length termination is genuinely learned, and
whether the observed behavior is induced by sequence-level learning rather than
by shortening alone.

Accordingly, we measure paired-view consistency, symbolic rate, vocabulary use
and collisions, positive--negative edit geometry, local objective geometry,
retrieval and comparison cost, learned termination, acoustic sensitivity,
temporal recoverability, robustness to external noise, and cross-lingual
transfer. Rearranged acoustic inputs test sensitivity to temporal organization,
while unconstrained decoding tests whether EOS is produced by the model rather
than imposed by the inference-time rate limit.

\paragraph{A lower-rate but coarser operating point.}

Because the decoder emits $\mathrm{EOS}$, output length is no longer tied
directly to encoder frame rate. On the clean TIMIT retrieval archive, PairAlign
operates at $8.28$ tokens/s compared with $28.21$ tokens/s for the geometric
Stage~I tokenizer, reducing archive token count by $70.65\%$. Once
tokenizations are available, the shorter sequences reduce the
sequence-length-dependent edit-distance work proxy by $91.87\%$.

This compactness changes representation granularity. Dense geometric and
pretrained SSL tokenizers preserve more fine-grained local acoustic evidence
and can retain stronger local persistence or top-rank retrieval. PairAlign
instead learns a coarser sequence-level representation with strong ordered
paired-view consistency, non-degenerate positive--negative geometry, and
input-dependent variable-length termination. Its contribution is therefore not
uniformly superior tokenization, but a different rate--granularity operating
point in which substantially fewer symbols retain useful sequence structure.

\paragraph{Low rate alone does not reproduce the learned geometry.}

A natural alternative explanation is that these properties arise simply
because PairAlign emits fewer symbols. We test this directly with deterministic
post-hoc BPE applied both to the geometric Stage~I stream and to substantially
stronger HuBERT and WavLM token streams. These rate-controlled baselines do not
recover PairAlign's low-rate paired-view sequence geometry even when initialized
from much stronger pretrained representations. Conversely, compressed SSL
streams can retain stronger retrieval performance.

The comparison therefore separates three properties that should not be
conflated: symbolic rate, sequence consistency, and retrieval discrimination.
PairAlign's operating point cannot be explained by post-hoc shortening alone.

\paragraph{Probing whether the Stage~III objective produces the intended geometry.}

We next test whether the two Stage~III terms produce the local relational
structure they are designed to encourage. On held-out TIMIT, the two
predictions associated with the same content are closer to one another than to
a hard mismatched target in
$99.898\pm0.009\%$ of evaluated comparisons.

The mechanism is straightforward. Shared-target fitting contracts the
anchor- and positive-conditioned predictions around the same EMA target,
whereas likelihood contrast separates that target from mismatched examples.
Likelihood separation, however, does not by itself imply Levenshtein
separation. We therefore test this remaining link using a
leave-one-candidate-out local calibration rather than assuming it from the
objective
(Appendix~\ref{sec:pairalign-local-theory}).

For the complete $15$-mismatch in-batch neighborhood, at least one
conditioning view provides a validated local edit lower bound for
$71.76\pm0.07\%$ of comparisons. Combining this bound with the realized
positive contraction certifies the desired relational ordering for
$59.21\pm0.08\%$ of comparisons using the observed candidate-specific score
gap, and for $52.92\pm0.20\%$ using only the more conservative margin implied
directly by the NCE objective. These certificates are sufficient rather than
necessary: the much higher $99.898\pm0.009\%$ observed ordering rate shows
that many uncertified comparisons nevertheless exhibit the intended local
geometry.

The analysis is deliberately restricted to the training-time candidate
structure. Final free-running EMA tokenizations are evaluated directly rather
than inferred from these local guarantees.

\paragraph{Recovering approximate temporal grounding.}

Compact autoregressive tokens are not natively time-stamped. When temporal
grounding is required, we interpret decoder cross-attention as a soft
token-to-frame association, optionally sharpen it with a weak diagonal Beta
prior, and recover a monotone token-to-time path using Viterbi decoding. This
procedure is entirely post hoc and does not alter either the generated sequence
or the training objective.

Evaluation against TIMIT phone boundaries shows that the recovered transitions
carry meaningful temporal structure, while PairAlign produces fewer and more
selective boundaries with less within-phone fragmentation. We therefore
interpret the resulting timestamps as approximate temporal grounding rather
than evidence of one-token--one-phone segmentation.

\paragraph{Summary of the learned interface.}

Taken together, the experiments characterize PairAlign as a compact,
input-conditioned, non-degenerate sequence representation. It substantially
reduces symbolic rate while preserving meaningful ordered and relational
structure, but does so by moving to a coarser representation regime. Dense
frame-derived tokenizers remain preferable when fine-grained local acoustic
resolution is the primary objective; PairAlign targets settings in which a
shorter learned symbolic trajectory is itself useful.

\paragraph{Contributions.}

The main contributions are:
\begin{itemize}

    \item \textbf{Autoregressive sequence tokenization.}
    We formulate audio tokenization as input-conditioned variable-length
    sequence generation, so token identity, ordering, sequence length, and
    termination are learned jointly rather than inherited from framewise
    assignment.

    \item \textbf{Shared-target self-alignment.}
    We introduce a paired-view objective in which an anchor and its
    content-preserving positive view predict the same anchor-derived symbolic
    sequence, providing a common reference for learning view-consistent ordered
    representations.

    \item \textbf{Adaptive sequence learning.}
    We develop a staged training procedure that begins from a geometric VQ
    tokenizer and replaces fixed geometric targets with sequences generated by
    a slowly updated encoder--decoder teacher, allowing the symbolic sequence
    itself to evolve during training.

    \item \textbf{Grounding and collapse control for autoregressive
    tokenization.}
    We combine prefix corruption, structured self-attention dropout,
    encoder-summary conditioning, and in-batch likelihood contrast to reduce
    decoder-history dependence, preserve acoustic conditioning, and prevent
    generic-sequence collapse.

    \item \textbf{Objective-linked local geometry analysis.}
    We characterize the complementary roles of shared-target contraction and
    likelihood-based negative separation and evaluate their transfer to local
    edit geometry with held-out leave-one-candidate-out analysis. On TIMIT,
    the desired positive-versus-hard-negative relational ordering is observed
    in more than $99.8\%$ of evaluated comparisons, while the corresponding
    sufficient condition remains non-vacuous for a substantial fraction of
    them.

\end{itemize}

\paragraph{Beyond speech tokenization.}

The present experiments focus on continuous speech, but the underlying
formulation is more general. PairAlign applies whenever continuous or weakly
structured inputs must be mapped into compact symbolic sequences that remain
predictable across content-preserving views while preserving useful relational
structure.

This connects PairAlign to the broader predictive-abstraction principle of
JEPA-style learning: prediction is performed in a learned representation space
rather than through reconstruction of low-level observations. PairAlign changes
the predictive object itself. Instead of a fixed-dimensional continuous latent,
the target is a learned variable-length symbolic sequence with its own
vocabulary, ordering, rate, and termination behavior. We refer to this view as
\emph{sequence-symbolic predictive learning}.

More broadly, the results suggest that predictive self-supervision can be used
not only to learn representations before discretization, but also to organize
the symbolic interface itself. This motivates extensions to longer-context
sequence compression, perceptual memory, generation, multimodal prediction,
and symbolic world models, where high-rate streams may be transformed into
shorter learned trajectories while retaining the relations required by
downstream sequence processing.

\section{Related Work}
\label{sec:rel_work}

\paragraph{Audio tokenizers as frame-derived symbolic interfaces.}
Discrete audio tokenization is now central to neural audio modeling. Neural
codec tokenizers such as SoundStream, EnCodec, and related high-fidelity codecs
learn discrete codes optimized for reconstruction, compression, and generation
\citep{zeghidour2021soundstream,defossez2022high,kumar2023high}. Semantic
speech tokenizers discretize representations from self-supervised models such
as wav2vec, HuBERT, and WavLM
\citep{baevski2020vq,hsu2021hubert,chung2021w2v,chen2022wavlm}, while recent
hierarchical and factorized tokenizers distribute acoustic, semantic, speaker,
prosodic, or residual information across multiple streams
\citep{zhang2023speechtokenizer,ju2024naturalspeech}. These systems provide
strong frame-derived symbolic interfaces for synthesis, compression,
audio-language modeling, and generation. PairAlign addresses a complementary
question: how to learn the emitted token string itself as a sequence-level
symbolic object with learned order, compactness, stability, length, termination,
and edit geometry.

\paragraph{From geometric assignment to learned sequence induction.}
Most audio tokenizers first encode audio into continuous frame or window
representations and then assign each local representation to a code by
nearest-centroid assignment, VQ, clustering, residual quantization, or related
local rules. This preserves timing and local detail, and it gives a stable
interface for downstream language modeling over audio tokens. However, the
token sequence structure is largely fixed before the language model is trained:
encoder stride, feature head, quantizer, codec rate, windowing, and
post-processing determine the temporal grid, ordering, length, and termination
behavior. PairAlign instead asks whether the emitted token string can itself be
the object of tokenization. It keeps geometric discretization as a stable
initialization, but then induces token identity, order, length, termination, and
cross-view stability through paired-view conditional likelihood.

\paragraph{Retrieval-oriented tokenizers and edit-distance comparison.}
Retrieval-oriented tokenizers such as wav2tok and BEST-STD motivate discrete
speech tokens as symbolic interfaces for query-by-example spoken term detection
and related matching tasks
\citep{banerjee2022wav2tok,singh2025best}. Edit distance is natural in this
setting because it compares token identity, order, substitutions, insertions,
deletions, and length, but it is combinatorial and non-differentiable. wav2tok
is the closest precursor to PairAlign: it uses a CTC-style pairwise alignment
constraint to encourage paired speech realizations to produce compatible
frame-indexed token sequences. PairAlign preserves this relational motivation,
but moves beyond retrieval-oriented frame tokenization by learning the emitted
sequence itself through input-conditioned autoregressive generation. Retrieval
is therefore used here as a probe of edit-distance structure, not as the sole
definition of tokenizer quality.

\paragraph{Sequence transduction aligns known labels; PairAlign induces them.}
CTC and RNN-T provide sequence-level likelihoods when the target label sequence
is known but its alignment to input frames is latent
\citep{graves2006connectionist,graves2012sequence}. CTC marginalizes monotone
alignments with frame-indexed posteriors and a collapse map, while RNN-T also
conditions on previously emitted labels. PairAlign is related in its use of
sequence-level likelihoods, but differs in the object being learned: the target
is not an external transcript or label string. It is a learned audio-token
sequence whose identity, order, length, and EOS behavior are induced during
training.

\paragraph{Predicting symbolic abstractions instead of continuous latents.}
Self-supervised predictive representation learning often predicts abstract
targets associated with another view, region, or time step rather than
reconstructing raw observations. JEPA-style methods formalize this principle by
predicting representations instead of pixels, samples, or other low-level
signals
\citep{lecun2022path,assran2023self,bardes2024revisiting,fei2023jepa}.
PairAlign follows this predictive-abstraction view, but changes the target
space from a fixed-dimensional continuous latent to a learned variable-length
symbolic sequence induced from a paired content-preserving view. It can
therefore be viewed as a sequence-symbolic analogue of predictive
self-supervision, where the representation and the discrete sequence interface
are learned jointly.

\paragraph{Alignment before the tokenizer is fixed.}
PairAlign uses the term \emph{alignment} differently from alignment in LLM
post-training. RLHF, RLAIF, Constitutional AI, DPO, IPO, KTO, RLVR, and related
methods reshape behavior over an already fixed tokenizer and vocabulary
\citep{stiennon2020learning,ouyang2022training,bai2022training,
bai2022constitutional,rafailov2023direct,azar2024general,ethayarajh2024kto,
shao2024deepseekmath,guo2025deepseek}. PairAlign instead learns the symbolic
interface itself: token identity, order, length, and termination evolve during
training, and the alignment signal comes from paired-view data geometry rather
than external preference labels. In this work, alignment is therefore a
representation-learning principle for inducing and stabilizing a token space.

\paragraph{A decoder, but not a prompt-continuation model.}
PairAlign uses conditional autoregressive likelihoods, but it is not a standard
language-modeling, prompt-continuation, or text sequence-to-sequence objective.
GPT-style models learn continuation, BERT predicts masked symbols, XLNet changes
the factorization, UniLM unifies attention masks, and MASS/BART perform
denoising sequence-to-sequence reconstruction
\citep{radford2018improving,radford2019language,devlin2019bert,yang2019xlnet,
dong2019unified,song2019mass,lewis2020bart}. PairAlign differs in both the
condition and the generated object: the input is continuous speech, the output
symbols are learned audio tokens, and inference starts from $\mathrm{BOS}$
without a text prompt, semantic prompt, acoustic-token prompt, or target-side
prefix. The acoustic condition must determine the token trajectory, identities,
length, and EOS placement.

\section{Methodology}

\subsection{Problem Formulation: Sequence Tokenization as Conditional Language Modeling}
\label{sec:problem-formulation}

Let $\mathcal{D}=\{x_i\}_{i=1}^{N}$ be a dataset of variable-length signals.
An encoder maps each input to a sequence of continuous latent states,
\begin{equation}
    Z = Enc(x), \qquad Z \in \mathbb{R}^{d\times T},
\end{equation}
where $d$ is the latent dimension and $T$ is the downsampled temporal length.
We define \emph{sequence tokenization} as mapping $Z$ to a compact discrete
string
\begin{equation}
    \mathcal{T}=[\tau_1,\dots,\tau_L],
    \qquad \tau_l\in\mathcal{A},
\end{equation}
where $\mathcal{A}$ is a finite alphabet and typically $L\ll T$. The goal is
not only to assign local frame labels, but to learn a symbolic sequence whose
identity, order, length, and edit geometry support comparison, retrieval,
alignment, memory, and downstream sequence modeling.

PairAlign formulates tokenization as conditional language modeling. Given
encoder states $Z$, an autoregressive decoder defines
\begin{equation}
    p(\mathcal{T}\mid Z;\theta_{\mathrm{AR}})
    =
    \prod_{l=1}^{L}
    p(\tau_l\mid \tau_{<l},Z;\theta_{\mathrm{AR}}).
    \label{eq:ar-tokenizer-factorization}
\end{equation}
Thus, sequence length and termination are not inherited from encoder stride or
framewise quantization; the tokenizer learns which symbols to emit, in what
order, and when to stop.

\paragraph{Target property: preserving sequence-level relational structure.}
The desired token space should preserve similarity relations between inputs.
For a content-preserving pair $(x,x^{+})$ and an unrelated example $x^{-}$,
with tokenizations $\mathcal{T}$, $\mathcal{T}^{+}$, and $\mathcal{T}^{-}$, we
would like
\begin{equation}
    ED(\mathcal{T},\mathcal{T}^{+})
    <
    \min\{
        ED(\mathcal{T},\mathcal{T}^{-}),
        ED(\mathcal{T}^{+},\mathcal{T}^{-})
    \},
    \label{edconstraint}
\end{equation}
where $ED(\cdot,\cdot)$ denotes edit distance. This captures the central
relational requirement: related views should induce compatible symbolic
strings, while unrelated inputs should remain separated.

\paragraph{Edit distance as the sequence geometry.}
Edit distance is appropriate because it compares token identity, order, length,
substitutions, insertions, and deletions. Unlike unordered overlap, it respects
the fact that the representation is a string, which is important for audio
where content and timing are expressed through ordered symbolic structure. The
constraint in Eq.~\ref{edconstraint} also excludes trivial stability: mapping
all inputs to the same generic string would make positive and negative pairs
equally close. The token space must therefore be stable across
content-preserving views while remaining discriminative across unrelated inputs.

\paragraph{Why edit distance is not optimized directly.}
Although Eq.~\ref{edconstraint} defines the desired geometry, direct
edit-distance optimization is not practical for neural tokenizer training. Edit
distance depends on discrete identities and hard alignment choices, is
piecewise constant with respect to model logits, and provides little useful
gradient. A direct ranking objective would also require decoding variable-length
strings and recomputing dynamic-programming alignments against many negatives
during training. PairAlign therefore uses edit distance as the evaluation
geometry, but optimizes a differentiable sequence-likelihood surrogate.

\paragraph{Canonical-target likelihood as a differentiable surrogate.}
For a paired example $(x_i,x_i^{+})$, let
\[
    Z_i=Enc(x_i), \qquad Z_i^{+}=Enc(x_i^{+}),
\]
and let $\mathcal{T}_i$ denote the token sequence associated with the anchor
$x_i$. Instead of directly minimizing the edit distance between independently
generated paired-view tokenizations, PairAlign uses the anchor sequence as a
shared symbolic target and makes it likely under both the anchor and
content-preserving positive representations:
\begin{equation}
    \log p(\mathcal{T}_i\mid Z_i),
    \qquad
    \log p(\mathcal{T}_i\mid Z_i^{+}).
\end{equation}
The first term preserves input-grounded prediction of the anchor tokenization,
while the second requires the perturbed positive view to recover the same
symbolic sequence. These autoregressive likelihoods provide differentiable
token-level credit assignment while remaining sensitive to token identity,
order, length, and EOS placement.

This surrogate is not equivalent to directly minimizing edit distance.
Rather, it encourages view-invariant tokenization by using a common
anchor-defined sequence as the prediction target for both content-preserving
views. If the two representations preserve the same underlying content, both
should support the same canonical symbolic description.

\paragraph{Canonicalization as a stronger view-invariance constraint.}
The shared-target formulation imposes a stricter invariance requirement than
symmetric cross-view prediction. To see this, suppose independently constructed
tokenizations of the anchor and positive views are
$\mathcal{T}_i$ and $\mathcal{T}_i^{+}$. A symmetric cross-view objective would
optimize
\[
    \log p(\mathcal{T}_i^{+}\mid Z_i),
    \qquad
    \log p(\mathcal{T}_i\mid Z_i^{+}).
\]
Even if both conditional predictions were solved perfectly, the resulting
decoded sequences need not coincide whenever
$\mathcal{T}_i\neq\mathcal{T}_i^{+}$: the objective establishes
cross-view predictability, but does not require the two views to adopt one
common symbolic realization.

PairAlign instead optimizes
\[
    \log p(\mathcal{T}_i\mid Z_i),
    \qquad
    \log p(\mathcal{T}_i\mid Z_i^{+}),
\]
so both conditioning views are trained toward the same anchor-defined sequence.
At the ideal likelihood solution, both conditions place their probability mass
on the same ordered target, giving
\[
    Dec(Z_i)=Dec(Z_i^{+})=\mathcal{T}_i
    \quad\Longrightarrow\quad
    ED\!\left(Dec(Z_i),Dec(Z_i^{+})\right)=0.
\]
Thus, zero positive-pair edit distance is the ideal symbolic configuration
induced by the objective. In practice, it is not guaranteed: PairAlign
optimizes teacher-forced sequence likelihood rather than discrete free-running
edit distance, and finite optimization, autoregressive decoding, target drift,
and model capacity can cause the decoded strings to deviate from the common
target. Positive-pair edit similarity and exact-match rate therefore measure
how closely the learned tokenizer realizes this canonicalization objective.

The distinction is consequently one of \emph{canonicalization} rather than
merely correspondence. Cross-view prediction asks each representation to
recover the symbolic realization produced by the other view, whereas
canonical-target prediction explicitly removes this target ambiguity by mapping
both content-preserving realizations toward a single symbolic description. This
is stronger specifically with respect to paired-view invariance; it does not
imply that canonical-target prediction is uniformly superior for every possible
representation-learning objective.

\paragraph{From canonical-target predictability to a discriminative symbolic space.}
Shared-target predictability alone is insufficient, since a collapsed tokenizer
could assign generic strings to many inputs and make those strings easy to
predict from both views. PairAlign therefore trains on minibatches of disjoint
positive pairs,
    $\{(x_i,x_i^{+})\}_{i=1}^{B}$,
and treats anchor token sequences from other batch elements as negatives.
For each anchor--positive pair, both conditioning representations should assign
high likelihood to the corresponding anchor-defined target
$\mathcal{T}_i$, while assigning lower likelihood to anchor targets belonging
to unrelated examples. This converts the desired edit-geometry structure into
a discriminative conditional sequence-likelihood objective.

\paragraph{Similarity in the audio setting.}
In this paper, two audio segments are treated as similar when they preserve the
same spoken content while differing in nuisance acoustics. We construct such
pairs through mild augmentations,
    $x^{+}=\mathrm{Aug}(x)$,
including gain changes, additive noise, filtering, reverberation, and short
masking. The learning signal is therefore consistency of a shared anchor-defined
tokenization across content-preserving acoustic views, not transcript
supervision.

\paragraph{Training path: from geometric tokens to self-aligned sequences.}
PairAlign is trained in three stages. Stage~I learns a contextual encoder and
nearest-centroid VQ tokenizer, providing a controlled VQ-style geometric
baseline. Stage~II freezes this tokenizer and trains an autoregressive decoder
to predict the deterministic anchor token sequence from both the anchor
representation and its content-preserving positive representation. Stage~III
replaces the fixed anchor target with an adaptive anchor tokenization generated
by an EMA teacher over the full encoder--decoder model, allowing the encoder,
decoder, output length, and generated token targets to co-evolve under
view-consistent self-alignment, hard-negative likelihood contrast, and
anti-collapse regularization. In JEPA-style terms, PairAlign uses a target
representation defined from the canonical anchor view and requires both the
anchor and its perturbed view to predict that same abstract target; unlike
standard continuous latent prediction, the target is a learned variable-length
symbolic sequence with its own vocabulary, order, and termination behavior.

\subsection{Stage I: Learning a Base Encoder and VQ Tokenizer}
\label{sec:stage1-full}
Stage~I provides the geometric starting point for PairAlign. It learns
contextual frame-level representations and converts them into discrete symbols
through nearest-centroid vector quantization. This stage is both a valid
tokenizer in its own right and the controlled baseline against which the later
sequence-level stages are compared.

\paragraph{Contextual sequence encoder.}
We use a unidirectional Mamba encoder $Enc(.)$ based on selective state-space models
\citep{dao2024transformers}. Structured state-space models provide efficient
sequence modeling through recurrent state dynamics
\citep{gu2021efficiently}, and Mamba makes the state-space parameters
input-dependent, allowing the encoder to model long-range temporal dependencies
efficiently. PairAlign is not tied to this specific encoder; the key
contribution lies in the sequence-level tokenization objective built on top of
the encoder states.

\paragraph{Self-supervised frame-level representation learning.}
The encoder is trained with the frame-level contrastive objective used in
BEST-STD~\citep{singh2025best}. Paired utterances are aligned with dynamic time
warping, aligned frames are treated as positives, and unrelated frames in the
batch provide negatives. A commitment term pulls encoder outputs toward their
assigned VQ centroids. The Stage~I objective is
\begin{equation}
    \mathcal{L}_{\mathrm{StageI}}
    =
    \mathcal{L}_{\mathrm{contrast}}
    +
    \lambda_{\mathrm{commit}}\mathcal{L}_{\mathrm{commit}}.
\end{equation}
This yields discriminative contextual embeddings suitable for geometric
discretization.

\paragraph{Nearest-centroid vector quantization.}
Given encoder embeddings \(Z_i=[z_{i,t}]_{t=1}^{T}\), a VQ codebook
\(C=\{c_a:a\in\mathcal{A}\}\) assigns each frame to its nearest centroid:
\begin{equation}
    \tau_{i,t}
    =
    \arg\min_{a\in\mathcal{A}}\|z_{i,t}-c_a\|_2^2 .
\end{equation}
This produces a raw frame-synchronous token sequence. The codebook centroids are
updated from the current encoder assignments using an exponential moving
average over assigned frame counts and assigned encoder states, following the
standard VQ update used to keep centroid estimates stable during training. For sequence-level
comparison, we apply run-length deduplication \(\phi(\cdot)\) to remove
consecutive repetitions:
\[
    \mathcal{T}_i=\phi([\tau_{i,1},\dots,\tau_{i,T}]).
\]
The resulting deduplicated VQ string is the Stage~I geometric tokenizer and
serves as the deterministic target source for Stage~II. Full implementation details for Stage~I, including the contrastive objective,
commitment loss, nearest-centroid assignment, EMA codebook update, and
run-length deduplication, are provided in Appendix~\ref{app:stage1-full}.

\paragraph{Relation to VQ-VAE-style tokenization.}
Stage~I deliberately matches the dominant geometric-tokenization paradigm:
continuous encoder latents are discretized by nearest-neighbor assignment to a
learned codebook, as in VQ-VAE-style tokenizers and related audio-tokenization
methods \citep{van2017neural,baevski2020vq,hsu2021hubert,chung2021w2v}. This
makes Stage~I a meaningful bridge to neural codecs, semantic speech units, and
retrieval-oriented VQ tokenizers. It also isolates the contribution of
PairAlign: later improvements are not obtained by replacing a weak tokenizer,
but by adding sequence-level conditional generation and self-alignment on top of
a strong VQ-style baseline.

\paragraph{Optional sequence-level strengthening of the VQ tokenizer.}
We also consider a Stage~I+ baseline that augments the geometric tokenizer with
a wav2tok-style pairwise no-blank CTC objective
\citep{banerjee2022wav2tok}. Stage~I first learns a stable encoder and VQ
codebook using the frame-level contrastive and commitment losses. Stage~I+ then
adds a soft sequence-level constraint: for paired views \((x_i,x_i^{+})\), the
deduplicated VQ sequence from one view is used as the target sequence for the
other view.

Concretely, the conditioning view's encoder latents are converted into
framewise token probabilities over the VQ alphabet,
\[
    p_{i,t}(a)=p(\tau=a\mid z_{i,t}), \qquad a\in\mathcal{A},
\]
using the same geometric token space as nearest-centroid assignment. The target
sequence is then scored with a no-blank CTC likelihood, which marginalizes over
monotonic frame-level paths that collapse to the target:
\[
    p_{\mathrm{CTC}}(\mathcal{T}\mid Z)
    =
    \sum_{\pi:\,\phi(\pi)=\mathcal{T}}
    \prod_{t=1}^{T} p(\pi_t\mid z_t).
\]
Thus, Stage~I+ encourages \(\mathcal{T}_i^{+}\) to be recoverable from
\(Z_i\), and \(\mathcal{T}_i\) to be recoverable from \(Z_i^{+}\), under a
monotonic alignment model. This injects order-sensitive sequence pressure into
the VQ tokenizer while keeping token identity frame-synchronous and
nearest-centroid based.

Because the CTC sequence likelihood operates on a different numerical scale
from the frame-level contrastive objective, we use an adaptive weight for its
contribution. The weight is set from the ratio of minibatch-averaged contrastive
and sequence losses, with a small stabilizer, so that the sequence-level term
remains proportional to the representation-learning objective rather than
dominating the second-phase optimization.

\paragraph{Why Stage~I+ uses cross-view prediction but PairAlign uses a shared canonical target.}
Stage~I+ and PairAlign intentionally use different paired-view objectives because the sequence loss plays a different role in the two formulations. Stage~I+ is an auxiliary comparison built on the discriminative Stage~I geometric tokenizer: the Stage~I contrastive and commitment objectives remain active, while symmetric cross-view no-blank CTC is added to encourage sequence-level consistency between paired geometric token streams. Importantly, Stage~I+ is \emph{not} part of the PairAlign training path; Stage~II is initialized directly from Stage~I.

PairAlign instead learns the autoregressive sequence itself as the final symbolic representation. Accordingly, from Stage~II onward, both the anchor and its perturbed view are trained toward a single anchor-defined target. This shared-target construction enforces a stronger invariance criterion than symmetric cross-prediction: exact recovery yields identical paired-view outputs, whereas cross-prediction can be perfectly satisfied even when the two view-specific target strings differ.

This distinction becomes particularly important in Stage~III, where targets are adaptive, variable-length EMA generations. Independently generating and cross-predicting two such targets would convert augmentation-, decoding-, and EOS-induced target disagreement into supervision. PairAlign therefore retains a single slowly evolving EMA-generated canonical target and complements shared-target fitting with in-batch likelihood contrast to preserve sequence-level discrimination. Full Stage~I+ details, including the no-blank CTC likelihood, symmetric cross-view loss, adaptive weighting, and complete objective, are provided in Appendix~\ref{app:stage1plus-full}.

\subsection{Stage II: Learning View-Consistent Anchor-Target Sequence Prediction}
\label{sec:stage2}

Stage~II introduces the autoregressive decoder on top of the frozen Stage~I
encoder and VQ codebook. The goal is not yet to adapt the symbolic targets, but
to learn an input-grounded sequence model over a stable token space. For each
paired example, the frozen Stage~I tokenizer defines a deterministic content-token
sequence from the anchor view. The decoder is then trained to predict this same
anchor sequence, including its terminal $\mathrm{EOS}$ symbol, from both the
anchor representation and the representation of its content-preserving positive
view. This establishes a stationary canonical-target sequence-prediction
problem before Stage~III introduces adaptive EMA-generated targets.

Throughout, we distinguish a \emph{content sequence}
\[
    \mathcal{T}
    =
    [\tau_1,\dots,\tau_M]
\]
from the corresponding \emph{autoregressive scored target}
\[
    \mathcal{Y}
    =
    [\mathcal{T},\mathrm{EOS}]
    =
    [\tau_1,\dots,\tau_M,\mathrm{EOS}].
\]
Thus, $\mathcal{T}$ excludes the terminal symbol, whereas $\mathcal{Y}$ includes
it. Token counts, token rates, and edit-based comparisons are defined on
$\mathcal{T}$ unless stated otherwise, while autoregressive likelihoods are
computed on the EOS-terminated target $\mathcal{Y}$. This distinction makes
termination part of the learned sequence-prediction objective rather than an
implicit post-processing convention.

For a scored target
$\mathcal{Y}'=[\mathcal{T}',\mathrm{EOS}]$, we define the length-normalized
conditional score
\begin{equation}
    \bar{s}(\mathcal{Y}'\mid Z;\theta_{\mathrm{AR}})
    =
    \frac{1}{|\mathcal{Y}'|}
    \sum_{l=1}^{|\mathcal{Y}'|}
    \log p_{\theta_{\mathrm{AR}}}
    (y'_l\mid y'_{<l},Z).
    \label{eq:ll-norm-main}
\end{equation}
Here $|\mathcal{Y}'|=|\mathcal{T}'|+1$, so the terminal $\mathrm{EOS}$
prediction contributes explicitly to both the likelihood and its length
normalization. Length normalization makes candidate comparisons less dominated
by sequence length, while preserving sensitivity to the complete
autoregressive string, including termination.

\paragraph{Sequence-likelihood surrogate.}
For a paired example $(x_i,x_i^{+})$, with representations $Z_i$ and
$Z_i^{+}$ and deterministic anchor content sequence $\mathcal{T}_i$, define
the EOS-terminated anchor target
\[
    \mathcal{Y}_i
    =
    [\mathcal{T}_i,\mathrm{EOS}].
\]
Stage~II maximizes
\[
    \bar{s}(\mathcal{Y}_i\mid Z_i)
    \quad \text{and} \quad
    \bar{s}(\mathcal{Y}_i\mid Z_i^{+}).
\]
The first term teaches the decoder to reproduce the anchor tokenizer's symbolic
description and its termination from its own acoustic representation. The
second requires the content-preserving positive view to support the same
EOS-terminated symbolic description. Thus, invariance is imposed through a
shared target rather than by exchanging independently generated token sequences
across the two views. This provides a differentiable surrogate for the desired
paired-view edit geometry without directly optimizing discrete edit distance.

\paragraph{Why begin with a deterministic teacher.}
Stage~I provides learned discrete units, but its sequence construction remains
locally defined by nearest-centroid assignment and run-length deduplication.
Stage~II uses the anchor sequence produced by this frozen rule as a deterministic
teacher target, giving the decoder a stable initial task: predict the same
anchor-defined VQ content string and its terminal $\mathrm{EOS}$ from both the
anchor and its content-preserving positive representation.

This separation is important because Stage~III jointly updates the encoder,
decoder, conditioning representation, generated strings, and target
distribution through EMA self-alignment. Starting directly in that regime would
entangle acoustic grounding with target drift, representation drift, and
possible collapse to generic strings. Stage~II therefore provides a grounded
autoregressive prior before the fixed-target constraint is relaxed.

\subsubsection{Deterministic Teacher from Frozen Encoder and Nearest-Centroid VQ}
\label{sec:stage2-deterministic}

In Stage~II, the Stage~I encoder and VQ codebook are kept fixed. For each
paired example, the frozen tokenizer can produce deduplicated nearest-centroid
content strings for both views,
\[
    \mathcal{T}_i=\phi(h(Enc(x_i))),
    \qquad
    \mathcal{T}_i^{+}=\phi(h(Enc(x_i^{+}))).
\]
PairAlign Stage~II, however, uses only the anchor-derived content sequence
$\mathcal{T}_i$ as autoregressive supervision. We therefore define the shared
EOS-terminated target
\[
    \mathcal{Y}_i
    =
    [\mathcal{T}_i,\mathrm{EOS}],
\]
and train both conditioning representations $Z_i$ and $Z_i^{+}$ to predict
$\mathcal{Y}_i$. The independently constructed positive-view sequence
$\mathcal{T}_i^{+}$ is not used as a Stage~II autoregressive training target.
No encoder or codebook parameters are updated in this stage.

\paragraph{Frozen VQ target as a stable bridge.}
The frozen Stage~I tokenizer supplies a learned symbolic alphabet and a fixed
frame-to-token assignment rule. Its anchor-derived output
$\mathcal{T}_i$ provides meaningful VQ-induced units, but its sequence
structure is still produced by local nearest-centroid assignment and
run-length deduplication rather than by conditional sequence generation.
Stage~II therefore uses the anchor-derived string as a stable bridge: the
decoder first learns to predict the same anchor-defined sequence and its
termination from both acoustic conditioning views before the canonical target
is allowed to adapt in Stage~III.

\paragraph{Shared-anchor teacher forcing.}
We implement $Dec_{\mathrm{AR}}$ as a Whisper-style Transformer decoder
\citep{radford2023robust}, with causal self-attention over previous target
tokens and cross-attention to encoder states. For a paired example
$(x_i,x_i^{+})$, let
\[
    \mathcal{T}_i=\phi(h(Enc(x_i))),
    \qquad
    \mathcal{Y}_i=[\mathcal{T}_i,\mathrm{EOS}]
\]
denote the deterministic anchor content sequence and its EOS-terminated scored
target produced from the frozen Stage~I tokenizer. Stage~II trains the decoder
using the same target under both conditioning views:
\begin{equation}
    \bar{s}(\mathcal{Y}_i \mid Z_i;\theta_{\mathrm{AR}})
    \qquad \text{and} \qquad
    \bar{s}(\mathcal{Y}_i \mid Z_i^{+};\theta_{\mathrm{AR}}).
\end{equation}
The anchor condition therefore predicts its own geometric content tokenization
and termination, while the positive condition is trained to recover that same
anchor-defined EOS-terminated target.

This construction lifts invariance from local frame assignments to ordered
symbolic sequences without requiring the two independently tokenized views to
serve as targets for one another. The anchor branch grounds the autoregressive
decoder in the deterministic Stage~I tokenization, while the positive branch
requires a nuisance-perturbed realization of the same content to map onto the
same ordered symbolic target. Consequently, the learned invariance is
canonicalizing: different content-preserving views are encouraged to support a
common anchor-defined sequence.

Shared-target prediction alone, however, does not guarantee acoustic grounding.
Under teacher forcing, the decoder also observes the target-side prefix, which
may explain next-token prediction through local continuation statistics while
using $Z$ weakly. The next section therefore introduces Stage~II regularization
so that high likelihood reflects input-conditioned sequence prediction rather
than primarily prefix-based continuation.

\subsubsection{Preventing Decoder Bypass in Shared-Target Teacher Forcing}
\label{sec:stage2-regularized-teacher-forcing}

Shared-anchor teacher forcing promotes sequence-level invariance across paired
conditions, but the likelihood is still computed while the decoder observes a
valid target prefix. Because fixed VQ-derived targets have regular local
structure, high teacher-forced likelihood does not by itself imply strong use
of the acoustic condition $Z$. This matters because the decoder is the
tokenizer: at inference, decoding starts from $\mathrm{BOS}$ and the complete
EOS-terminated sequence must be generated from $Z$ and the model's own previous
outputs. Stage~II therefore regularizes teacher forcing so that likelihood
reflects input-conditioned sequence prediction rather than prefix continuation.

\paragraph{Decoder bypass as the main failure mode.}
The main failure mode is that the decoder becomes insensitive to $Z$. Under
teacher forcing, next-token prediction can be explained by token-transition
statistics; if the BOS-induced state and prefix history dominate, the decoder
behaves like an unconditional language model over token strings, emitting common
prefixes and continuing them through self-attention with weak dependence on the
encoder states.

This produces misleading invariance: acoustically different inputs may map to
the same or nearly same generic string, causing Jaccard similarity, edit
similarity, and exact-match rate to rise without preserving content-dependent
distinctions. We call this \emph{decoder bypass}: the target-token pathway
explains the objective while the acoustic pathway that should define the
tokenizer is bypassed. Structured self-attention dropout weakens this
continuation path and increases prediction pressure on decoder--encoder
cross-attention to $Z$.

\paragraph{Three mechanisms for input grounding.}
We use three complementary controls. Prefix corruption masks selected
teacher-forced prefix tokens while keeping the prediction target clean, reducing
the reliability of local token history. Encoder-summary conditioning adds a
projected global summary $c(Z)$ to every decoder input, providing a direct
input-dependent pathway in addition to cross-attention. Structured
self-attention dropout stochastically removes the decoder self-attention
residual branch during training, weakening token-side continuation. Together,
these controls preserve causal autoregressive training while making high
likelihood harder to obtain from the target prefix alone.

\paragraph{Scheduled relaxation of bypass regularization.}
Decoder-side bypass controls are applied as an annealed curriculum. Prefix
corruption and structured self-attention dropout are strongest early and are
linearly relaxed over training. Early training therefore makes the target prefix
and self-attention continuation path unreliable, encouraging acoustic grounding;
later relaxation reduces train--test mismatch as the decoder approaches its
full-capacity inference configuration. Stage-specific start and end values are
reported in Appendix~\ref{app:architecture-training}.

\paragraph{Masked teacher-forcing score.}
Let
\[
    \mathcal{T}=[y_1,\dots,y_M]
\]
denote the clean content sequence,
\[
    \mathcal{Y}
    =
    [\mathcal{T},\mathrm{EOS}]
    =
    [y_1,\dots,y_M,\mathrm{EOS}]
\]
the corresponding scored target, and
\[
    \mathcal{U}
    =
    [\mathrm{BOS},y_1,\dots,y_M]
\]
the shifted teacher-forcing prefix. We sample a mask set $\mathcal{M}$ over
valid prediction positions and construct a corrupted prefix
$\widetilde{\mathcal{U}}$ by replacing selected non-special prefix tokens with
$\mathrm{MASK}$. Targets remain clean, including the terminal $\mathrm{EOS}$,
so the decoder must predict the same EOS-terminated sequence from a partially
degraded history, encoder states $Z$, and summary bias $c(Z)$.

Masking is biased toward early positions because early tokens establish the
symbolic trajectory from $\mathrm{BOS}$ and can become self-reinforcing during
autoregressive decoding. We score masked and unmasked positions separately,
obtaining
$\bar{s}_{m}(\mathcal{Y}\mid Z;\theta_{\mathrm{AR}})$ and
$\bar{s}_{u}(\mathcal{Y}\mid Z;\theta_{\mathrm{AR}})$, each normalized over its
own valid prediction positions. The terminal $\mathrm{EOS}$ prediction is part
of the clean target and is included in the teacher-forcing likelihood. The
final score is
\begin{equation}
    \bar{s}_{\mathrm{MTF}}(\mathcal{Y}\mid Z;\theta_{\mathrm{AR}})
    =
    \alpha_{MTF}\,\bar{s}_{m}(\mathcal{Y}\mid Z;\theta_{\mathrm{AR}})
    +
    (1-\alpha_{MTF})\,\bar{s}_{u}(\mathcal{Y}\mid Z;\theta_{\mathrm{AR}}),
    \qquad
    \alpha_{MTF}\in[0,1].
    \label{eq:mtf-score-stage2}
\end{equation}
Masked positions test recovery when prefix information is unreliable, while
unmasked positions retain ordinary teacher-forcing stability. Unless otherwise
stated, $\alpha_{MTF}=0.5$.

\paragraph{Encoder-summary conditioning.}
Prefix corruption weakens the target-side shortcut, but does not by itself
ensure use of the acoustic condition. We therefore add a global conditioning
pathway in parallel with cross-attention. Cross-attention provides time-resolved
access to $Z$, while the summary pathway provides an input-dependent acoustic
bias at every decoder position, including early steps where the prefix is short.

Given encoder states $Z=[z_1,\dots,z_T]$, we compute
$\bar{z}=\frac{1}{T}\sum_{t=1}^{T}z_t$, or the mask-normalized average when
padding is present, and project it to the decoder dimension:
\[
    c(Z)
    =
    W_c\,\mathrm{LayerNorm}(\bar{z}) + b_c,
    \qquad
    c(Z)\in\mathbb{R}^{d_{\mathrm{dec}}}.
\]
For the Whisper-style decoder, the input at position $l$ is
\[
    \widetilde{e}_l
    =
    E(\widetilde{u}_{l-1}) + p_l + c(Z),
\]
where $E(\widetilde{u}_{l-1})$ is the corrupted-prefix token embedding and
$p_l$ is the learned positional embedding. The summary is not a prompt token and
is not prepended; it is an additive input-dependent bias applied to each decoder
input, with scale controlled by layer normalization and projection.

This conditioning is not tied to absolute positional embeddings. Since $c(Z)$
is added to the token stream, the same idea applies to decoders using rotary,
relative, ALiBi-style, or other position-aware attention mechanisms
\citep{shaw2018self,su2024roformer,press2021train}; positional information is
then supplied by the decoder's native positional mechanism.

\paragraph{Structured self-attention dropout.}
To weaken the autoregressive shortcut directly, we apply structured dropout to
the decoder self-attention residual branch during training:
\[
    h^{(r+1)}
    =
    h^{(r)}
    +
    g^{(r)}\odot \mathrm{SA}^{(r)}(h^{(r)})
    +
    \mathrm{CA}^{(r)}(h^{(r)},Z)
    + \cdots,
\]
where $g^{(r)}\sim\mathrm{Bernoulli}(1-p_{\mathrm{sa}})$. The gate acts on the
self-attention residual branch at decoder layer $r$, while cross-attention to
$Z$ remains active. Dropping this branch weakens token-side continuation and
routes more predictive information through cross-attention and
encoder-summary conditioning.

\paragraph{Regularized Stage~II objective.}
The final Stage~II objective uses the same EOS-terminated deterministic anchor
target under both conditioning views:
\begin{equation}
    \mathcal{L}_{\mathrm{StageII}}
    =
    -\frac{1}{B}
    \sum_{i=1}^{B}
    \Big[
        \bar{s}_{\mathrm{MTF}}
        (\mathcal{Y}_i\mid Z_i;\theta_{\mathrm{AR}})
        +
        \bar{s}_{\mathrm{MTF}}
        (\mathcal{Y}_i\mid Z_i^{+};\theta_{\mathrm{AR}})
    \Big].
    \label{eq:stage2-final}
\end{equation}
The first term preserves faithful prediction of the anchor tokenizer's content
sequence and termination from the anchor representation, while the second
imposes view consistency by requiring the positive representation to predict
the same anchor-defined EOS-terminated sequence. This stage therefore learns an
input-conditioned decoder over fixed VQ targets while discouraging
teacher-forced likelihood from becoming a prefix-continuation shortcut.

\subsection{Stage III: Full-Model EMA Self-Alignment for Adaptive Tokenization}
\label{sec:stage3}

Stage~II learns an input-grounded decoder over fixed anchor VQ targets, but
those target strings remain determined by the frozen geometric tokenizer.
Stage~III removes this constraint by replacing the deterministic anchor VQ
content sequence $\mathcal{T}_i$ with an adaptive anchor content sequence
$\widehat{\mathcal{T}}_i$ generated by an EMA teacher over the full
encoder--decoder tokenizer. Its corresponding scored target is
\[
    \widehat{\mathcal{Y}}_i
    =
    [\widehat{\mathcal{T}}_i,\mathrm{EOS}].
\]
The encoder and decoder are then optimized jointly, allowing the conditioning
space, generated content string, output length, EOS placement, and target
distribution to co-evolve.

The difficulty is that Stage~III is self-referential: the model family both
generates the symbolic anchor targets and learns from them. Without additional
structure, target generation, target fitting, and negative comparison can drift
together, leading to unstable targets, decoder bypass, or many-to-one collapse.
We therefore separate these roles. The EMA teacher generates a free-running
EOS-terminated target from the anchor view at full decoding capacity; the
student fits that same anchor target from both the anchor and positive
conditions using the bypass-resistant $\bar{s}_{\mathrm{MTF}}$ score from
Stage~II; and hard-negative comparison uses clean-prefix likelihoods so that the
likelihood matrix measures symbolic confusability rather than robustness to a
sampled mask pattern.

Thus, Stage~III is not ordinary self-training on decoded strings. It is a
controlled canonical-target refinement step that decouples adaptive anchor
target generation, grounded multi-view fitting, and discriminative negative
comparison. This lets PairAlign move beyond frame-derived VQ targets while
preserving stable target generation, input grounding, view consistency, and
sequence-level separability.

\subsubsection{Full-Model EMA Teacher and Joint Encoder--Decoder Optimization}

\paragraph{EMA teacher over the complete tokenizer.}
Let $\theta=(\theta_{\mathrm{enc}},\theta_{\mathrm{AR}})$ denote the student
parameters. Stage~III maintains an EMA teacher $\tilde{\theta}$ updated as
\[
    \tilde{\theta}
    \leftarrow
    \alpha\tilde{\theta}+(1-\alpha)\theta .
\]
Unlike the Stage~I deterministic teacher, this teacher contains both the encoder
and autoregressive decoder. It provides adaptive symbolic targets whose
evolution is slower than the student updates, allowing the target distribution
to change without training directly on rapidly moving student-generated strings.

\paragraph{Full-capacity EMA target generation.}
For each paired example, the EMA teacher generates the adaptive symbolic target
from the anchor view. Given the teacher representation
$\widetilde{Z}_i=\widetilde{Enc}(x_i)$, the EMA decoder generates a
variable-length sequence by free-running autoregressive decoding from
$\mathrm{BOS}$ until $\mathrm{EOS}$. Generation uses the full decoder
computation: prefix corruption is disabled, self-attention dropout is disabled,
and cross-attention and encoder-summary conditioning remain active. The teacher
therefore defines the current adaptive anchor tokenizer rather than a
regularized training-time approximation to it.

To avoid repetition loops, generic starts, and uncontrolled continuation,
teacher decoding uses top-$p$ sampling~\citep{holtzman2019curious}, repetition
control, early-step stochasticity, and a sample-specific length cap. We write
the generated EOS-terminated anchor target as
\begin{equation}
    \widehat{\mathcal{Y}}_i
    =
    [\widehat{\mathcal{T}}_i,\mathrm{EOS}]
    =
    \mathrm{Sample}
    (\widetilde{Enc}(x_i);\tilde{\theta}),
    \label{eq:stage3-ema-target}
\end{equation}
where $\widehat{\mathcal{T}}_i$ denotes only the generated content tokens
preceding $\mathrm{EOS}$. The same anchor-generated target
$\widehat{\mathcal{Y}}_i$ is subsequently used as the student target under both
the anchor representation $Z_i$ and the positive representation $Z_i^{+}$.
These adaptive targets replace the fixed Stage~I anchor targets
$\mathcal{Y}_i$ used in Stage~II. Sampling details are given in
\S~\ref{sec:ar-target-generation-decoding}.

\paragraph{Bypass-resistant student fitting.}
The EMA teacher generates an adaptive target from the anchor view, while the
student fits that same EOS-terminated target under both the anchor and positive
conditioning representations using $\bar{s}_{\mathrm{MTF}}$. Prefix corruption
and structured self-attention dropout remain active for this fitting step, as in
Stage~II, so positive alignment cannot be solved only through teacher-forced
prefix continuation. Encoder-summary conditioning remains part of the decoder
throughout Stages~II--III, keeping the acoustic condition directly available at
every prediction step. Thus, even with adaptive targets, fitting remains an
input-grounded canonical-target sequence-prediction problem in which
content-preserving views are required to support a common symbolic sequence and
its termination.

\paragraph{Stable joint refinement.}
Stage~III updates both encoder and decoder, but the encoder also defines the
conditioning space used for teacher generation and student scoring. To reduce
simultaneous drift in representations, targets, and likelihood scores, we update
the encoder more conservatively than the decoder:
\[
    \eta_{\mathrm{enc}} = 0.1\,\eta_{\mathrm{dec}}.
\]
This allows the full tokenizer to move beyond the fixed Stage~I geometry while
limiting instability in the moving target distribution.

\paragraph{Shared-anchor positive alignment.}
Using student representations $Z_i$ and $Z_i^{+}$, Stage~III scores the same
EMA-generated EOS-terminated anchor target under both conditioning views:
\begin{equation}
    \mathcal{P}^{\mathrm{anc}}_{ii}
    =
    \bar{s}_{\mathrm{MTF}}
    (\widehat{\mathcal{Y}}_i\mid Z_i;\theta),
    \qquad
    \mathcal{P}^{\mathrm{pos}}_{ii}
    =
    \bar{s}_{\mathrm{MTF}}
    (\widehat{\mathcal{Y}}_i\mid Z_i^{+};\theta).
    \label{eq:stage3-positive-scores}
\end{equation}
The anchor term requires the student to reproduce the EMA teacher's current
anchor content tokenization and termination from the anchor representation
itself. The positive term requires a content-preserving perturbed view to
recover that same adaptive EOS-terminated anchor target. Stage~III therefore
retains the Stage~II shared-target principle while allowing the canonical
symbolic target itself to evolve with the EMA tokenizer.

\paragraph{Uncorrupted likelihoods for negative comparison.}
Positive fitting uses $\bar{s}_{\mathrm{MTF}}$ because paired-target alignment
should remain bypass-resistant. Negative comparison has a different role: it
should measure which teacher-generated strings are genuinely confusable under a
given acoustic condition. We therefore compute the hard-negative likelihood
matrix using clean teacher-forced prefixes and the length-normalized score
$\bar{s}(\cdot\mid Z)$.

Here, ``uncorrupted'' refers to the negative-scoring path as a whole: candidate
EOS-terminated strings are scored without prefix masking and without structured
self-attention dropout. Thus, confusability is evaluated under a deterministic
clean-prefix decoder configuration rather than depending on sampled masking or
self-attention-dropout patterns. The remaining decoder architecture and
acoustic conditioning pathways are unchanged. The resulting matrix tests
whether each acoustic condition assigns higher clean-prefix likelihood to its
paired EMA-generated target than to the most confusable mismatched targets in
the minibatch.

\subsubsection{Preventing Tokenization Collapse with In-Batch Likelihood Contrast}
\label{sec:stage3-collapse-prevention}

Positive self-alignment alone does not ensure a discriminative token space. A
collapsed tokenizer can make paired views mutually predictable while also making
unrelated examples highly likely. Stage~III therefore adds sequence-level
discriminative pressure: under each conditioning representation, the paired
teacher target should outrank the most confusable teacher targets from other
examples in the minibatch.

\paragraph{Likelihood matrices for symbolic confusability.}
For negative comparison, we construct likelihood matrices using the clean-prefix
score $\bar{s}(\cdot\mid Z)$ from Eq.~\ref{eq:ll-norm-main}. All candidate
strings are EOS-terminated EMA-generated anchor targets
$\{\widehat{\mathcal{Y}}_j\}_{j=1}^{B}$; the two matrices differ only in
whether the conditioning representation comes from the anchor or positive view.

For anchor conditioning,
\begin{equation}
    M^{\mathrm{anc}}_{ij}
    =
    \bar{s}(\widehat{\mathcal{Y}}_j\mid Z_i;\theta),
    \qquad i,j\in\{1,\dots,B\}.
    \label{eq:matrix-anchor}
\end{equation}
For positive conditioning,
\begin{equation}
    M^{\mathrm{pos}}_{ij}
    =
    \bar{s}(\widehat{\mathcal{Y}}_j\mid Z_i^{+};\theta).
    \label{eq:matrix-positive}
\end{equation}
In both matrices, column $j$ denotes the adaptive EOS-terminated anchor target
belonging to example $j$. The diagonal therefore corresponds to the correct
shared target $\widehat{\mathcal{Y}}_i$, while off-diagonal entries correspond
to anchor targets from unrelated examples. The desired behavior is row-wise
separation under both conditioning views: the correct anchor-defined sequence
should score above mismatched anchor sequences whether the condition is $Z_i$
or $Z_i^{+}$.

\paragraph{Hardest-$K$ confusable negatives.}
For each row, we retain the $K$ largest off-diagonal entries:
\begin{equation}
    \mathcal{H}^{\mathrm{anc}}_K(i)
    =
    \operatorname{TopK}
    \big(\{M^{\mathrm{anc}}_{ij}:j\neq i\},K\big),
    \qquad
    \mathcal{H}^{\mathrm{pos}}_K(i)
    =
    \operatorname{TopK}
    \big(\{M^{\mathrm{pos}}_{ij}:j\neq i\},K\big).
\end{equation}
These are the mismatched anchor targets that currently score highest under the
anchor and positive conditioning representations, respectively. They are the
relevant negatives for collapse prevention: high likelihood for such strings
indicates loss of input-dependent discriminability.

\paragraph{Row-wise contrast over sequence likelihoods.}
We apply a row-wise contrastive loss between the correct anchor target and its
hardest mismatched competitors. For anchor conditioning,
\begin{equation}
    \mathcal{L}^{\mathrm{anc}}_{\mathrm{NCE}}
    =
    -\frac{1}{B}
    \sum_{i=1}^{B}
    \log
    \frac{
        \exp(M^{\mathrm{anc}}_{ii}/\tau_{\ell})
    }{
        \exp(M^{\mathrm{anc}}_{ii}/\tau_{\ell})
        +
        \sum_{j\in\mathcal{H}^{\mathrm{anc}}_K(i)}
        \exp(M^{\mathrm{anc}}_{ij}/\tau_{\ell})
    },
    \label{eq:nce-anchor}
\end{equation}
where $\tau_{\ell}>0$ is a temperature over likelihood scores. The corresponding
positive-conditioning loss is
\begin{equation}
    \mathcal{L}^{\mathrm{pos}}_{\mathrm{NCE}}
    =
    -\frac{1}{B}
    \sum_{i=1}^{B}
    \log
    \frac{
        \exp(M^{\mathrm{pos}}_{ii}/\tau_{\ell})
    }{
        \exp(M^{\mathrm{pos}}_{ii}/\tau_{\ell})
        +
        \sum_{j\in\mathcal{H}^{\mathrm{pos}}_K(i)}
        \exp(M^{\mathrm{pos}}_{ij}/\tau_{\ell})
    }.
    \label{eq:nce-positive}
\end{equation}
We define
\[
    \mathcal{L}_{\mathrm{NCE}}
    =
    \mathcal{L}^{\mathrm{anc}}_{\mathrm{NCE}}
    +
    \mathcal{L}^{\mathrm{pos}}_{\mathrm{NCE}}.
\]
This term complements shared-target fitting: both conditioning views should
predict the same correct EOS-terminated anchor target, while targets belonging
to other anchors must not become equally likely.

\paragraph{Objective-level effect of likelihood contrast.}
The row-wise contrastive objective admits a direct likelihood-separation
interpretation. For conditioning view
$V\in\{\mathrm{anc},\mathrm{pos}\}$, let $\ell_i^{V}$ denote the
per-example negative log-softmax contrastive loss, let $M_{ii}^{V}$ denote
the clean conditional log-likelihood score of the paired EMA target, and let
$M_{ij}^{V}$ denote the corresponding score of a selected hard negative.
Rewriting the row-wise negative log-softmax gives
\begin{align}
    \ell_i^{V}
    &=
    -\log
    \frac{
        \exp(M_{ii}^{V}/\tau_{\ell})
    }{
        \exp(M_{ii}^{V}/\tau_{\ell})
        +
        \sum_{j\in\mathcal{H}^{V}_K(i)}
        \exp(M_{ij}^{V}/\tau_{\ell})
    }
    \nonumber\\
    &=
    \log
    \left[
        1+
        \sum_{j\in\mathcal{H}^{V}_K(i)}
        \exp
        \left(
            \frac{
                M_{ij}^{V}-M_{ii}^{V}
            }{
                \tau_{\ell}
            }
        \right)
    \right].
    \label{eq:rowwise-nce-form}
\end{align}
If $\ell_i^{V}\leq\varepsilon$, then every selected hard negative satisfies
\begin{equation}
    M_{ii}^{V}-M_{ij}^{V}
    \geq
    \tau_{\ell}
    \log
    \left(
        \frac{1}{
            e^{\varepsilon}-1
        }
    \right),
    \qquad
    j\in\mathcal{H}^{V}_K(i).
    \label{eq:nce-likelihood-margin}
\end{equation}
The right-hand side is strictly positive whenever
$\varepsilon<\log 2$. Since $\mathcal{H}^{V}_K(i)$ contains the
highest-scoring mismatched candidates, separating the hardest negative by this
margin also separates all lower-scoring mismatches in the same minibatch.
Thus, sufficiently small contrastive loss directly implies
paired-versus-mismatched separation in clean conditional likelihood. This
statement concerns the likelihood geometry optimized by
$\mathcal{L}_{\mathrm{NCE}}$; we do not interpret it as a numerical
edit-distance guarantee.

The same objective also penalizes complete identical-sequence collapse. If all
candidate targets in the minibatch are identical,
\begin{equation}
    \widehat{\mathcal{Y}}_j
    =
    \mathcal{Y}_c,
    \qquad
    \forall j,
\end{equation}
then paired and mismatched candidates receive identical scores within each
conditioning row, giving
\begin{equation}
    \ell_i^{V}
    =
    \log(K+1).
    \label{eq:nce-complete-collapse}
\end{equation}
Complete identical-target collapse therefore cannot minimize the contrastive
term. This observation is deliberately limited: it does not exclude partial
many-to-one mappings, subgroup collisions, or other low-diversity solutions.
We therefore assess these weaker failure modes empirically through
active-vocabulary usage, low-diversity diagnostics, and cross-example sequence
collisions.

\paragraph{Entropy regularization.}
We additionally apply a negative-entropy penalty to the student next-token
distribution, averaged over batch elements and decoder positions. This term
discourages early concentration of predictive mass on a small set of
high-prior symbols and provides a complementary pressure toward broader
predictive support during EMA self-alignment.

Predictive entropy alone, however, does not imply input-dependent
tokenization. A decoder may satisfy
\begin{equation}
    p_{\theta}
    \left(
        \tau_l
        \mid
        \tau_{<l},Z
    \right)
    =
    q_l
    \left(
        \tau_l
        \mid
        \tau_{<l}
    \right)
    \qquad
    \forall Z,
    \label{eq:input-independent-high-entropy}
\end{equation}
while $q_l$ remains high-entropy. Entropy regularization can therefore
discourage concentration of the predictive distribution, but cannot by itself
establish acoustic dependence or prevent all forms of sequence collapse.
PairAlign consequently uses likelihood contrast for explicit cross-example
discrimination and evaluates realized non-collapse directly through vocabulary
usage, sequence diversity, and cross-example collision statistics.

\paragraph{Final Stage~III objective.}
The positive self-alignment loss is
\begin{equation}
    \mathcal{L}_{\mathrm{pos}}
    =
    -\frac{1}{B}
    \sum_{i=1}^{B}
    \left(
        \mathcal{P}^{\mathrm{anc}}_{ii}
        +
        \mathcal{P}^{\mathrm{pos}}_{ii}
    \right),
\end{equation}
where both positive scores use the same EMA-generated EOS-terminated anchor
target $\widehat{\mathcal{Y}}_i$ and are computed with
$\bar{s}_{\mathrm{MTF}}$. The full objective is
\begin{equation}
    \mathcal{L}_{\mathrm{StageIII}}
    =
    \mathcal{L}_{\mathrm{pos}}
    +
    \lambda_{\mathrm{NCE}}
    \left(
        \mathcal{L}^{\mathrm{anc}}_{\mathrm{NCE}}
        +
        \mathcal{L}^{\mathrm{pos}}_{\mathrm{NCE}}
    \right)
    +
    \lambda_{\mathrm{entropy}}\mathcal{L}_{\mathrm{entropy}}.
    \label{eq:stage3-final-rev}
\end{equation}
The terms have distinct roles: positive alignment requires both
content-preserving conditioning views to reproduce the same adaptive
EOS-terminated anchor target; NCE prevents many-to-one collapse by ranking the
correct anchor target above hard mismatched anchor targets under both
conditioning views; and entropy regularization preserves predictive diversity.

\paragraph{From fixed VQ targets to adaptive symbolic self-alignment.}
Stage~II trains on stationary anchor VQ content sequences
$\mathcal{T}_i$, using their EOS-terminated forms
$\mathcal{Y}_i=[\mathcal{T}_i,\mathrm{EOS}]$ as the common prediction target
for both the anchor and positive representations. Stage~III replaces this fixed
target with an adaptive anchor content sequence $\widehat{\mathcal{T}}_i$
generated by the EMA encoder--decoder tokenizer and trains against its
EOS-terminated form
$\widehat{\mathcal{Y}}_i=[\widehat{\mathcal{T}}_i,\mathrm{EOS}]$ while jointly
refining the student. The transition is controlled by separating three
computations: full-capacity EMA anchor-target generation, bypass-resistant
shared-target fitting, and clean-prefix likelihood contrast against confusable
anchor targets from other examples. The resulting content token sequence is no
longer restricted to deduplicated nearest-centroid assignment; it becomes an
adaptive symbolic string whose stability across content-preserving views is
learned through canonical-target prediction and whose discriminability is
constrained by in-batch likelihood contrast.

\begin{figure*}[t]
    \centering
    \includegraphics[width=\textwidth]{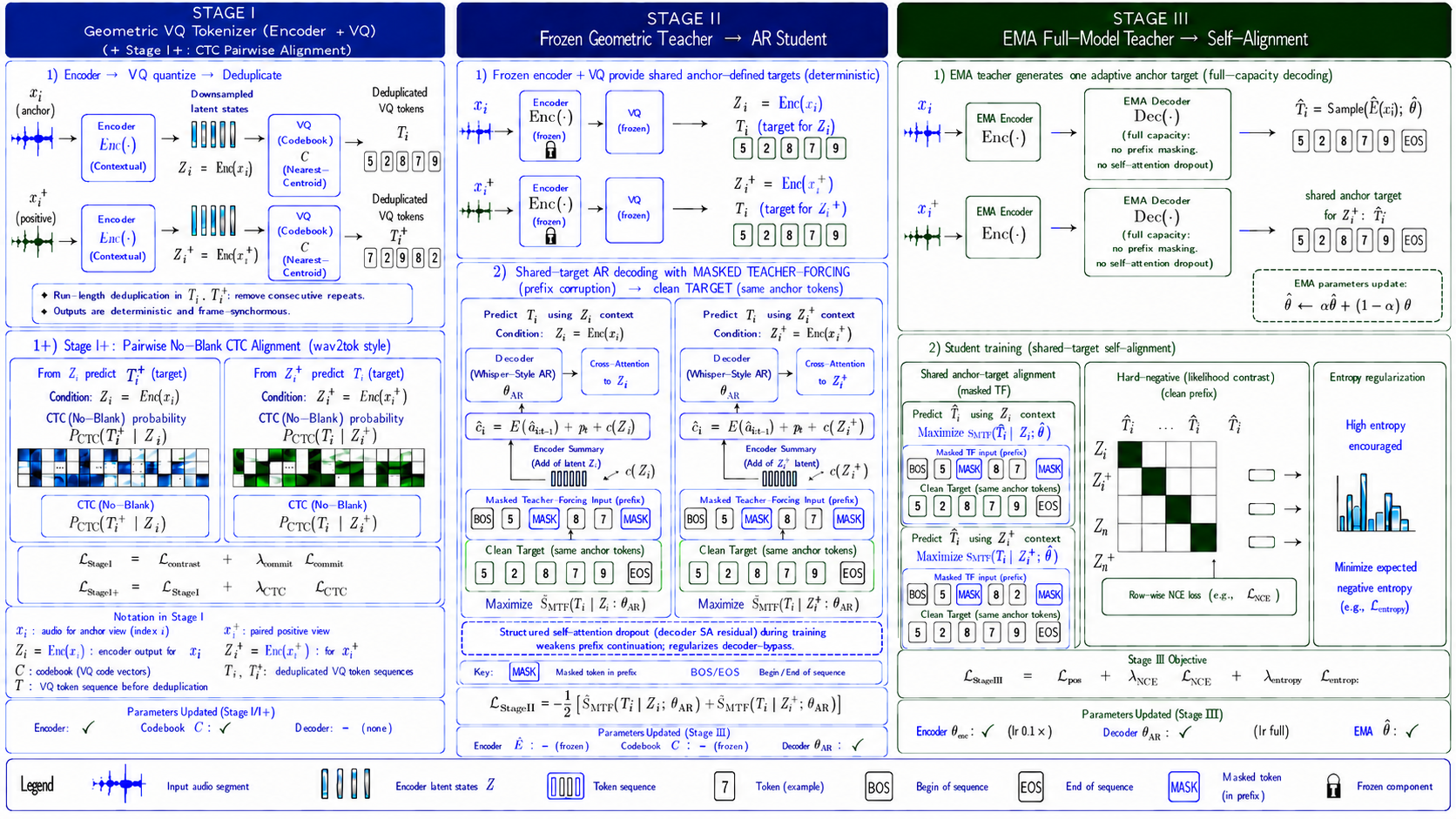}
    \caption{
    Overview of the staged PairAlign training framework.
    }
    \label{fig:pairalign_method_overview}
\end{figure*}

Figure~\ref{fig:pairalign_method_overview} consolidates the complete staged
progression. Stage~I learns the geometric encoder--VQ tokenizer and its
deduplicated content-token sequences. The optional Stage~I+ variant strengthens
the geometric tokenizer using symmetric cross-view no-blank CTC: the anchor
representation predicts the positive-view geometric sequence, while the
positive representation predicts the anchor-view geometric sequence. Stage~II
instead introduces PairAlign's canonical shared-target formulation, using
$\mathcal{Y}_i=[\mathcal{T}_i,\mathrm{EOS}]$, derived from the deterministic
anchor sequence, as the common autoregressive target for both views. Finally,
Stage~III replaces the fixed geometric anchor target with an adaptive content
sequence $\widehat{\mathcal{T}}_i$ generated from the anchor by a full-capacity
EMA copy of the complete encoder--decoder tokenizer, with
$\widehat{\mathcal{Y}}_i=
[\widehat{\mathcal{T}}_i,\mathrm{EOS}]$ used for autoregressive scoring.
The student is jointly refined by requiring both the anchor and positive
conditions to predict this shared EMA anchor target, while hard in-batch
likelihood contrast separates it from EOS-terminated anchor targets belonging
to unrelated examples and entropy regularization discourages predictive
collapse.

\subsubsection{Autoregressive Target Generation and Decoding}
\label{sec:ar-target-generation-decoding}

PairAlign uses the autoregressive tokenizer in two modes. During Stage~III, the
EMA teacher generates adaptive self-alignment targets with stochastic top-$p$
sampling. After Stage~III, the \emph{final EMA teacher} is used as the tokenizer
for all reported PairAlign evaluations and is decoded deterministically with
beam search; the student is used only for optimization during training. This
separates training-time target exploration from evaluation-time tokenization
stability.
 This separates training-time target
exploration from evaluation-time tokenization stability.

\paragraph{EMA-teacher targets.}
For an input segment $x$, the EMA teacher encoder produces
$\widetilde{Z}=\widetilde{Enc}(x)$, and the EMA decoder generates an
EOS-terminated variable-length sequence autoregressively from $\mathrm{BOS}$.
We denote its content portion by
\[
    \widehat{\mathcal{T}}
    =
    [\widehat{\tau}_1,\dots,\widehat{\tau}_M]
\]
and the complete generated target by
\[
    \widehat{\mathcal{Y}}
    =
    [\widehat{\mathcal{T}},\mathrm{EOS}].
\]
The content sequence $\widehat{\mathcal{T}}$ is the symbolic tokenization used
for downstream token counts and sequence comparisons, whereas
$\widehat{\mathcal{Y}}$ is the sequence used whenever the autoregressive model
is scored. The target is therefore not an external label or a fixed VQ string;
it is the current EMA tokenizer's own input-conditioned prediction. During
Stage~III, targets are sampled with nucleus sampling
\citep{holtzman2019curious}. To keep this free-running process useful, teacher
decoding combines repetition control, position-dependent sampling parameters,
and an explicit length cap to avoid generic prefixes, short loops, and
unbounded continuation.

\paragraph{Repetition control and sampling schedule.}
Before sampling or beam expansion, repeated non-special tokens receive a
count-dependent multiplicative penalty. If token $a$ has appeared $n_a$ times in the current partial content sequence,
its next-token probability is downweighted by a factor proportional to
$\gamma_{\mathrm{rep}}^{-n_a}$, or equivalently its logit receives a
count-dependent negative shift. Unseen tokens are unchanged, and
$\mathrm{BOS}$, $\mathrm{EOS}$, and $\mathrm{PAD}$ are excluded from the count.
The penalty only affects next-token selection; it does not edit the generated
sequence after decoding. During teacher sampling it shapes the stochastic
top-$p$ distribution, while during inference it shapes deterministic
beam-search scores.

Teacher sampling then applies nucleus filtering and temperature scaling. The
decoder keeps the smallest token set whose cumulative probability exceeds the
top-$p$ threshold, preserves the highest-probability token, and samples from the
filtered distribution. Early steps use higher temperature, larger top-$p$, and
weaker repetition penalty because they establish the trajectory from
$\mathrm{BOS}$; later steps use sharper settings to stabilize continuation.

\paragraph{Length-constrained EOS handling.}
The decoded tokenization is not required to emit one content token per encoder
frame. The decoder learns when to emit $\mathrm{EOS}$, so the content length is
adaptive. We define
\[
    L_{\mathrm{cap}}(x)
    =
    \lfloor \rho T_x \rfloor,
\]
where $T_x$ is the valid encoder length and $\rho\in[0,1]$. Crucially,
$L_{\mathrm{cap}}(x)$ counts autoregressive \emph{output slots after
$\mathrm{BOS}$, including the terminal $\mathrm{EOS}$ slot}. Therefore the
content sequence satisfies
\begin{equation}
    |\mathcal{T}(x)|
    \leq
    L_{\mathrm{cap}}(x)-1.
    \label{eq:content-length-cap}
\end{equation}
If $\mathrm{EOS}$ is emitted before the final permitted output slot, decoding
stops normally. If it has not been emitted earlier, the final permitted slot is
forced to $\mathrm{EOS}$. For example, an output cap of
$L_{\mathrm{cap}}=45$ permits at most $44$ content tokens followed by
$\mathrm{EOS}$.

The cap therefore prevents unbounded generation and controls the maximum
symbolic bitrate, but it is not a supervised length target and does not use
transcript, boundary, or Stage~I VQ length information. Within this upper bound,
the model determines the termination position by predicting $\mathrm{EOS}$.

\paragraph{Inference-time beam search.}
After Stage~III, all reported PairAlign tokenizations are produced by the final
EMA teacher rather than the student. Given an input $x$, the final EMA encoder
produces
$\widetilde{Z}=\widetilde{Enc}(x)$,
and the corresponding EMA decoder performs deterministic beam search from
$\mathrm{BOS}$. It maintains $K$ partial hypotheses until $\mathrm{EOS}$ is
emitted. At most $L_{\mathrm{cap}}(x)$ output positions after $\mathrm{BOS}$ are
permitted, including $\mathrm{EOS}$; if no hypothesis terminates earlier,
$\mathrm{EOS}$ is forced in the final permitted slot.

Evaluation uses the full EMA decoder: prefix corruption and self-attention
dropout are disabled, while cross-attention and encoder-summary conditioning
remain active. Repetition control and the length cap are retained. The
highest-scoring completed beam defines the EOS-terminated decoded sequence
$\mathcal{Y}$, and the reported tokenizer output $\mathcal{T}$ is obtained by
removing the terminal $\mathrm{EOS}$. Thus, downstream sequence metrics operate
on content tokens, while beam completion and sequence likelihood explicitly
model termination. The beam width is $K$, $\rho$ controls the output-slot cap,
$\gamma_{\mathrm{rep}}$ the repetition penalty, and
$\alpha_{\mathrm{len}}$ optional beam-score length normalization.

\paragraph{Summary.}
Stage~III uses stochastic EMA-teacher sampling to obtain adaptive,
input-conditioned EOS-terminated targets
$\widehat{\mathcal{Y}}=[\widehat{\mathcal{T}},\mathrm{EOS}]$ without external
symbolic supervision. After training, the final EMA teacher is retained as the
PairAlign tokenizer for all reported evaluations and is decoded
deterministically with beam search; the student is not used for evaluation.
The resulting EOS-terminated sequence $\mathcal{Y}$ yields the reported content
tokenization $\mathcal{T}$ after removing $\mathrm{EOS}$. Pseudocode for
teacher sampling and EMA-teacher beam-search decoding is provided in
Appendices~\ref{app:teacher-sampling-full} and
\ref{app:beam-search-full}.

\subsection{Extracting Timing Information from Autoregressive Decoding}
\label{sec:timing-from-ar}

Autoregressive tokenization produces a compact symbolic sequence rather than a
frame-synchronous label stream. This supports storage and sequence-level
comparison, but removes native token-to-frame timing. When approximate temporal
grounding is needed, we extract timestamps from decoder cross-attention after
decoding. This is an inference-time post-processing step: it does not modify the
training objective, constrain the decoder, or backpropagate through recovered
alignments.

The procedure uses monotonicity only as a weak post-hoc bias. Speech-to-symbol
correspondence is expected to be approximately ordered in time, and prior work
has incorporated this bias directly through local, Gaussian, hard monotonic,
monotonic multi-head, location-aware, and attention-prior-guided mechanisms
\citep{tjandra2017local,hou2017gaussian,merboldt2019analysis,wu2019exact,
ma2019monotonic,chowdhury2023monotonic,zhao2020cross,neekhara2024improving}.
PairAlign does not impose such constraints during training; monotonicity is used
only to interpret the trained decoder's cross-attention.

Given encoder latents $Z$ and a decoded sequence, we extract decoder--encoder
cross-attention from a selected layer or from an average over layers and heads.
After removing $\mathrm{BOS}$, $\mathrm{EOS}$, decoder padding, and padded
encoder frames, we obtain $W\in\mathbb{R}^{L\times T}$, where $L$ is the number
of decoded non-special tokens and $T$ is the number of valid encoder frames.
Because raw attention can be diffuse or multi-modal, we apply a weak diagonal
2D Beta-shaped prior inspired by beta-binomial attention guidance
\citep{neekhara2024improving}. There, the prior is applied to cross-attention
during training with an alignment loss to encourage monotonicity in speech-LLM
generation. Here, it is used only after decoding to sharpen monotonic structure
already present in the learned attention.

The prior-reweighted token-to-frame attention is renormalized across decoded
positions for each encoder frame, yielding a frame-to-token posterior
$A\in\mathbb{R}^{T\times L}$. We then decode a monotone Viterbi path in which
each frame either remains at the current token or advances to the next token,
producing a contiguous monotone token-to-frame segmentation. The frame support
assigned to each token is converted to start and end times using the window
start time, sampling rate, and encoder downsampling factor, giving
$\{(u_l,t_l^{\mathrm{start}},t_l^{\mathrm{end}})\}_{l=1}^{L}$.

\paragraph{Evaluating recovered temporal grounding.}
We evaluate the recovered alignment at two levels. First, internal diagnostics
measure whether the raw and prior-reweighted frame-to-token posteriors become
more monotone and whether the resulting Viterbi path receives higher alignment
score. Second, because improved monotonicity alone does not establish temporal
accuracy, we compare recovered token transitions against sample-level TIMIT
phone-boundary timestamps. We report boundary precision, recall, and F1 under
multiple temporal tolerances, together with matched-boundary localization
error. This external check is intentionally diagnostic: reference phone
boundaries provide a reproducible timing target, but PairAlign is not trained
to produce phonemes or one-token--one-phone segments.

The recovered timestamps should therefore be interpreted as approximate temporal
grounding for learned token positions, not supervised boundaries. Unlike CTC,
RNN-T, monotonic attention, monotonic chunkwise attention, monotonic multi-head
attention, or training-time attention-prior methods
\citep{graves2006connectionist,graves2012sequence,raffel2017online,
chiu2017monotonic,ma2019monotonic,wu2019exact,sak2017recurrent,
zhao2020cross,neekhara2024improving}, this procedure introduces no alignment
latent variable and makes no training-time change. Its reliability depends on
the structure of the learned cross-attention; diffuse or non-monotone attention
yields less reliable timestamps. Empirically, the Beta prior improves the
internal monotonicity diagnostics and yields small but consistent improvements
in TIMIT phone-boundary F1, while matched-boundary localization error changes
little. We therefore treat the procedure as an auxiliary mechanism for
approximate post-hoc temporal grounding rather than evidence of supervised
alignment or phonetic segmentation. The full algorithm, raw-versus-prior
diagnostics, boundary-based evaluation, and visualizations are given in
Appendix~\ref{app:timing-recovery-full}.

\section{Experiments}
\label{sec:experiments}

\paragraph{Experimental scope.}
PairAlign is an empirical framework for sequence-level audio tokenization. We
therefore evaluate the learned representation directly through paired-view
consistency, symbolic rate, collapse and inventory usage, positive--negative
edit geometry, retrieval and comparison cost, local acoustic sensitivity,
learned termination, temporal recoverability, acoustic--linguistic structure,
computational cost, and distribution shift. We additionally probe how the
Stage~III training objective relates to the local edit geometry that emerges
after training. The objective is to characterize the resulting
compactness--granularity trade-off rather than require PairAlign to dominate
every metric.

\paragraph{Continuous speech operating regime.}
PairAlign targets continuous speech, not isolated lexical crops. All learned
tokenizers and SSL references use LibriSpeech-trained representations and are
evaluated without adaptation on TIMIT or Shrutilipi-Tamil; fixed $3$ s windows
form the primary regime. In our LibriSpeech word-crop extraction, retained
words were shorter than $0.5$ s and provided insufficient conditioning for
autoregressive generation, producing severe generic-string collapse. A $3$ s
window instead provides phonetic, speaker, prosodic, and local lexical context
while remaining tractable for cross-attention, teacher-forced likelihoods,
EMA-teacher decoding, and in-batch contrast; comparable short-context speech
modeling is common in codec language models
\citep{borsos2023audiolm,wang2023neural}.

\paragraph{Evaluation questions.}
We test whether PairAlign:
\emph{(i)} produces consistent but non-collapsed paired-view strings;
\emph{(ii)} moves to a substantially lower symbolic rate;
\emph{(iii)} contracts positives relative to meaningful negatives;
\emph{(iv)} preserves useful retrieval geometry while reducing repeated
sequence-comparison cost;
\emph{(v)} evolves coherently under small acoustic context shifts;
\emph{(vi)} learns input-conditioned EOS and admits approximate post-hoc timing;
\emph{(vii)} retains measurable phonetic, boundary, silence, and speaker
structure;
\emph{(viii)} maintains a structured operating point under external noise,
cross-corpus English speech, and lexically disjoint Tamil; and
\emph{(ix)} exhibits the local contraction--separation geometry encouraged by
the Stage~III objective.
Component ablations, rate-controlled BPE, and high-exposure HuBERT/WavLM
tokenizers test alternative explanations.

\paragraph{Evaluation domains.}
PairAlign is trained on LibriSpeech \texttt{train-clean-360} and
\texttt{train-other-500} ($860$ h)
\citep{panayotov2015librispeech}. Reported checkpoints were selected using
held-out LibriSpeech \texttt{test-clean}; this official test split is not used
otherwise in the reported tables, and its use for checkpoint selection is a
limitation. Disjoint \texttt{train-clean-100} is used for LibriSpeech
consistency, inventory, and rate-control analyses.

TIMIT is the primary cross-corpus English testbed
\citep{garofolo1993darpa}, differing from LibriSpeech in speakers, recording,
utterance structure, and phonetic coverage. It supports consistency, external
noise, relational geometry, retrieval, rate control, continuous sweeps,
unconstrained termination, timing recovery, and phonetic/boundary/silence/
speaker probes. Its retrieval archive contains $9{,}461$ overlapping $3$ s
segments ($\sim3.94$ h unique speech) with $1.5$ s hop; continuous sweeps use a
$100$ ms hop.

Shrutilipi-Tamil provides lexically disjoint cross-lingual transfer
\citep{bhogale2023effectiveness}. Tamil and English differ in lexical inventory,
morphology, phonotactics, prosody, and word order
\citep{krishnamurti2003dravidian,wals,Steever_2017}; because language, speakers,
recording conditions, and corpus change jointly, we interpret this as combined
cross-lingual/corpus-domain shift rather than language invariance. We evaluate
consistency, collapse, rate, retrieval, and positional inventory behavior. The
Tamil retrieval archive contains $\sim10.00$ h unique speech using the same
$3$ s/$1.5$ s protocol.

\paragraph{Training and paired-view construction.}
All controlled stages use $3.0$ s LibriSpeech segments and the same paired-view
pipeline. Longer utterances are cropped; shorter/boundary-adjacent examples are
padded only when required, with padded regions masked from conditioning,
tokenization, and loss computation. Silence-aware sampling avoids mostly
non-speech segments.

Each anchor $x_i$ is paired with
$x_i^{+}=\mathrm{Aug}(x_i)$ using conservative gain perturbation, additive
noise, filtering, reverberation, short time masking, and mild time stretching.
Noise/RIRs come from MUSAN ($\sim6$ h noise)
\citep{snyder2015musan}, ETSI ($\sim0.39$ h binaural environmental noise)
\citep{etsi202396}, and AIR measured room impulse responses
\citep{jeub2009binaural}. Time stretching may change positive duration, so
anchor and positive need not be frame-aligned; batching padding is masked
throughout.

All controlled stages use training-dataloader seed $240$, hence later stages
revisit the same deterministic sampled LibriSpeech-window stream rather than
independent unique-audio sets. Stage~I receives $\sim667$ sampled anchor-hours;
Stages~II and III each receive $\sim400$ h. These are optimization exposures,
not unique durations. Augmentations are generated online, so repeated speech
segments may receive different nuisance conditions.

The data stream is fixed across stages; only supervision changes. Stage~I
learns geometric VQ tokens, Stage~II learns an autoregressive bridge to frozen
VQ-derived strings, and Stage~III replaces them with adaptive EMA-teacher
sequences plus in-batch likelihood contrast. No frame alignment between paired
views is required.

\paragraph{Controlled and external baselines.}
Stage~I Geometric is the primary controlled baseline: the same encoder followed
by nearest-centroid VQ and consecutive deduplication. Stage~I+ is the
wav2tok-style intermediate, adding symmetric no-blank CTC sequence consistency
while retaining frame-synchronous geometric induction
\citep{banerjee2022wav2tok}. Both share PairAlign's preprocessing, encoder
family, source pool, VQ alphabet, and evaluation protocol and therefore isolate
the effect of autoregressive sequence induction and self-alignment.

HuBERT+VQ and WavLM+VQ are deliberately unmatched external references
\citep{hsu2021hubert,chen2022wavlm}. We discretize fixed layer-\(6\)
representations with \(K=512\) KMeans and consecutive deduplication. For
HuBERT, layer~6 is a principled clustering layer: first-iteration HuBERT-BASE
features peak around the middle layers, near layer~6, while the final layers
degrade in clustering quality; layer~6 is also used in HuBERT target refinement.
For WavLM, we use the same absolute intermediate layer index (6) as a standardized intermediate
extraction point, avoiding model-specific layer tuning and keeping the two SSL
references comparable at the representation-selection level. We set \(K=512\)
to match PairAlign's nominal alphabet and remain close to the
\(\sim500\)-cluster regime used in HuBERT-style acoustic-unit discovery.

HuBERT is pretrained by masked prediction of cluster-derived acoustic units,
initially obtained from MFCC KMeans and subsequently refined from learned
intermediate representations; WavLM inherits HuBERT-style clustered targets
and adds denoising-oriented pretraining. Their much greater capacity, exposure,
and explicit speech-unit-biased target construction make them stress-test
references rather than compute- or data-matched baselines.
\paragraph{Paired-view consistency and repeated deterministic inference.}
For independently tokenized anchors and positives, we report unigram Jaccard,
normalized Levenshtein similarity, exact match, sequence length, active
vocabulary, and edit decomposition. Jaccard measures unordered reuse; edit
similarity additionally reflects order, substitution, insertion, deletion, and
length mismatch.

All controlled models use training seed $240$. For clean LibriSpeech-100 and
TIMIT, validation uses fixed deterministic seed $1234$, giving identical
anchor--positive views. Stage~I/Stage~I+ use evaluation seed $13$; PairAlign is
re-evaluated with seeds $13$, $97$, and $313$ on the same views. Final PairAlign
uses deterministic beam search without sampling. TIMIT outputs are identical
across repeats; on LibriSpeech only $4$ of $28{,}539$ examples differ in one
run, consistent with floating-point/GPU nondeterminism near beam ties.
Accordingly, these are repeated deterministic evaluations, not stochastic
decoding runs. Supporting edit/inventory analyses use evaluation seed $13$.

\paragraph{Held-out external noise.}
External-noise TIMIT uses \texttt{noise\_only} recordings from the Speech and
Noise Dataset\footnote{\label{fn:speech-noise-dataset}\url{https://www.kaggle.com/datasets/abdullahhaydarkadolu/speech-noise-dataset}},
which is disjoint from tokenizer training sources. Noise is RMS-normalized and
mixed at SNR $\mathcal{U}[15,30]$ dB with no additional training-style
augmentation; relational controls use the same noise realization and SNR where
required. Both validation-view and evaluation seeds are fixed to $13$. We
measure consistency, compactness, collapse, and relational geometry under this
held-out corruption.

\paragraph{Token inventory and collapse.}
Agreement is meaningful only if outputs remain input-dependent. We therefore
measure active/dead vocabulary, normalized entropy, effective vocabulary,
top-token concentration, native- and normalized-position usage, low-diversity
collapse, and exact cross-example collision. For a token sequence $\mathcal{T}$, we use the
unique-token ratio
\[
r_{\mathrm{uniq}}(\mathcal{T})
=
\frac{|\mathrm{unique}(\mathcal{T})|}{|\mathcal{T}|},
\]
and classify sequences with $r_{\mathrm{uniq}}\leq0.2$ as low-diversity. Collapsed-pair rate counts pairs where either
sequence satisfies this criterion; exact collision measures identical complete
strings assigned to distinct examples. These diagnostics separate useful
cross-view agreement from repetitive or many-to-one collapse.

\paragraph{Relational edit geometry.}
Because training optimizes shared-target conditional sequence likelihood rather than edit distance directly, we test the induced geometry explicitly. For each anchor $A$ and
content-preserving positive $P$, negatives $N$ are phone-disjoint,
trigram-disjoint, or \emph{rearranged-context}. Rearranged negatives permute
temporal chunks before the complete tokenizer, retaining constituent acoustic
material while changing temporal organization.

We report normalized $d(A,P)$, $d(A,N)$, absolute margin, and
negative-to-positive ratio $d(A,N)/d(A,P)$, and repeat the analysis under held-out external noise with
matched  noise realization and SNR for the positive and negative inputs. This tests whether the likelihood surrogate contracts
positives relative to meaningful negatives and whether generation remains
sensitive to acoustic temporal order. Evaluation uses seed $13$.

A separate held-out analysis after the component ablations examines how the
Stage~III shared-target and likelihood-contrast objectives relate to this local
relational geometry; its derivation and calibration protocol are given in
Appendix~\ref{sec:pairalign-local-theory}.

\paragraph{Augmented-source retrieval and comparison cost.}
We evaluate \emph{augmented-source retrieval} on TIMIT using $9{,}461$ clean
overlapping $3$ s archive windows with a $1.5$ s hop. The same $300$ source
segments are transformed into content-preserving augmented queries, while the
archive contains only the corresponding clean windows. Thus, the task measures
recovery of the matching clean source region under augmentation rather than
source-excluded or source-independent retrieval. Queries and archive windows are
tokenized independently and ranked by normalized Levenshtein distance; forced
alignment is used only to define relevance. Both the validation-dataloader and
evaluation-script seeds are fixed to $13$.

We report segment-overlap, phoneme-exact, and phoneme-relaxed relevance.
Phoneme-exact requires identical silence-filtered phone strings;
phoneme-relaxed permits normalized phone edit distance $\leq0.35$. Metrics are
Recall@$K$, MRR, and mean first relevant rank. Archive statistics include total
tokens, mean/range length, token rate, relative reduction, and serialized size.

Systems measurements separate representation construction from reuse.
Tokenizer throughput, wall time, CUDA-synchronized batch-amortized model
latency, end-to-end latency, and peak allocated memory are measured with batch
size $16$ on the same NVIDIA L4. Search measurements include the
$\sum_{q,a}|q||a|$ length-product proxy, ranking/full-stage wall time, mean
ranking latency, and comparisons/s. These quantify symbolic storage and
sequence-comparison cost, not codec bitrate.

\paragraph{Rate-controlled post-hoc compression.}
To test whether shortening alone accounts for PairAlign's operating point, we
apply deterministic integer BPE to frozen dense token streams. Stage~I BPE is
fitted on $25{,}000$ distinct $3$ s LibriSpeech training segments
($20.83$ h): $12{,}500$ clean and $12{,}500$ mildly augmented segments drawn
from disjoint sets. PairAlign's LibriSpeech-100 rate of $11.85$ tok/s is used
as a fixed target, while the number of BPE merges is determined entirely from
the LibriSpeech training corpus, yielding $122{,}820$ merges. Independent
HuBERT and WavLM BPE inventories are fitted on the same LibriSpeech training
data using the same fixed target, yielding $4{,}190$ and $9{,}895$ merges,
respectively. The resulting transforms are then frozen and evaluated unchanged
on LibriSpeech-100 and TIMIT.

We deliberately avoid selecting a separate number of BPE merges for each
evaluation corpus. Such per-dataset adjustment could force a closer numerical
rate match, but would make the post-hoc compressor depend on the evaluation
distribution. Our control instead asks whether a single BPE transformation,
fitted on LibriSpeech training data toward a fixed in-domain target rate,
preserves its compression level when applied unchanged. Because the effective
compression of BPE depends on the occurrence of mergeable token subsequences,
the same frozen inventory need not realize the same rate on a different sample
distribution. Consequently, the observed LibriSpeech-100 and TIMIT rates
reflect the behavior of the fixed post-hoc compressor rather than
evaluation-specific retuning.

Paired-view evaluation uses validation seed $1234$ and evaluation seed $13$;
retrieval and timing use seed $13$. We report token rate, Jaccard/edit
similarity, exact agreement, active vocabulary, retrieval, and measured
construction cost. Since BPE introduces composite symbols, its active
vocabulary is not capacity-matched to PairAlign. The control therefore tests
the specific hypothesis that deterministic post-hoc shortening can reproduce
PairAlign's compactness--consistency operating point under a fixed
rate-control procedure.

\paragraph{Continuous-sweep tokenization.}
Local symbolic evolution is probed by shifting a $3$ s TIMIT window in
$100$ ms increments; adjacent windows share $2.9$ s ($96.7\%$). Stage~I and
PairAlign are independently decoded for every window with both validation and
evaluation seeds fixed to $13$.

For adjacent windows we measure Jaccard/edit similarity, sequence length/rate,
$|\Delta L|$, length ratio, relative length change, edit distance,
substitution/insertion/deletion counts, and their normalized rates. The probe
does not assume shift-equivariant subsequence preservation; it distinguishes
absolute edit budgets from normalized changes in PairAlign's much shorter
sequence.

\paragraph{Learned termination.}
Training ratio $\rho=0.15$ bounds target-length support; with $301$ conditioning
frames, training targets are at most $44$ tokens. To distinguish learned EOS
from inference truncation, we remove the $\rho$ restriction, permit EOS at
every step, and extend the decoding horizon to $301$ steps. Forced-EOS rate and
termination-length distributions then test input-dependent termination within
the support encountered during training; absence of outputs beyond $44$ is
interpreted as learned support, not evidence for unconstrained length
generalization.

\paragraph{Post-hoc temporal recovery.}
PairAlign has no native timestamps. We therefore convert decoder--encoder
cross-attention to frame-to-token posteriors and recover monotone paths using
the procedure in \S~\ref{sec:timing-from-ar}, with and without the
post-hoc diagonal Beta prior. The prior changes neither token identity nor
order.

Internal diagnostics are expected/peak monotonicity violation, diagonal error,
and Viterbi log-score. Externally, recovered transitions are compared with
sample-level TIMIT \texttt{.PHN} boundaries on $4{,}620$ clean $3$ s segments,
reporting boundary precision/recall/F1 at multiple tolerances and
matched-boundary localization error with paired bootstrap intervals. Both
validation and evaluation seeds are fixed to $13$. Phone boundaries are
temporal references only; PairAlign is not trained to emit phones.

\paragraph{Phonetic, boundary, silence, and speaker probes.}
TIMIT annotations characterize information retained after converting dense
Stage~I tokens to PairAlign's lower-rate sequence. We measure boundary
precision/recall, tokens per phone, phone/token segment purity, phone purity,
PNMI, leave-one-speaker-out token-to-phone prediction, discrete ABX, and
silence/non-silence association. Unigram/bigram/trigram probes test whether
phonetic information is redistributed into short compositions.

Speaker structure is measured through unconditional and phone-conditioned
speaker association and lightweight unigram--trigram speaker identification and
verification probes. These measure recoverability, not privacy: PairAlign has
no anonymization, speaker-adversarial, differential-privacy, or explicit
speaker-information objective. Timing, phonetic, boundary, and speaker analyses
use fixed validation and evaluation seed $13$.

\paragraph{Lexically disjoint Tamil transfer.}
Tamil evaluates transfer of the complete learned similarity relation, not
word-level correspondence. Because language and corpus change jointly, observed
differences are not attributed uniquely to linguistic structure. Paired-view
validation uses deterministic seed $1234$; Stage~I/Stage~I+ use evaluation seed
$13$, while PairAlign uses seeds $13$, $97$, and $313$ on the same views.
Tamil retrieval uses seed $13$ for both validation and evaluation. We report
consistency, collapse, active vocabulary, rate, retrieval, and native-position
entropy shift relative to LibriSpeech.

\paragraph{Ablations.}
Ablations isolate adaptive target learning, decoder grounding, prediction
difficulty, and collapse prevention. \emph{Stage~II AR Prior} evaluates the
fixed-teacher bridge before EMA self-alignment. Removing structured
self-attention dropout tests decoder-history bypass; removing prefix masking
tests harder partial-prefix prediction; removing encoder-summary conditioning
removes the persistent global acoustic pathway in both Stages~II and III while
retaining cross-attention; removing in-batch likelihood contrast removes only
negative-sequence discrimination while retaining entropy regularization and the
remaining Stage~III stabilizers.

The distinction between entropy and contrast is important: entropy encourages
broad token usage but does not require different inputs to receive different
sequence-level codes, whereas likelihood contrast explicitly requires paired
sequences to outrank mismatched sequences under the same acoustic condition.
Clean consistency ablations use validation seed $1234$ and evaluation seed
$13$; retrieval ablations use seed $13$ for both. Agreement is reported jointly
with active vocabulary, collapse, exact collision, retrieval, and archive rate.

\paragraph{Architecture, optimization, and reproducibility.}
All controlled PairAlign models use training seed $240$. Audio is sampled at
$16$ kHz and represented by $96$-bin log-Mel features with $25$ ms window and
$10$ ms hop. Stage~I uses a Mamba2 encoder with $512$-dimensional output and
$512$-entry nearest-centroid codebook. Stages~II--III use a Whisper-style
$6$-layer, $8$-head autoregressive Transformer with hidden size $512$ and
$516$ decoder symbols: $512$ VQ entries plus
$\mathrm{BOS}$/$\mathrm{EOS}$/$\mathrm{PAD}$/$\mathrm{MASK}$.

Stage~I optimizes
$\mathcal{L}_{\mathrm{contrast}}+
\lambda_{\mathrm{commit}}\mathcal{L}_{\mathrm{commit}}$
(\S~\ref{sec:stage1-full}); codebook centroids are updated separately by
EMA assignments. Stage~I uses $10\times5000$ steps, batch size $16$
($\sim667$ sampled anchor-hours). Stage~II uses $6\times5000$ steps at the same
batch size ($\sim400$ h); Stage~III uses the same duration while updating the
full encoder--decoder, with encoder LR one tenth of decoder LR.

Optimization uses Adam, base/decoder LR $5\times10^{-5}$, Stage~III encoder LR
$5\times10^{-6}$, weight decay $10^{-4}$, and LR scheduling. The EMA teacher is
updated every $1000$ steps with decay $0.995$. Teacher targets use scheduled
top-$p$ sampling with $\rho=0.15$; final inference uses deterministic beam
search with beam size $4$, $\rho=0.15$, minimum cap $2$, repetition penalty
$1.30$, and no length-normalization correction. With batch size (B=16), each Stage~III conditioning row has (B-1=15) in-batch distractor candidates, and all (K=15) are used as negatives in the likelihood-contrast loss. Full architecture, schedules,
sampling, seeds, and protocols are given in
Appendix~\ref{app:architecture-training}.

\paragraph{How to read the experiments.}
The evaluation separates stability, discrimination, granularity, and cost.
Consistency and positive--negative geometry test sequence-level organization;
inventory and collapse diagnostics verify input dependence; rate-controlled BPE
tests shortening as an alternative explanation; retrieval and systems metrics
measure utility and reuse cost; continuous sweeps expose local persistence;
unconstrained decoding tests learned EOS; timing and acoustic--linguistic probes
characterize temporal, phonetic, and speaker structure; and external noise,
Tamil, and high-exposure SSL references delimit robustness and scope. After the
component ablations, we additionally probe how shared-target contraction and
likelihood contrast relate to the local relational geometry observed in the
trained Stage~III model.

Accordingly, the central empirical claim is not uniform dominance. PairAlign
occupies a substantially lower-rate, input-conditioned, non-degenerate
sequence-level operating point that preserves meaningful ordered and relational
structure, while sacrificing part of the dense local acoustic and retrieval
resolution retained by geometric and pretrained frame-synchronous tokenizers.

\section{Results and Discussion}

\subsection{Probing Consistency, Collapse, and Token-Inventory Usage in Learned Token Sequences}
\label{sec:token_consistency_results}

\begin{table*}[t]
\centering
\scriptsize
\setlength{\tabcolsep}{4pt}
\begin{tabular}{llccccc}
\toprule
\multirow{3}{*}{\textbf{Dataset}}
& \multirow{3}{*}{\textbf{Model}}
& \textbf{Jaccard}
& \textbf{Edit}
& \textbf{Exact}
& \textbf{Mean}
& \textbf{Active} \\
& &
\textbf{Similarity}
& \textbf{Similarity}
& \textbf{Match Rate}
& \textbf{Sequence}
& \textbf{Vocabulary} \\
& &
&
&
&
\textbf{Length}
& \textbf{Size} \\
\midrule

\multirow{3}{*}{LibriSpeech-100}
& Stage I Geometric
  & $0.718$
  & $0.609$
  & $0.264$
  & $92.09$
  & $512$ \\

& Stage I+ Geometric
  & $0.738$
  & $0.629$
  & $0.265$
  & $75.61$
  & $512$ \\

& PairAlign
  & $0.719 \pm 9.7{\times}10^{-7}$
  & $0.629 \pm 2.7{\times}10^{-6}$
  & $0.291 \pm 0.000$
  & $35.55 \pm 3.0{\times}10^{-5}$
  & $512 \pm 0.0$ \\
\midrule

\multirow{3}{*}{TIMIT}
& Stage I Geometric
  & $0.742$
  & $0.616$
  & $0.267$
  & $78.65$
  & $456$ \\

& Stage I+ Geometric
  & $0.750$
  & $0.643$
  & $0.267$
  & $58.79$
  & $420$ \\

& PairAlign
  & $0.747 \pm 0.0$
  & $0.687 \pm 0.0$
  & $0.302 \pm 0.0$
  & $26.10 \pm 0.0$
  & $440 \pm 0.0$ \\

\midrule
TIMIT
& Stage I Geometric
  & $0.616$
  & $0.459$
  & $0.000$
  & $87.27$
  & $502$ \\

+ & Stage I+ Geometric
  & $0.687$
  & $0.530$
  & $0.000$
  & $69.81$
  & $505$ \\

External Noise & PairAlign
  & $0.651$
  & $0.554$
  & $0.017$
  & $34.58$
  & $508$ \\

\bottomrule
\end{tabular}

\caption{
Discrete token consistency and compactness on LibriSpeech-100, TIMIT, and
TIMIT with external real noise. Evaluation uses $28{,}539$ anchor--positive
pairs for LibriSpeech-100 and $4{,}620$ for each TIMIT condition. Each model
independently tokenizes a $3$-second anchor and a content-preserving positive.
For external noise, TIMIT segments are mixed with RMS-normalized
\texttt{noise\_only} recordings from the Speech and Noise
Dataset\textsuperscript{\ref{fn:speech-noise-dataset}}, disjoint from tokenizer
training data, at SNR uniformly sampled from $15$--$30$ dB and without
additional augmentation. Jaccard Similarity is unordered unigram-set overlap;
Edit Similarity is normalized Levenshtein similarity; Exact Match Rate is the
fraction of identical anchor--positive strings; Mean Sequence Length averages
anchor and positive lengths; Active Vocabulary Size counts codebook entries
used across both streams. All models are trained with seed $240$. For
LibriSpeech-100 and clean TIMIT, the validation dataloader uses fixed seed
$1234$; Stage~I and Stage~I+ use evaluation seed $13$, while PairAlign uses
seeds $13$, $97$, and $313$ on the same fixed views. For TIMIT + External
Noise, both the validation-view and evaluation seeds are fixed to $13$.
}
\label{tab:token_space_comparison}
\end{table*}
We first test whether the tokenizers produce stable, compact, and
non-degenerate symbolic sequences under content-preserving perturbations. Each
anchor is a $3$ s segment and its positive is an augmented view of the same
segment, so agreement measures nuisance robustness at fixed acoustic content.
We compare Stage~I Geometric, Stage~I+ Geometric, and PairAlign on
LibriSpeech-100 and TIMIT. Table~\ref{tab:token_space_comparison} reports
paired-view consistency, compactness, and active vocabulary; 
Table~\ref{tab:collapse_behavior} reports low-diversity and exact-collision
collapse.

All models are trained with seed $240$. For clean LibriSpeech-100 and TIMIT,
the validation dataloader uses fixed seed $1234$, giving identical
anchor--positive views across evaluations. Stage~I and Stage~I+ use evaluation
seed $13$; PairAlign uses seeds $13$, $97$, and $313$ on those same views.
Stage~I and Stage~I+ use deterministic nearest-codebook assignment followed by
consecutive-token deduplication; PairAlign uses deterministic beam search
without sampling. For TIMIT + External Noise, both the validation-view and
evaluation seeds are fixed to $13$. Appendix~\ref{app:token-inventory-diagnostics}
reports edit-operation, vocabulary-usage, native-position entropy, and
relative-position entropy diagnostics using evaluation seed $13$.

\paragraph{Repeated-inference variability is negligible.}
With fixed views and deterministic decoding, PairAlign should reproduce nearly
identical outputs across runs. It does: TIMIT tokenizations and aggregate
metrics are identical for evaluation seeds $13$, $97$, and $313$.
On LibriSpeech-100, run~2 differs from runs~1 and~3 on only $4$ of
$28{,}539$ examples, yielding deviations of only
$9.7\times10^{-7}$ in Jaccard similarity and
$2.7\times10^{-6}$ in edit similarity. Because neither views nor decoding
rules change, these four cases are consistent with low-level floating-point or
GPU-kernel nondeterminism near beam-search ties. The reported deviations
therefore reflect negligible computational variability, not uncertainty from
different evaluation-view draws.

\paragraph{Stage~I+ is a strong geometric intermediate.}
Stage~I+ improves Stage~I while preserving frame-synchronous token induction:
edit similarity rises from $0.609$ to $0.629$ on LibriSpeech-100 and from
$0.616$ to $0.643$ on TIMIT, while mean length falls from $92.09$ to $75.61$
and from $78.65$ to $58.79$. It is therefore a substantive intermediate,
already adding order-sensitive pairwise pressure to deduplicated frame-derived
assignments. PairAlign tests the stronger hypothesis that the symbolic
trajectory itself should be generated and self-aligned at sequence level.

\paragraph{PairAlign moves to a lower-rate regime while preserving ordered consistency.}
PairAlign reduces mean length to $35.55$ tokens on LibriSpeech-100 and $26.10$
on TIMIT: approximately $61\%$/$67\%$ below Stage~I and $53\%$/$56\%$ below
Stage~I+, respectively. Despite this compression, it matches Stage~I+ edit
similarity on LibriSpeech-100 ($0.629$) and improves it on TIMIT
($0.687$ versus $0.643$), while exact-match rate increases to $0.291$ and
$0.302$.

Jaccard must be read at this changed granularity. The denser geometric streams
provide more opportunities for shared token identities, whereas each token
change affects a larger fraction of PairAlign's shorter set. The later
$100$ ms continuous-context sweep shows the same pattern: dense geometric
tokens retain stronger normalized local overlap. The relevant result is thus
strong order-sensitive consistency at substantially lower sequence rate, not
uniform superiority on every overlap measure.

\begin{table*}[t]
\centering
\scriptsize
\setlength{\tabcolsep}{3.5pt}
\begin{tabular}{llccccc}
\toprule
\multirow{5}{*}{\textbf{Dataset}}
& \multirow{5}{*}{\textbf{Model}}
& \textbf{Anchor}
& \textbf{Positive}
& \textbf{Collapsed}
& \textbf{Anchor}
& \textbf{Positive} \\
& &
\textbf{Low-}
& \textbf{Low-}
& \textbf{Pair}
& \textbf{Exact-}
& \textbf{Exact-} \\
& &
\textbf{Diversity}
& \textbf{Diversity}
& \textbf{Rate}
& \textbf{Collision}
& \textbf{Collision} \\
& &
\textbf{Sequence}
& \textbf{Sequence}
&
& \textbf{Rate}
& \textbf{Rate} \\
& &
\textbf{Rate}
& \textbf{Rate}
&
&
\\
\midrule

\multirow{3}{*}{LibriSpeech-100}
& Stage I Geometric
  & $0.0269$
  & $0.0381$
  & $0.0500$
  & $0.0000$
  & $0.0000$ \\

& Stage I+ Geometric
  & $0.0263$
  & $0.0329$
  & $0.0450$
  & $0.0000$
  & $0.0000$ \\

& PairAlign
  & $0.0$
  & $0.0$
  & $0.0$
  & $0.00004 \pm 0.0$
  & $0.00007 \pm 0.0$ \\

\midrule

\multirow{3}{*}{TIMIT}
& Stage I Geometric
  & $0.0240$
  & $0.0719$
  & $0.0803$
  & $0.0000$
  & $0.0000$ \\

& Stage I+ Geometric
  & $0.0136$
  & $0.0223$
  & $0.0290$
  & $0.0000$
  & $0.0000$ \\

& PairAlign
  & $0.0 \pm 0.0$
  & $0.0 \pm 0.0$
  & $0.0 \pm 0.0$
  & $0.0 \pm 0.0$
  & $0.0022 \pm 0.0$ \\
\midrule

\multirow{3}{*}{TIMIT + External Noise}
& Stage I Geometric
  & $0.0219$
  & $0.0738$
  & $0.0833$
  & $0.0000$
  & $0.0000$ \\

& Stage I+ Geometric
  & $0.0177$
  & $0.0437$
  & $0.0532$
  & $0.0000$
  & $0.0000$ \\

& PairAlign
  & $0.0000$
  & $0.0000$
  & $0.0000$
  & $0.0000$
  & $0.0000$ \\

\bottomrule
\end{tabular}

\caption{
Collapse diagnostics over $28{,}539$ LibriSpeech-100 and $4{,}620$ pairs per
TIMIT condition. Low-Diversity Sequence Rate is the fraction with unique-token
ratio $\leq0.2$; Collapsed Pair Rate counts pairs where either sequence is
low-diversity; Exact-Collision Rate is one minus the fraction of unique
sequence-level tokenizations within each stream. Thus, low diversity measures
internal repetitive degeneration, while exact collision measures many-to-one
assignment across examples. External-noise TIMIT uses RMS-normalized
\texttt{noise\_only} recordings from the Speech and Noise
Dataset\textsuperscript{\ref{fn:speech-noise-dataset}}, mixed at uniformly
sampled $15$--$30$ dB SNR without additional augmentation. All models are
trained with seed $240$. For LibriSpeech-100 and clean TIMIT, validation uses
fixed seed $1234$; Stage~I and Stage~I+ use evaluation seed $13$, and PairAlign
uses seeds $13$, $97$, and $313$ on the same views. For TIMIT + External
Noise, both validation-view and evaluation seeds are fixed to $13$.
}
\label{tab:collapse_behavior}
\end{table*}

\paragraph{Higher exact agreement is not explained by collapse.}
PairAlign's higher exact-match rates ($0.291$ on LibriSpeech-100 and $0.302$ on
TIMIT, versus $0.264$--$0.267$ geometrically) do not arise from repetitive
strings. At $r_{\mathrm{uniq}}\leq0.2$, its collapsed-pair rate is zero on both
datasets, versus $0.0500/0.0803$ for Stage~I and $0.0450/0.0290$ for
Stage~I+. Exact-collision rates are likewise near zero:
$0.00004/0.00007$ for LibriSpeech-100 anchor/positive streams and
$0.0000/0.0022$ on TIMIT. The latter reflects only rare positive-stream
duplicates. Thus, increased paired-view exact agreement is not accompanied by
broad low-diversity or many-to-one collapse.

\paragraph{Inventory and edit diagnostics rule out trivial compactness.}
Appendix~\ref{app:token-inventory-diagnostics}, evaluated with seed $13$,
tests whether PairAlign obtains its lower rate through unstable edits, restricted
inventory use, or generic positional structure. Mean anchor--positive edit
distance falls from $39.25/31.17$ for Stage~I/Stage~I+ to $14.58$ for
PairAlign on LibriSpeech-100, and from $33.72/23.90$ to $9.54$ on TIMIT. On
TIMIT, substitutions fall from $18.36/11.81$ to $4.37$, insertions from
$8.57/5.72$ to $1.77$, and deletions from $6.79/6.36$ to $3.39$. Because
PairAlign is much shorter, these absolute counts do not imply uniformly
stronger normalized invariance; they show instead that paired compact strings
remain related by small absolute edit scripts, consistent with the later
continuous sweep where fewer absolute edits can still modify a larger fraction
of the shorter sequence.

Inventory usage also remains broad. All models activate all $512$ symbols on
LibriSpeech-100; PairAlign has slightly lower normalized entropy, as expected
at lower autoregressive rate, but top-$10$ mass remains below $0.09$. On
TIMIT, PairAlign has normalized entropy $0.813$ and effective vocabulary
$\approx160$, versus $0.786/135$ for Stage~I and $0.766/119$ for Stage~I+.
Native- and relative-position entropy further show an early entropy ramp rather
than persistent prefix collapse: all models begin with constrained usage that
broadens over the trajectory, with PairAlign showing the strongest initial
constraint from $\mathrm{BOS}$ and autoregressive history but distributed
usage across normalized beginning, middle, and end positions. Combined with
broad inventory use, zero low-diversity collapse, and near-zero collisions,
these results argue against generic-prefix or code-collapse explanations of
PairAlign's compactness.

\paragraph{Held-out external noise preserves the compactness--consistency trade-off.}
TIMIT + External Noise uses a corruption source disjoint from tokenizer
training, with both validation-view and evaluation seeds fixed to $13$. All
methods degrade, but PairAlign remains substantially shorter
($34.58$ tokens versus $87.27$ for Stage~I and $69.81$ for Stage~I+) while
achieving edit similarity $0.554$, versus $0.459$ and $0.530$. Its Jaccard
similarity remains below Stage~I+ ($0.651$ versus $0.687$), again separating
dense unigram overlap from lower-rate ordered consistency. Exact agreement is
rare but non-zero for PairAlign ($0.017$ versus $0.000$ for both geometric
variants), with no measured low-diversity or exact-collision collapse. Thus,
the compactness--consistency operating point persists under this held-out
corruption, without implying general noise robustness from a single condition.

\paragraph{The likelihood surrogate induces the intended relational edit geometry.}
Appendix~\ref{app:sequence-geometry-diagnostics},
Table~\ref{tab:pairalign_timit_edit_geometry}, directly tests the
positive-versus-negative edit relation optimized indirectly by shared-target conditional likelihood and likelihood contrast. Under the original content-preserving
positive condition, PairAlign gives the smallest $d(A,P)$ for all three
negative constructions and the largest negative-to-positive ratio:
$2.782\times$ for phone-disjoint, $2.819\times$ for trigram-disjoint, and
$2.568\times$ for rearranged-context negatives. PairAlign need not maximize
absolute negative distance because denser streams can accumulate larger
normalized separation; the relevant criterion is contraction of positives
relative to negatives. PairAlign has the largest ratio in all three settings
and the largest absolute margin for trigram-disjoint and rearranged-context
negatives, supporting the intended relational geometry rather than mere
sequence shortening.

\paragraph{Rearranged acoustic inputs directly probe input grounding.}
The rearranged-context test in
Appendix~\ref{app:sequence-geometry-diagnostics},
Table~\ref{tab:pairalign_timit_edit_geometry}, permutes $12$ temporal chunks
of the input before the complete tokenizer, preserving acoustic material while
disrupting its temporal organization. PairAlign separates these negatives from
content-preserving positives with margin $0.484$, versus $0.342$ for Stage~I
and $0.355$ for Stage~I+, and raises the negative-to-positive ratio from
$1.890\times$/$1.997\times$ to $2.568\times$. Because the intervention is
applied directly to the conditioning audio, the generated sequence remains
sensitive to input temporal organization rather than being dominated by its
autoregressive prior.

\paragraph{The relational separation persists under held-out external noise.}
The same ordering holds under external noise
(Appendix~\ref{app:sequence-geometry-diagnostics},
Table~\ref{tab:pairalign_timit_edit_geometry}). Positive and negative inputs
receive the same noise recording at the same SNR, ruling out corruption mismatch
as the source of separation. PairAlign gives the largest negative-to-positive
ratios for phone-disjoint, trigram-disjoint, and rearranged-context negatives
($2.003\times$, $2.088\times$, and $1.968\times$), versus
$1.615\times$--$1.777\times$ for Stage~I and
$1.711\times$--$2.023\times$ for Stage~I+, and also the largest margin in all
three cases. The likelihood-induced relational geometry therefore persists
under the held-out corruption source.

\paragraph{Termination is model-predicted within the training-supported length range.}
Appendix~\ref{app:termination-diagnostics},
Figure~\ref{fig:unconstrained-token-length-distributions}, and
Table~\ref{tab:unconstrained-termination} disentangle the training-time rate
constraint from inference-time stopping. During training, \(\rho=0.15\) only
bounds teacher-target length to at most \(44\) tokens for a \(301\)-frame
conditioning sequence; it is not a length loss and does not impose a fixed
inference length.
At evaluation, we remove the \(\rho\)-based cap, allow EOS at every decoding
step, and extend the horizon to \(301\) steps. The forced-EOS rate is \(0\%\)
on both datasets. TIMIT anchor/positive means are \(27.01/25.40\), with
\(84.00\%/87.01\%\) terminating before \(44\); LibriSpeech-100 means are
\(35.97/35.13\), with \(49.73\%/53.66\%\) terminating before \(44\).
The resulting distributions are not concentrated solely at the training
maximum: TIMIT assigns substantial mass to shorter lengths, whereas
LibriSpeech-100 exhibits a broader intermediate-length distribution together
with a pronounced mode at \(44\). Observed lengths extend down to \(6\) tokens.
These results show that \(44\) is the maximum target length supported during
training, not an inference-time stopping boundary. Within this support,
PairAlign predicts input-dependent EOS positions without requiring forced
termination.

\paragraph{Exact agreement is not attributable to output brevity.}
Building on the termination analysis above,
Figure~\ref{fig:exact-match-token-length-distributions} and
Table~\ref{tab:unconstrained-termination} directly test whether exact matches
arise preferentially from short sequences. They do not. Table~\ref{tab:unconstrained-termination}
shows that TIMIT exact matches span \(6\)--\(44\) tokens with mean \(26.41\),
while LibriSpeech-100 exact matches have mean \(35.30\); \(84.18\%\) and
\(51.54\%\), respectively, terminate before \(44\). Figure~\ref{fig:exact-match-token-length-distributions}
further shows that these exact-match length distributions closely mirror the
corresponding unconstrained output distributions, retaining substantial mass
across intermediate lengths and the same pronounced mode at \(44\), rather
than shifting toward shorter outputs.
Thus, exact agreement occurs throughout the learned length range, including at
the maximum supported length. The length-resolved statistics and distributions
rule out the explanation that PairAlign's higher exact-match rate is simply a
consequence of output brevity.

\paragraph{Overall interpretation.}
Together, Tables~\ref{tab:token_space_comparison}
and~\ref{tab:collapse_behavior} place PairAlign at a different operating point
from the geometric tokenizers. Stage~I and Stage~I+ retain denser local
redundancy, benefiting normalized local overlap and top-rank retrieval.
PairAlign instead produces substantially shorter sequences while preserving or
improving order-sensitive paired-view consistency, retaining broad vocabulary
usage, and exhibiting essentially no low-diversity collapse.

The supporting diagnostics show that this compactness remains structured:
PairAlign contracts positives relative to phone-disjoint, trigram-disjoint, and
temporally rearranged negatives, attains the strongest negative-to-positive
ratios across all three constructions, and remains sensitive to temporal
organization of the conditioning audio. Together with the later retrieval and
continuous-sweep analyses, this identifies a granularity trade-off rather than
uniform dominance: PairAlign sacrifices some dense local redundancy while
retaining meaningful ordered and relational structure at much lower rate.

The unconstrained-decoding analysis further shows that $\rho$ bounds training
support rather than fixing inference length. For $301$ conditioning frames,
training targets are at most $44$ tokens; after removing the inference-time
restriction and permitting up to $301$ decoding steps, forced-EOS remains
$0\%$, all terminations are model-predicted, and outputs span $6$--$44$ tokens
without extending beyond the maximum training target. PairAlign therefore
learns input-dependent termination within its learned rate support.

Overall, PairAlign is best characterized as a compact, input-conditioned,
variable-length symbolic interface: not a uniformly finer local representation,
but a substantially coarser token trajectory that remains consistent across
content-preserving views, separated from meaningful negatives, grounded in the
conditioning audio, and non-degenerate despite the reduction in symbolic rate.

\subsection{Probing Augmented-Source Edit-Distance Retrieval Under Geometric and Self-Aligned Tokenization}
\label{sec:retrieval-geom-vs-pairalign}

\begin{table*}[t]
\centering
\scriptsize
\setlength{\tabcolsep}{3.5pt}

\begin{tabular}{llcccccc}
\toprule
\textbf{Relevance definition}
& \textbf{Model}
& \textbf{R@1}
& \textbf{R@5}
& \textbf{R@10}
& \textbf{R@20}
& \textbf{MRR}
& \textbf{Mean FRR} \\
\midrule

\multirow{3}{*}{Segment overlap}
& Stage~I Geometric
    & $0.75$ & $0.83$ & $0.86$ & $0.88$ & $0.79$ & $36.45$ \\
& Stage~I+ Geometric
    & $0.68$ & $0.75$ & $0.79$ & $0.81$ & $0.72$ & $45.86$ \\
& PairAlign
    & $0.71$ & $0.78$ & $0.81$ & $0.83$ & $0.74$ & $44.67$ \\

\midrule

\multirow{3}{*}{Phoneme exact}
& Stage~I Geometric
    & $0.75$ & $0.83$ & $0.85$ & $0.88$ & $0.79$ & $49.02$ \\
& Stage~I+ Geometric
    & $0.68$ & $0.75$ & $0.79$ & $0.81$ & $0.72$ & $69.69$ \\
& PairAlign
    & $0.71$ & $0.78$ & $0.80$ & $0.83$ & $0.74$ & $61.60$ \\

\midrule

\multirow{3}{*}{Phoneme relaxed}
& Stage~I Geometric
    & $0.76$ & $0.83$ & $0.85$ & $0.88$ & $0.79$ & $35.19$ \\
& Stage~I+ Geometric
    & $0.69$ & $0.75$ & $0.79$ & $0.82$ & $0.72$ & $53.76$ \\
& PairAlign
    & $0.71$ & $0.78$ & $0.81$ & $0.84$ & $0.74$ & $39.54$ \\

\bottomrule
\end{tabular}

\caption{
TIMIT retrieval on $3$ s segments using normalized Levenshtein distance between
augmented query tokenizations and the clean archive. Segment-overlap relevance
accepts temporally overlapping windows; phoneme-exact requires identical
non-silence phone sequences; phoneme-relaxed permits normalized phone edit
distance $\leq0.35$. Stage~I+ denotes wav2tok. All models use the same $300$
query sources, with validation-dataloader and evaluation-script seeds fixed to
$13$. HitRate is $1.0$ for all systems because the closed-set archive contains
a relevant source segment for every query.
}
\label{tab:retrieval_geom_vs_pairalign}
\end{table*}
\paragraph{Retrieval setup.}
We evaluate whether PairAlign's lower-rate symbolic space retains useful
cross-corpus retrieval structure. All tokenizers are trained on LibriSpeech and
evaluated on the same clean TIMIT archive of $9{,}461$ overlapping $3$ s
segments extracted with a $1.5$ s hop. The same $300$ archive segments provide
query sources for all models; their content-preserving augmented views are
absent from the archive. Both validation-dataloader and evaluation-script seeds
are fixed to $13$, fixing archive construction, query selection, and query
augmentations across tokenizers. Ranking uses normalized Levenshtein distance
in token space. Table~\ref{tab:retrieval_geom_vs_pairalign} reports retrieval,
Table~\ref{tab:TIMIT_archive_compactness} archive compactness,
Table~\ref{tab:TIMIT_tokenizer_efficiency} tokenization cost, and
Table~\ref{tab:TIMIT_edit_distance_efficiency} search cost. Tokenization timing
uses the same NVIDIA L4 GPU with batch size $16$.

Archive-rate statistics use only clean, unaugmented archive tokenizations;
augmented queries are used for ranking but excluded from archive mean length and
rate. This differs from \S~\ref{sec:token_consistency_results}, where mean
length averages clean anchors and augmented positives, with validation seed
$1234$ and evaluation seed $13$. Retrieval instead uses seed $13$ for both.
Hence absolute lengths differ across protocols without any model or decoding
change: Stage~I changes from $78.65$ to $84.62$ tokens per $3$ s segment and
PairAlign from $26.10$ to $24.84$.

\paragraph{The comparison spans distinct symbolic rates.}
The systems intentionally occupy different granularities. Stage~I is a dense
deduplicated trace of frame-level nearest-codebook assignments; Stage~I+ adds
sequence-consistency pressure while retaining geometric induction; PairAlign
generates a much shorter autoregressive sequence by sequence-level
self-alignment. Rate therefore affects both accuracy and cost: denser streams
retain more local evidence for edit-distance discrimination, whereas shorter
streams reduce storage and sequence-length-dependent comparison cost. The
comparison is thus a rate--precision--computation trade-off, not a
rate-matched benchmark.

\begin{table}[t]
\centering
\scriptsize
\setlength{\tabcolsep}{3.5pt}
\begin{tabular}{lccc}
\toprule
\textbf{Archive statistic}
& \textbf{Stage I}
& \textbf{Stage I+}
& \textbf{PairAlign} \\
\midrule

Total tokens \(N_{\mathrm{tok}}\)
    & 800{,}609
    & 578{,}500
    & 234{,}998 \\

Average tokens/segment \(\overline{L}\)
    & 84.62
    & 61.15
    & 24.84 \\

Token rate \(R_{\mathrm{tok}}\) (tok/s)
    & 28.21
    & 20.38
    & 8.28 \\

Token-count compression
    & \(1.00{\times}\)
    & \(1.38{\times}\)
    & \(3.41{\times}\) \\

Relative token reduction
    & --
    & 27.74\%
    & 70.65\% \\

Minimum tokens/segment
    & 12
    & 8
    & 3 \\

Maximum tokens/segment
    & 154
    & 139
    & 44 \\

\bottomrule
\end{tabular}
\caption{
Compactness of the $9{,}461$ clean, unaugmented TIMIT archive segments
\citep{garofolo1993darpa}. Stage~I+ denotes wav2tok. Augmented queries are
excluded. Compression is relative to Stage~I; all vocabularies have
$\lvert\mathcal{A}\rvert=512$. Both validation-dataloader and evaluation-script
seeds are fixed to $13$. These values differ from paired-view consistency,
which uses validation seed $1234$, evaluation seed $13$, and averages clean and
augmented views; such differences reflect evaluation population and averaging,
not model changes.
}
\label{tab:TIMIT_archive_compactness}
\end{table}

\paragraph{Retrieval exhibits a rate--precision trade-off.}
Stage~I gives the strongest top-rank retrieval: under segment-overlap relevance,
R@1 is $0.75$, $0.68$, and $0.71$ and MRR is $0.79$, $0.72$, and $0.74$ for
Stage~I, Stage~I+, and PairAlign, with broadly the same ordering under both
phoneme criteria. PairAlign therefore does not surpass dense Stage~I on raw
retrieval precision.

This gap is consistent with rate. Stage~I provides edit distance with a dense
local trace, whereas PairAlign uses only $8.28$ tokens/s on the clean archive
versus $28.21$ for Stage~I, making each edit a larger fraction of its sequence.
The same granularity effect appears in the continuous-context sweep, where dense
geometric tokens preserve stronger normalized local overlap. Nevertheless,
PairAlign retains substantial retrieval structure and consistently exceeds
Stage~I+ at lower rate: segment-overlap R@20 rises from $0.81$ to $0.83$ and
phoneme-relaxed R@20 from $0.82$ to $0.84$. It therefore occupies a different
rate--precision operating point rather than optimizing accuracy independently
of representation size.

\paragraph{The main degradation is ranking precision, not retrieval coverage.}
All systems have HitRate $1.0$ because every closed-set query has a relevant
archive segment; the distinction is ranking sharpness. Mean FRR for
Stage~I/Stage~I+/PairAlign is $36.45/45.86/44.67$ for segment overlap,
$49.02/69.69/61.60$ for phoneme exact, and $35.19/53.76/39.54$ for phoneme
relaxed. PairAlign therefore has a heavier tail than Stage~I but improves over
Stage~I+ on the phoneme criteria; under relaxed relevance, its Mean FRR
($39.54$) is close to Stage~I ($35.19$) despite the rate gap. Thus, PairAlign
primarily sacrifices fine-grained ranking resolution rather than archive
coverage, consistent with a compact but non-collapsed representation having
fewer symbolic degrees of freedom for nearest-neighbor discrimination.

\paragraph{PairAlign substantially reduces archive representation size.}
Across the same $9{,}461$ archive segments, Stage~I, Stage~I+, and PairAlign
produce $800{,}609$, $578{,}500$, and $234{,}998$ tokens. PairAlign therefore
reduces token count by $70.65\%$ relative to Stage~I, a $3.41\times$
compression, while mean length falls from $84.62$ to $24.84$ tokens and rate
from $28.21$ to $8.28$ tokens/s. This is not explained by consecutive
deduplication, which both geometric baselines already apply: PairAlign changes
the representation granularity itself. Its archive lengths span $3$--$44$
tokens, versus $12$--$154$ for Stage~I and $8$--$139$ for Stage~I+.

\paragraph{Archive and paired-view rates use different evaluation populations.}
Retrieval and paired-view means should not be equated. Stage~I changes from
$78.65$ tokens per $3$ s segment in paired-view TIMIT to $84.62$ on the clean
archive; PairAlign changes from $26.10$ to $24.84$. Paired-view consistency
uses validation seed $1234$, evaluation seed $13$, and averages clean anchors
with augmented positives; retrieval uses seed $13$ for both and computes rate
only on the clean $9{,}461$-segment archive. Since sequence length is
input-dependent, the protocol change can shift means in either direction
without any tokenizer or decoding change. The qualitative ordering is
unchanged: Stage~I remains much denser than PairAlign ($78.65$ versus $26.10$
in paired-view evaluation; $84.62$ versus $24.84$ in retrieval).

\paragraph{Autoregressive generation dominates PairAlign tokenization cost.}
Stage~I and Stage~I+ tokenize the archive at $174.78$ and $173.23$ segments/s,
requiring $54.13$ and $54.62$ s, whereas PairAlign reaches $5.50$ segments/s
and requires $1719.47$ s. CUDA-synchronized model latency is
$2.26/1.57/194.27$ ms/query for Stage~I/Stage~I+/PairAlign; end-to-end latency
is $12.77/12.44/206.09$ ms/query, and peak memory is
$361.41/361.41/570.58$ MB. PairAlign's cost is therefore dominated by
autoregressive beam search: compact token sequences do not imply cheap token
generation, and their computational benefit appears after materialization.

\paragraph{Compact sequences substantially reduce edit-distance search cost.}
For the same $2{,}838{,}300$ query--archive comparisons, the length-product
proxy $\sum_{q,a}|q||a|$ decreases from $2.00\times10^{10}$ for Stage~I to
$1.05\times10^{10}$ for Stage~I+ and $1.62\times10^{9}$ for PairAlign, a
$91.87\%$ PairAlign reduction relative to Stage~I. Ranking wall time similarly
falls from $11.30$ to $8.25$ and $5.41$ s, respectively, giving PairAlign a
$52.16\%$ reduction. Mean ranking latency drops from $37.66$ to $18.02$
ms/query, throughput rises from about $251{,}194$ to $525{,}063$
comparisons/s, and full retrieval-stage time falls from $17.68$ to $11.55$ s.
The measured wall-time gain is smaller than the length-product reduction because
sorting, implementation, and other sequence-independent overheads remain.

\begin{table*}[t]
\centering
\scriptsize
\setlength{\tabcolsep}{3.5pt}
\begin{tabular}{lcccccc}
\toprule

\multirow{3}{*}{\textbf{Model}}
& \multirow{3}{*}{\textbf{GPU}}
& \multicolumn{2}{c}{\textbf{Archive tokenization}}
& \multicolumn{2}{c}{\textbf{Query tokenization}}
& \multirow{2}{*}{\textbf{Peak GPU}} \\

\cmidrule(lr){3-4}
\cmidrule(lr){5-6}

&
& \textbf{Throughput}
& \textbf{Wall time}
& \textbf{Model latency}
& \textbf{End-to-end}
& \textbf{memory} \\

&
& \textbf{segments/s}
& \textbf{s}
& \textbf{ms/query}
& \textbf{ms/query}
& \textbf{MB} \\

\midrule

Stage~I Geometric
    & NVIDIA L4
    & 174.78
    & 54.13
    & 2.26
    & 12.77
    & 361.41 \\

Stage~I+ Geometric
    & NVIDIA L4
    & 173.23
    & 54.62
    & 1.57
    & 12.44
    & 361.41 \\

PairAlign
    & NVIDIA L4
    & 5.50
    & 1719.47
    & 194.27
    & 206.09
    & 570.58 \\

\bottomrule
\end{tabular}
\caption{
TIMIT tokenizer efficiency with validation-dataloader and evaluation-script
seeds fixed to $13$, batch size $16$, and the same NVIDIATIMIT tokenizer efficiency with validation-dataloader and evaluation-script
seeds fixed to $13$, batch size $16$, and the same NVIDIA L4 GPU. Archive
throughput includes the full tokenization loop. Query Model latency is
CUDA-synchronized, batch-amortized \texttt{model.tokenize()} time over $300$
queries; PairAlign includes autoregressive beam search. End-to-end latency also
includes loading, feature extraction, transfer, and post-processing. Peak GPU
memory is maximum allocated CUDA memory during archive tokenization.
}
\label{tab:TIMIT_tokenizer_efficiency}
\end{table*}

\paragraph{Token generation and token use favor different operating regimes.}
Combining end-to-end query tokenization with ranking-only latency gives
approximately $50.43$ ms for Stage~I, $39.94$ ms for Stage~I+, and $224.11$ ms
for PairAlign. Thus, the current PairAlign implementation is slower for one-shot
online retrieval: cheaper edit-distance search does not offset autoregressive
decoding. Its advantage emerges when representations are computed once and
reused, where the $70.65\%$ archive-token reduction and $52.16\%$ ranking-time
reduction directly benefit storage and repeated comparisons. Faster or
parallel decoding would be needed to convert these representational savings
into lower latency for online retrieval.

\begin{table*}[t]
\centering
\scriptsize
\setlength{\tabcolsep}{4pt}
\begin{tabular}{lccc}
\toprule
\textbf{Retrieval statistic}
& \textbf{Stage I}
& \textbf{Stage I+}
& \textbf{PairAlign} \\
\midrule

Query--archive comparisons
    & 2{,}838{,}300
    & 2{,}838{,}300
    & 2{,}838{,}300 \\

Length-product proxy
\(\sum_{q,a}|q||a|\)
    & \(2.00{\times}10^{10}\)
    & \(1.05{\times}10^{10}\)
    & \(1.62{\times}10^{9}\) \\

Relative edit-distance work
    & \(1.00{\times}\)
    & \(0.52{\times}\)
    & \(0.08{\times}\) \\

Edit-distance work reduction
    & --
    & 47.56\%
    & 91.87\% \\

Ranking wall time
    & 11.30 s
    & 8.25 s
    & 5.41 s \\

Ranking-time reduction
    & --
    & 26.98\%
    & 52.16\% \\

Mean ranking latency/query
    & 37.66 ms
    & 27.50 ms
    & 18.02 ms \\

Full retrieval-stage time
    & 17.68 s
    & 14.47 s
    & 11.55 s \\

Comparisons/s
    & 251{,}193.64
    & 343{,}996.93
    & 525{,}063.40 \\

\bottomrule
\end{tabular}
\caption{
TIMIT edit-distance efficiency for $300$ queries against $9{,}461$ archive
segments. Stage~I+ denotes wav2tok. Validation-dataloader and evaluation-script
seeds are fixed to $13$. All systems perform the same $2{,}838{,}300$
comparisons. The length-product proxy captures the dominant
sequence-length dependence of dynamic-programming Levenshtein computation;
relative work is measured against Stage~I. Ranking wall time includes
token-distance computation and ranking; Full retrieval-stage time additionally
includes relevance computation. Comparisons/s uses ranking wall time. All
systems use the same RapidFuzz Levenshtein implementation and hardware.
}
\label{tab:TIMIT_edit_distance_efficiency}
\end{table*}

\paragraph{Interpretation.}
With a $1.5$ s hop, adjacent $3$ s archive windows replace half of their
acoustic context, so retrieval is not exact-token lookup. The four tables expose
a consistent granularity trade-off. Stage~I retains the densest symbolic trace
and strongest top-rank retrieval, but also the largest archive and highest
edit-distance cost. PairAlign instead reduces archive token count by $70.65\%$
and the edit-distance work proxy by $91.87\%$ while retaining meaningful
retrieval accuracy and complete closed-set coverage.

This is consistent with the preceding representation analysis: PairAlign
preserves positive-view consistency and nontrivial positive--negative geometry
but less fine-grained local redundancy than Stage~I. Normalized edit distance
therefore has fewer symbols with which to resolve nearby candidates, reducing
top-rank precision. The result is evidence that substantial symbolic
compression can preserve useful retrieval geometry, not that PairAlign
dominates Stage~I in accuracy.

Finally, representation construction and use have opposite cost profiles.
PairAlign's autoregressive beam search makes one-shot retrieval slower, whereas
its materialized sequences require substantially less storage and cheaper
repeated edit-distance comparison. Its operating point is therefore lower
symbolic rate and cheaper repeated search, traded against some fine-grained
retrieval precision and substantially higher token-generation cost.

\subsection{Probing Local Symbolic Compositionality Under Continuous Acoustic Context Shifts}
\label{sec:continuous_sweep_results}

\paragraph{Continuous-sweep setup.}
We probe local symbolic continuity under a $100$ ms acoustic-context shift,
complementing the preceding paired-view and retrieval experiments. On TIMIT, we
tokenize $3$ s windows with a $100$ ms hop, so adjacent windows share $2.9$ s
($96.7\%$) and differ only at their entering/exiting boundaries. This is a much
finer intervention than the $1.5$ s retrieval hop and tests how symbolic
sequences evolve under continuous context displacement.

We compare the endpoint representations: deduplicated Stage~I Geometric and
PairAlign. Both the validation dataloader and evaluation script use fixed seed
$13$, ensuring identical sweep windows and evaluation conditions. We report
adjacent-window overlap, absolute/relative length changes, and Levenshtein edit
decomposition in Tables~\ref{tab:continuous_sweep_summary},
\ref{tab:continuous_sweep_distribution}, and
\ref{tab:continuous_sweep_edit_ops}.

\begin{table}[t]
\centering
\scriptsize
\setlength{\tabcolsep}{5pt}
\begin{tabular}{lcc}
\toprule
\textbf{Metric} & \textbf{Geometric} & \textbf{PairAlign} \\
\midrule
Mean sequence length & 85.08 & 25.48 \\
Approx. tokens / s & 28.36 & 8.49 \\
Edit similarity & $0.536 \pm 0.217$ & $0.414 \pm 0.218$ \\
Unigram Jaccard & $0.595 \pm 0.227$ & $0.479 \pm 0.211$ \\
Adjacent length change $|\Delta L|$ & $10.30 \pm 11.18$ & $6.21 \pm 6.95$ \\
Length ratio & $0.888 \pm 0.111$ & $0.793 \pm 0.182$ \\
Relative length change & 12.1\% & 24.4\% \\
\bottomrule
\end{tabular}
\caption{
Continuous-sweep statistics on TIMIT \citep{garofolo1993darpa} for adjacent
$3$ s windows separated by $100$ ms and sharing $2.9$ s. Relative length
change is mean $|\Delta L|$ divided by mean sequence length. Geometric tokens
are consecutively deduplicated. Both validation-dataloader and evaluation-script
seeds are fixed to $13$.
}
\label{tab:continuous_sweep_summary}
\end{table}

\begin{table}[t]
\centering
\scriptsize
\setlength{\tabcolsep}{5pt}
\begin{tabular}{lcc}
\toprule
\textbf{Distributional statistic} & \textbf{Geometric} & \textbf{PairAlign} \\
\midrule
Median edit similarity & 0.596 & 0.409 \\
Edit similarity $\geq 0.5$ & 70.1\% & 36.9\% \\
Edit similarity $\geq 0.4$ & 80.3\% & 52.2\% \\
Median $|\Delta L|$ & 7 & 4 \\
\bottomrule
\end{tabular}
\caption{
Distributional adjacent-window similarity and length change. Both
validation-dataloader and evaluation-script seeds are fixed to $13$.
}
\label{tab:continuous_sweep_distribution}
\end{table}

\paragraph{The sweep recovers the same low-rate PairAlign operating regime.}
Stage~I produces $85.08$ tokens per $3$ s ($28.36$ tokens/s), versus
$25.48$ ($8.49$ tokens/s) for PairAlign. These closely match the independent
TIMIT retrieval rates of $28.21$ and $8.28$ tokens/s despite its $1.5$ s hop.
Thus, the approximately $3.3\times$ granularity gap persists across distinct
windowing protocols, and local-overlap metrics must be interpreted at these
different rates.

\paragraph{Dense geometric tokens preserve stronger fine-grained local overlap.}
Stage~I is more persistent under normalized local overlap: mean edit similarity
is $0.536$ versus $0.414$, Jaccard is $0.595$ versus $0.479$, and $70.1\%$
versus $36.9\%$ of transitions achieve edit similarity $\geq0.5$. This is
consistent with its dense frame-derived trace: highly overlapping windows retain
many local assignments, whereas each changed symbol occupies a larger fraction
of PairAlign's approximately three-times-shorter sequence. PairAlign therefore
does not provide stronger fine-grained local invariance; Stage~I is stronger
when normalized persistence under small context shifts is the objective.

\paragraph{PairAlign changes fewer symbols absolutely, but more relatively.}
PairAlign has smaller adjacent length changes (median $4$ versus $7$; mean
$6.21$ versus $10.30$) and smaller edit scripts (mean $17.48$ versus $42.57$;
median $15$ versus $36$). It also uses fewer substitutions
($9.22$ versus $23.22$), insertions ($3.91$ versus $9.56$), and deletions
($4.35$ versus $9.79$).

These lower absolute counts do not imply greater invariance. Because PairAlign
is much shorter, relative length change is higher ($24.4\%$ versus $12.1\%$),
as are all normalized edit-operation rates. A $100$ ms shift therefore changes
fewer PairAlign symbols in absolute number, but a larger fraction of its emitted
sequence.

\paragraph{The edit pattern is consistent with coarse, context-dependent sequence updates.}
PairAlign's edits are dominated by substitutions ($9.22$), with fewer
insertions ($3.91$) and deletions ($4.35$), consistent with the earlier
positive-view decomposition. The deterministic $100$ ms boundary intervention
therefore extends that result: PairAlign typically revises a compact subset of
sequence elements rather than maintaining a dense frame-level trace. This
complements, rather than overturns, the normalized-overlap result: Stage~I is
more persistent relatively, while PairAlign expresses the context change with a
smaller absolute symbolic edit budget.

\paragraph{Granularity-aware interpretation.}
Together with the consistency, sequence-geometry, retrieval, and rate-controlled
results, the sweep isolates PairAlign's rate--granularity trade-off. Earlier
experiments show strong paired-view edit consistency, meaningful
positive--negative geometry, and useful retrieval structure at substantially
lower rate. At the finer $100$ ms scale, however, Stage~I preserves more of its
normalized sequence, while PairAlign changes fewer symbols absolutely but a
larger fraction of its shorter sequence.

PairAlign should therefore not be described as uniformly more stable under
small context shifts. It trades dense frame-local persistence for a lower-rate
sequence-level interface: neighboring contexts remain related by relatively
small absolute edit scripts, but individual coarse symbols are more sensitive
in normalized sequence space. This behavior is consistent with the broader
rate--granularity trade-off and argues against viewing PairAlign either as a
dense frame-synchronous code or as an unrelated holistic identifier for each
$3$ s segment.

\begin{table}[t]
\centering
\scriptsize
\setlength{\tabcolsep}{4.5pt}
\begin{tabular}{lcc}
\toprule
\textbf{Edit-operation statistic} & \textbf{Geometric} & \textbf{PairAlign} \\
\midrule
Edit distance $ED$ & $42.57 \pm 24.06$ & $17.48 \pm 10.47$ \\
Median $ED$ & 36 & 15 \\
\midrule
Substitutions $S$ & $23.22 \pm 20.43$ & $9.22 \pm 7.32$ \\
Insertions $I$ & $9.56 \pm 8.84$ & $3.91 \pm 5.90$ \\
Deletions $D$ & $9.79 \pm 9.46$ & $4.35 \pm 5.64$ \\
\midrule
Substitution rate $r_S$ & $0.253 \pm 0.204$ & $0.314 \pm 0.197$ \\
Insertion rate $r_I$ & $0.104 \pm 0.088$ & $0.127 \pm 0.165$ \\
Deletion rate $r_D$ & $0.107 \pm 0.095$ & $0.145 \pm 0.157$ \\
\bottomrule
\end{tabular}
\caption{
Levenshtein decomposition between adjacent $100$ ms-shifted windows.
Operation rates are normalized by the maximum adjacent sequence length; because
PairAlign operates at much lower rate, absolute counts and normalized rates
must be interpreted jointly. Both validation-dataloader and evaluation-script
seeds are fixed to $13$.
}
\label{tab:continuous_sweep_edit_ops}
\end{table}
\subsection{Ablation Study: Stage Progression, Acoustic Conditioning, and Decoder-Bypass Controls}
\label{sec:pairalign_ablation}

The ablation study isolates the components required for compact,
input-dependent, paired-view-consistent, non-degenerate sequences. Stage~II
tests fixed-target autoregressive learning before adaptive self-alignment;
structured self-attention dropout tests decoder-side reliance on token history;
prefix masking tests harder partial-prefix prediction; encoder-summary
conditioning tests persistent global acoustic context; and in-batch likelihood
contrast tests explicit separation from alternative sequences.

For clean LibriSpeech-100 and TIMIT consistency, the validation dataloader uses
fixed deterministic seed $1234$ and the evaluation script seed $13$, giving all
systems identical anchor--positive views. Retrieval ablations use seed $13$ for
both validation and evaluation, matching
\S~\ref{sec:retrieval-geom-vs-pairalign}. Full PairAlign corresponds to
Table~\ref{tab:token_space_comparison}: Jaccard/edit/exact-match
$0.719/0.629/0.291$ on LibriSpeech-100 and $0.747/0.687/0.302$ on TIMIT.

\paragraph{Stage~II is a stable autoregressive bridge, not the final token space.}
Stage~II predicts deterministic Stage~I strings with the encoder and VQ
codebook frozen. It remains input-dependent---$506$ active symbols, zero
low-diversity collapse, and zero exact collisions---but underperforms Stage~I+:
Jaccard/edit/exact match fall from $0.750/0.643/0.267$ to
$0.676/0.590/0.249$. Thus, autoregressive reproduction of fixed geometric
targets is insufficient; Stage~II provides a non-collapsed initialization from
which the targets can subsequently evolve.

\paragraph{Adaptive self-alignment provides the main Stage~II--Stage~III gain.}
Replacing fixed VQ-derived targets with EMA-teacher sequences raises
Jaccard/edit/exact match from $0.676/0.590/0.249$ to
$0.747/0.687/0.302$, while retaining zero measured low-diversity collapse:
absolute gains of $0.071$, $0.097$, and $0.053$. The largest improvement is in
edit similarity, which is order-sensitive. The final token space therefore
emerges from adaptive co-evolution of teacher targets and encoder-conditioned
trajectories, not fixed-target autoregressive imitation alone.

\paragraph{Structured self-attention dropout prevents decoder-side bypass.}
Without structured self-attention dropout, Jaccard, edit similarity, and exact
match all reach $1.0$, but only six symbols remain active, every pair is
low-diversity collapsed, and both collision rates are $1.0$. Perfect agreement
therefore reflects generic collapse. The failure is consistent with
teacher-forced token history becoming an overly strong pathway that reduces
reliance on acoustic conditioning. Structured dropout weakens this pathway;
its role is specifically bypass prevention, which cannot be assessed from
agreement metrics without collapse diagnostics.

\paragraph{Prefix masking improves the sequence-prediction objective, not collapse prevention.}
Prefix masking replaces intact-prefix next-token prediction with recovery from
partially masked history, forcing the decoder to combine incomplete symbolic
context with acoustic conditioning. Removing it remains non-collapsed with
$505$ active symbols and zero collapse/collision, but reduces
Jaccard/edit/exact match from Stage~II's $0.676/0.590/0.249$ to
$0.656/0.552/0.239$. Its benefit is therefore improved cross-view sequence
learning under a harder prediction task, not prevention of degeneracy.

\paragraph{Encoder-summary conditioning improves both stability and compactness.}
Removing the encoder summary lowers Jaccard/edit/exact match from
$0.747/0.687/0.302$ to $0.680/0.514/0.252$, including a $0.173$ edit-similarity
drop. This is not caused by stronger compression: Appendix~\ref{app:retrieval-ablation},
Table~\ref{tab:pairalign_ablation_compactness}, shows archive length increasing
from $24.84$ to $38.55$ tokens and rate from $8.28$ to $12.85$ tok/s.
The variant still uses $508$ symbols with zero measured collapse, isolating a
conditioning failure rather than degeneracy. Cross-attention preserves input
dependence, but the per-step encoder summary provides a persistent global
acoustic reference needed for a compact, stable free-running trajectory.

\paragraph{Likelihood contrast prevents many-to-one agreement collapse.}
Removing in-batch likelihood contrast again yields nominally perfect
Jaccard/edit/exact agreement ($1.0$), but only eight active symbols, collapsed
pair rate $1.0$, and anchor/positive collision rates $1.0/1.0$. Importantly,
this variant still retains the entropy regularizer. The failure therefore shows
that entropy regularization alone provides only a soft pressure toward broader
inventory usage and is insufficient to prevent the dominant autoregressive
modes from converging to a generic many-to-one solution. Positive agreement can
thus be satisfied by assigning generic strings to unrelated inputs despite the
entropy term. Likelihood contrast supplies the explicit negative pressure needed
to separate alternative sequences, consistent with
Appendix~\ref{app:sequence-geometry-diagnostics}, where Full PairAlign contracts
positives relative to meaningful negatives rather than collapsing the entire
sequence space.

\begin{table*}[t]
\centering
\scriptsize
\setlength{\tabcolsep}{4pt}
\renewcommand{\arraystretch}{1.05}

\begin{tabular}{lcccccc}
\toprule
\multirow{2}{*}{\textbf{Variant}}
& \textbf{Jaccard}
& \textbf{Edit}
& \textbf{Exact}
& \textbf{Active}
& \textbf{Collapsed}
& \textbf{Exact} \\
&
\textbf{Similarity}
& \textbf{Similarity}
& \textbf{Match}
& \textbf{Vocabulary}
& \textbf{Pair}
& \textbf{Collision} \\
\midrule

Stage~I Geometric
    & 0.742 & 0.616 & 0.267 & 456 & 0.0803 & 0.0000/0.0000 \\

Stage~I+ Geometric
    & 0.750 & 0.643 & 0.267 & 420 & 0.0290 & 0.0000/0.0000 \\

\midrule

Stage~II AR Prior
    & 0.676 & 0.590 & 0.249 & 506 & 0.0000 & 0.0000/0.0000 \\

\quad w/o structured self-attn dropout
    & 1.000 & 1.000 & 1.000 & 6 & 1.0000 & 1.0000/1.0000 \\

\quad w/o prefix masking
    & 0.656 & 0.552 & 0.239 & 505 & 0.0000 & 0.0000/0.0000 \\

\midrule

\textbf{Full PairAlign}
    & \textbf{0.747}
    & \textbf{0.687}
    & \textbf{0.302}
    & 440
    & \textbf{0.0000}
    & \textbf{0.0000/0.0022} \\

\quad w/o encoder-summary conditioning
    & 0.680 & 0.514 & 0.252 & 508 & 0.0000 & 0.0000/0.0000 \\

\quad w/o in-batch likelihood contrast
    & 1.000 & 1.000 & 1.000 & 8 & 1.0000 & 1.0000/1.0000 \\

\bottomrule
\end{tabular}

\caption{
PairAlign ablation on clean TIMIT. The validation dataloader uses fixed
deterministic seed $1234$ and the evaluation script seed $13$, matching the
fixed-view protocol of Table~\ref{tab:token_space_comparison}. Collapsed Pair
is the fraction of pairs for which either sequence has
$r_{\mathrm{uniq}}\leq0.2$; Exact Collision gives anchor/positive within-stream
duplication across examples. Perfect agreement alone is not useful invariance:
removing structured self-attention dropout or in-batch likelihood contrast
produces generic low-diversity strings with complete cross-example collision.
}
\label{tab:pairalign_consistency_ablation}
\end{table*}

\paragraph{High agreement is meaningful only under non-degeneracy.}
The two catastrophic controls show why Jaccard or exact match cannot be read in
isolation: both reach $1.0$ while collapsing vocabulary and cross-example
diversity. Full PairAlign instead attains edit similarity $0.687$ and exact
match $0.302$ with $440$ active symbols, zero measured low-diversity collapse,
and only $0.0022$ positive-stream collision. The relevant objective is strong
cross-view agreement subject to input dependence, diversity, and separation
between unrelated inputs.

\paragraph{The full-model operating point holds on LibriSpeech-100 and TIMIT.}
Table~\ref{tab:token_space_comparison} shows the same qualitative behavior on
both corpora. PairAlign obtains Jaccard/edit/exact match
$0.719/0.629/0.291$ at mean length $35.55$ on LibriSpeech-100 and
$0.747/0.687/0.302$ at $26.10$ tokens on TIMIT. The compact-consistent regime
is therefore not specific to the TIMIT ablation set: corpus values differ, but
strong ordered agreement is retained at substantially lower rate than the
geometric streams.

\paragraph{Retrieval exposes a different trade-off from paired-view consistency.}
Appendix~\ref{app:retrieval-ablation} evaluates retrieval with validation and
evaluation seed $13$ throughout. Removing the encoder summary does not yield a
uniform retrieval gain: R@1 falls from $0.71$ to $0.68$, MRR from $0.74$ to
$0.73$, and Mean FRR worsens from $44.67$ to $53.50$, although
Recall@$5$--Recall@$20$ increases slightly. Relevant candidates therefore
remain recoverable at broader ranks but are less sharply ranked. Since the
no-summary model is also much less consistent and compact, Full PairAlign gives
the stronger joint operating point rather than exchanging a large consistency
loss for systematic retrieval improvement.

\paragraph{Compactness alone cannot explain the consistency gain.}
Full PairAlign is more compact than the no-summary variant:
$24.84$ versus $38.55$ archive tokens per segment, or $8.28$ versus
$12.85$ tok/s. Its $0.173$ edit-similarity advantage therefore cannot come
from retaining a denser trace. \S~\ref{sec:rate_controlled_bpe} provides an
independent control: post-hoc shortening of fixed geometric tokens likewise
fails to reproduce PairAlign's consistency. The improvement therefore reflects
learned sequence organization rather than sequence length alone.

\paragraph{Ablation conclusion.}
The ablations assign distinct roles to PairAlign's components. Stage~II provides
a stable autoregressive bridge; Stage~III adaptive targets deliver the main
ordered-consistency gain; structured self-attention dropout prevents
autoregressive bypass; prefix masking strengthens partial-prefix sequence
prediction without serving as the primary collapse control; encoder-summary
conditioning supplies persistent global acoustic context and improves both
stability and compactness; and in-batch likelihood contrast provides the
explicit negative pressure required to prevent generic many-to-one collapse,
which the entropy regularizer alone does not prevent.

Full PairAlign is therefore supported by the joint evidence, not any single
metric: it combines low symbolic rate, broad vocabulary use, negligible
cross-example collision, strong ordered paired-view agreement, meaningful
positive--negative geometry, and useful edit-distance retrieval without 
degenerate agreement.

\subsection{Probing the Local Objective Geometry}
\label{sec:results-local-objective-geometry}

We probe the local Stage~III objective geometry on all $4{,}620$ examples of
the held-out TIMIT validation set using the trained PairAlign checkpoint. For
each content-preserving pair $(x_i,x_i^{+})$, the EMA teacher generates a
single anchor-derived target $\widehat{\mathcal Y}_i$, and both the anchor and
positive conditioning paths are trained to predict this same sequence.
Shared-target fitting therefore provides a common symbolic reference for
positive contraction, while likelihood contrast encourages
$\widehat{\mathcal Y}_i$ to score above EMA targets
$\widehat{\mathcal Y}_j$ from other minibatch examples. We evaluate three
independently generated EMA target banks; the full derivation, protocol, and
neighborhood-width sensitivity are given in
Appendix~\ref{sec:pairalign-local-theory}.

\paragraph{Likelihood-space separation.}
For conditioning view $V\in\{A,P\}$, where $A$ denotes the anchor condition
$Z_i$ and $P$ the positive condition $Z_i^{+}$, let $S_{ii,V}$ be the clean
length-normalized likelihood score assigned to the paired target
$\widehat{\mathcal Y}_i$, and $S_{ij,V}$ the corresponding score assigned to
a mismatched target $\widehat{\mathcal Y}_j$. We define the
candidate-specific score gap
\[
    \delta_{ij,V}
    =
    S_{ii,V}-S_{ij,V},
\]
and the minimum realized gap over all mismatches in the row,
\[
    G_{i,V}
    =
    \min_{j\neq i}\delta_{ij,V}.
\]
The Stage~III NCE loss additionally induces a conservative row-wise lower
margin, denoted $\Delta_{i,V}^{\mathrm{NCE}}$. The resulting hierarchy
\[
    \delta_{ij,V}
    \geq
    G_{i,V}
    \geq
    \Delta_{i,V}^{\mathrm{NCE}}
\]
holds with zero observed violations on TIMIT. The NCE-implied margin is
positive for $99.827\pm0.043\%$ of anchor-conditioned rows and
$98.759\pm0.066\%$ of positive-conditioned rows. Its median magnitude is
$4.520\pm0.023$ and $3.689\pm0.026$, respectively, compared with realized
minimum row gaps of $4.725\pm0.019$ and $3.892\pm0.035$.

\paragraph{Held-out transfer from score to edit space.}
Likelihood-space separation alone does not imply edit-distance separation. We
therefore test this link using the leave-one-candidate-out (LOO) monotone
calibration in Appendix~\ref{sec:pairalign-local-theory}, without imposing an
affine or other parametric score--edit relationship. For each score row, the
$k$ highest-scoring mismatched EMA targets define the local hard-negative
neighborhood. The evaluation batch size is $16$, so each row contains $15$
mismatched targets; consequently, $k=15$ uses the complete in-batch mismatch
set and is our primary setting, while $k=5$ and $k=8$ test sensitivity to
neighborhood width.

For each held-out candidate, the monotone score--edit lower envelope is
estimated from the remaining $k-1$ candidates. A query is termed
\emph{supported} only when its score gap lies within the observed,
non-extrapolated range of these calibration points. At $k=15$, among supported
held-out candidates, the resulting lower envelope is valid for
$73.05\pm0.08\%$ of anchor-conditioned comparisons and
$73.13\pm0.05\%$ of positive-conditioned comparisons. The corresponding
median Spearman association between score gap and normalized edit distance is
$0.591\pm0.003$ and $0.589\pm0.003$, respectively. Considering both
conditioning views, at least one view provides a validated local edit-distance
lower bound for $71.76\pm0.07\%$ of all hard-candidate comparisons.

\paragraph{Joint contraction--separation geometry.}
We next test whether this negative separation combines with the contraction
induced by shared canonical prediction. Let
$\widetilde{\mathcal Y}_i^A$ and
$\widetilde{\mathcal Y}_i^P$ denote the teacher-forced student predictions
under the anchor and positive conditions, respectively. Because both paths are
fitted to the same anchor-derived EMA target $\widehat{\mathcal Y}_i$, their
mutual edit distance is bounded by the sum of their individual errors to this
common symbolic reference. Stage~III does not directly optimize their pairwise
edit distance; the contraction follows from shared-target fitting.

Combining this contraction with the validated target--negative edit lower bound
gives the sufficient condition derived in
Appendix~\ref{sec:pairalign-local-theory} for
\[
    ED(
        \widetilde{\mathcal Y}_i^A,
        \widetilde{\mathcal Y}_i^P
    )
    <
    \min
    \left\{
        ED(
            \widetilde{\mathcal Y}_i^A,
            \widehat{\mathcal Y}_j
        ),
        ED(
            \widetilde{\mathcal Y}_i^P,
            \widehat{\mathcal Y}_j
        )
    \right\}.
\]
This is the desired local relational ordering: the two predictions associated
with the same underlying content are closer to one another than either is to a
hard mismatched target.

On TIMIT, at the full $k=15$ neighborhood, this ordering holds directly in
$99.898\pm0.009\%$ of evaluated hard-candidate comparisons. The explicit
sufficient condition certifies $59.21\pm0.08\%$ of all comparisons when using
the observed candidate-specific score gap. Importantly, replacing that gap
with only the more conservative margin implied directly by the NCE objective
still certifies $52.92\pm0.20\%$ of all comparisons. Conditional on successful
held-out calibration, the corresponding rates are
$82.51\pm0.05\%$ and $73.74\pm0.21\%$, respectively.

\paragraph{Implication.}
The TIMIT audit supports the intended decomposition of Stage~III.
Shared-target prediction supplies positive contraction through a common
anchor-derived symbolic reference, while likelihood contrast supplies local
negative separation. The LOO analysis further shows that this score-space
separation transfers to edit space for a substantial fraction of held-out hard
candidates without assuming a parametric score--edit mapping. Their combination
therefore yields a non-vacuous sufficient condition for the desired local
relational geometry. The much higher directly observed ordering rate indicates
that the sufficient condition is conservative rather than that uncertified
comparisons violate the desired ordering.

All uncertainties above are mean $\pm$ standard deviation across three
independently generated EMA target banks for the same trained checkpoint and
therefore quantify target-sampling variability rather than training-run
variability. Full neighborhood-width sensitivity, calibration statistics, and
consistency checks are reported in
Appendix~\ref{sec:pairalign-local-theory},
Table~\ref{tab:local-monotone-joint-audit}.
\subsection{Stress Tests and Rate-Controlled Probes of the Learned Similarity Geometry}
\label{sec:additional_experiments}

We summarize three complementary appendix analyses. First,
Appendix~\ref{app:tamil-transfer} probes transfer from LibriSpeech English to
lexically disjoint Shrutilipi-Tamil, jointly introducing cross-lingual
acoustic--phonetic, short-range linguistic, and corpus-domain shift. Second,
Appendix~\ref{app:ssl-tokenizer-comparison} stress-tests PairAlign against
native-rate HuBERT+VQ and WavLM+VQ, which are high-capacity, high-exposure,
speech-unit-biased external references rather than controlled ablations. Third,
\S~\ref{sec:rate_controlled_bpe} asks whether PairAlign's low-rate
consistency can be reproduced by deterministic post-hoc BPE applied to
Stage~I, HuBERT, or WavLM token streams. Stage~I/Stage~I+ remain the controlled
in-pipeline baselines because they share PairAlign's preprocessing, encoder
family, VQ alphabet, source pool, and training pipeline.

For clean paired-view evaluations in the Tamil and SSL analyses, the validation
dataloader uses fixed deterministic seed $1234$; Stage~I/Stage~I+ and SSL
references use evaluation seed $13$, while PairAlign uses seeds $13$, $97$,
and $313$ on the same fixed views. Tamil and TIMIT retrieval use seed $13$ for
both dataloader and evaluation. All controlled PairAlign-family models use
training seed $240$. The rate-controlled paired-view experiment likewise uses
validation seed $1234$ and evaluation seed $13$; its retrieval and timing
measurements use seed $13$.

\paragraph{Lexically disjoint Tamil transfer preserves a compact, non-degenerate representation but weakens fine-grained geometry.}
Appendix~\ref{app:tamil-transfer} transfers all LibriSpeech-trained tokenizers
unchanged to Shrutilipi-Tamil; because language and corpus vary jointly, this is
a combined acoustic--linguistic/domain stress test rather than isolated
linguistic generalization. In
Table~\ref{tab:pairalign_tamil_consistency_ablation}, PairAlign drops from
LibriSpeech-100 Jaccard/edit $0.719/0.629$ to $0.662/0.553$ on Tamil, while
Stage~I changes from $0.718/0.609$ to $0.694/0.587$ and Stage~I+ from
$0.738/0.629$ to $0.699/0.600$. The denser geometric streams therefore retain
more graded overlap; the result does not support language-invariant PairAlign
geometry.

PairAlign nevertheless remains non-degenerate: Tamil exact match is $0.265$
versus $\approx0.243$ for Stage~I/Stage~I+, $510/512$ symbols remain active,
collapsed-pair rate is zero, and exact collision is only
$3.3\times10^{-4}/0$ for anchor/positive streams
(Table~\ref{tab:pairalign_tamil_consistency_ablation}). Native-position entropy
is also non-invariant, with signed shift $-0.381$ nats, but PairAlign has the
smallest mean/max absolute displacement ($0.462/0.660$ versus
$0.594/1.609$ for Stage~I and $0.649/2.303$ for Stage~I+);
Figure~\ref{fig:tamil_entropy_comparison} and
Table~\ref{tab:tamil_librispeech_entropy_shift} measure this only at each
tokenizer's native output scale, not language-invariant positional structure.

Retrieval exposes the corresponding resolution cost
(Table~\ref{tab:pairalign_tamil_retrieval_compactness}). Stage~I,
Stage~I+, and PairAlign obtain R@1/MRR $0.87/0.90$, $0.85/0.87$, and
$0.68/0.73$, respectively, while emitting $37.72$, $33.72$, and
$12.64$ tok/s. PairAlign therefore uses approximately $66.5\%$ fewer symbols
than Stage~I and $62.5\%$ fewer than Stage~I+, but sacrifices fine-grained
edit-distance discrimination. Its rate also rises from $8.28$ tok/s on the
clean TIMIT retrieval archive to $12.64$ tok/s on Tamil, consistent with
input-conditioned autoregressive termination rather than fixed stride, although
the increase cannot be attributed uniquely to linguistic or acoustic factors.

\paragraph{Native-rate HuBERT and WavLM define a substantially stronger high-exposure reference regime.}
Appendix~\ref{app:ssl-tokenizer-comparison} compares PairAlign with discrete
streams from HuBERT Base and WavLM Large. These are deliberately unmatched
external references: HuBERT begins with KMeans pseudo-labels derived from
39-dimensional MFCCs and iteratively refines them using learned intermediate
representations; WavLM inherits HuBERT-style masked speech-unit prediction,
uses HuBERT-derived clustered targets, and adds denoising-oriented learning
\citep{hsu2021hubert,chen2022wavlm}. Thus, before our final $K=512$ KMeans
discretization, both SSL representations have already been optimized against
explicit cluster-derived acoustic-unit targets, unlike PairAlign's
SimCLR+VQ Stage~I.

Scale is likewise highly asymmetric
(Table~\ref{tab:ssl_pairalign_capacity_exposure}). HuBERT Base has $94.372$M
parameters and approximately $506$k sampled optimization-hours; WavLM Large has
$315.453$M parameters, $94$k h of unique audio, and approximately $8.96$M h
of direct optimization exposure ($\sim9.47$M h pipeline-inclusive). PairAlign
uses a $2.076$M Stage~I encoder and $27.822$M Stage~II/III
encoder--decoder, repeatedly sampling the same $\sim860$-h pool for
approximately $667/400/400$ stage-local hours
($\sim1{,}467$ h cumulative). These systems therefore differ simultaneously in
capacity, exposure, target construction, and symbolic rate.

At native rate, the SSL references preserve stronger graded structure
(Table~\ref{tab:pairalign_SSL_consistency}). On clean TIMIT,
HuBERT/WavLM/PairAlign obtain Jaccard/edit
$0.767/0.726$, $0.773/0.730$, and $0.747/0.687$; on Tamil,
$0.726/0.681$, $0.766/0.718$, and $0.662/0.553$; under external noise,
$0.805/0.749$, $0.794/0.774$, and $0.651/0.554$. PairAlign instead has
higher complete-string agreement on clean TIMIT ($0.302$ versus
$0.263/0.264$) and Tamil ($0.265$ versus $0.255/0.262$), without measured
collapse. Exact equality is therefore complementary to, not a replacement for,
graded local similarity.

Native retrieval strongly favors SSL
(Tables~\ref{tab:ssl_retrieval} and~\ref{tab:ssl_compactness}):
HuBERT/WavLM/PairAlign obtain R@1/MRR
$0.943/0.961$, $0.977/0.985$, and $0.710/0.740$, respectively, but operate at
$28.99$, $32.06$, and $8.28$ tok/s. The SSL systems therefore expose
approximately $3.5$--$3.9\times$ more symbols per unit time. Their advantage
should consequently be attributed to the combined effect of stronger
speech-unit-biased pretraining, much greater scale, and denser local symbolic
resolution, not to any single factor.

\paragraph{Rate-controlled BPE separates dense representation quality from compact sequence geometry.}
Appendix~\ref{sec:rate_controlled_bpe} tests whether PairAlign's operating
point can be recovered by shortening dense token streams post hoc. Stage~I
serves as the controlled baseline: deterministic BPE is fitted on $25{,}000$
distinct $3$ s LibriSpeech training segments ($20.83$ h; $12{,}500$ clean
and $12{,}500$ mildly augmented segments from disjoint sets). PairAlign's
LibriSpeech-100 rate of $11.85$ tok/s is used as a fixed target, while the
number of BPE merges is determined on the LibriSpeech training corpus,
yielding $122{,}820$ merges. Independent HuBERT and WavLM BPE inventories are
fitted on LibriSpeech training data using the same fixed target, yielding
$4{,}190$ and $9{,}895$ merges, respectively. All transforms are then frozen
and evaluated unchanged on LibriSpeech-100 and TIMIT; no TIMIT data are used
for fitting, merge selection, or rate calibration.

Table~\ref{tab:rate_controlled_bpe_consistency} shows that deterministic
post-hoc shortening does not reproduce PairAlign's
compactness--consistency operating point. On LibriSpeech-100, Stage~I+BPE
reduces the mean sequence length from $92.07$ to $43.54$ tokens, but
Jaccard/edit similarity falls from $0.718/0.609$ to $0.421/0.412$; PairAlign
is shorter at $35.55$ tokens while reaching $0.719/0.629$. Under TIMIT
transfer, Stage~I+BPE, HuBERT+BPE, and WavLM+BPE remain denser than PairAlign
($12.29/13.08/12.77$ versus $8.70$ tok/s), yet attain only
$0.422/0.407$, $0.462/0.496$, and $0.429/0.465$ Jaccard/edit similarity,
respectively, compared with PairAlign's $0.747/0.687$. Thus, across both the
controlled Stage~I stream and substantially stronger pretrained SSL streams,
a fixed deterministic BPE transform does not recover PairAlign's joint
low-rate and paired-view-consistency operating point.

Exact match is not the discriminating quantity because deterministic lossless
BPE preserves equality of the underlying source strings. The relevant effect
is therefore in the graded geometry among non-identical paired sequences.
Retrieval provides an important qualification
(Table~\ref{tab:ssl_bpe_retrieval}): HuBERT+BPE retains R@1/MRR
$0.807/0.823$, above PairAlign's $0.710/0.740$, while WavLM+BPE reaches
$0.760/0.777$. Hence degradation in paired-view sequence geometry does not
imply loss of all discriminative structure. The native SSL retrieval advantage
nevertheless contracts substantially after rate control, and both SSL+BPE
systems exhibit heavier Mean-FRR tails ($138.47/241.77$ versus PairAlign's
$44.67$).

Construction cost forms a separate axis
(Table~\ref{tab:rate_controlled_bpe_timing}). HuBERT+BPE and WavLM+BPE are
faster end-to-end than PairAlign ($33.75/89.30$ versus $206.09$ ms/segment),
whereas the evaluated Stage~I+BPE implementation is slower
($793.73$ ms/segment) because sequential BPE post-processing dominates its
measured runtime. PairAlign's neural latency is dominated by autoregressive
decoding ($194.27$ ms), while its peak GPU memory ($570.58$ MB) lies above
Stage~I+BPE ($126.93$ MB) but below HuBERT+BPE and WavLM+BPE
($1129.17/2272.30$ MB). These measurements characterize the evaluated
implementations rather than intrinsic complexity bounds.

\paragraph{Takeaway.}
The three appendix analyses separate representation scale, symbolic rate,
cross-view geometry, and retrieval discrimination. Tamil transfer shows that
PairAlign is not language invariant: dense geometric streams retain stronger
graded similarity and retrieval, while PairAlign remains much more compact,
input-conditioned, broadly utilizes its inventory, and shows no measured
low-diversity collapse
(Appendix~\ref{app:tamil-transfer};
Tables~\ref{tab:pairalign_tamil_consistency_ablation} and
\ref{tab:pairalign_tamil_retrieval_compactness}). Native HuBERT/WavLM establish
an intentionally stronger reference regime, preserving substantially more local
structure and retrieval information at approximately $3.5$--$3.9\times$ the
PairAlign rate
(Appendix~\ref{app:ssl-tokenizer-comparison};
Tables~\ref{tab:pairalign_SSL_consistency},
\ref{tab:ssl_retrieval}, and~\ref{tab:ssl_compactness}).

The rate-controlled experiment provides the complementary result:
deterministically shortening Stage~I, HuBERT, or WavLM does not reproduce
PairAlign's compactness--consistency operating point
(Appendix~\ref{sec:rate_controlled_bpe};
Table~\ref{tab:rate_controlled_bpe_consistency}), although compressed SSL
streams can retain stronger retrieval cues
(Table~\ref{tab:ssl_bpe_retrieval}). PairAlign should therefore be characterized
as a lower-rate sequence-level operating point, not a universally stronger
tokenizer: it trades part of the dense local acoustic and retrieval resolution
for compact, non-degenerate, input-conditioned sequence geometry.

These results also separate two natural scaling axes. Stronger pretrained SSL
features could improve the initial acoustic--phonetic representation, while
PairAlign's autoregressive bridge and EMA self-alignment target adaptive
low-rate sequence organization. Replacing or initializing Stage~I with
HuBERT/WavLM features would directly test whether these advantages can be
combined; the present results motivate, but do not establish, that outcome.

\subsection{Temporal Recoverability and Acoustic--Linguistic Structure of PairAlign Tokens}
\label{sec:temporal_linguistic_summary}

Because PairAlign emits compact autoregressive sequences rather than
frame-synchronous labels, we probe two properties not captured by consistency or
retrieval: post-hoc temporal recoverability and retained acoustic--linguistic
structure. Full timing results appear in
Appendix~\ref{sec:timing-recovery-results}; phonetic, boundary, and speaker
analyses appear in Appendix~\ref{sec:phonetic_boundary_analysis}. Both the
validation dataloader and evaluation script use fixed seed $13$ throughout.

\paragraph{Decoder cross-attention provides a usable post-hoc timing signal.}
We derive frame-to-token posteriors from decoder cross-attention and recover a
monotone alignment with Viterbi decoding. A weak diagonal Beta prior is also
applied post hoc; it changes neither token identities nor order. In
Table~\ref{tab:frame_to_token_timing_primary}, Beta reweighting reduces expected
frame-to-token violation from $0.4734$ to $0.3769$ ($20.38\%$ relative),
reduces mean absolute diagonal error by $5.10\%$, and improves mean Viterbi
log-score by $5.85\%$, while peak assignments change only marginally. Thus,
the prior mainly sharpens temporal structure already present in cross-attention.

Against sample-level TIMIT phone boundaries
(Table~\ref{tab:pairalign_timit_boundary_alignment}), Beta yields
approximately $0.6$--$0.7\%$ relative F1 gains at $10$--$30$ ms tolerance,
with paired bootstrap intervals excluding zero, while matched-boundary
localization error is essentially unchanged. We therefore treat Beta alignment
as approximate post-hoc temporal grounding, not evidence that PairAlign tokens
correspond to phones, syllables, or other predefined linguistic units.

\paragraph{PairAlign changes temporal granularity rather than becoming a phonemic tokenizer.}
Relative to Stage~I Geometric, PairAlign exhibits a clear boundary
precision--recall trade-off
(Table~\ref{tab:pairalign_beta_vs_geometric_boundaries}). At $20$ ms tolerance,
precision rises from $0.319$ to $0.458$, while recall falls from $0.705$ to
$0.347$: PairAlign emits fewer, more selectively placed transitions. Mean
tokens per reference phone likewise decrease from $2.258$ to $1.656$, reducing
within-phone fragmentation, while individual PairAlign segments more often span
multiple phones. PairAlign therefore learns a coarser segmentation, not a
one-token--one-phone mapping.

\paragraph{Fine-grained phone identity remains stronger in Stage~I.}
Neither representation uses phone supervision, yet both show emergent phonetic
association. Stage~I retains substantially more local phone information:
phone purity decreases from $0.292$ to $0.247$ under PairAlign, PNMI from
$0.147$ to $0.080$, and leave-one-speaker-out token-to-phone accuracy from
$0.289$ to $0.243$; Stage~I also gives lower within- and across-speaker discrete
ABX error. PairAlign therefore does not improve phonetic discriminability; its
lower symbolic rate discards part of the fine-grained acoustic--phonetic
resolution present in the dense Stage~I stream.

\paragraph{Short token context does not recover the phonetic resolution lost under compression.}
A possible alternative is that PairAlign redistributes phonetic information
from individual tokens into short compositions. Bigram/trigram probes in
Table~\ref{tab:pairalign_context_privacy_probes} do not support this: PNMI
increases with context for both models, but Stage~I remains substantially
stronger at both orders, while leave-one-speaker-out context-to-phone accuracy
stays near $0.29$ for Stage~I and $0.24$ for PairAlign. PairAlign therefore
retains short-context phone-correlated structure, but its weaker phonetic
resolution is not merely a shift from unigram to bigram/trigram coding.

\paragraph{The phonetic association is emergent rather than imposed by a speech-unit prior.}
Unlike HuBERT- or WavLM-style systems, whose masked-prediction objectives use
clustered or discretized acoustic targets that bias representations toward
segmental structure, Stage~I and PairAlign receive no pseudo-label speech-unit
sequence. Their phone purity, PNMI, ABX, and boundary association therefore
emerge indirectly from recurrent acoustic structure and self-supervised
learning. These symbols should not be identified with phonemes; they are
acoustic--symbolic sequence elements that correlate with phonetic structure
without being constrained to reproduce a phoneme inventory.

\paragraph{The same granularity change explains the retrieval operating point.}
Stage~I's denser stream preserves more local acoustic--phonetic evidence,
providing normalized edit distance with more discriminative cues; PairAlign
retains fewer such local distinctions because each symbol plays a coarser role
in the sequence. This does not make PairAlign acoustically unstructured:
selective boundaries, non-trivial phone association, paired-view consistency,
and positive--negative edit geometry all remain. The result is a different
rate--granularity operating point, not uniformly stronger local invariance or
phonetic discrimination.

\paragraph{PairAlign reduces measured speaker recoverability, but does not remove speaker information.}
Information-theoretic speaker measures are mixed:
unconditional speaker NMI decreases from $0.249$ for Stage~I to $0.233$ for
PairAlign, whereas speaker NMI conditioned on phone increases from $0.351$ to
$0.362$. These metrics alone therefore do not support speaker removal.

Lightweight attacker-style probes show lower recoverability under PairAlign
(Table~\ref{tab:pairalign_context_privacy_probes}). Speaker-ID accuracy falls
from $0.202$ to $0.124$ with unigrams, $0.181$ to $0.068$ with bigrams, and
$0.131$ to $0.036$ with trigrams. Trigram verification ROC-AUC decreases from
$0.730$ to $0.545$, while EER increases from $0.331$ to $0.472$, approaching
non-discriminative performance for this count-based probe. These results support
only reduced \emph{measured speaker recoverability}: PairAlign has no
anonymization, speaker-adversarial, differential-privacy, or explicit
speaker-information objective, and stronger or longer-context attackers may
recover additional information.

\paragraph{Beta affects temporal grounding, not representation content.}
PairAlign Raw and PairAlign+Beta have identical sequence-level speaker-probe
results because Beta modifies only the post-hoc token-to-frame alignment, not
the generated sequence. It can therefore affect timing and frame-aligned
phonetic diagnostics but not bag-of-$n$-gram speaker probes. Conceptually,
Raw--Beta comparisons evaluate temporal recovery, whereas
Stage~I--PairAlign comparisons characterize the representation itself.

\paragraph{Summary.}
The appendix analyses show that PairAlign does not become a more phonemic
tokenizer. It converts the dense Stage~I stream into a much lower-rate symbolic
trajectory with fewer, more selective recovered transitions and less
within-phone fragmentation, while sacrificing part of Stage~I's fine-grained
phonetic and speaker-discriminative information.

PairAlign is therefore neither a frame-synchronous linguistic code nor an
anonymized sequence. It is a coarser acoustic--symbolic interface with
approximate post-hoc temporal grounding, emergent phonetic structure, reduced
local acoustic resolution, and lower measured speaker recoverability under the
evaluated probes. This matches the broader PairAlign trade-off: useful
sequence-level relational structure is retained at markedly lower rate, with
compactness obtained partly by moving away from dense local acoustic detail.

\section{Conclusion}
\label{sec:conclusion}

This paper introduced PairAlign, a sequence-level approach to discrete audio
tokenization. Rather than learning only token identity, PairAlign learns the
structure of the emitted sequence---its ordering, length, termination, and
sequence-level geometry---through shared-target predictive self-alignment.
Starting from a geometric VQ tokenizer, an autoregressive decoder is first
trained on fixed anchor-derived targets and is then refined using adaptive
sequences generated by a slowly updated teacher, with both the anchor and its
content-preserving view trained toward the same symbolic target. Likelihood
contrast preserves discrimination across different examples, while
decoder-side grounding mechanisms reduce reliance on token history and prevent
generic sequence collapse.

PairAlign can be viewed as \emph{sequence-symbolic predictive learning}. Like
JEPA-style objectives, it predicts an abstract representation rather than
reconstructing the raw input. Here, however, the predictive object is itself a
learned variable-length symbolic sequence. PairAlign therefore extends
predictive self-supervision from learning latent representations to learning the
discrete sequence interface through which acoustic structure is serialized and
compared.

Empirically, PairAlign moves tokenization to a substantially lower-rate regime
while retaining meaningful ordered and relational structure. On the clean TIMIT
retrieval archive, it operates at $8.28$ tokens/s compared with $28.21$
tokens/s for the geometric Stage~I tokenizer, reducing archive token count by
$70.65\%$. The resulting sequences remain broadly distributed over the token
inventory, exhibit negligible measured collapse, and preserve strong
paired-view and positive--negative edit structure. Rate-controlled post-hoc BPE
applied to Stage~I, HuBERT, and WavLM streams does not reproduce the same
compactness--consistency operating point, indicating that PairAlign's behavior
is not explained by sequence shortening alone.

The principal trade-off is granularity. Dense geometric and pretrained SSL
tokenizers retain more fine-grained local acoustic, phonetic, and retrieval
information, whereas PairAlign represents the signal with substantially fewer
symbols. This compactness reduces repeated sequence-comparison cost: relative to
Stage~I, the edit-distance work proxy falls by $91.87\%$ once tokenizations are
available. The benefit does not extend to token construction itself, since the
current autoregressive beam-search decoder is substantially more expensive than
frame-derived tokenization. PairAlign should therefore be interpreted as a
different rate--granularity--computation operating point rather than a uniformly
stronger tokenizer.

The ablations identify complementary roles for the main components. Adaptive
teacher-generated targets provide the transition from fixed geometric strings
to learned sequence organization; structured self-attention dropout and prefix
corruption weaken decoder-history shortcuts; encoder-summary conditioning
provides persistent acoustic context; and in-batch likelihood contrast prevents
many-to-one sequence collapse. Approximate timing can additionally be recovered
from decoder cross-attention, while phonetic analysis shows that PairAlign
retains acoustic--linguistic structure at a coarser granularity rather than
inducing one-token--one-phone units.

The local objective-geometry analysis further links Stage~III to the relational
structure observed after training. Shared-target fitting contracts the positive
pair around a common EMA target, while likelihood contrast separates that target
from hard mismatches. On held-out TIMIT, the desired local ordering holds in
$99.898\pm0.009\%$ of comparisons; for the full $15$-mismatch neighborhood,
validated local edit bounds support sufficient certificates for
$59.21\pm0.08\%$ of comparisons using observed score gaps and
$52.92\pm0.20\%$ using only the NCE-implied margin. These are local sufficient
conditions, not a global edit-geometry guarantee.

The cross-lingual and pretrained-SSL comparisons further bound the claim.
PairAlign does not dominate dense pretrained tokenizers on fine-grained local
consistency, phonetic detail, noise robustness, or top-rank retrieval. Instead,
it provides a complementary mechanism for transforming a frame-derived acoustic
representation into a compact, adaptive-length, input-conditioned symbolic
trajectory while retaining useful sequence-level structure.

More broadly, PairAlign suggests that predictive self-supervision need not end
with continuous embeddings or frame-synchronous codes. The symbolic interface
itself---its vocabulary, ordering, length, termination, and relational
geometry---can be learned as the predictive object. This sequence-symbolic view
extends the JEPA-style intuition toward learning not only
\emph{what representation should be predicted}, but also \emph{how that
representation should be serialized as a discrete sequence} for storage,
comparison, composition, and downstream reasoning.

\section{Broader Impact}
\label{sec:broader_impact}

PairAlign studies self-supervised induction of compact symbolic sequences from
continuous audio. Its most immediate uses are retrieval, indexing, matching,
and repeated comparison over large audio collections, particularly when
transcripts or reliable ASR are unavailable. More broadly, lower-rate symbolic
interfaces may be useful for long-context audio modeling, memory, generation,
editing, and multimodal systems in settings where preserving sequence-level
structure is more important than waveform reconstruction or dense temporal
resolution.

The methodological implication is that symbolic interfaces need not be fixed by
frame stride or manually specified units. Sequence-level predictive learning
could be extended to other continuous domains---including music, video,
robotics, sensor streams, biomedical signals, and scientific time
series---where appropriate discrete units are not known in advance. In such
settings, jointly learning token identity, ordering, sequence length, and
termination may provide compact interfaces for storage, comparison, and
downstream sequence modeling.

Compact searchable representations also introduce risks. Lower-rate audio
tokens can make large-scale indexing and comparison easier, including for
private or sensitive recordings. PairAlign is not an anonymization or
privacy-preserving method. The speaker probes in this work show reduced
recoverability under the evaluated lightweight attackers relative to the dense
Stage~I representation, but they do not establish speaker invariance or prevent
stronger models from recovering identity or other sensitive attributes.
Applications involving human recordings should therefore consider consent,
access control, retention policy, and privacy evaluation appropriate to the
deployment setting. If the representation is incorporated into generative or
editing systems, additional safeguards may also be needed against unauthorized
impersonation, deceptive synthesis, or manipulation.

The current system also has practical limitations relevant to deployment.
PairAlign trades fine-grained frame-local information and native timestamps for
a substantially lower-rate symbolic sequence, and its autoregressive decoder is
more expensive to run than the geometric tokenizers considered here. The
learned symbols are not shown to correspond directly to phonemes, words, or
other human-interpretable units; the phonetic and temporal analyses instead
support a coarser acoustic--symbolic representation with approximate temporal
recoverability and reduced fine-grained phonetic detail.

Accordingly, PairAlign should be viewed as a representation-learning framework,
not as a deployment-ready codec, privacy mechanism, or universally preferable
audio tokenizer. Applications beyond the present setting require separate
validation of task utility, computational cost, robustness, interpretability,
privacy, and safety under the intended data and deployment conditions.

\section{Future Work}
\label{sec:future_work}

The present work already characterizes PairAlign's low-rate operating point,
compactness--granularity trade-off, component roles, learned termination,
relational sequence geometry, and the training-time mechanisms underlying
shared-target fitting and negative separation. Future work should therefore
focus on extending the operating regime, reducing token-generation cost,
strengthening the underlying acoustic representation, and testing whether the
resulting compact sequences provide practical advantages in larger
sequence-modeling systems.

\paragraph{Controllable and adaptive symbolic rate.}

PairAlign currently learns an input-dependent sequence length within one
training-defined rate regime. Rate conditioning, explicit information budgets,
or multi-rate training could expose multiple operating points within a single
tokenizer, allowing downstream systems to select the required balance between
symbolic rate and acoustic granularity without retraining.

A stronger extension is to make the allocation itself adaptive. Rather than
varying only a global rate, the tokenizer could allocate more symbolic capacity
to acoustically or linguistically complex regions and fewer symbols to
redundant regions under a fixed overall budget. This would extend the learned
EOS behavior studied here toward explicit, content-dependent allocation of
symbolic capacity.

\paragraph{Stronger acoustic foundations and higher-capacity sequence models.}

The HuBERT and WavLM comparisons identify a concrete opportunity to combine
stronger pretrained acoustic representations with PairAlign's low-rate
sequence-level learning. Initializing the encoder from such representations
would test whether more fine-grained acoustic--phonetic information can be
retained without giving up the compact sequence geometry learned by PairAlign.

The decoder can likewise be scaled beyond the controlled architecture used
here. Higher-capacity sequence models, including architectures and training
techniques developed for modern LLMs, could provide stronger long-range
sequence modeling while the grounding mechanisms introduced here maintain
dependence on the acoustic condition. This would test whether model capacity
can shift the compactness--locality frontier rather than simply increasing
decoder-side predictive power.

\paragraph{Efficient, streaming, and long-context tokenization.}

The main systems limitation is token construction: PairAlign's materialized
sequences are compact and inexpensive to compare, but autoregressive beam
decoding is substantially slower than frame-derived tokenization. Future work
could use faster decoder architectures, more efficient decoding methods
developed for large language models, non-autoregressive alternatives, or
distilled student tokenizers that reproduce PairAlign's learned low-rate
sequences at lower inference cost.

Streaming and long-context variants provide a complementary direction.
Causal emission, bounded revision of previously generated symbols, and
hierarchical token rates could extend PairAlign beyond short segments to
long-form audio. An important systems-level target is to preserve the learned
sequence geometry while improving latency, throughput, and context scalability.

\paragraph{Sequence-to-sequence compression.}

PairAlign's formulation naturally extends from continuous-to-symbolic
tokenization to compression of already discrete high-rate streams. Codec
tokens, geometric units, or pretrained SSL tokens could be mapped into shorter
variable-length sequences whose ordering, rate, and termination are learned
rather than inherited from frame stride.

Such models could optimize reconstruction, prediction, or downstream utility
under an explicit symbolic budget. This direction is particularly relevant to
neural codecs, audio-language models, and multimodal LLMs, where high-rate
acoustic token streams directly consume context length and increase
sequence-modeling cost.




\paragraph{Intrinsic temporal structure.}

The present analysis shows that approximate timing can be recovered after token
generation from decoder cross-attention. Future models could instead make
temporal support intrinsic to sequence induction through monotonic alignment,
duration modeling, transducer-style prediction, or explicit latent
token-to-time variables.

This would provide compact variable-length symbols with native temporal extent,
supporting localization, segmentation, streaming retrieval, and time-aware
editing without returning to dense frame-synchronous tokenization.

\paragraph{Retrieval-aware and downstream sequence modeling.}

The retrieval experiments show that compact sequence geometry and fine-grained
ranking precision are distinct objectives. One direction is therefore to make
retrieval structure explicit during sequence learning, for example through
retrieval-aware negative selection or neighborhood-preserving sequence
training. A coarse-to-fine system is particularly natural: PairAlign sequences
could provide inexpensive first-stage indexing, followed by reranking with a
denser acoustic representation.

More broadly, the key downstream question is whether lower symbolic rate
translates into larger usable context or lower computation for models that
consume the tokens. PairAlign-style interfaces should be evaluated under
matched model, memory, and context budgets in long-context audio modeling,
retrieval-augmented systems, audio question answering, symbolic editing, and
audio-language modeling.

\paragraph{Integration with LLMs and multimodal sequence models.}

PairAlign's compact variable-length representation is particularly relevant to
LLM-based audio systems, where acoustic token rate directly determines how much
audio can be represented inside a finite context window. Future work could
integrate PairAlign tokens directly into audio-language and multimodal LLMs,
either as an input vocabulary, an intermediate acoustic memory, or a compressed
representation consumed alongside text.

Such experiments could test whether lower-rate symbolic trajectories improve
long-context audio understanding, cross-modal reasoning, retrieval-augmented
generation, or persistent acoustic memory, while determining how much local
acoustic information must be retained for these downstream capabilities.

\paragraph{Broader domains, information allocation, and privacy.}

Broader multilingual evaluation should disentangle language, speaker, channel,
and corpus variation more systematically. Beyond speech, the same
sequence-symbolic learning principle can be tested on music, environmental
audio, mixtures, multimodal streams, robotics, and other continuous signals
where appropriate symbolic units are not predefined.

The speaker-recoverability results also motivate explicit control over what
information survives compression. Future models could allocate symbolic
capacity according to task requirements while explicitly retaining or
suppressing factors such as speaker, prosody, environment, or other nuisance
attributes. In privacy-oriented settings, speaker suppression would need to be
trained explicitly and evaluated against substantially stronger identification,
verification, reconstruction, and membership-style attacks; reduced
recoverability under the lightweight probes studied here should not itself be
interpreted as anonymity.

\paragraph{Sequence-symbolic predictive learning and the JEPA connection.}
PairAlign follows the broader spirit of JEPA-style predictive learning in
predicting an abstract target representation rather than directly
reconstructing the input. In PairAlign, this predictive target takes the form
of a learned, variable-length symbolic sequence.

A direct comparison with continuous predictive-learning approaches under
matched data, encoders, augmentations, and model capacity would help clarify
when discrete sequence prediction provides a useful alternative predictive
interface. Of particular interest is how continuous and symbolic formulations
distribute information across representation dimensionality, symbol identity,
and sequence length, and how these choices affect downstream context length
and computation.

\paragraph{Toward predictive symbolic world models.}

A longer-term direction is to investigate whether compact learned symbolic
trajectories can serve as state interfaces over extended multimodal experience.
Audio, vision, text, and action need not share a single vocabulary;
modality-specific symbolic representations could instead be related through
temporal, cross-modal, and action-conditioned prediction.

An important open question is whether such low-rate symbolic trajectories can
support persistent state, memory, prediction, planning, and reasoning over
long horizons. Exploring these capabilities could connect predictive learning
over compact symbolic sequences with broader efforts toward predictive
multimodal world models.

\section{Code Availability}
The source code for PairAlign is publicly available at \url{https://github.com/adhirajAI/pairAlign}.

\section{Declaration of LLM Usage} \label{llm_usage}

LLM is used only to aid or polish writing and does not impact the core methodology, scientific rigorousness, or originality of the research.

\bibliography{tmlr}
\bibliographystyle{tmlr}

 \appendix

\section{Stage I Details: Encoder Pretraining and Geometric VQ Tokenization}
\label{app:stage1-full}

Stage~I trains the encoder and VQ codebook before any autoregressive
sequence-level modeling. The encoder produces contextual frame embeddings
$Z_i=[z_{i,t}]_{t=1}^{T_i}$ and $Z_i^{+}=[z^{+}_{i,t}]_{t=1}^{T_i^{+}}$ for
paired views.

\paragraph{Frame-level contrastive loss.}
Following BEST-STD~\citep{singh2025best}, paired views are aligned with DTW.
Let $t^{+}(t)$ denote the frame in the positive view aligned to anchor frame
$t$. The contrastive loss for example $i$ is
\begin{equation}
\mathcal{L}_{\mathrm{contrast}}^{(i)}
=
-\frac{1}{T_i}
\sum_{t=1}^{T_i}
\log
\frac{
\exp\!\left(z_{i,t}^{\top} z^{+}_{i,t^{+}(t)} / \tau\right)
}{
\exp\!\left(z_{i,t}^{\top} z^{+}_{i,t^{+}(t)} / \tau\right)
+
\sum_{n\in\mathcal{N}_{i,t}}
\exp\!\left(z_{i,t}^{\top} z_n / \tau\right)
}.
\end{equation}
Here $\tau$ is the temperature and $\mathcal{N}_{i,t}$ denotes unrelated
negative frames from the minibatch.

\paragraph{Nearest-centroid VQ assignment.}
Let $C=\{c_a:a\in\mathcal{A}\}$ be the VQ codebook. Each frame is assigned to
its nearest centroid:
\begin{equation}
\tau_{i,t}
=
\arg\min_{a\in\mathcal{A}}
\|z_{i,t}-c_a\|_2^2,
\qquad
z^q_{i,t}=c_{\tau_{i,t}} .
\end{equation}

\paragraph{Commitment loss.}
The commitment loss is
\begin{equation}
\mathcal{L}_{\mathrm{commit}}^{(i)}
=
\frac{1}{T_i}
\sum_{t=1}^{T_i}
\|z_{i,t}-z^q_{i,t}\|_2^2
+
\frac{1}{T_i^{+}}
\sum_{t=1}^{T_i^{+}}
\|z^{+}_{i,t}-z^{q,+}_{i,t}\|_2^2 .
\end{equation}

\paragraph{Stage I objective.}
The Stage~I training objective is
\begin{equation}
\mathcal{L}_{\mathrm{StageI}}
=
\frac{1}{B}
\sum_{i=1}^{B}
\left(
\mathcal{L}_{\mathrm{contrast}}^{(i)}
+
\lambda_{\mathrm{commit}}
\mathcal{L}_{\mathrm{commit}}^{(i)}
\right).
\end{equation}

\paragraph{EMA codebook update.}
For token $a$, let $n_a$ be the number of assigned frames in the minibatch and
$m_a$ the sum of their encoder states. The EMA statistics are updated as
\begin{equation}
N_a \leftarrow \mu N_a + (1-\mu)n_a,
\qquad
M_a \leftarrow \mu M_a + (1-\mu)m_a,
\end{equation}
and the centroid is updated by
\begin{equation}
c_a \leftarrow \frac{M_a}{N_a+\varepsilon}.
\end{equation}

\paragraph{Deduplicated geometric token sequence.}
The raw frame-synchronous token sequence is
\begin{equation}
\mathcal{T}^{\mathrm{raw}}_i
=
[\tau_{i,1},\ldots,\tau_{i,T_i}].
\end{equation}
For sequence-level comparison and for Stage~II targets, we apply run-length
deduplication:
\begin{equation}
\mathcal{T}_i
=
\phi(\mathcal{T}^{\mathrm{raw}}_i).
\end{equation}

\section{Stage I+ Details: Sequence-Consistent Geometric Tokenization}
\label{app:stage1plus-full}

Stage~I+ keeps the Stage~I frame-synchronous VQ tokenizer but adds a symmetric
no-blank CTC sequence-consistency term between paired views.

\paragraph{Framewise token posteriors.}
For each encoder state, we compute a posterior over the VQ alphabet:
\begin{equation}
p_{i,t}(a)
=
\frac{
\exp(z_{i,t}^{\top}c_a)
}{
\sum_{a'\in\mathcal{A}}
\exp(z_{i,t}^{\top}c_{a'})
},
\qquad
a\in\mathcal{A}.
\end{equation}
The positive-view posteriors $p^{+}_{i,t}(a)$ are defined analogously.

\paragraph{No-blank CTC likelihood.}
Let $\mathcal{T}=(\tau_1,\ldots,\tau_L)$ be a deduplicated target sequence and
let $P=\{p_t(a)\}_{t=1}^{T}$ be framewise token posteriors. The no-blank CTC
likelihood is
\begin{equation}
p_{\mathrm{CTC}}(\mathcal{T}\mid P)
=
\sum_{\pi:\phi(\pi)=\mathcal{T}}
\prod_{t=1}^{T}p_t(\pi_t),
\end{equation}
where $\phi(\cdot)$ removes consecutive repetitions.

\paragraph{Forward recursion.}
Let $\alpha_t(l)$ be the probability of all paths up to frame $t$ that collapse
to prefix $(\tau_1,\ldots,\tau_l)$. The recursion is
\begin{equation}
\alpha_1(1)=p_1(\tau_1),
\qquad
\alpha_1(l)=0 \;\; \text{for } l>1,
\end{equation}
\begin{equation}
\alpha_t(l)
=
\left[
\alpha_{t-1}(l)
+
\alpha_{t-1}(l-1)
\right]
p_t(\tau_l),
\end{equation}
with out-of-range terms set to zero. The likelihood is
$p_{\mathrm{CTC}}(\mathcal{T}\mid P)=\alpha_T(L)$.
The recursion is implemented in log space.

\paragraph{Backward recursion.}
Let $\beta_t(l)$ be the probability of all suffix paths from frame $t$ to $T$
that collapse to suffix $(\tau_l,\ldots,\tau_L)$. The initialization is
\begin{equation}
\beta_T(L)=p_T(\tau_L),
\qquad
\beta_T(l)=0 \quad \text{for } l<L .
\end{equation}
For $t\le T-1$,
\begin{equation}
\beta_t(l)
=
p_t(\tau_l)
\left[
\beta_{t+1}(l)
+
\beta_{t+1}(l+1)
\right],
\end{equation}
again with out-of-range terms set to zero. The recursions are implemented in
log space for numerical stability.

\paragraph{Symmetric pairwise sequence loss.}
For paired views, Stage~I+ scores each deduplicated target sequence under the
other view's framewise posteriors:
\begin{equation}
\mathcal{L}_{\mathrm{seq}}^{(i)}
=
\frac{1}{2}
\left[
-\log p_{\mathrm{CTC}}(\mathcal{T}_i^{+}\mid P_i)
-
\log p_{\mathrm{CTC}}(\mathcal{T}_i\mid P_i^{+})
\right].
\end{equation}

\paragraph{Adaptive CTC weight.}
The sequence term is weighted by
\begin{equation}
\lambda_{\mathrm{CTC}}
=
\gamma
\frac{
\overline{\mathcal{L}}_{\mathrm{contrast}}
}{
\overline{\mathcal{L}}_{\mathrm{seq}}+\varepsilon
},
\end{equation}
where $\gamma$ controls the relative strength of the sequence term and
$\varepsilon$ is a numerical stabilizer.

\paragraph{Stage I+ objective.}
The complete Stage~I+ objective is
\begin{equation}
\mathcal{L}_{\mathrm{StageI+}}
=
\frac{1}{B}
\sum_{i=1}^{B}
\left(
\mathcal{L}_{\mathrm{contrast}}^{(i)}
+
\lambda_{\mathrm{commit}}
\mathcal{L}_{\mathrm{commit}}^{(i)}
+
\lambda_{\mathrm{CTC}}
\mathcal{L}_{\mathrm{seq}}^{(i)}
\right).
\end{equation}

\section{Teacher Target Generation via Top-p Sampling with Compound Repetition Penalty and Differential Stochasticity}
\label{app:teacher-sampling-full}

\begin{algorithm}[t]
\caption{EMA-Teacher Target Generation}
\label{alg:surprisal-top-p}
\begin{algorithmic}[1]
\Require Teacher encoder states $\widetilde{Z}$, teacher decoder
$\widetilde{Dec}$, valid encoder length $T_x$, rate cap $\rho$, schedules
$(p_l,\tau_l,\gamma_l)$
\Ensure Sampled teacher target $\widehat{\mathcal{T}}$

\State Initialize prefix $\widehat{\mathcal{T}}\gets[\mathrm{BOS}]$
\State Set length cap $L_{\mathrm{cap}}\gets \lfloor \rho T_x\rfloor$

\For{$l=1,\ldots,L_{\mathrm{cap}}$}
    \State Compute next-token logits
    $\ell_l\gets \widetilde{Dec}(\widehat{\mathcal{T}}_{<l},\widetilde{Z})$

    \State Count prefix occurrences $c_l(v)$ for each token $v$
    \State Set $c_l(v)=0$ for special symbols excluded from repetition control

    \State Apply compound repetition penalty:
    \[
    \widetilde{\ell}_l(v)
    \gets
    \ell_l(v)-c_l(v)\log\gamma_l
    \]

    \If{$l=L_{\mathrm{cap}}$}
        \State Force $\hat{\tau}_l\gets \mathrm{EOS}$
    \Else
        \State Sort $\widetilde{\ell}_l$ in descending order
        \State Retain the smallest nucleus whose cumulative probability mass
        exceeds $p_l$
        \State Always retain the highest-probability token
        \State Mask all non-retained logits to $-\infty$
        \State Apply temperature scaling by $\tau_l$
        \State Sample $\hat{\tau}_l$ from the resulting categorical distribution
    \EndIf

    \State Append $\hat{\tau}_l$ to $\widehat{\mathcal{T}}$

    \If{$\hat{\tau}_l=\mathrm{EOS}$}
        \State \textbf{break}
    \EndIf
\EndFor

\State Remove $\mathrm{BOS}$ and return tokens up to and including
$\mathrm{EOS}$
\end{algorithmic}
\end{algorithm}
This appendix gives the Stage~III EMA-teacher sampling procedure. The role of
teacher targets, top-$p$ sampling, repetition control, and length-constrained
EOS handling is described in \S~\ref{sec:ar-target-generation-decoding}.

\paragraph{Compound repetition penalty.}
At decoding step $l$, let $\ell_l(v)$ be the teacher logit for token $v$ and
let $c_l(v)$ be the number of times token $v$ appears in the generated prefix.
Special symbols such as $\mathrm{BOS}$, $\mathrm{EOS}$, and $\mathrm{PAD}$ are
excluded by setting their counts to zero. Given repetition factor
$\gamma_l>1$, repeated tokens are downweighted in probability space as
\[
\widetilde{p}_l(v)
\propto
p_l(v)\gamma_l^{-c_l(v)} .
\]
Since $p_l(v)\propto \exp(\ell_l(v))$, this is implemented by the equivalent
logit shift
\[
\widetilde{\ell}_l(v)
=
\ell_l(v)-c_l(v)\log\gamma_l .
\]
The penalty therefore compounds with repeated use of the same token, leaves
unseen tokens unchanged, and modifies only the next-token sampling
distribution.

\paragraph{Step-dependent schedule.}
The sampling parameters are position dependent:
\[
(\tau_l,p_l,\gamma_l)
=
\begin{cases}
(\tau_{\mathrm{early}},p_{\mathrm{early}},\gamma_{\mathrm{early}}),
& l\le K_{\mathrm{early}},\\
(\tau_{\mathrm{late}},p_{\mathrm{late}},\gamma_{\mathrm{late}}),
& l>K_{\mathrm{early}}.
\end{cases}
\]
We use more stochastic early decoding and sharper later decoding:
\[
\tau_{\mathrm{early}}>\tau_{\mathrm{late}},
\qquad
p_{\mathrm{early}}>p_{\mathrm{late}},
\qquad
\gamma_{\mathrm{early}}<\gamma_{\mathrm{late}}.
\]

\section{Inference-Time Beam Search}
\label{app:beam-search-full}
\begin{algorithm}[t]
\caption{Inference-Time Beam Search with Repetition Control}
\label{alg:beam-search}
\begin{algorithmic}[1]
\Require Conditioning latent $Z$, decoder $Dec$, beam size $K$, valid encoder
length $T_x$, rate cap $\rho$, repetition factor $\gamma_{\mathrm{rep}}$,
optional length-normalization exponent $\alpha_{\mathrm{len}}$
\Ensure Decoded token sequence $\widehat{\mathcal{T}}$

\State Initialize beam set with one active hypothesis $[\mathrm{BOS}]$ of score
$0$ and $K-1$ inactive hypotheses of score $-\infty$
\State Set length cap $L_{\mathrm{cap}}\gets \lfloor \rho T_x\rfloor$

\For{$l=1,\ldots,L_{\mathrm{cap}}$}
    \State Initialize candidate set $\mathcal{C}\gets\emptyset$

    \For{each beam hypothesis $k=1,\ldots,K$}
        \If{beam $k$ is finished}
            \State Add the unchanged finished beam to $\mathcal{C}$
            \State \textbf{continue}
        \EndIf

        \State Compute next-token logits
        $\ell_l^{(k)}\gets Dec(\mathcal{T}^{(k)}_{<l},Z)$

        \State Count prefix occurrences $c_l^{(k)}(v)$ for each token $v$
        \State Set $c_l^{(k)}(v)=0$ for special symbols excluded from repetition control

        \State Apply repetition penalty:
        \[
        \widetilde{\ell}_l^{(k)}(v)
        \gets
        \ell_l^{(k)}(v)-c_l^{(k)}(v)\log\gamma_{\mathrm{rep}} .
        \]

        \State Disallow $\mathrm{PAD}$ for unfinished beams

        \If{$l=L_{\mathrm{cap}}$}
            \State Force $\mathrm{EOS}$ by masking all non-$\mathrm{EOS}$ tokens
        \EndIf

        \State Convert adjusted logits to log-probabilities:
        \[
        \log p_l^{(k)}(v)
        =
        \log\mathrm{softmax}(\widetilde{\ell}_l^{(k)})(v).
        \]

        \For{each vocabulary token $v$}
            \State Add candidate $\mathcal{T}^{(k)}\!\circ v$ with score
            $S^{(k)}+\log p_l^{(k)}(v)$ to $\mathcal{C}$
        \EndFor
    \EndFor

    \State Retain the top $K$ candidates from $\mathcal{C}$, using
    length-normalized ranking when $\alpha_{\mathrm{len}}>0$
    \State Mark beams ending in $\mathrm{EOS}$ as finished

    \If{all retained beams are finished}
        \State \textbf{break}
    \EndIf
\EndFor

\State Select the highest-ranking completed beam; if none is completed, select
the highest-ranking current beam
\State Remove $\mathrm{BOS}$ and return tokens up to and including
$\mathrm{EOS}$ if present
\end{algorithmic}
\end{algorithm}
This appendix gives the inference-time beam-search procedure. The role of beam
search, EOS constraints, and length normalization is described in
\S~\ref{sec:ar-target-generation-decoding}. Beam search uses the same compound
repetition penalty as Appendix~\ref{app:teacher-sampling-full}: repeated
non-special tokens receive the count-dependent logit shift
$-c_l(v)\log\gamma_{\mathrm{rep}}$ before beam expansion. In teacher sampling,
this penalty shapes the stochastic top-$p$ distribution; at inference, it
shapes deterministic beam scores. In both cases, it affects only next-token
selection and does not edit the decoded sequence post hoc.

\paragraph{Length-normalized ranking.}
When length normalization is enabled, a beam with cumulative log-probability
$S$ and decoded length $L$ is ranked using
$S_{\mathrm{rank}}
=
\frac{S}{L^{\alpha_{\mathrm{len}}}} $.
Setting $\alpha_{\mathrm{len}}=0$ recovers standard unnormalized beam search.
Length normalization changes only beam ranking and final selection; the beam
score remains the cumulative sum of token log-probabilities.

\section{Timing Recovery from Autoregressive Decoding}
\label{app:timing-recovery-full}

\begin{algorithm}[t]
\caption{Post-Hoc Timing Recovery from Decoder Cross-Attention}
\label{alg:timing-recovery}
\begin{algorithmic}[1]
\Require Input-window start time $t^{\mathrm{win\text{-}start}}$, decoded token
sequence $\mathcal{T}$, decoder--encoder cross-attention tensor, encoder-valid
mask, sampling rate $F_s$, encoder downsampling factor $M$, prior strength
$\omega$
\Ensure Time-stamped token sequence
$\{(u_l,t_l^{\mathrm{start}},t_l^{\mathrm{end}})\}_{l=1}^{L}$

\State Remove $\mathrm{BOS}$, $\mathrm{EOS}$, decoder padding, and padded
encoder frames
\State Average selected attention heads and layers to obtain
$W\in\mathbb{R}^{L\times T}$

\State Construct the diagonal Beta attention prior
$P\in\mathbb{R}^{L\times T}$

\State Apply prior reweighting:
\[
\widetilde{W}_{l,t}
=
\frac{
W_{l,t}P_{l,t}^{\omega}
}{
\sum_{t'=1}^{T}W_{l,t'}P_{l,t'}^{\omega}+\varepsilon
}.
\]

\State Convert token-to-frame attention into a frame-to-token posterior:
\[
A_{t,l}
=
\frac{
\widetilde{W}_{l,t}
}{
\sum_{l'=1}^{L}\widetilde{W}_{l',t}+\varepsilon
}.
\]

\State Decode a monotone Viterbi path $\pi^\ast$ through $A$:
\[
\pi^\ast
=
\arg\max_{\pi\in\mathcal{M}}
\sum_{t=1}^{T}
\log(A_{t,\pi_t+1}+\varepsilon),
\]
where $\mathcal{M}$ contains paths that either stay at the current token or
advance to the next token.

\State Set encoder frame duration $\Delta t=M/F_s$

\For{$l=1,\ldots,L$}
    \State Collect assigned frames
    $\mathcal{I}_l=\{t:\pi^\ast_t=l-1\}$
    \State Convert assigned frame support to timestamps:
    \[
    t_l^{\mathrm{start}}
    =
    t^{\mathrm{win\text{-}start}}
    +
    \left(\min\mathcal{I}_l-1\right)\Delta t,
    \qquad
    t_l^{\mathrm{end}}
    =
    t^{\mathrm{win\text{-}start}}
    +
    \max\mathcal{I}_l\,\Delta t .
    \]
\EndFor

\State Return
$\{(u_l,t_l^{\mathrm{start}},t_l^{\mathrm{end}})\}_{l=1}^{L}$
\end{algorithmic}
\end{algorithm}

This appendix describes the inference-time timestamp extraction procedure used
for the timing diagnostics in \S~\ref{sec:timing-recovery-results}. The
method is purely post hoc: it does not modify the training objective, introduce
an alignment loss, constrain the decoder during learning, or backpropagate
through recovered alignments. It uses decoder--encoder cross-attention as a
soft token-to-frame association signal and converts it into an approximate
monotone token segmentation.

\paragraph{Diagonal Beta attention prior.}
Let $W\in\mathbb{R}^{L\times T}$ denote the token-to-frame cross-attention
matrix after removing $\mathrm{BOS}$, $\mathrm{EOS}$, decoder padding, and
padded encoder frames, where $L$ is the number of decoded non-special tokens
and $T$ is the number of valid encoder frames. Since raw cross-attention may be
diffuse or multi-modal, we apply a weak diagonal prior before extracting a hard
monotone path.

For token position $l$ and encoder frame $t$, define normalized coordinates
\[
\bar{l}=\frac{l-1}{L-1},
\qquad
\bar{t}=\frac{t-1}{T-1},
\]
for $L,T>1$. We define a row-wise Beta-shaped prior over encoder frames as
\[
P_{l,t}
\propto
\bar{t}^{\,\alpha_l-1}
(1-\bar{t})^{\,\beta_l-1},
\qquad
\alpha_l = 1 + a\bar{l},
\qquad
\beta_l = 1 + b(1-\bar{l}),
\]
followed by row normalization:
$\sum_{t=1}^{T} P_{l,t}=1$.

This prior assigns earlier decoded token positions more mass near earlier
encoder frames and later decoded token positions more mass near later encoder
frames. The parameters $a$ and $b$ control the strength and asymmetry of this
diagonal preference.

We apply the prior multiplicatively:
\[
\widetilde{W}_{l,t}
=
\frac{
W_{l,t}P_{l,t}^{\omega}
}{
\sum_{t'=1}^{T}W_{l,t'}P_{l,t'}^{\omega}+\varepsilon
},
\]
where $\omega$ controls the strength of the post-hoc bias and $\varepsilon$ is
a numerical stabilizer. This reweighting is applied only during timestamp
extraction; it does not alter the learned decoder attention mechanism.

The recovered timestamps should be interpreted as approximate temporal
grounding for learned autoregressive token positions, not as supervised phone
boundaries or semantic labels for the token inventory. The method can summarize
near-monotone structure already present in decoder cross-attention, but it
cannot create reliable timing structure when the attention map is diffuse or
strongly non-monotone.

\subsection{Post-Hoc Timing Recoverability}
\label{sec:timing-recovery-results}

PairAlign emits compact autoregressive sequences rather than frame-synchronous
labels, so token timing must be recovered post hoc. We test whether decoder--encoder
cross-attention contains a usable temporal signal and whether a weak diagonal
Beta prior improves its recovery.

The prior is applied only after decoding, before monotone Viterbi alignment; it
reweights attention toward a weak forward diagonal but is absent from training
and leaves token identities and order unchanged. This analysis therefore probes
only approximate temporal grounding. Phonetic and speaker-content properties are
analyzed separately in \S~\ref{sec:phonetic_boundary_analysis}.

\paragraph{Experimental setting.}
The attention-internal analysis uses $1{,}000$ TIMIT anchors and their augmented
positives ($2{,}000$ decoded sequences), each a $3$ s window with $301$ valid
encoder frames. Mean decoded length is $26.66$ tokens for anchors and $24.89$
for positives. Cross-attention is converted to frame-to-token posteriors and
evaluated before and after the Beta prior of
Appendix~\ref{app:timing-recovery-full}.

We additionally compare recovered timing against sample-level TIMIT
\texttt{.PHN} boundaries. Because augmentation can alter local timing, this
external check uses only the $4{,}620$ clean $3$ s TIMIT segments. Both the
validation dataloader and evaluation script use fixed seed $13$.

\paragraph{Metrics.}
Because timestamps assign encoder frames to decoded token steps, diagnostics are
computed frame-to-token: expected/peak monotonicity violations, mean
absolute/peak diagonal error, and mean monotone-Viterbi log-score. Lower
violations/errors and higher Viterbi score indicate stronger compatibility with
a forward temporal path.

As an independent check, recovered token transitions are matched to TIMIT phone
boundaries at $10$, $20$, and $30$ ms tolerances. This measures timing quality
only and is not evidence that PairAlign tokens correspond to phones.


\begin{table*}[t]
\centering
\small
\setlength{\tabcolsep}{4.5pt}
\renewcommand{\arraystretch}{1.05}

\begin{tabularx}{\textwidth}{
    >{\raggedright\arraybackslash}X
    c
    c
    c
    c
    c
}
\toprule

\textbf{Frame-to-token metric}
&
\textbf{Raw}
&
\textbf{+ Beta}
&
\makecell{\textbf{$\Delta$}\\\textbf{(Beta $-$ Raw)}}
&
\makecell{\textbf{Median}\\\textbf{change}}
&
\makecell{\textbf{Relative}\\\textbf{gain (\%)}}
\\

\midrule

Expected violation rate $\downarrow$
    & $0.4734$
    & $\mathbf{0.3769}$
    & $-0.0966$
    & $-0.0467$
    & $+20.38$ \\

Peak violation rate $\downarrow$
    & $0.0609$
    & $\mathbf{0.0605}$
    & $-0.0004$
    & $0.0000$
    & $+0.66$ \\

Mean absolute diagonal error $\downarrow$
    & $0.2626$
    & $\mathbf{0.2492}$
    & $-0.0134$
    & $-0.0129$
    & $+5.10$ \\

Mean peak diagonal error $\downarrow$
    & $0.2862$
    & $\mathbf{0.2809}$
    & $-0.0053$
    & $-0.0011$
    & $+1.85$ \\

Mean Viterbi log-score $\uparrow$
    & $-1.4895$
    & $\mathbf{-1.4024}$
    & $+0.0870$
    & $+0.0792$
    & $+5.85$ \\

\bottomrule
\end{tabularx}

\caption{
Frame-to-token timing diagnostics over $1{,}000$ TIMIT anchors and their
augmented positives. Raw uses decoder cross-attention; \textbf{+ Beta} applies
the diagonal two-dimensional Beta prior. Expected violations decrease by
$20.38\%$, mean absolute diagonal error by $5.10\%$, and mean Viterbi
log-score improves by $5.85\%$, while peak metrics change little, indicating
redistribution around existing dominant assignments. Relative gain follows the
preferred metric direction: $100(\mathrm{Raw}-\mathrm{Beta})/|\mathrm{Raw}|$
for lower-is-better metrics and
$100(\mathrm{Beta}-\mathrm{Raw})/|\mathrm{Raw}|$ otherwise.
}
\label{tab:frame_to_token_timing_primary}
\end{table*}


\begin{table*}[t]
\centering
\small
\setlength{\tabcolsep}{5.5pt}
\renewcommand{\arraystretch}{1.05}

\begin{tabular}{lccccc}
\toprule

\textbf{Ground-truth timing metric}
&
\textbf{Raw}
&
\textbf{+ Beta}
&
\makecell{\textbf{$\Delta$}\\\textbf{(Beta $-$ Raw)}}
&
\makecell{\textbf{Paired 95\% CI}\\\textbf{for $\Delta$}}
&
\makecell{\textbf{Relative}\\\textbf{gain (\%)}}
\\

\midrule

Boundary F1 @ 10 ms $\uparrow$
    & $0.2934$
    & $\mathbf{0.2952}$
    & $+0.0018^{\dagger}$
    & $[+0.0008,+0.0027]$
    & $+0.61$ \\

Boundary F1 @ 20 ms $\uparrow$
    & $0.3732$
    & $\mathbf{0.3758}$
    & $+0.0025^{\dagger}$
    & $[+0.0017,+0.0035]$
    & $+0.70$ \\

Boundary F1 @ 30 ms $\uparrow$
    & $0.4170$
    & $\mathbf{0.4195}$
    & $+0.0025^{\dagger}$
    & $[+0.0017,+0.0034]$
    & $+0.60$ \\

Boundary MAE @ 20-ms tolerance $\downarrow$
    & $\mathbf{8.606}$ ms
    & $8.626$ ms
    & $+0.020$ ms
    & $[-0.022,+0.063]$ ms
    & $-0.23$ \\

\bottomrule
\end{tabular}

\caption{
External timing check on the $4{,}620$ clean $3$ s TIMIT segments using
sample-level \texttt{.PHN} boundaries. Token-transition times come from
monotone Viterbi decoding; F1 uses one-to-one matching at each tolerance and
MAE uses matches within $20$ ms. Beta yields small but reliable F1 gains at all
tolerances, while the $0.020$ ms MAE increase is not reliable. This validates
temporal recovery only; phonetic interpretation is deferred to
\S~\ref{sec:phonetic_boundary_analysis}. $\dagger$ denotes a paired
bootstrap $95\%$ CI excluding zero. Both validation-dataloader and
evaluation-script seeds are fixed to $13$.
}
\label{tab:pairalign_timit_boundary_alignment}
\end{table*}

\begin{figure*}[t]
\centering
\begin{subfigure}{0.48\textwidth}
    \centering
    \includegraphics[width=\linewidth]{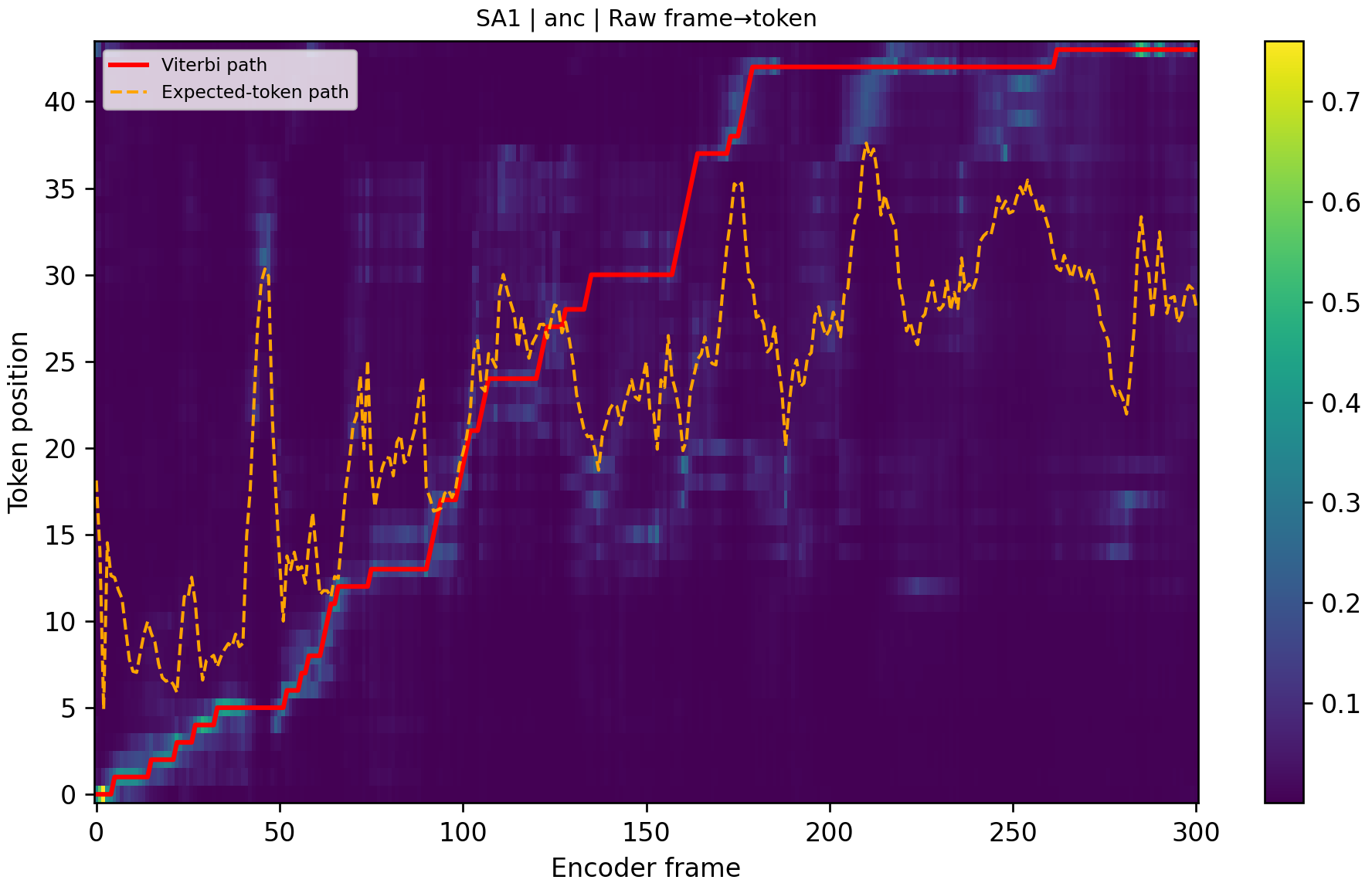}
    \caption{\texttt{SA1.wav}: raw posterior.}
    \label{fig:sa1_raw_frame_to_token}
\end{subfigure}
\hfill
\begin{subfigure}{0.48\textwidth}
    \centering
    \includegraphics[width=\linewidth]{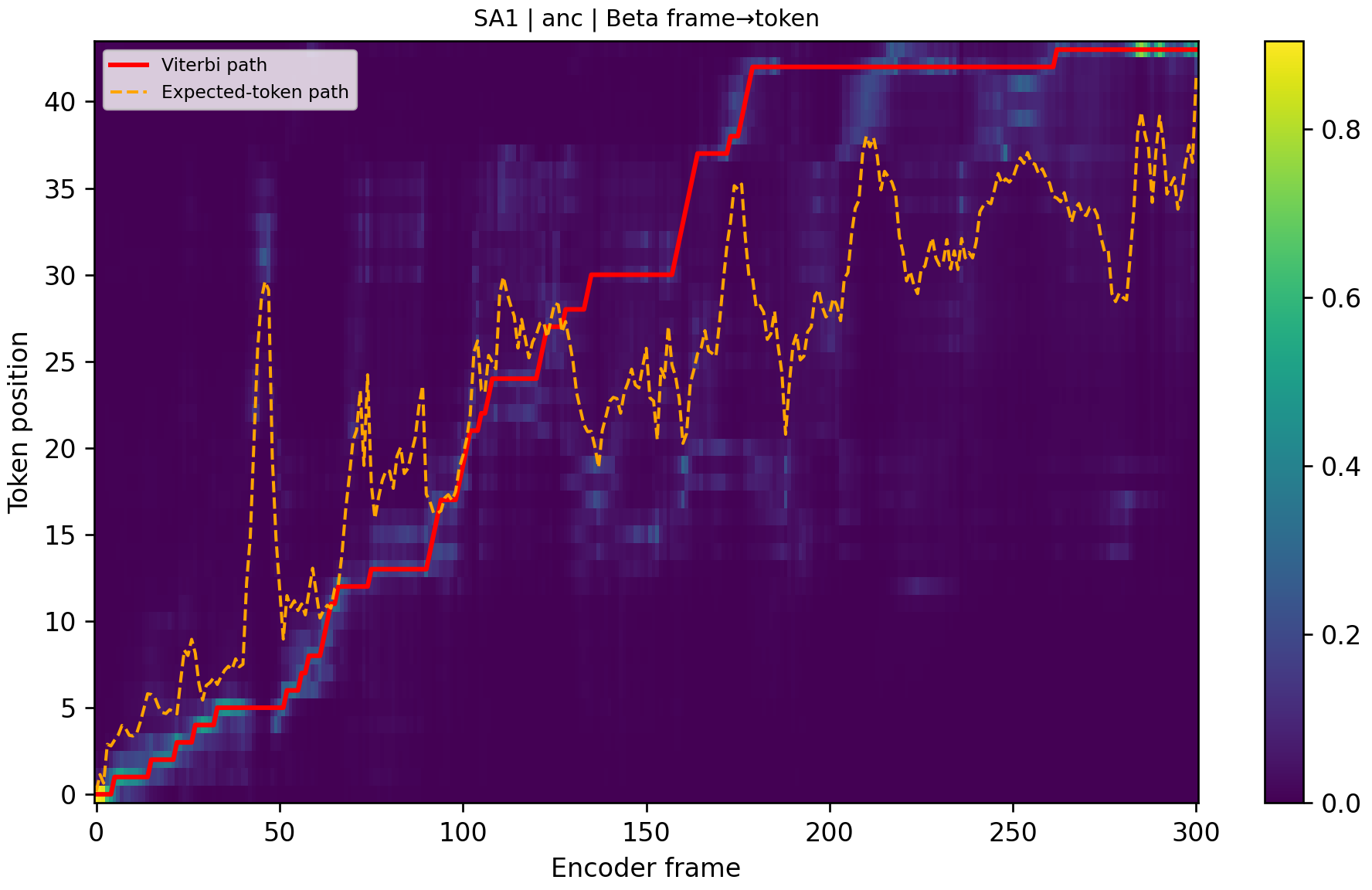}
    \caption{\texttt{SA1.wav}: Beta-reweighted posterior.}
    \label{fig:sa1_beta_frame_to_token}
\end{subfigure}

\vspace{0.8em}

\begin{subfigure}{0.48\textwidth}
    \centering
    \includegraphics[width=\linewidth]{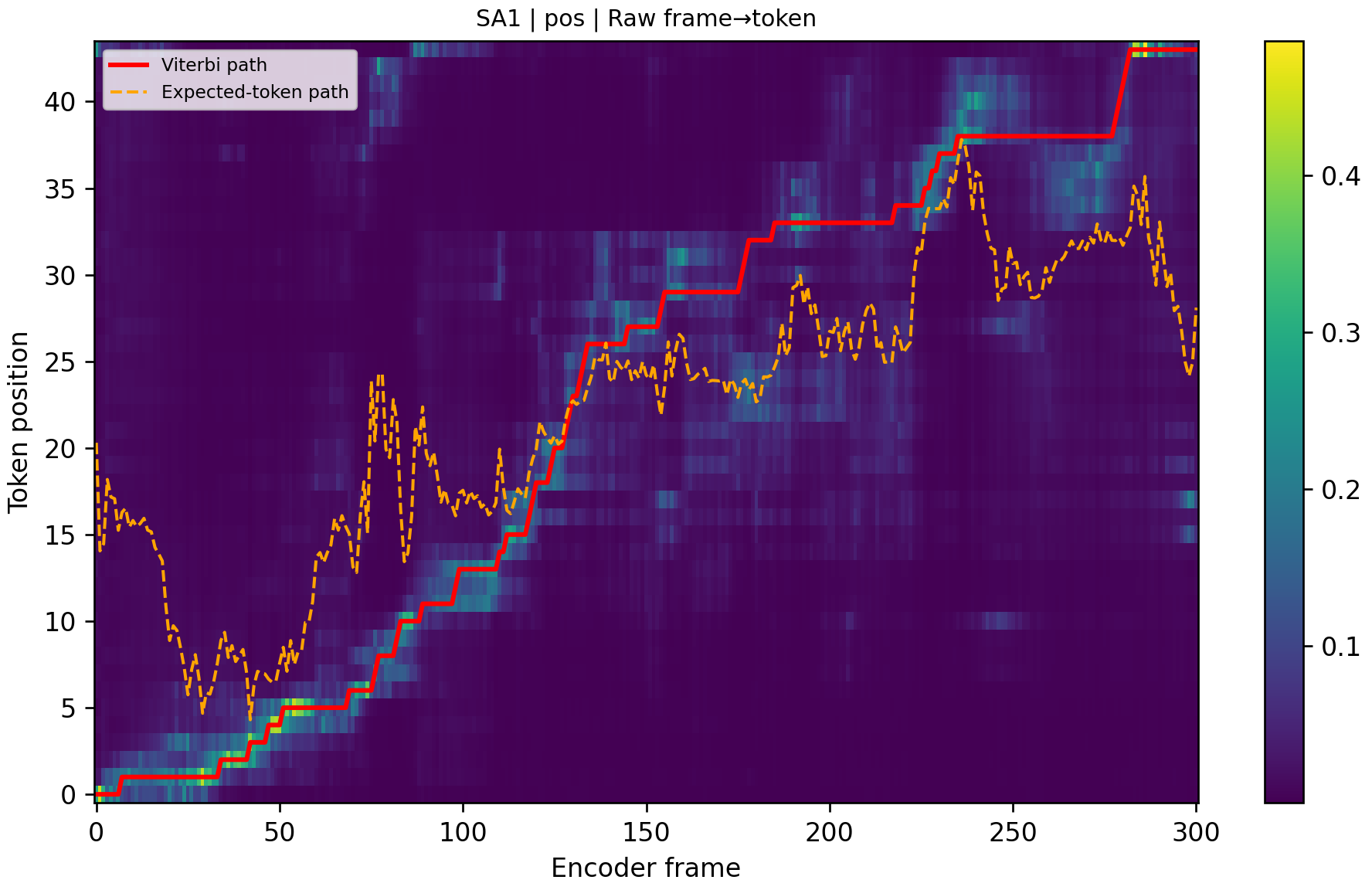}
    \caption{\texttt{SA1.wav} + augmentation: raw posterior.}
    \label{fig:sa1_aug_raw_frame_to_token}
\end{subfigure}
\hfill
\begin{subfigure}{0.48\textwidth}
    \centering
    \includegraphics[width=\linewidth]{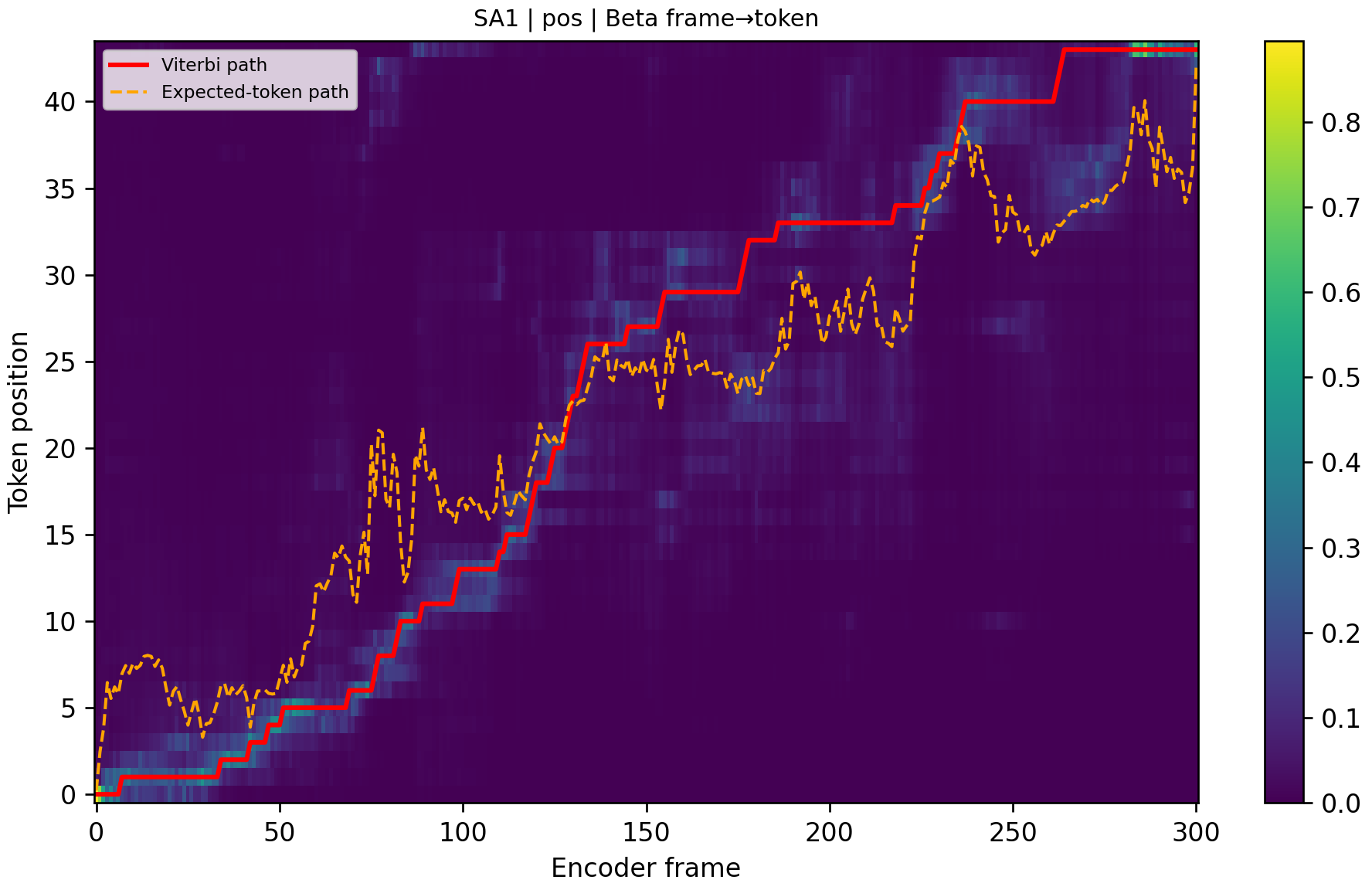}
    \caption{\texttt{SA1.wav} + augmentation: Beta-reweighted posterior.}
    \label{fig:sa1_aug_beta_frame_to_token}
\end{subfigure}

\caption{
Frame-to-token recovery for a $3$ s TIMIT segment and its augmented view.
Heatmaps show the frame-to-token posterior; solid/dashed curves denote the
monotone Viterbi and expected-token paths. Beta sharpens diagonal structure
without changing the decoded token sequence.
}
\label{fig:sa1_frame_to_token_timing}
\end{figure*}

\paragraph{Results.}
Beta primarily improves soft monotonic organization. Expected violation falls
from $0.4734$ to $0.3769$, mean absolute diagonal error from $0.2626$ to
$0.2492$, and mean Viterbi log-score improves from $-1.4895$ to $-1.4024$;
peak violation changes only from $0.0609$ to $0.0605$. The prior therefore
moves posterior mass toward a smoother monotone trajectory rather than replacing
dominant framewise assignments.

The external check yields the same, narrower conclusion. Boundary F1 improves
by about $0.6$--$0.7\%$ relative at $10$--$30$ ms, with paired bootstrap
intervals excluding zero, whereas the $0.020$ ms MAE increase has a confidence
interval spanning zero. Beta therefore makes recovered transitions slightly
more likely to fall near annotated boundaries without improving localization
error for already matched transitions.

Figure~\ref{fig:sa1_frame_to_token_timing} visualizes this effect: Beta sharpens
diagonal posterior structure and monotone-path support while preserving the
generated sequence. Recovered token times remain approximate and should not be
identified with phones, syllables, or other predefined linguistic units.

Overall, PairAlign cross-attention provides a usable but imperfect timing signal.
Beta improves its monotonic organization and slightly improves external boundary
agreement without altering the token sequence. We therefore use the Beta-aligned
path only for auxiliary temporal grounding, visualization, and alignment
analysis; phone, segmentation, and speaker structure are evaluated separately in
\S~\ref{sec:phonetic_boundary_analysis}.

\section{Full Experimental Details}
\label{app:full_experimental_details}

\subsection{Discrete Token Consistency}
\label{app:exp_discrete_token_consistency}

For each anchor segment $x_i$ and augmented positive view
$x_i^{+}=\mathrm{Aug}(x_i)$, we independently decode token sequences
$\mathcal{T}_i$ and $\mathcal{T}_i^{+}$. All metrics are averaged over
$N$ anchor--positive pairs.

\paragraph{Unigram Jaccard similarity.}
Let $\mathcal{G}^{(1)}(\mathcal{T})$ denote the set of unique tokens appearing
in sequence $\mathcal{T}$. We compute
\begin{equation}
\mathrm{Jaccard}(x_i,x_i^{+})
=
\frac{
\left|
\mathcal{G}^{(1)}(\mathcal{T}_i)
\cap
\mathcal{G}^{(1)}(\mathcal{T}_i^{+})
\right|
}{
\left|
\mathcal{G}^{(1)}(\mathcal{T}_i)
\cup
\mathcal{G}^{(1)}(\mathcal{T}_i^{+})
\right|
}.
\end{equation}
This measures unordered token-identity overlap and ignores order and repeated
occurrences.

\paragraph{Normalized edit similarity.}
Let $ED(\mathcal{T}_i,\mathcal{T}_i^{+})$ be the Levenshtein edit distance. We
define
\begin{equation}
\label{eq:norm_edit_similarity}
\mathrm{Sim}_{\mathrm{edit}}(\mathcal{T}_i,\mathcal{T}_i^{+})
=
1-
\frac{
ED(\mathcal{T}_i,\mathcal{T}_i^{+})
}{
\max\left\{|\mathcal{T}_i|,|\mathcal{T}_i^{+}|\right\}
}.
\end{equation}
This penalizes substitutions, insertions, deletions, ordering differences, and
length mismatch.

\paragraph{Exact match rate.}
We report
\begin{equation}
\mathrm{ExactMatch}
=
\frac{1}{N}
\sum_{i=1}^{N}
\mathbb{1}\left[\mathcal{T}_i=\mathcal{T}_i^{+}\right].
\end{equation}

\paragraph{Mean sequence length.}
We report the average decoded length over both streams:
\begin{equation}
\overline{L}
=
\frac{1}{2N}
\sum_{i=1}^{N}
\left(
|\mathcal{T}_i|+
|\mathcal{T}_i^{+}|
\right).
\end{equation}

\paragraph{Edit-operation decomposition.}
For an optimal Levenshtein alignment, let $S_i$, $I_i$, and $D_i$ denote the
numbers of substitutions, insertions, and deletions. Then
\begin{equation}
ED(\mathcal{T}_i,\mathcal{T}_i^{+})
=
S_i+I_i+D_i .
\end{equation}
We report these counts to separate token relabeling from token insertion or
deletion.

\subsection{Token Inventory and Collapse Analysis}
\label{app:exp_token_inventory_collapse}

Let $\mathcal{D}_{\mathrm{eval}}$ be an evaluation set and let
$\mathcal{T}_i=[\tau_{i,1},\ldots,\tau_{i,L_i}]$ denote the decoded token
sequence for example $i$. Let $\mathcal{A}$ be the tokenizer vocabulary.

\paragraph{Global token distribution.}
The empirical token frequency is
\begin{equation}
p(a)
=
\frac{
\sum_i \sum_{l=1}^{L_i}
\mathbb{1}[\tau_{i,l}=a]
}{
\sum_i L_i
},
\qquad a\in\mathcal{A}.
\end{equation}
We compute global entropy and normalized entropy as
\begin{equation}
H_{\mathrm{global}}
=
-\sum_{a\in\mathcal{A}} p(a)\log p(a),
\qquad
\widetilde{H}_{\mathrm{global}}
=
\frac{H_{\mathrm{global}}}{\log|\mathcal{A}|}.
\end{equation}
The effective vocabulary size is
\begin{equation}
V_{\mathrm{eff}}
=
\exp(H_{\mathrm{global}}).
\end{equation}

\paragraph{Active vocabulary and concentration.}
The active vocabulary size is
\begin{equation}
V_{\mathrm{active}}
=
\left|\{a\in\mathcal{A}:p(a)>0\}\right|,
\end{equation}
and the dead-token rate is
\begin{equation}
r_{\mathrm{dead}}
=
1-\frac{V_{\mathrm{active}}}{|\mathcal{A}|}.
\end{equation}
For the top-$q$ most frequent tokens, we report
\begin{equation}
M_q
=
\sum_{a\in \mathrm{Top}q}p(a).
\end{equation}
In the main results, we use this statistic for top-token concentration.

\paragraph{Native-position token usage.}
For absolute output position $l$, define
\begin{equation}
\mathcal{I}_l=\{i:L_i\geq l\}.
\end{equation}
The token distribution at position $l$ is
\begin{equation}
p_l(a)
=
\frac{
\sum_{i\in\mathcal{I}_l}
\mathbb{1}[\tau_{i,l}=a]
}{
|\mathcal{I}_l|
}.
\end{equation}
We compute
\begin{equation}
H_l
=
-\sum_{a\in\mathcal{A}}p_l(a)\log p_l(a),
\qquad
\widetilde{H}_l
=
\frac{H_l}{\log|\mathcal{A}|}.
\end{equation}
\

\paragraph{Relative-position token usage.}
To compare tokenizers with different sequence lengths, each position $l$ in
sequence $i$ is assigned to a relative-position bin
\begin{equation}
b(l,i)
=
\left\lfloor
B_{\mathrm{pos}}\frac{l-1}{L_i}
\right\rfloor,
\end{equation}
where $B_{\mathrm{pos}}$ is the number of bins. For each bin $b$, we compute
the corresponding token distribution $p_b(a)$, entropy $H_b$, and normalized
entropy $\widetilde{H}_b$ using the same
definitions as above.

\paragraph{Low-diversity collapse.}
For a decoded sequence $\mathcal{T}$,we define the unique-token ratio 
\begin{equation}
r_{\mathrm{uniq}}(\mathcal{T})
=
\frac{
|\mathrm{unique}(\mathcal{T})|
}{
|\mathcal{T}|
}.
\end{equation}
A sequence is marked low-diversity collapsed when
$r_{\mathrm{uniq}}(\mathcal{T})\leq 0.2$.

For anchor and positive streams, we report the fraction of decoded sequences
satisfying this criterion. We also report the collapsed-pair rate:
\begin{equation}
\mathrm{CollapsedPair}
=
\frac{1}{N}
\sum_{i=1}^{N}
\mathbb{1}
\left[
r_{\mathrm{uniq}}(\mathcal{T}_i)\leq 0.2
\;\;\mathrm{or}\;\;
r_{\mathrm{uniq}}(\mathcal{T}_i^{+})\leq 0.2
\right].
\end{equation}

\paragraph{Within-stream exact collisions.}
For a set of decoded sequences
$\mathcal{S}=\{\mathcal{U}_i\}_{i=1}^{N}$, let
$N_{\mathrm{unique}}(\mathcal{S})$ denote the number of distinct sequences in
the stream. We define the within-stream exact-collision rate as
\begin{equation}
\mathrm{Collision}(\mathcal{S})
=
\frac{N-N_{\mathrm{unique}}(\mathcal{S})}{N}
=
1-\frac{N_{\mathrm{unique}}(\mathcal{S})}{N}.
\end{equation}
This measures the fraction of redundant sequence realizations under exact
sequence equality. We report it separately for the anchor stream
$\mathcal{S}_{\mathrm{anchor}}=\{\mathcal{T}_i\}_{i=1}^{N}$ and the positive
stream
$\mathcal{S}_{\mathrm{positive}}=\{\mathcal{T}_i^{+}\}_{i=1}^{N}$.
These metrics distinguish desirable anchor--positive robustness from undesirable
many-to-one reuse of identical sequences across distinct examples.

\subsection{Long-Form Augmented-Source Segment Retrieval on Continuous Audio}
\label{app:exp_long_form_retrieval}

This appendix gives only the ranking and compactness metrics used for the
continuous-audio augmented-source retrieval experiments. The archive construction, query
construction, relevance notions, and token-space ranking protocol are defined
in the main text.

\paragraph{Recall@$K$.}
Let $\mathcal{R}_i$ be the set of relevant archive indices for query $i$, and
let $\mathrm{TopK}(i)$ be the top-$K$ retrieved archive indices. We compute
\begin{equation}
\mathrm{Recall@}K
=
\frac{1}{Q}
\sum_{i=1}^{Q}
\mathbb{1}
\left[
\mathcal{R}_i \cap \mathrm{TopK}(i) \neq \emptyset
\right].
\end{equation}

\paragraph{Mean reciprocal rank.}
Let $r_i$ be the rank of the first relevant archive segment for query $i$. We
compute
\begin{equation}
\mathrm{MRR}
=
\frac{1}{Q}
\sum_{i=1}^{Q}
\frac{1}{r_i}.
\end{equation}

\paragraph{Mean first relevant rank.}
We report
\begin{equation}
\mathrm{MeanFRR}
=
\frac{1}{Q}
\sum_{i=1}^{Q} r_i .
\end{equation}

\paragraph{Archive token count and token rate.}
For archive segment $n$, let $L_n=|\mathcal{T}_n|$. The total archive token
count and average segment length are
\begin{equation}
N_{\mathrm{tok}}
=
\sum_{n=1}^{N}L_n,
\qquad
\overline{L}
=
\frac{1}{N}\sum_{n=1}^{N}L_n .
\end{equation}
For segment duration $T_{\mathrm{ctx}}=3\,\mathrm{s}$, the token rate is
\begin{equation}
R_{\mathrm{tok}}
=
\frac{\overline{L}}{T_{\mathrm{ctx}}}.
\end{equation}

\paragraph{Compression and relative token reduction.}
Given a baseline token count $N_{\mathrm{tok}}^{\mathrm{base}}$ and a PairAlign
token count $N_{\mathrm{tok}}^{\mathrm{pa}}$, we compute
\begin{equation}
C_{\mathrm{tok}}
=
\frac{
N_{\mathrm{tok}}^{\mathrm{base}}
}{
N_{\mathrm{tok}}^{\mathrm{pa}}
},
\qquad
r_{\mathrm{red}}
=
1-
\frac{
N_{\mathrm{tok}}^{\mathrm{pa}}
}{
N_{\mathrm{tok}}^{\mathrm{base}}
}.
\end{equation}
We also report the minimum and maximum segment lengths and the serialized
archive size in the implemented cache format.

\subsection{Continuous-Sweep Tokenization Analysis}
\label{app:exp_continuous_sweep}

This appendix defines only sweep-specific quantities. Unigram Jaccard
similarity, normalized edit similarity, mean sequence length, edit distance, and
the substitution/insertion/deletion decomposition are defined in
Appendix~\ref{app:exp_discrete_token_consistency}.

For the $m$-th $3$ s sweep window, let
$\mathcal{T}^{(m)}=[\tau^{(m)}_1,\ldots,\tau^{(m)}_{L^{(m)}}]$ be the decoded
non-special token sequence, with length
$L^{(m)}=|\mathcal{T}^{(m)}|$. The adjacent sequence
$\mathcal{T}^{(m+1)}$ is decoded independently from the next sweep window,
whose start time is shifted by $100$ ms.

\paragraph{Adjacent length change.}
For each adjacent pair, we compute
\begin{equation}
|\Delta L|^{(m)}
=
\left|L^{(m+1)}-L^{(m)}\right|.
\end{equation}

\paragraph{Length ratio.}
We report the adjacent length ratio
\begin{equation}
\rho_L^{(m)}
=
\frac{
\min\left(L^{(m)},L^{(m+1)}\right)
}{
\max\left(L^{(m)},L^{(m+1)}\right)
}.
\end{equation}

\paragraph{Relative length change.}
The relative length change reported in
Table~\ref{tab:continuous_sweep_summary} is
\begin{equation}
\frac{
\frac{1}{N_{\mathrm{pairs}}}\sum_m |\Delta L|^{(m)}
}{
\frac{1}{N_{\mathrm{pairs}}}\sum_m L^{(m)}
}.
\end{equation}

\paragraph{Edit-operation rates.}
For each adjacent pair, let $S^{(m)}$, $I^{(m)}$, and $D^{(m)}$ denote the
substitution, insertion, and deletion counts in the optimal Levenshtein edit
script. We report length-normalized operation rates:
\begin{equation}
r_S^{(m)}
=
\frac{S^{(m)}}{\max\left(L^{(m)},L^{(m+1)}\right)},
\qquad
r_I^{(m)}
=
\frac{I^{(m)}}{\max\left(L^{(m)},L^{(m+1)}\right)},
\end{equation}
\begin{equation}
r_D^{(m)}
=
\frac{D^{(m)}}{\max\left(L^{(m)},L^{(m+1)}\right)}.
\end{equation}

\paragraph{Distributional summaries.}
All sweep statistics are aggregated over adjacent-window pairs. We report
means, standard deviations, medians, and thresholded fractions for the reported
quantities in Tables~\ref{tab:continuous_sweep_summary},
\ref{tab:continuous_sweep_distribution}, and
\ref{tab:continuous_sweep_edit_ops}.

\subsection{Architecture and Training Details}
\label{app:architecture-training}

Table~\ref{tab:architecture_training_summary} reports the architecture,
optimization, decoding, sampling, and implementation details for the controlled
PairAlign experiments. These details are deferred from the main text to preserve
readability while maintaining reproducibility.

\begin{table*}[t]
\centering
\scriptsize
\setlength{\tabcolsep}{4pt}
\begin{tabular}{p{0.18\textwidth}p{0.74\textwidth}}
\toprule
\textbf{Component} & \textbf{Configuration} \\
\midrule

Input
& $16$ kHz audio; $96$-bin log-Mel; $25$ ms window; $10$ ms hop; $3.0$ s
segments. \\

Training data
& LibriSpeech \texttt{train-clean-360}+\texttt{train-other-500} ($860$ h).
Training dataloader seed $240$ fixes the sampled speech-window stream.
No TIMIT or Shrutilipi-Tamil speech is used for training/adaptation. \\

Augmentation sources
& Positive-view nuisance conditions are sampled online from
MUSAN ($\sim6$ h assorted noise)~\citep{snyder2015musan},
ETSI background noise ($\sim0.39$ h binaural recordings)
~\citep{etsi202396}, and Aachen AIR measured RIRs
~\citep{jeub2009binaural}. Additional transforms include gain/filtering,
time masking, and mild time stretching. \\

Training exposure
& Stage~I: $\sim667$ sampled anchor-h; Stage~II: $\sim400$ h;
Stage~III: $\sim400$ h. Because all stages use training dataloader seed $240$,
later stages revisit the deterministic sampled speech stream rather than adding
independent unique speech hours; noise/RIR augmentation is resampled online. \\

Checkpoint selection
& Reported checkpoints selected on held-out LibriSpeech \texttt{test-clean};
no reported evaluation is performed on this split. \\

Seeds
& \textbf{Training dataloader/model seed:} $240$ for all controlled models.
\textbf{Clean consistency:} dataloader seed $1234$; evaluation-script seed
$13$ for Stage~I/Stage~I+, and $13/97/313$ for PairAlign.
\textbf{Retrieval, continuous sweep, timing/phonetic probes, retrieval
ablations, efficiency, and external-noise evaluation:} dataloader seed $13$;
evaluation-script seed $13$.
\textbf{Rate-controlled BPE:} dataloader seed $1234$; evaluation-script seed
$13$.
\textbf{Tamil consistency:} dataloader seed $1234$; evaluation-script seed
$13$ for Stage~I/Stage~I+ and $13/97/313$ for PairAlign.
\textbf{Tamil retrieval:} dataloader seed $13$; evaluation-script seed $13$. \\

Evaluation determinism
& Stage~I/Stage~I+: deterministic nearest-centroid VQ + run-length
deduplication. PairAlign: deterministic beam search; repeated evaluation seeds
probe numerical repeatability, not stochastic decoding. \\

Stage~I encoder
& Mamba2; $96\!\rightarrow\!512$ input projection; $8$ layers; state dim.\ $128$;
conv.\ width $4$; expansion $8$; head dim.\ $48$; $\ell_2$-normalized output. \\

Stage~I VQ
& $512$ codewords of dim.\ $512$; nearest-centroid assignment; run-length
deduplication; EMA codebook decay $0.7$. \\

Stages~II--III decoder
& Whisper-style AR Transformer; context $512$; hidden dim.\ $512$;
$8$ heads; $6$ layers; dropout $0.1$; vocabulary $516$
($512+\mathrm{BOS/EOS/PAD/MASK}$). \\

Conditioning
& Decoder causal self-attention + encoder cross-attention + feed-forward layers;
mean-pooled, normalized, projected encoder summary added to every decoder input. \\

Optimization
& Adam; weight decay $10^{-4}$; LR scheduling; decoder/base LR
$5\times10^{-5}$; Stage~III encoder LR $5\times10^{-6}$. \\

Stage~I training
& $\mathcal{L}_{\mathrm{contrast}}
+\lambda_{\mathrm{commit}}\mathcal{L}_{\mathrm{commit}}$,
$\lambda_{\mathrm{commit}}=0.1$; VQ centroids updated separately by EMA.
$10$ epochs, $5000$ steps/epoch, batch $16$; $\sim667$ sampled anchor-h. \\

Stage~II training
& Shared-anchor masked teacher forcing on deterministic Stage~I anchor targets; Stage~I encoder/codebook frozen.
 $6$ epochs, $5000$ steps/epoch, batch $16$;
$\sim400$ sampled anchor-h. \\

Stage~III training
& Adaptive EMA targets + paired self-alignment + hard in-batch likelihood
contrast over $15$ negatives + entropy regularization + repetition/length controls;
$\sim400$ sampled anchor-h. EMA update every $1000$ steps, decay $0.995$. \\

EMA target decoding
& $\rho=0.15$; global top-$p=0.88$, temperature $0.90$,
$\gamma_{\mathrm{rep}}=1.15$; first $4$ steps:
top-$p$ $0.97\!\rightarrow\!0.88$, temperature
$1.30\!\rightarrow\!1.00$, repetition penalty
$1.03\!\rightarrow\!1.30$. \\

Inference decoding
& Deterministic beam search; beam $4$; $\rho=0.15$; min cap $2$;
$\gamma_{\mathrm{rep}}=1.30$; $\alpha_{\mathrm{len}}=0$. \\

Anti-bypass
& Self-attention dropout: Stage~II $0.6\!\rightarrow\!0.2$,
Stage~III $0.2\!\rightarrow\!0.1$.
Prefix masking: Stage~II $0.6\!\rightarrow\!0.3$,
Stage~III $0.3\!\rightarrow\!0.1$.
Encoder-summary conditioning retained unless ablated. \\

Anti-collapse
& Hard in-batch likelihood contrast + entropy regularization + repetition-aware
teacher generation + length constraints. Entropy supplies a soft vocabulary-use
pressure; likelihood contrast supplies explicit sequence-level discrimination.
In the w/o-contrast ablation, entropy remains active but collapse occurs. \\

External-noise evaluation
& TIMIT + RMS-normalized Speech and Noise Dataset\textsuperscript{\ref{fn:speech-noise-dataset}} \texttt{noise\_only};
SNR $\mathcal{U}[15,30]$ dB; no additional augmentation; validation/eval
seed $13$. This corruption source is disjoint from the training augmentation
pools above. \\

Efficiency
& Batch $16$, NVIDIA L4. Report archive throughput/wall time,
CUDA-synchronized batch-amortized model and end-to-end latency, peak GPU memory,
and RapidFuzz edit-distance ranking cost. \\

\bottomrule
\end{tabular}
\caption{
Architecture, optimization, augmentation, training exposure, decoding, seeds,
and evaluation configuration for the controlled PairAlign experiments.
Stage~I, Stage~I+, and PairAlign share the controlled LibriSpeech training
pool; later stages revisit the seeded training stream while nuisance
augmentations are sampled online. HuBERT+VQ and WavLM+VQ are higher-exposure
external references rather than data- or compute-matched baselines.
}
\label{tab:architecture_training_summary}
\end{table*}

\section{Edit, Inventory, and Position-Entropy Diagnostics}
\label{app:token-inventory-diagnostics}

This appendix provides the diagnostic plots summarized in
\S~\ref{sec:token_consistency_results}. The goal is to check whether the
compact PairAlign strings remain meaningful symbolic sequences rather than
arising from unstable edit behavior, narrow vocabulary use, generic prefixes, or
position-wise entropy collapse. The diagnostics are complementary: edit
operations test how anchor--positive strings differ, global inventory statistics
test vocabulary concentration, and position-wise entropy tests whether token
usage remains diverse across the decoded trajectory.

\begin{figure*}[t]
\centering
\includegraphics[width=0.75\textwidth]{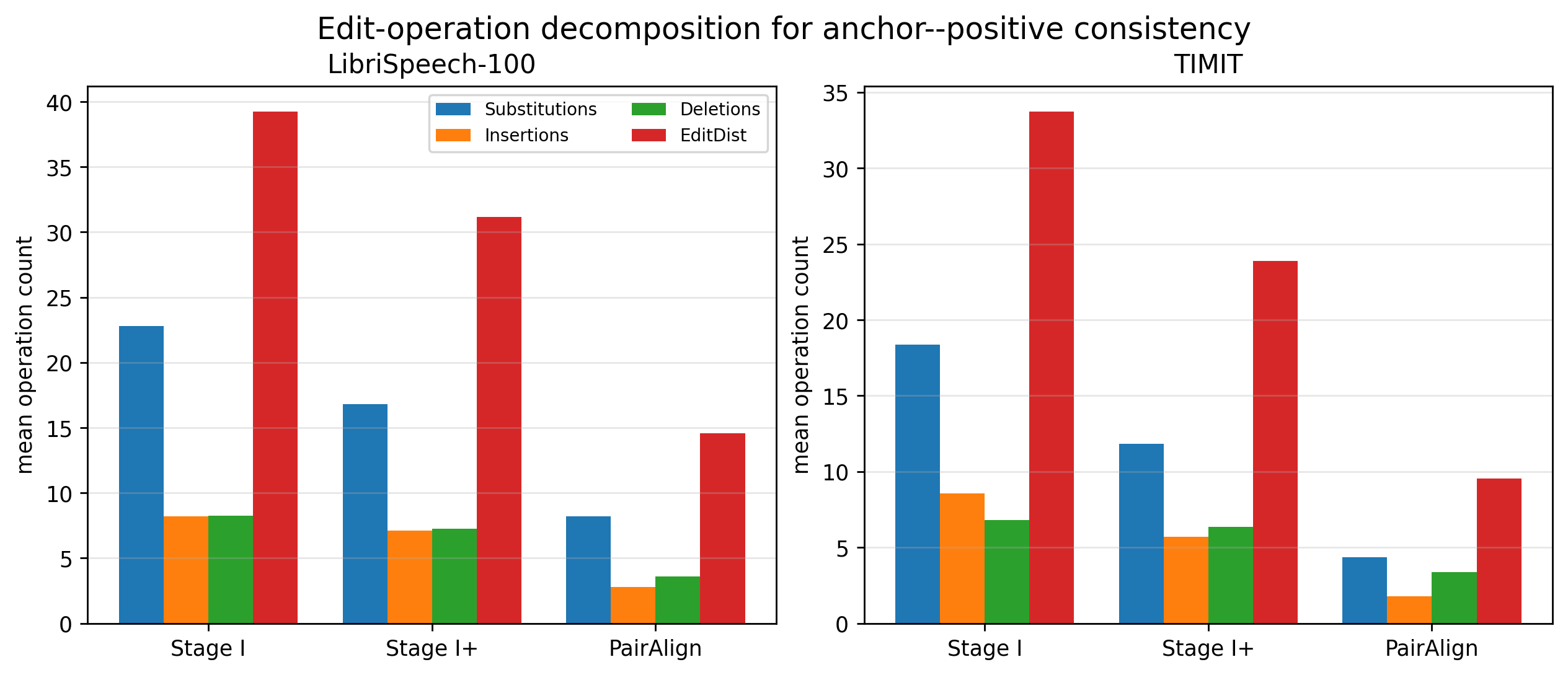}
\caption{Edit-operation decomposition for anchor--positive token consistency.
PairAlign requires fewer absolute edit operations across substitutions,
insertions, and deletions, indicating that compactness is not obtained through
unstable token birth/death.}
\label{fig:edit_ops_consistency}
\end{figure*}

\paragraph{Edit-operation decomposition.}
Figure~\ref{fig:edit_ops_consistency} decomposes normalized edit behavior into
absolute substitutions, insertions, and deletions. This distinguishes genuine
symbolic stability from a degenerate shortening strategy: a compact tokenizer
should not obtain lower edit distance merely by deleting unstable regions or by
creating large insertion/deletion imbalance across paired views.

\begin{figure*}[t]
\centering
\includegraphics[width=0.75\textwidth]{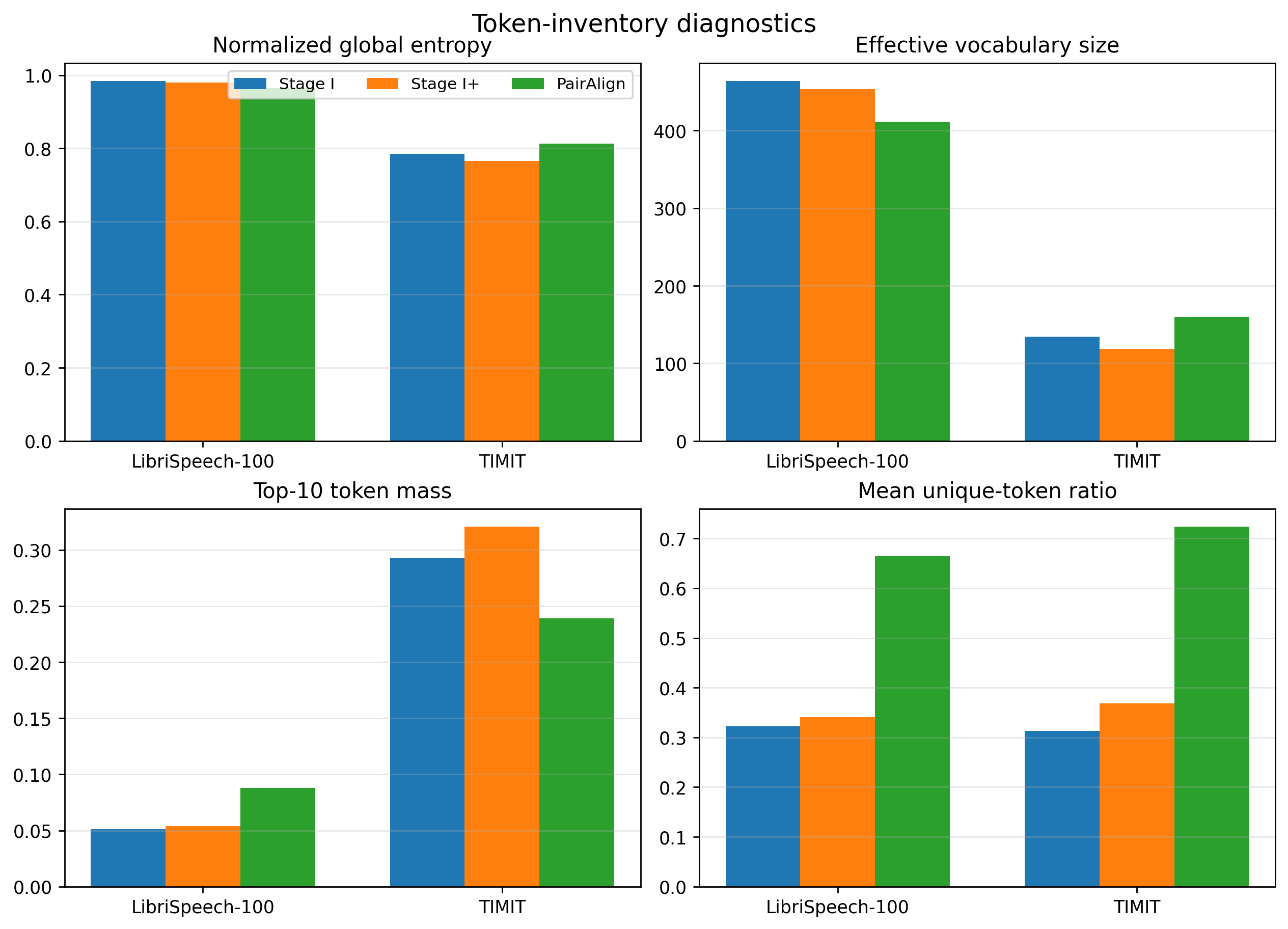}
\caption{Global token-inventory diagnostics. PairAlign remains broad-vocabulary
rather than collapsed: it uses the full vocabulary on LibriSpeech-100 and has
higher normalized entropy and effective vocabulary than the geometric systems on
TIMIT despite shorter sequences.}
\label{fig:inventory_diagnostics}
\end{figure*}

\paragraph{Global token-inventory usage.}
Figure~\ref{fig:inventory_diagnostics} tests whether compactness is achieved
without vocabulary collapse. High paired-view consistency is only meaningful if
the tokenizer remains input-dependent rather than assigning generic strings to
many examples. We therefore report normalized global token entropy, effective
vocabulary size, top-10 token mass, and mean unique-token ratio, which jointly
measure whether token usage is broad or concentrated in a small dominant set.

\begin{figure*}[t]
\centering
\includegraphics[width=0.75\textwidth]{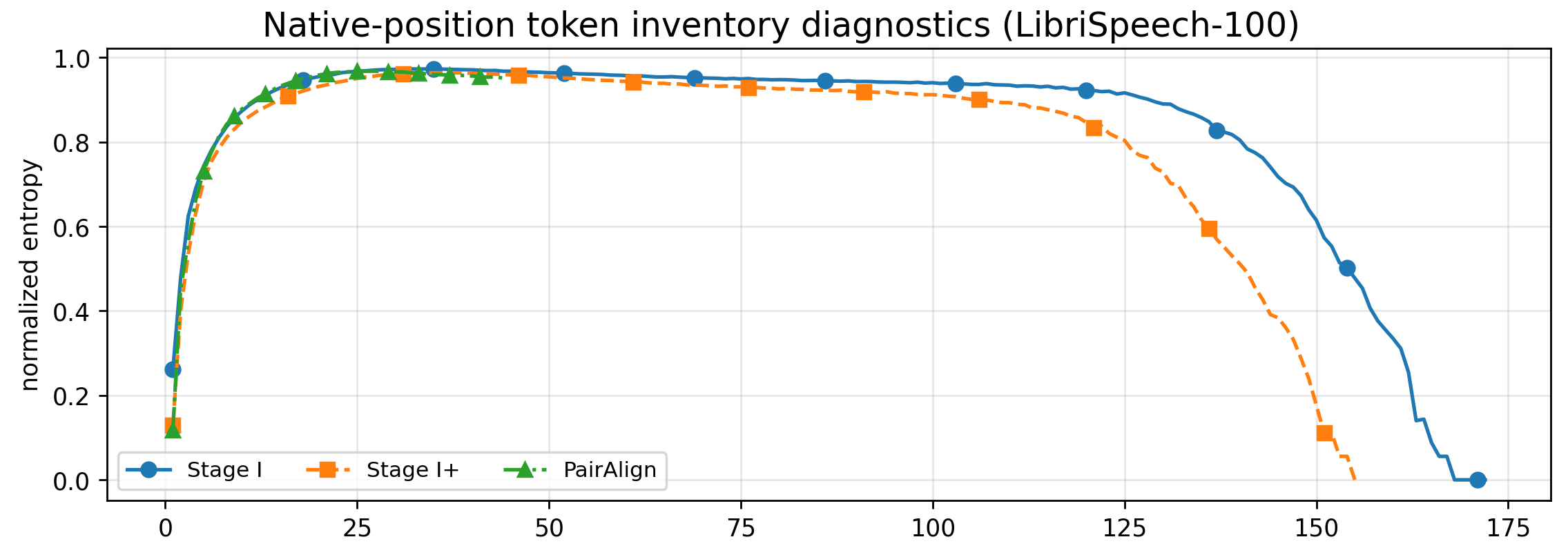}

\vspace{0.8em}

\includegraphics[width=0.75\textwidth]{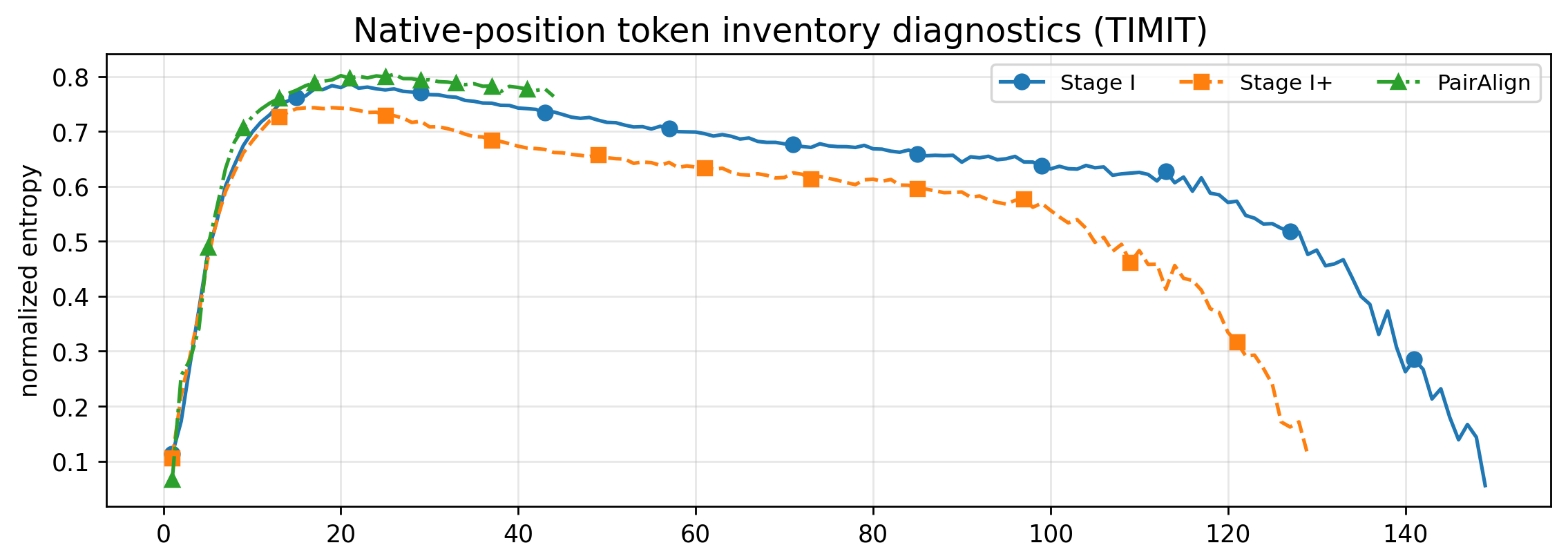}
\caption{Native-position token-entropy diagnostics on LibriSpeech-100 and
TIMIT. The top panel shows LibriSpeech-100 and the bottom panel shows TIMIT.
Models are shown at their own absolute output-length scales.}
\label{fig:native_position_diagnostics}
\end{figure*}
\paragraph{Native-position token entropy.}
Figure~\ref{fig:native_position_diagnostics} measures token entropy at each
absolute decoded position. These plots are shown at each model's own output
length scale, because the geometric tokenizers produce longer deduplicated
frame-derived streams while PairAlign produces shorter autoregressive strings.
The diagnostic tests whether early decoded positions become generic or
low-entropy in a way that would indicate prefix collapse.

\begin{figure*}[t]
\centering
\includegraphics[width=0.75\textwidth]{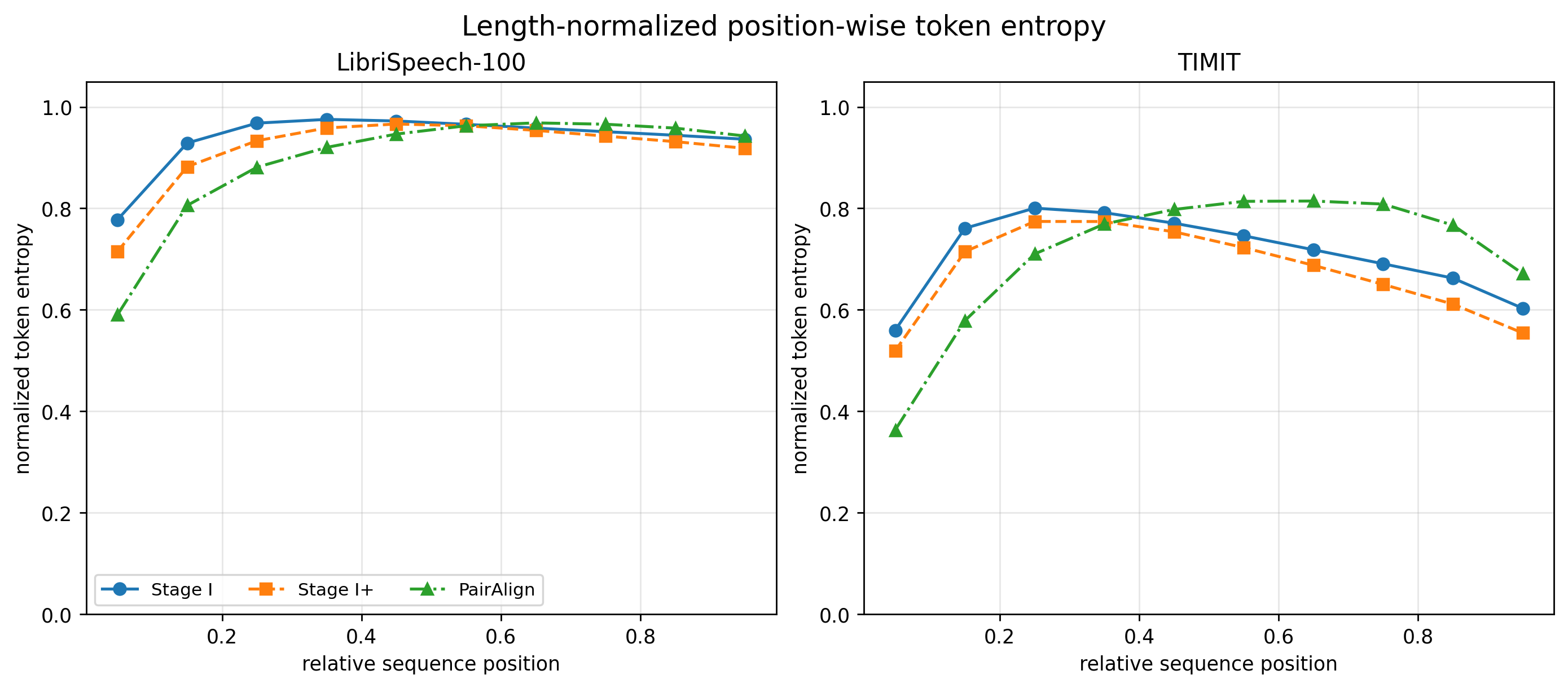}
\caption{Length-normalized position-wise token entropy. Relative-position bins
compare the beginning, middle, and end of each sequence independently of
absolute sequence length.}
\label{fig:relative_position_entropy}
\end{figure*}

\paragraph{Length-normalized position-wise token entropy.}
Figure~\ref{fig:relative_position_entropy} complements the native-position view
by mapping each decoded sequence to a common beginning--middle--end axis. This
removes the confound of absolute sequence length and tests whether diversity is
distributed across the compact sequence rather than concentrated only in a late
suffix.

\section{Positive--Negative Sequence-Geometry Diagnostics}
\label{app:sequence-geometry-diagnostics}

This appendix reports the detailed edit-distance geometry underlying the
analysis in \S~\ref{sec:token_consistency_results}. We include the
positive--negative distances, separation margins, and distance ratios for
phone-disjoint, trigram-disjoint, and temporally rearranged negatives under
both the standard augmented-view and external-noise conditions.

\begin{table*}[t]
\centering
\scriptsize
\setlength{\tabcolsep}{3.5pt}
\begin{tabular}{lccccccc}
\toprule

\multirow{3}{*}{\textbf{Variant}}
&
\multirow{3}{*}{\textbf{\# Pairs}}
&
\textbf{Mean Sequence}
&
\multicolumn{3}{c}{\textbf{Normalized Edit Distance}}
&
\textbf{Separation}
&
\textbf{Negative/Positive}
\\

&
&
\textbf{Length}
&
$\boldsymbol{d(A,P)}$
&
$\boldsymbol{d(A,N)}$
&
$\boldsymbol{d(P,N)}$
&
\textbf{Margin}
&
\textbf{Distance Ratio}
\\

&
&
\textbf{(A/P/N)}
&
$\boldsymbol{\downarrow}$
&
$\boldsymbol{\uparrow}$
&
$\boldsymbol{\uparrow}$
&
$\boldsymbol{d(A,N)-d(A,P)\uparrow}$
&
$\boldsymbol{d(A,N)/d(A,P)\uparrow}$
\\

\midrule

\multicolumn{8}{p{0.97\textwidth}}{
\textbf{Dataset and positive condition: TIMIT with the original
content-preserving augmented positive view.}
}
\\

\midrule

\multicolumn{8}{l}{
\textbf{Phone-disjoint negatives:}
no shared non-silence phone labels between anchor and negative
}
\\
\addlinespace[2pt]

Stage~I Geometric
    & $848$
    & $68.71/73.00/39.86$
    & $0.374$
    & $\mathbf{0.875}$
    & $\mathbf{0.881}$
    & $0.500$
    & $2.336{\times}$ \\

Stage~I+ Geometric
    & $848$
    & $51.99/53.60/31.54$
    & $0.343$
    & $0.872$
    & $0.872$
    & $\mathbf{0.529}$
    & $2.546{\times}$ \\

PairAlign
    & $848$
    & $25.72/24.20/18.25$
    & $\mathbf{0.288}$
    & $0.801$
    & $0.799$
    & $0.513$
    & $\mathbf{2.782{\times}}$ \\

\midrule

\multicolumn{8}{l}{
\textbf{Trigram-disjoint negatives:}
no shared non-silence phone trigrams between anchor and negative
}
\\
\addlinespace[2pt]

Stage~I Geometric
    & $4,620$
    & $77.78/79.13/77.24$
    & $0.384$
    & $\mathbf{0.910}$
    & $\mathbf{0.909}$
    & $0.527$
    & $2.372{\times}$ \\

Stage~I+ Geometric
    & $4,620$
    & $58.84/58.56/57.44$
    & $0.356$
    & $0.896$
    & $0.893$
    & $0.539$
    & $2.515{\times}$ \\

PairAlign
    & $4,620$
    & $26.63/24.84/26.23$
    & $\mathbf{0.309}$
    & $0.871$
    & $0.870$
    & $\mathbf{0.562}$
    & $\mathbf{2.819{\times}}$ \\

\midrule

\multicolumn{8}{l}{
\textbf{Rearranged-context negatives:}
12 temporal chunks from the anchor are permuted to alter their ordering
}
\\
\addlinespace[2pt]

Stage~I Geometric
    & $4,620$
    & $77.78/79.13/82.89$
    & $0.384$
    & $0.725$
    & $0.740$
    & $0.342$
    & $1.890{\times}$ \\

Stage~I+ Geometric
    & $4,620$
    & $58.84/58.56/59.66$
    & $0.356$
    & $0.711$
    & $0.720$
    & $0.355$
    & $1.997{\times}$ \\

PairAlign
    & $4,620$
    & $26.63/24.84/24.51$
    & $\mathbf{0.309}$
    & $\mathbf{0.793}$
    & $\mathbf{0.790}$
    & $\mathbf{0.484}$
    & $\mathbf{2.568{\times}}$ \\

\midrule
\midrule

\multicolumn{8}{p{0.97\textwidth}}{
\textbf{Dataset and positive condition: TIMIT + External Noise.}
The positive is formed by mixing the clean anchor with an RMS-normalized
external \texttt{noise\_only} recording at an SNR uniformly sampled between
$15$ and $30$ dB. The negative receives the same external noise segment at
the same SNR. No additional augmentation is applied.
}
\\

\midrule

\multicolumn{8}{l}{
\textbf{Phone-disjoint negatives:}
no shared non-silence phone labels between anchor and negative
}
\\
\addlinespace[2pt]

Stage~I Geometric
    & $848$
    & $73.29/91.47/86.32$
    & $0.594$
    & $0.959$
    & $0.933$
    & $0.365$
    & $1.615{\times}$ \\

Stage~I+ Geometric
    & $848$
    & $59.81/71.61/62.28$
    & $0.505$
    & $0.937$
    & $0.919$
    & $0.432$
    & $1.855{\times}$ \\

PairAlign
    & $848$
    & $33.52/32.74/23.26$
    & $0.444$
    & $0.890$
    & $0.871$
    & $0.446$
    & $2.003{\times}$ \\

\midrule

\multicolumn{8}{l}{
\textbf{Trigram-disjoint negatives:}
no shared non-silence phone trigrams between anchor and negative
}
\\
\addlinespace[2pt]

Stage~I Geometric
    & $4,620$
    & $82.73/91.80/91.99$
    & $0.541$
    & $0.962$
    & $0.948$
    & $0.420$
    & $1.777{\times}$ \\

Stage~I+ Geometric
    & $4,620$
    & $66.87/72.75/72.45$
    & $0.470$
    & $0.950$
    & $0.940$
    & $0.480$
    & $2.023{\times}$ \\

PairAlign & $4,620$ & $35.00/34.16/33.98$ & ${0.446}$ & $0.931$ & $0.921$ & ${0.485}$ & ${2.088{\times}}$ \\
\midrule

\multicolumn{8}{l}{
\textbf{Rearranged-context negatives:}
12 temporal chunks from the anchor are permuted to alter their ordering
}
\\
\addlinespace[2pt]

Stage~I Geometric
    & $4,620$
    & $82.73/91.80/98.02$
    & $0.541$
    & $0.808$
    & $0.744$
    & $0.267$
    & $1.493{\times}$ \\

Stage~I+ Geometric
    & $4,620$
    & $66.87/72.75/77.51$
    & $0.470$
    & $0.803$
    & $0.773$
    & $0.334$
    & $1.711{\times}$ \\

PairAlign & $4,620$ & $35.00/34.16/34.39$ & ${0.446}$ & ${0.878}$ & ${0.866}$ & ${0.432}$ & ${1.968{\times}}$ \\
\bottomrule
\end{tabular}

\caption{
Edit-distance geometry of Stage~I Geometric, Stage~I+ Geometric, and
PairAlign token sequences on 3-second TIMIT segments under three negative
settings and two positive-view conditions. Here, $A$, $P$, and $N$ denote the
clean anchor, its content-preserving positive, and the corresponding negative,
respectively. In the external-noise condition, the positive is formed by
mixing the anchor with an RMS-normalized external \texttt{noise\_only}
recording at an SNR uniformly sampled between $15$ and $30$ dB. The negative
receives the same noise segment at the same SNR, ensuring that negative
separation cannot be attributed to mismatched noise conditions. Phone-disjoint
negatives share no non-silence phone labels with the anchor and could be
constructed for $848$ of $4,620$ anchors. Trigram-disjoint negatives share no
contiguous non-silence phone trigrams with the anchor and could be constructed
for all $4,620$ anchors. For both PHN-based settings, \texttt{h\#},
\texttt{pau}, and \texttt{epi} are excluded when measuring phone overlap.
Rearranged-context negatives are constructed by applying a Sattolo permutation
to $12$ temporal chunks of the acoustic input and then tokenizing the resulting
reordered signal with the complete tokenizer. This preserves the constituent
acoustic material while disrupting its temporal organization and therefore
directly probes sensitivity to the acoustic conditioning input.
Distances are Levenshtein edit distances normalized by the maximum length of
the two sequences. The separation margin is $d(A,N)-d(A,P)$, and the distance
ratio is $d(A,N)/d(A,P)$. All tokenizer variants are evaluated with evaluation
seed $13$. For clean TIMIT, the validation dataloader uses a fixed deterministic
seed of $1234$. For TIMIT + External Noise, the validation-view seed is fixed
to $13$, matching the evaluation seed. Bold values indicate the best result within each
negative setting when all three tokenizer variants are available.
}
\label{tab:pairalign_timit_edit_geometry}
\end{table*}

\section{Unconstrained-Decoding and Termination Diagnostics}
\label{app:termination-diagnostics}

This appendix provides the detailed sequence-length and termination statistics
for the unconstrained-decoding analysis discussed in
\S~\ref{sec:token_consistency_results}. During training, $\rho$ controls
the maximum target rate and therefore limits the target length to at most
$44$ tokens for the $301$-frame conditioning context used here. At evaluation,
this inference-time restriction is removed entirely, allowing us to test whether
PairAlign nevertheless learns to predict EOS within the sequence-length support
encountered during training.



\begin{figure*}[t]
    \centering

    \begin{subfigure}[t]{0.485\textwidth}
        \centering
        \includegraphics[
            width=\linewidth,
            keepaspectratio
        ]{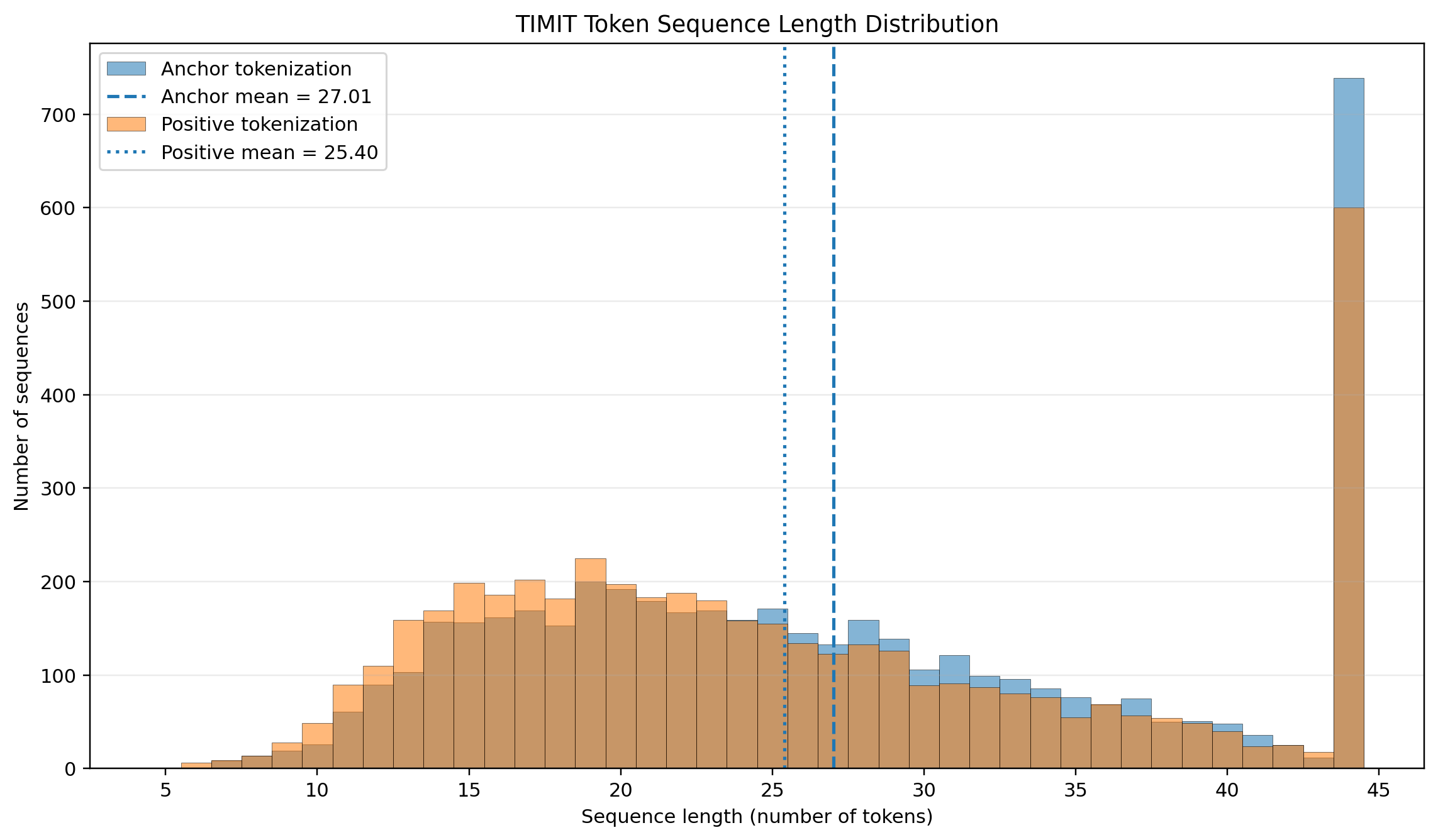}
        \caption{TIMIT}
        \label{fig:timit-unconstrained-length}
    \end{subfigure}
    \hfill
    \begin{subfigure}[t]{0.485\textwidth}
        \centering
        \includegraphics[
            width=\linewidth,
            keepaspectratio
        ]{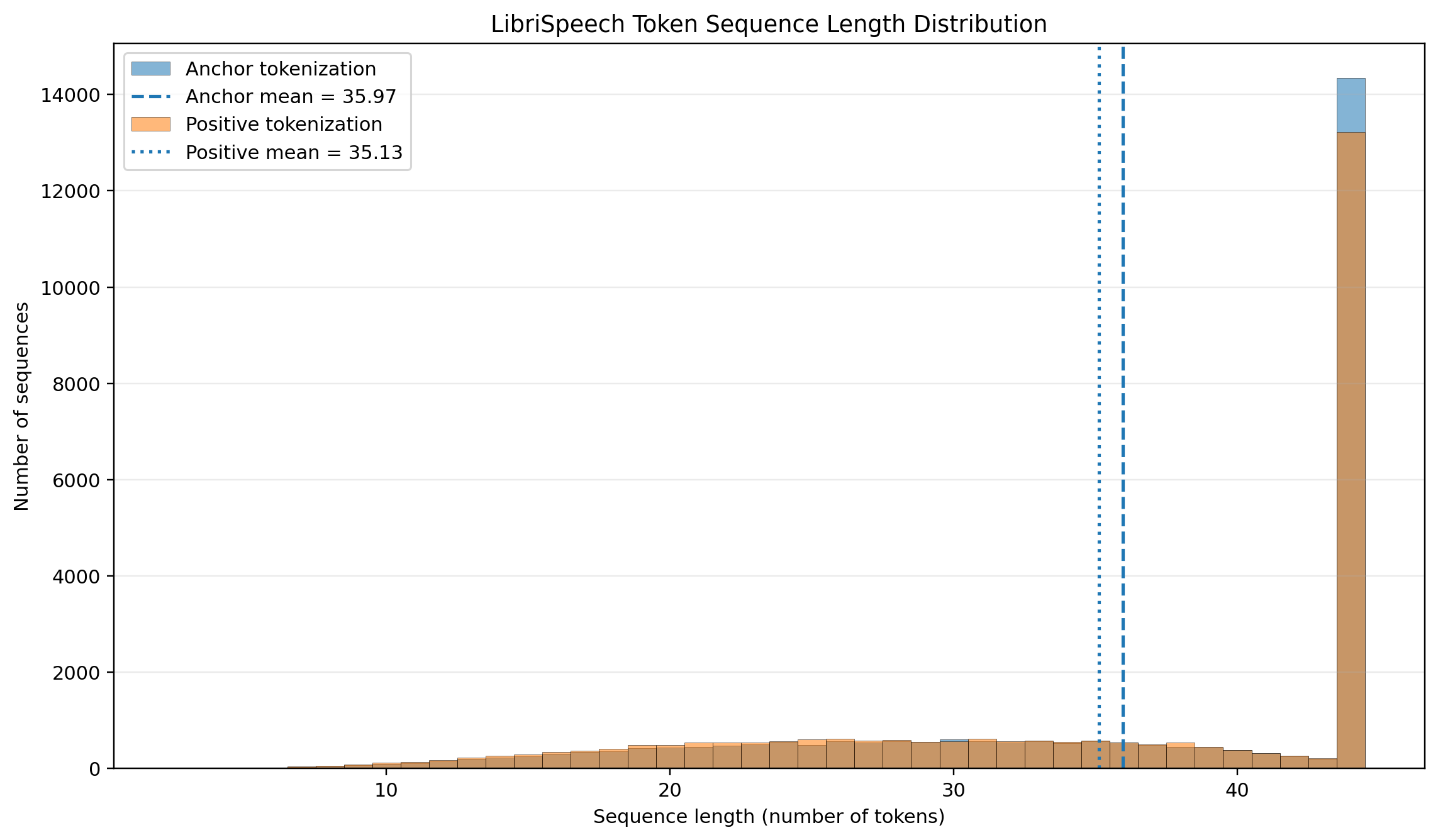}
        \caption{LibriSpeech-100}
        \label{fig:librispeech-unconstrained-length}
    \end{subfigure}

    \caption{
    Token-sequence length distributions under unconstrained PairAlign
    inference on TIMIT and LibriSpeech-100. Both datasets use fixed
    deterministic validation dataloader seed $1234$, and evaluation uses seed
    $13$. During training, $\rho$ limits the maximum target length to $44$
    tokens for the $301$-frame conditioning context. At evaluation, the
    $\rho$-based inference restriction is removed, EOS is permitted at every
    decoding step, and the decoding horizon is extended to $301$ steps.
    Consequently, termination both before and exactly at length $44$ is
    model-predicted. No sequence exceeds $44$ tokens, no sequence reaches the
    $301$-step horizon, and the forced-EOS rate is $0\%$. The minimum observed
    sequence length is $6$ tokens.
    }
    \label{fig:unconstrained-token-length-distributions}
\end{figure*}


\begin{table*}[t]
\centering
\scriptsize
\setlength{\tabcolsep}{4.5pt}
\renewcommand{\arraystretch}{1.08}

\begin{tabular}{llrrrrrr}
\toprule

\textbf{Dataset}
&
\textbf{Sequence}
&
\textbf{Number}
&
\makecell{\textbf{Mean}\\\textbf{length}}
&
\makecell{\textbf{EOS}\\$\mathbf{<44}$ (\%)}
&
\makecell{\textbf{EOS}\\$\mathbf{=44}$ (\%)}
&
\makecell{\textbf{EOS}\\$\mathbf{>44}$ (\%)}
&
\makecell{\textbf{Forced}\\\textbf{EOS} (\%)}
\\

\midrule

\multirow{3}{*}{TIMIT}
& Anchor
& 4,620
& 27.01
& 84.00
& 16.00
& 0.00
& 0.00 \\

& Positive
& 4,620
& 25.40
& 87.01
& 12.99
& 0.00
& 0.00 \\

& Exact anc $=$ pos
& 1,391
& 26.41
& 84.18
& 15.82
& 0.00
& 0.00 \\

\midrule

\multirow{3}{*}{LibriSpeech-100}
& Anchor
& 28,539
& 35.97
& 49.73
& 50.27
& 0.00
& 0.00 \\

& Positive
& 28,539
& 35.13
& 53.66
& 46.34
& 0.00
& 0.00 \\

& Exact anc $=$ pos
& 8,295
& 35.30
& 51.54
& 48.46
& 0.00
& 0.00 \\

\bottomrule
\end{tabular}

\caption{
Termination statistics under unconstrained PairAlign inference over
$4{,}620$ anchor--positive pairs for TIMIT and $28{,}539$ pairs for
LibriSpeech-100. Both datasets use fixed deterministic validation dataloader
seed $1234$, and evaluation uses seed $13$. During training, $\rho$ limits the
maximum target length to $44$ tokens for a $301$-frame conditioning context.
At evaluation, this inference-time restriction is removed, EOS is available at
every decoding step, and the decoding horizon is extended to $301$ steps.
Hence, both EOS $<44$ and EOS $=44$ correspond to model-predicted termination;
length $44$ is not an active inference-time stopping boundary. No sequence
exceeds $44$ tokens and the forced-EOS rate is $0\%$. ``Exact anc $=$ pos''
denotes independently decoded anchor--positive pairs with identical token
strings; these comprise $1{,}391/4{,}620$ pairs ($30.11\%$) on TIMIT and
$8{,}295/28{,}539$ pairs ($29.07\%$) on LibriSpeech-100.
}
\label{tab:unconstrained-termination}
\end{table*}


\begin{figure*}[t]
    \centering

    \begin{subfigure}[t]{0.485\textwidth}
        \centering
        \includegraphics[
            width=\linewidth,
            keepaspectratio
        ]{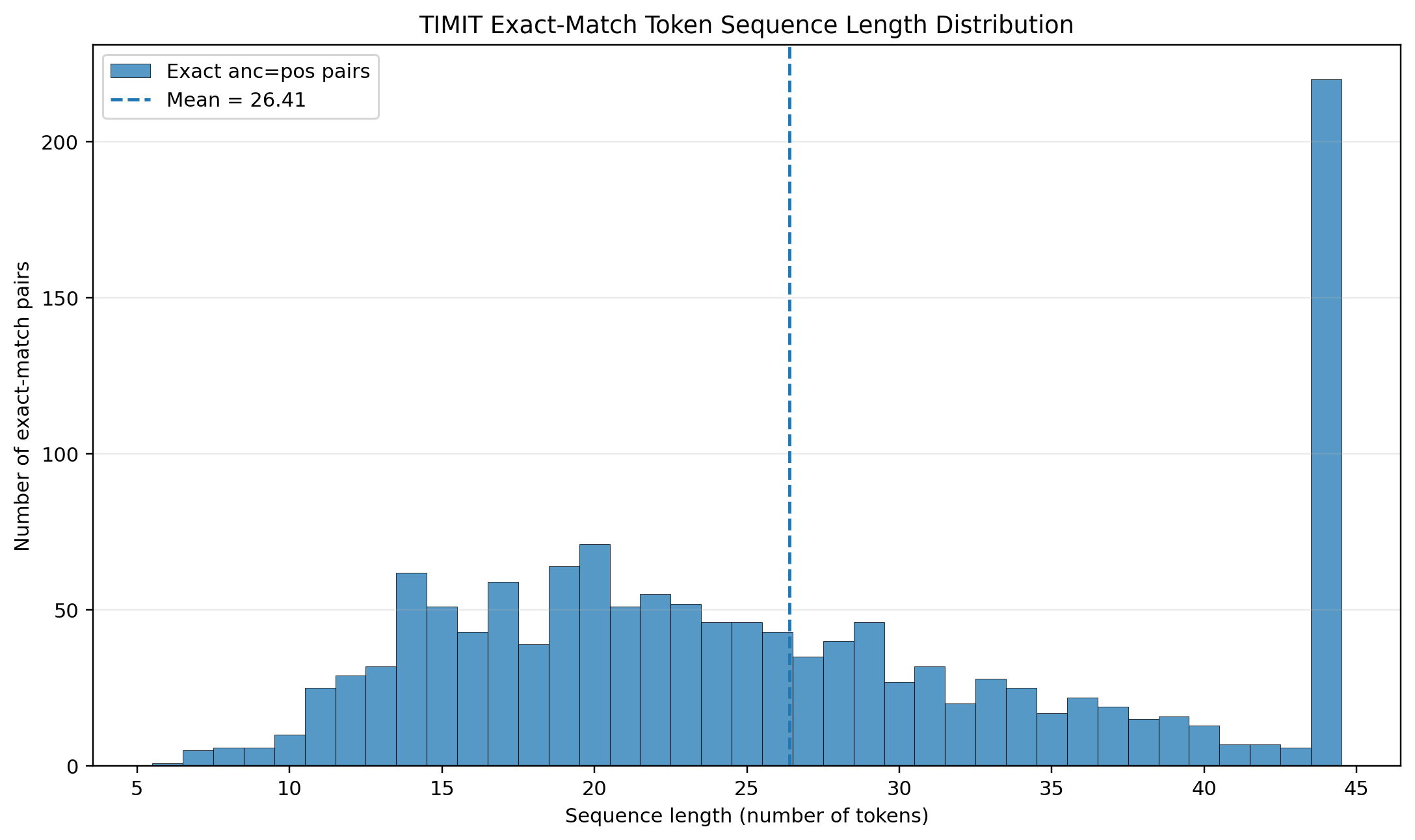}
        \caption{TIMIT}
        \label{fig:timit-exact-match-length}
    \end{subfigure}
    \hfill
    \begin{subfigure}[t]{0.485\textwidth}
        \centering
        \includegraphics[
            width=\linewidth,
            keepaspectratio
        ]{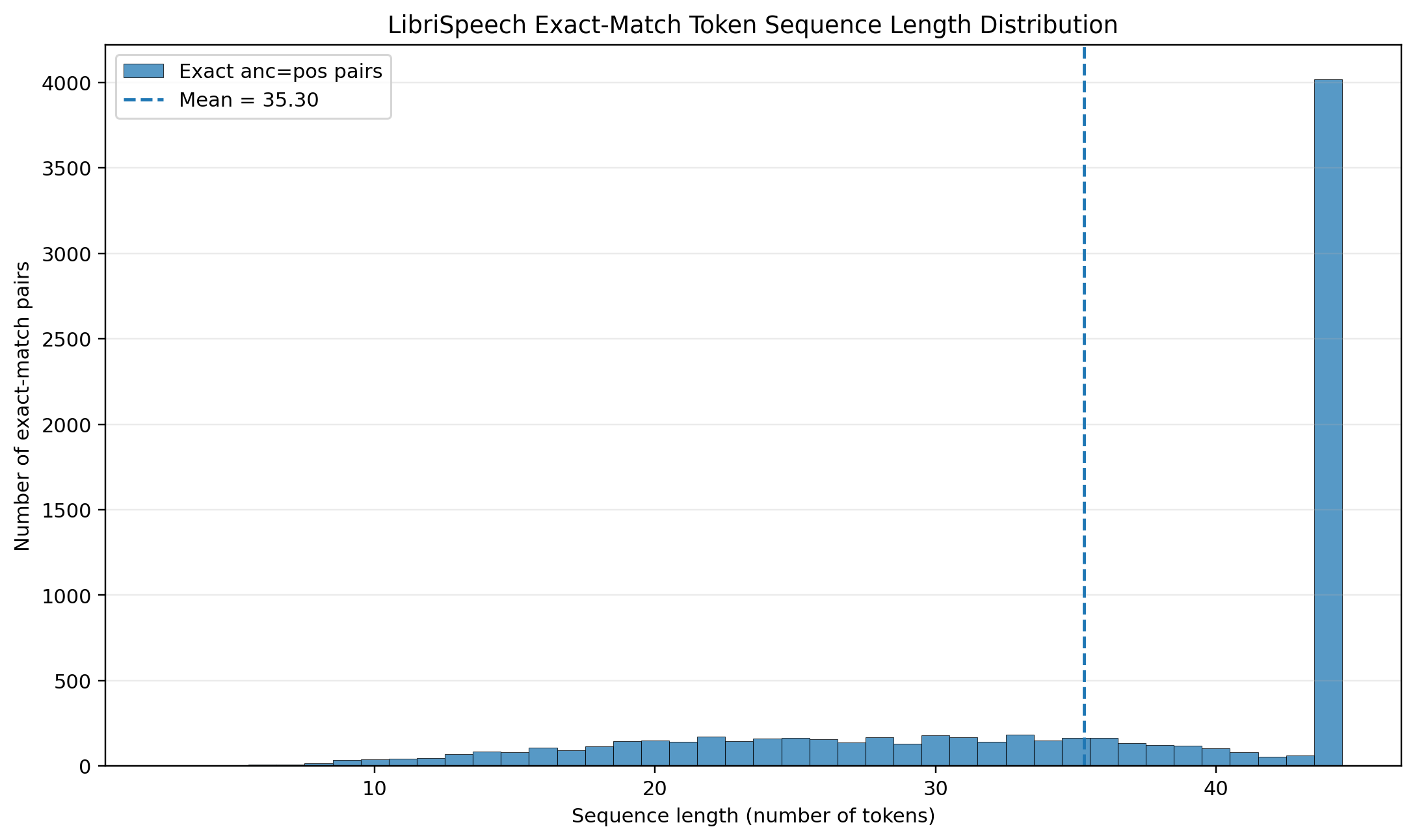}
        \caption{LibriSpeech-100}
        \label{fig:librispeech-exact-match-length}
    \end{subfigure}

    \caption{
    Token-sequence length distributions restricted to exact-match
    anchor--positive pairs under unconstrained PairAlign inference. On TIMIT,
    exact-match lengths span $6$--$44$ tokens with mean $26.41$; on
    LibriSpeech-100, the mean is $35.30$ with a similarly broad distribution
    and a dominant mode at $44$. During training, $\rho$ limits the maximum
    target length to $44$ tokens for the $301$-frame conditioning context.
    During evaluation, the corresponding inference-time restriction is removed,
    EOS is permitted at every step, and decoding may continue for up to $301$
    steps. Therefore, termination both before and exactly at $44$ is
    model-predicted rather than forced by an inference-time cap. Validation
    dataloader seed is $1234$ and evaluation seed is $13$.
    }
    \label{fig:exact-match-token-length-distributions}
\end{figure*}


\section{Retrieval and Compactness Ablations}
\label{app:retrieval-ablation}

This appendix complements \S~\ref{sec:pairalign_ablation} by testing how
encoder-summary conditioning changes the retrieval--compactness operating
point. We use the same TIMIT protocol as
\S~\ref{sec:retrieval-geom-vs-pairalign}: $300$ augmented queries are
ranked against $9{,}461$ overlapping $3$ s segments using normalized
Levenshtein distance and segment-overlap relevance. Both the validation
dataloader and evaluation script use fixed seed $13$.

We focus on Full PairAlign and the no-summary variant. Stage~II is only a
fixed-target intermediate, while removing structured self-attention dropout or
in-batch likelihood contrast causes catastrophic collapse
(Table~\ref{tab:pairalign_consistency_ablation}), making retrieval from those
generic strings uninformative.

\begin{table*}[t]
\centering
\scriptsize

\begin{subtable}[t]{\textwidth}
\centering
\setlength{\tabcolsep}{4pt}
\begin{tabular}{lcccccc}
\toprule
\textbf{Variant}
& \textbf{R@1}
& \textbf{R@5}
& \textbf{R@10}
& \textbf{R@20}
& \textbf{MRR}
& \textbf{Mean FRR} \\
\midrule

Stage~I Geometric
    & $0.75$
    & $0.83$
    & $0.86$
    & $0.88$
    & $0.79$
    & $36.45$ \\

Stage~I+ Geometric
    & $0.68$
    & $0.75$
    & $0.79$
    & $0.81$
    & $0.72$
    & $45.86$ \\

\textbf{Full PairAlign}
    & $\mathbf{0.71}$
    & $0.78$
    & $0.81$
    & $0.83$
    & $\mathbf{0.74}$
    & $\mathbf{44.67}$ \\

\quad w/o encoder-summary conditioning
    & $0.68$
    & $\mathbf{0.79}$
    & $\mathbf{0.83}$
    & $\mathbf{0.85}$
    & $0.73$
    & $53.50$ \\

\bottomrule
\end{tabular}

\caption{
TIMIT segment-overlap retrieval. Boldface within PairAlign rows marks the
better value between Full PairAlign and the no-summary variant. Removing the
summary slightly improves Recall@$K$ for $K\geq5$ but reduces R@1/MRR and
worsens Mean FRR, indicating rank redistribution rather than uniform retrieval
improvement.
}
\label{tab:pairalign_retrieval_ablation_main}
\end{subtable}

\vspace{0.8em}

\begin{subtable}[t]{\textwidth}
\centering
\setlength{\tabcolsep}{6pt}
\begin{tabular}{lcc}
\toprule
\textbf{Variant}
& \textbf{Avg. Length}
& \textbf{Token Rate} \\
\midrule

Stage~I Geometric
    & $84.62$
    & $28.21$ tok/s \\

Stage~I+ Geometric
    & $61.15$
    & $20.38$ tok/s \\

\textbf{Full PairAlign}
    & $\mathbf{24.84}$
    & $\mathbf{8.28}$ tok/s \\

\quad w/o encoder-summary conditioning
    & $38.55$
    & $12.85$ tok/s \\

\bottomrule
\end{tabular}

\caption{
Compactness on the same archive. Removing encoder-summary conditioning raises
mean length from $24.84$ to $38.55$ tokens and rate from $8.28$ to
$12.85$ tok/s. Full PairAlign is therefore both more compact and more
paired-view consistent.
}
\label{tab:pairalign_ablation_compactness}
\end{subtable}

\caption{
Retrieval and compactness ablations on TIMIT: $300$ queries against
$9{,}461$ archive segments. Mean FRR is the mean first-relevant rank
($\downarrow$). Both validation-dataloader and evaluation-script seeds are
fixed to $13$.
}
\label{tab:pairalign_retrieval_compactness_ablation}
\end{table*}

\paragraph{Removing encoder-summary conditioning redistributes ranks rather than uniformly improving retrieval.}
Relative to Full PairAlign, the no-summary variant reduces R@1 from $0.71$ to
$0.68$ and MRR from $0.74$ to $0.73$, while Mean FRR worsens from $44.67$ to
$53.50$. Conversely, R@5/R@10/R@20 increase slightly from
$0.78/0.81/0.83$ to $0.79/0.83/0.85$. Relevant candidates therefore remain
recoverable at broader ranks but are less concentrated near the top; the
ablation is neither uniformly better nor uniformly worse on retrieval alone.

\paragraph{The consistency loss dominates the small broader-recall gains.}
Table~\ref{tab:pairalign_consistency_ablation} shows that removing the summary
reduces Jaccard/edit/exact match from $0.747/0.687/0.302$ to
$0.680/0.514/0.252$. The $0.173$ edit-similarity loss is large relative to the
$0.01$--$0.02$ gains at R@5--R@20, while R@1, MRR, and Mean FRR also worsen.
Thus, the variant exchanges substantial paired-view stability for a small
redistribution of recall toward deeper ranks, not a superior retrieval point.

\paragraph{Encoder-summary conditioning also improves compactness.}
The stronger consistency of Full PairAlign is not obtained by using more
symbols. Removing the summary increases mean archive length from $24.84$ to
$38.55$ tokens ($55.2\%$) and rate from $8.28$ to $12.85$ tok/s. Full
PairAlign therefore achieves a $0.173$ edit-similarity advantage while using
approximately $35.6\%$ fewer tokens, improving consistency and compactness
simultaneously.

\paragraph{The summary pathway supplies persistent global acoustic context.}
Cross-attention alone remains sufficient for non-collapsed, retrieval-capable
generation, but without the summary the trajectory is longer and far less
stable across content-preserving views. Adding the encoder summary at every
decoding step provides a persistent global acoustic reference alongside local
cross-attention.

Its benefit is joint rather than metric-wise universal: Full PairAlign provides
much stronger paired-view agreement, lower symbolic rate, higher R@1/MRR, and
lower Mean FRR, while sacrificing only the small no-summary gains at
R@5--R@20. This is the more favorable operating point for a compact reusable
sequence representation.

\paragraph{Overall interpretation.}
The ablation separates sequence stability, ranking distribution, and symbolic
rate. Removing encoder-summary conditioning slightly improves broad-cutoff
recall, but weakens top-rank placement, lengthens the representation
substantially, and sharply reduces ordered paired-view consistency.

Full PairAlign is therefore preferred on joint evidence: it combines lower
rate, stronger sequence stability, and competitive retrieval with better R@1,
MRR, and first-relevant-rank behavior than the no-summary variant. Together
with the collapse controls in \S~\ref{sec:pairalign_ablation}, this
supports encoder-summary conditioning as a global acoustic reference that
improves the compactness--consistency trade-off without removing the token
space's usefulness for edit-distance retrieval.

\section{Probing the Learned Similarity Geometry Under Lexically Disjoint Tamil Transfer}
\label{app:tamil-transfer}

\begin{table*}[t]
\centering
\scriptsize
\setlength{\tabcolsep}{3.5pt}
\begin{tabular}{lccccccc}
\toprule

\multirow{3}{*}{\textbf{Dataset}}
& \multirow{3}{*}{\textbf{Variant}}
& \textbf{Jaccard}
& \textbf{Edit}
& \textbf{Exact}
& \textbf{Active}
& \textbf{Collapsed}
& \textbf{Exact} \\
&
& \textbf{Similarity}
& \textbf{Similarity}
& \textbf{Match}
& \textbf{Vocabulary}
& \textbf{Pair}
& \textbf{Collision} \\
&
&
&
&
&
&
&
$\mathbf{\times 10^{-4}\;(A/P)}$ \\

\midrule

\multirow{3}{*}{LibriSpeech-100}
& Stage~I
    & 0.718
    & 0.609
    & 0.264
    & 512
    & 0.0500
    & $0/0$ \\

& Stage~I+
    & 0.738
    & 0.629
    & 0.265
    & 512
    & 0.0450
    & $0/0$ \\

& PairAlign
    & $0.719 \pm 9.7{\times}10^{-7}$
    & $0.629 \pm 2.7{\times}10^{-6}$
    & $0.291 \pm 0.0$
    & $512 \pm 0.0$
    & $0.0 \pm 0.0$
    & $0.4/0.7$ \\

\midrule

\multirow{3}{*}{Shrutilipi-Tamil}
& Stage~I
    & 0.6937
    & 0.5867
    & 0.2425
    & 512
    & 0.0843
    & $0/0$ \\

& Stage~I+
    & 0.6991
    & 0.5995
    & 0.2421
    & 512
    & 0.1051
    & $0/0$ \\

& PairAlign
    & $0.662 \pm 2.8{\times}10^{-6}$
    & $0.553 \pm 4.4{\times}10^{-6}$
    & $0.265 \pm 0.0$
    & $510 \pm 0.0$
    & $0.0 \pm 0.0$
    & $3.3/0$ \\

\bottomrule
\end{tabular}

\caption{
Paired-view consistency and collapse under lexically disjoint Tamil transfer.
All tokenizers are trained on LibriSpeech and transferred unchanged to
Shrutilipi-Tamil. Exact Collision is reported in units of $10^{-4}$ as
anchor/positive (A/P). For both LibriSpeech-100 and Tamil, the validation
dataloader uses fixed deterministic seed $1234$; Stage~I/Stage~I+ use
evaluation seed $13$, while PairAlign uses seeds $13$, $97$, and $313$ on the
same fixed views and reports mean and across-run deviation. All models use
training seed $240$.
}
\label{tab:pairalign_tamil_consistency_ablation}
\end{table*}

\begin{figure*}[t]
    \centering

    \begin{subfigure}[t]{0.48\textwidth}
        \centering
        \includegraphics[width=\linewidth]{
            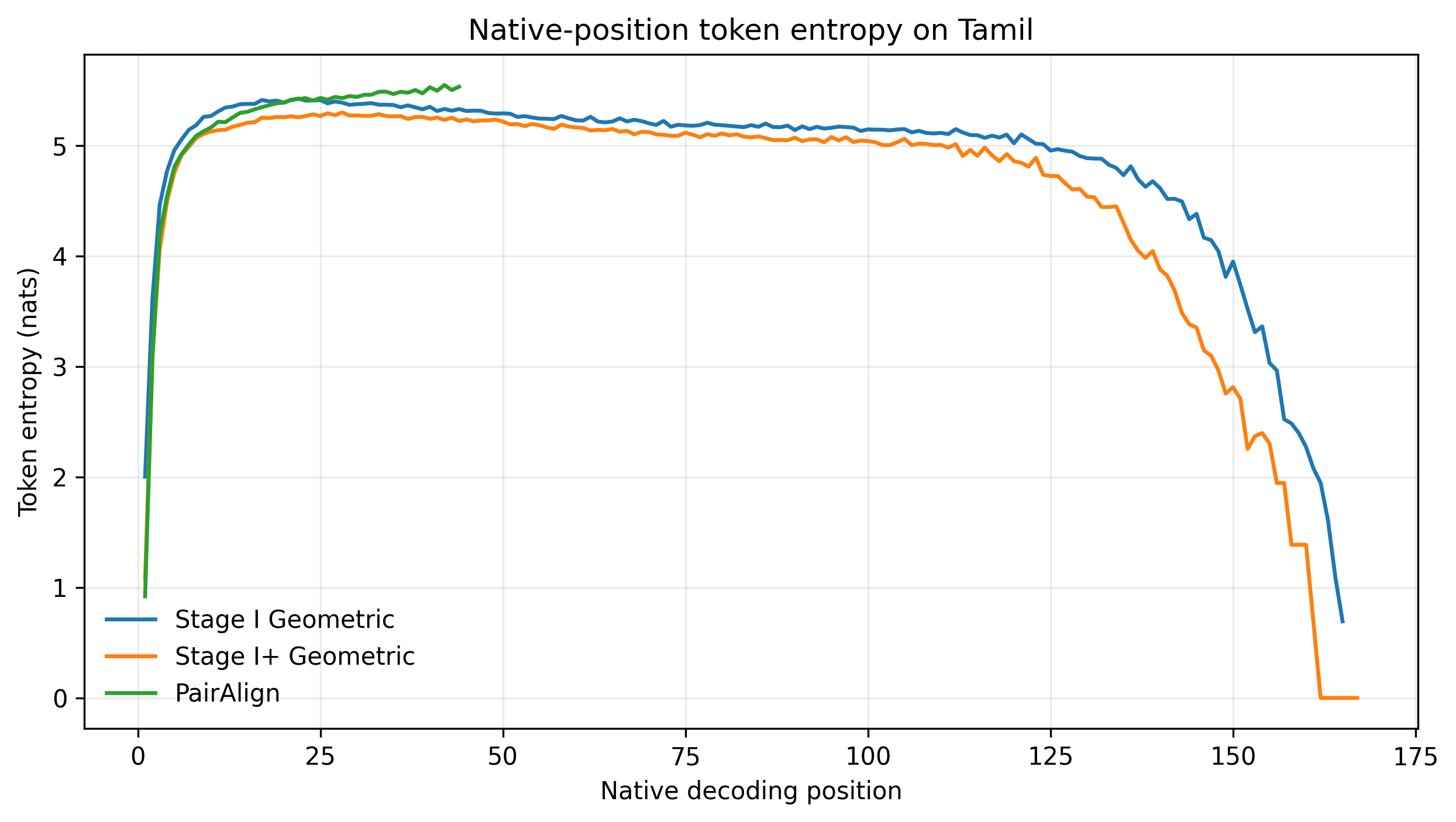
        }
        \caption{
            Native-position token entropy on Tamil for Stage~I Geometric,
            Stage~I+ Geometric, and PairAlign.
        }
        \label{fig:tamil_native_entropy_all}
    \end{subfigure}
    \hfill
    \begin{subfigure}[t]{0.48\textwidth}
        \centering
        \includegraphics[width=\linewidth]{
            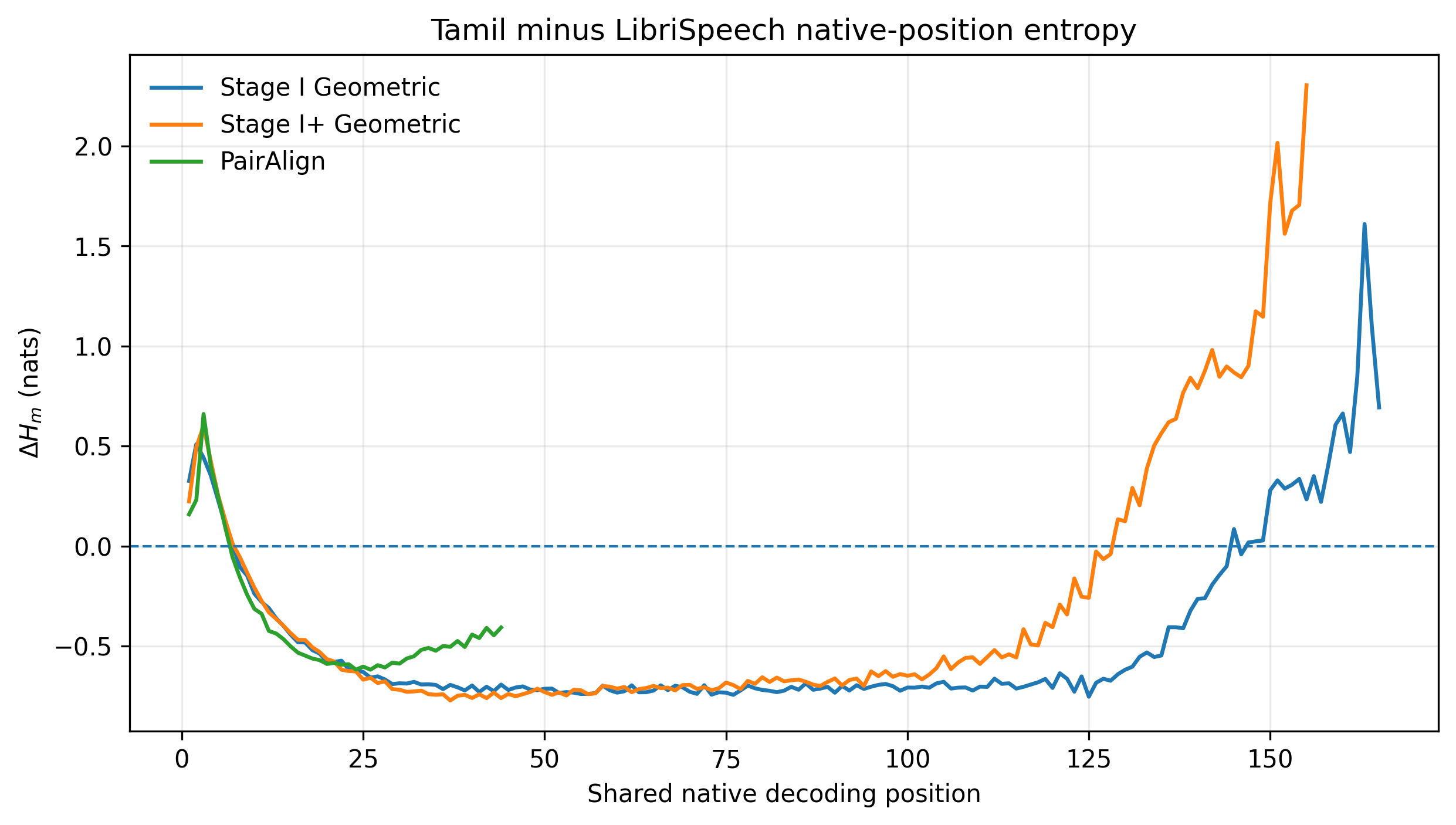
        }
        \caption{
            Native-position entropy shift
            $\Delta H_m=H_m^{\mathrm{Tamil}}-H_m^{\mathrm{LibriSpeech}}$
            over positions present in both datasets.
        }
        \label{fig:tamil_minus_libri_entropy_shift}
    \end{subfigure}

    \caption{
        Native-position entropy under Tamil transfer, reflecting both
        linguistic and corpus-level acoustic shift relative to LibriSpeech.
        Left: Tamil entropy at each tokenizer's native positions. Right:
        position-wise Tamil-minus-LibriSpeech residual at matched native
        positions; positive values denote higher Tamil entropy. The validation
        dataloader uses fixed deterministic seed $1234$ and the evaluation
        script seed $13$.
    }
    \label{fig:tamil_entropy_comparison}
\end{figure*}

\begin{table}[t]
\centering
\scriptsize
\setlength{\tabcolsep}{3.5pt}
\renewcommand{\arraystretch}{1.08}

\begin{tabular}{lcccc}
\toprule

\textbf{Tokenizer}
&
\makecell{\textbf{\# Common Native}\\\textbf{Positions}}
&
\makecell{\textbf{Mean Signed}\\\textbf{Entropy Shift}}
&
\makecell{\textbf{Mean Absolute}\\\textbf{Entropy Shift}}
&
\makecell{\textbf{Maximum Absolute}\\\textbf{Entropy Shift}}
\\

&
$\boldsymbol{M}$
&
$\displaystyle
\frac{1}{M}\sum_{m=1}^{M}\Delta H_m$
&
$\displaystyle
\frac{1}{M}\sum_{m=1}^{M}\left|\Delta H_m\right|$
&
$\displaystyle
\max_m\left|\Delta H_m\right|$
\\

\midrule

Stage~I Geometric
    & 165
    & -0.463
    & 0.594
    & 1.609 \\

Stage~I+ Geometric
    & 155
    & -0.295
    & 0.649
    & 2.303 \\

PairAlign
    & 44
    & -0.381
    & \textbf{0.462}
    & \textbf{0.660} \\

\bottomrule
\end{tabular}

\caption{
Native-position entropy shift
$\Delta H_m=H_m^{\mathrm{Tamil}}-H_m^{\mathrm{LibriSpeech}}$, computed over the
$M$ positions present in both datasets for each tokenizer. Signed shift gives
direction; mean/max absolute shift give magnitude. PairAlign has the smallest
absolute shifts, but only $44$ common positions versus $165/155$ for
Stage~I/Stage~I+, reflecting different sequence granularities. Thus, this
measures inventory displacement at each tokenizer's native output scale, not
language-invariant positional structure. Validation uses fixed deterministic
seed $1234$ and evaluation seed $13$. Values are in nats.
}
\label{tab:tamil_librispeech_entropy_shift}
\end{table}

\begin{table*}[t]
\centering
\scriptsize

\begin{subtable}[t]{0.64\textwidth}
\centering
\setlength{\tabcolsep}{4pt}
\begin{tabular}{lcccccc}
\toprule
\textbf{Variant}
& \textbf{R@1}
& \textbf{R@5}
& \textbf{R@10}
& \textbf{R@20}
& \textbf{MRR}
& \textbf{Mean FRR} \\
\midrule

Stage~I Geometric
    & 0.87 & 0.93 & 0.96 & 0.97 & 0.90 & 11.93 \\

Stage~I+ Geometric
    & 0.85 & 0.90 & 0.91 & 0.93 & 0.87 & 9.19 \\

PairAlign
    & 0.68 & 0.79 & 0.84 & 0.87 & 0.73 & 63.60 \\

\bottomrule
\end{tabular}
\caption{Retrieval under segment-overlap relevance.}
\label{tab:pairalign_tamil_retrieval_ablation_main}
\end{subtable}
\hfill
\begin{subtable}[t]{0.31\textwidth}
\centering
\setlength{\tabcolsep}{4pt}
\begin{tabular}{lcc}
\toprule
\textbf{Variant}
& \textbf{Avg. Len.}
& \textbf{Tok/s} \\
\midrule

Stage~I
    & 113.15 & 37.72 \\

Stage~I+
    & 101.15 & 33.72 \\

PairAlign
    & 37.93 & 12.64 \\

\bottomrule
\end{tabular}
\caption{Token-stream compactness.}
\label{tab:pairalign_tamil_ablation_compactness}
\end{subtable}

\caption{
Tamil retrieval and archive compactness under the same configuration.
Both retrieval validation-dataloader and evaluation-script seeds are fixed to
$13$.
}
\label{tab:pairalign_tamil_retrieval_compactness}
\end{table*}

\paragraph{Lexically disjoint linguistic shift.}
We use Shrutilipi-Tamil to probe transfer of LibriSpeech-trained token geometry,
not component ablations or word-level cross-lingual correspondence. Tamil
(Dravidian) and English (Indo-European) differ substantially in lexicon,
morphology, phonotactics, prosody, and word order; Tamil is predominantly SOV
and agglutinative, so a $3$ s window may differ in both acoustic--phonetic
realization and short-range linguistic organization
\citep{krishnamurti2003dravidian,wals,Steever_2017}. Because the corpora also
differ in speakers, recording, channel, and other acoustics, this probe measures
combined acoustic--linguistic and corpus-domain shift rather than language
alone.

We evaluate paired-view consistency, collapse, Tamil-archive retrieval,
native-position inventory entropy, and symbolic rate. Consistency measures
content-preserving proximity; retrieval tests edit-distance ranking after
transfer; entropy measures position-wise inventory displacement relative to
LibriSpeech at each tokenizer's native output scale. For paired-view
consistency, validation uses fixed deterministic seed $1234$; Stage~I/Stage~I+
use evaluation seed $13$, while PairAlign uses seeds $13$, $97$, and $313$ on
the same fixed views. Tamil retrieval uses seed $13$ for both validation and
evaluation.

\paragraph{Tamil transfer degrades graded similarity but does not induce PairAlign collapse.}
PairAlign drops from LibriSpeech-100 to Tamil from $0.719$ to $0.662$ Jaccard
and $0.629$ to $0.553$ edit similarity, so transfer is not invariant.
Stage~I changes from $0.718/0.609$ to $0.694/0.587$, and Stage~I+ from
$0.738/0.629$ to $0.699/0.600$. Because language and corpus change together,
these drops cannot be uniquely assigned to linguistic or acoustic shift.

Despite the larger graded-similarity loss, PairAlign remains non-degenerate:
exact match is $0.265$ versus $\approx0.243$ for both geometric systems,
collapsed-pair rate is zero, $510/512$ symbols remain active, and exact
collision is only $3.3\times10^{-4}/0$ for anchor/positive streams. Transfer
therefore weakens partial correspondence without collapsing PairAlign into a
narrow inventory or generic sequence family.

\paragraph{Dense geometric streams preserve more partial overlap under transfer.}
On Tamil, Stage~I and Stage~I+ retain higher Jaccard/edit similarity
($0.694/0.587$ and $0.699/0.600$) than PairAlign ($0.662/0.553$), consistent
with denser frame-derived streams providing more opportunities for local
agreement under perturbation or shift. They are nevertheless affected:
collapsed-pair rate rises from $0.0500$ to $0.0843$ for Stage~I and from
$0.0450$ to $0.1051$ for Stage~I+, despite both retaining all $512$ symbols.
Thus, full vocabulary coverage does not preclude low-diversity sequences.
PairAlign shows the complementary pattern: weaker graded overlap but zero
measured collapse and near-complete inventory coverage.

\paragraph{PairAlign shows the smallest native-position entropy displacement.}
PairAlign's mean signed Tamil-minus-LibriSpeech entropy shift is $-0.381$ nats,
so its positional inventory is not invariant. However, mean/max absolute shift
is $0.462/0.660$ nats, below Stage~I's $0.594/1.609$ and Stage~I+'s
$0.649/2.303$. This comparison is tokenizer-native: PairAlign has only $44$
common positions versus $165/155$ for Stage~I/Stage~I+, so it does not establish
language-invariant positional structure. More narrowly, among positions present
in both corpora at each tokenizer's own scale, PairAlign exhibits the smallest
inventory displacement. Together with $510$ active symbols and zero collapse,
its transfer degradation is not accompanied by severe inventory contraction or
positional destabilization.

\paragraph{Tamil augmented-source retrieval exposes the cost of coarse symbolic granularity.}
Retrieval strongly favors the dense streams: Stage~I gives R@1/MRR
$0.87/0.90$, Stage~I+ $0.85/0.87$, and PairAlign $0.68/0.73$; Mean FRR is
$11.93$, $9.19$, and $63.60$, respectively. The rate gap is large:
Stage~I/Stage~I+/PairAlign emit $113.15/101.15/37.93$ tokens per $3$ s, or
$37.72/33.72/12.64$ tok/s. PairAlign therefore uses about $66.5\%$ fewer
symbols than Stage~I and $62.5\%$ fewer than Stage~I+.

As on TIMIT, normalized edit-distance ranking benefits from denser local
resolution. PairAlign retains non-trivial retrieval structure but fewer local
degrees of freedom for separating nearby candidates, exposing the same
rate--precision trade-off under a substantially different language and corpus
domain.

\paragraph{PairAlign's sequence rate adapts under Tamil transfer.}
PairAlign is not fixed-rate: it emits $24.84$ tokens per $3$ s
($8.28$ tok/s) on the clean TIMIT retrieval archive but $37.93$
($12.64$ tok/s) on Tamil, while remaining much shorter than Stage~I/Stage~I+.
This is consistent with input-conditioned autoregressive termination rather than
stride-determined output length. The increase should be interpreted as the
model's response to the combined cross-lingual/corpus shift, not as evidence
that linguistic structure alone causes the rate change.

\paragraph{Consistency, retrieval, inventory behavior, and rate separate under transfer.}
The Tamil results separate several axes of representation quality. Stage~I+
has the strongest edit similarity ($0.600$), whereas Stage~I has the strongest
top-rank retrieval ($0.87$ R@1, $0.90$ MRR); improved paired-view consistency
therefore does not imply sharper archive discrimination.

PairAlign occupies a different point: transfer causes a larger graded-similarity
drop and a higher rate than on TIMIT, yet it retains higher exact agreement,
$510/512$ active symbols, zero measured low-diversity collapse, the smallest
native-position entropy displacement, and roughly one-third the symbolic rate
of the geometric streams.

The experiment therefore establishes neither language-invariant PairAlign
geometry nor isolated linguistic transfer. It shows that a LibriSpeech-trained
tokenizer can be transferred unchanged to a linguistically distant Tamil corpus
with substantial acoustic--phonetic, linguistic, and corpus-domain shift.
Under this combined shift, PairAlign remains compact, input-conditioned, and
non-degenerate, while sacrificing some graded paired-view similarity and
retrieval discrimination relative to denser geometric representations.

\section{PairAlign Versus Pretrained Semantic Tokenizers}
\label{app:ssl-tokenizer-comparison}

\paragraph{Pretrained SSL tokenizers as high-exposure, speech-unit-biased external references.}
The main text uses Stage~I/Stage~I+ as controlled PairAlign baselines because
they share preprocessing, encoder family, source pool, VQ alphabet, and training
pipeline. Here we instead stress-test PairAlign against discrete streams derived
from substantially larger pretrained speech encoders whose objectives also
provide a much stronger explicit speech-unit prior.

HuBERT is bootstrapped from offline acoustic-unit pseudo-labels: its first
training iteration predicts KMeans assignments obtained from 39-dimensional
MFCC features, and subsequent iterations replace these targets with KMeans
assignments from intermediate HuBERT representations
\citep{hsu2021hubert}. This iterative refinement substantially increases the
phone-correlated structure of the clustering targets. WavLM inherits this
HuBERT-style masked speech-unit prediction framework while adding
denoising-oriented learning and gated relative position bias
\citep{chen2022wavlm}; WavLM Large is trained using pseudo-labels obtained by
clustering an intermediate representation of the released second-iteration
HuBERT Base model.

HuBERT+VQ and WavLM+VQ therefore differ from PairAlign along more than capacity
and exposure. Before the downstream $K=512$ KMeans discretization used in our
comparison, their continuous representations have already been optimized
against explicit cluster-derived acoustic-unit targets---and, for later
iterations, targets derived from previously learned speech representations.
PairAlign Stage~I instead learns through acoustic contrastive consistency and
VQ, without an offline cluster-derived speech-unit prediction target. The SSL
references consequently enter discretization with a substantially stronger
speech-unit-biased prior, in addition to much larger models and far greater
optimization exposure.

These systems should therefore be interpreted as high-capacity,
high-exposure, speech-unit-biased external operating points rather than
controlled architectural ablations of PairAlign. The comparison below uses
each system at its native symbolic rate; the subsequent
\S~\ref{sec:rate_controlled_bpe} separately tests whether the SSL advantage
persists when Stage~I, HuBERT, and WavLM streams are shortened post hoc toward
PairAlign's operating rate.

\paragraph{SSL layer and discrete vocabulary.}
Both SSL references use a fixed intermediate Transformer representation,
followed by \(K=512\) KMeans and consecutive duplicate removal. For HuBERT, we
use the \textbf{6th Transformer layer} because first-iteration HuBERT-BASE
features are strongest for clustering in the middle layers, peaking around
layer~6, whereas the final layers degrade substantially in phone purity, cluster
purity, and PNMI \citep{hsu2021hubert}. Layer~6 therefore provides a
higher-level acoustic representation than the earliest layers while avoiding
the degradation observed toward the top of BASE-it1. It is also directly used
for HuBERT target refinement, where the \(768\)-dimensional layer-6
representation is clustered to construct subsequent training targets.
For WavLM, we use the same relative layer as a standardized intermediate
extraction point, avoiding model-specific layer tuning and keeping the two SSL
references comparable at the representation-selection level. This choice is
also consistent with WavLM's use of HuBERT-derived intermediate
representations in its target construction \citep{chen2022wavlm}. Thus,
layer~6 serves as a principled intermediate operating point rather than a
post-hoc downstream choice.

We set \(K=512\) to match PairAlign's nominal alphabet and remain close to the
\(\sim500\)-cluster regime used in HuBERT-style acoustic-unit discovery. This
controls only the scale of the \emph{final} discrete inventory; it does not
equate the continuous representations entering KMeans. Encoder architecture,
model capacity, pretraining exposure, training objective, source-data scale,
and the SSL models' prior optimization against cluster-derived speech-unit
targets remain intentionally unmatched.


\begin{table*}[t]
\centering
\scriptsize

\begin{subtable}[t]{\textwidth}
\centering
\setlength{\tabcolsep}{3pt}
\begin{tabular}{llcccccc}
\toprule
\multirow{2}{*}{\textbf{Dataset}}
& \multirow{2}{*}{\textbf{Variant}}
& \textbf{Jaccard}
& \textbf{Edit}
& \textbf{Exact}
& \textbf{Active}
& \textbf{Collapsed}
& \textbf{Exact Collision} \\
& &
\textbf{Similarity}
& \textbf{Similarity}
& \textbf{Match}
& \textbf{Vocabulary}
& \textbf{Pair}
& $\mathbf{\times 10^{-4}\;(A/P)}$ \\
\midrule

\multirow{3}{*}{TIMIT}
& HuBERT
    & $0.767$
    & $0.726$
    & $0.263$
    & $464$
    & $0$
    & $0/0$ \\

& WavLM
    & $0.773$
    & $0.730$
    & $0.264$
    & $397$
    & $0$
    & $0/0$ \\

& PairAlign
    & $0.747 \pm 0.0$
    & $0.687 \pm 0.0$
    & $0.302 \pm 0.0$
    & $440 \pm 0.0$
    & $0$
    & $0/22$ \\

\midrule

TIMIT
& HuBERT
    & $0.805$
    & $0.749$
    & $0.0002$
    & $470$
    & $0$
    & $0/0$ \\

+
& WavLM
    & $0.794$
    & $0.774$
    & $0.0000$
    & $399$
    & $0$
    & $0/0$ \\

External Noise
& PairAlign
    & $0.6512$
    & $0.5540$
    & $0.0173$
    & $508$
    & $0$
    & $0/0$ \\

\midrule

\multirow{3}{*}{Shrutilipi-Tamil}
& HuBERT
    & $0.726$
    & $0.681$
    & $0.255$
    & $460$
    & $0$
    & $0/0$ \\

& WavLM
    & $0.766$
    & $0.718$
    & $0.262$
    & $340$
    & $0$
    & $0/0$ \\

& PairAlign
    & $0.662 \pm 2.8{\times}10^{-6}$
    & $0.553 \pm 4.4{\times}10^{-6}$
    & $0.265 \pm 0.0$
    & $510 \pm 0.0$
    & $0$
    & $3.3/0$ \\

\bottomrule
\end{tabular}

\caption{
Paired-view consistency and collapse for HuBERT+VQ, WavLM+VQ, and PairAlign
at native rate. HuBERT Base and WavLM Large use fixed intermediate
representations, $K=512$ KMeans, and consecutive deduplication. For clean TIMIT
and Shrutilipi-Tamil, validation uses fixed deterministic seed $1234$;
HuBERT/WavLM use evaluation seed $13$, while PairAlign averages evaluation
seeds $13$, $97$, and $313$ on the same fixed views. For TIMIT + External
Noise, both validation-view and evaluation seeds are fixed to $13$. Exact
Collision is reported as $10^{-4}$ anchor/positive (A/P).
}
\label{tab:pairalign_SSL_consistency}
\end{subtable}

\vspace{0.8em}

\begin{subtable}[t]{0.62\textwidth}
\centering
\begin{tabular}{lcccccc}
\toprule
\textbf{Variant}
& \textbf{R@1}
& \textbf{R@5}
& \textbf{R@10}
& \textbf{R@20}
& \textbf{MRR}
& \textbf{Mean FRR} \\
\midrule

HuBERT
    & $0.943$
    & $0.977$
    & $0.990$
    & $0.997$
    & $0.961$
    & $1.29$ \\

WavLM
    & $0.977$
    & $0.997$
    & $1.000$
    & $1.000$
    & $0.985$
    & $1.05$ \\

PairAlign
    & $0.710$
    & $0.780$
    & $0.810$
    & $0.830$
    & $0.740$
    & $44.67$ \\

\bottomrule
\end{tabular}
\caption{Native-rate TIMIT segment-overlap retrieval.}
\label{tab:ssl_retrieval}
\end{subtable}
\hfill
\begin{subtable}[t]{0.32\textwidth}
\centering
\begin{tabular}{lcc}
\toprule
\textbf{Variant}
& \textbf{Avg. Len.}
& \textbf{Tok/s} \\
\midrule

HuBERT
    & $86.97$
    & $28.99$ \\

WavLM
    & $96.17$
    & $32.06$ \\

PairAlign
    & $24.84$
    & $8.28$ \\

\bottomrule
\end{tabular}
\caption{
Compactness of the clean, unaugmented TIMIT retrieval archive; augmented
queries are excluded.
}
\label{tab:ssl_compactness}
\end{subtable}

\caption{
Native-rate comparison against pretrained SSL tokenizers. Top: paired-view
consistency and collapse. Bottom left: TIMIT segment-overlap retrieval. Bottom
right: compactness on the same clean retrieval archive. Paired-view validation
uses fixed seed $1234$; SSL evaluation uses seed $13$, while PairAlign uses
seeds $13$, $97$, and $313$ on the same fixed views. For retrieval and archive
compactness, both dataloader and evaluation script use fixed seed $13$.
Archive Avg. Len. and Tok/s use only clean archive tokenizations; augmented
queries contribute to retrieval but not the reported archive rate.
}
\label{tab:pairalign_ssl_full_comparison}
\end{table*}


\begin{table*}[t]
\centering
\scriptsize
\setlength{\tabcolsep}{3pt}
\renewcommand{\arraystretch}{1.15}

\begin{tabularx}{\textwidth}{
    >{\raggedright\arraybackslash}p{2.25cm}
    >{\raggedright\arraybackslash}p{2.35cm}
    >{\centering\arraybackslash}p{1.35cm}
    >{\centering\arraybackslash}p{1.80cm}
    >{\centering\arraybackslash}p{1.85cm}
    >{\raggedright\arraybackslash}X
}
\toprule

\textbf{System / stage}
&
\textbf{Model capacity}
&
\textbf{Unique audio}
&
\textbf{Stage-local exposure}
&
\textbf{Cumulative exposure}
&
\textbf{Training configuration}
\\

\midrule

HuBERT Base
&
94.372M
&
960 h
&
Iter.~1: $\sim194$k h;
Iter.~2: $\sim311$k h
&
$\sim506$k h
&
32 GPUs; up to 87.5 s/GPU per update;
250k + 400k updates.
\\

\addlinespace[3pt]

WavLM Large
&
315.453M
&
94k h
&
$\sim8.96$M h
&
$\sim9.47$M h$^{\dagger}$
&
64 GPUs; 720 s batch duration;
700k updates.
\\

\midrule

\multicolumn{6}{l}{\textit{PairAlign training pipeline}}
\\

\addlinespace[2pt]

Stage~I Geometric
&
Encoder: 2.076M
&
860 h
&
$\sim667$ h
&
$\sim667$ h
&
Seed 240; SimCLR + VQ.
\\

\addlinespace[2pt]

Stage~II AR bridge
&
Encoder: 2.076M;
Decoder: 25.745M;
Total: 27.822M
&
Same 860-h pool
&
$\sim400$ h
&
$\sim1{,}067$ h$^{\ddagger}$
&
Seed 240; frozen Stage~I encoder and VQ;
autoregressive decoder training.
\\

\addlinespace[2pt]

Stage~III PairAlign
&
Encoder: 2.076M;
Decoder: 25.745M;
Total: 27.822M
&
Same 860-h pool
&
$\sim400$ h
&
$\sim1{,}467$ h$^{\ddagger}$
&
Seed 240; joint encoder--decoder
EMA self-alignment.
\\

\bottomrule
\end{tabularx}

\caption{
Model capacity, unique source audio, and optimization exposure for the SSL
references and PairAlign stages. Cumulative exposure counts repeated
optimization-time segment presentations; $^{\dagger}$ is the
pipeline-inclusive WavLM estimate and $^{\ddagger}$ denotes cumulative
PairAlign exposure from the same seeded source pool.
}
\label{tab:ssl_pairalign_capacity_exposure}
\end{table*}

\paragraph{Capacity, exposure, and target construction contextualize the comparison.}
HuBERT Base has $94.372$M parameters and two LibriSpeech-960 iterations,
corresponding to approximately $194$k and $311$k sampled hours
($\sim506$k h cumulative) \citep{hsu2021hubert}. Its first iteration predicts
KMeans targets derived from MFCCs; the second uses refined targets obtained by
clustering an intermediate representation of the first-iteration HuBERT model.
WavLM Large has $315.453$M parameters, $94$k h of unique audio, and
$\sim8.96$M h of direct optimization exposure; the $\sim9.47$M-h
pipeline-inclusive estimate additionally accounts for the HuBERT training
underlying the clustered targets used by WavLM
\citep{chen2022wavlm}.

PairAlign operates at much smaller scale and without this cluster-derived
speech-unit supervision pipeline. Stage~I has a $2.076$M-parameter encoder;
Stages~II/III use a $27.822$M encoder--decoder
($2.076$M+$25.745$M). With training seed $240$, all stages repeatedly sample
the same $\sim860$-h LibriSpeech pool, with stage-local exposures
$\sim667/\sim400/\sim400$ h and cumulative exposures
$\sim667/\sim1{,}067/\sim1{,}467$ h. These cumulative values count repeated
presentations, not distinct audio. Thus, HuBERT and WavLM combine substantially
greater capacity and exposure with an explicit iterative speech-unit
pseudo-label prior; they are deliberately strong external references, not
controlled alternatives to PairAlign.

\paragraph{Native SSL streams preserve stronger graded paired-view structure.}
On clean TIMIT, HuBERT/WavLM/PairAlign achieve Jaccard/edit similarity
$0.767/0.726$, $0.773/0.730$, and $0.747/0.687$; on Shrutilipi-Tamil,
$0.726/0.681$, $0.766/0.718$, and $0.662/0.553$. Under external noise the
separation widens to $0.805/0.749$, $0.794/0.774$, and $0.651/0.554$.
The native SSL advantage is consistent with three reinforcing factors:
substantially greater pretraining exposure and capacity, explicit optimization
against cluster-derived speech-unit targets, and denser frame-derived token
streams; WavLM additionally includes denoising-oriented pretraining. The
comparison therefore bounds PairAlign against substantially stronger
speech-specialized references rather than attributing the gap to any one of
these factors.

\paragraph{Complete-string agreement captures a different stability property.}
PairAlign instead has higher exact match on clean TIMIT
($0.302$ versus $0.263/0.264$ for HuBERT/WavLM), remains slightly higher on
Tamil ($0.265$ versus $0.255/0.262$), and is non-zero under external noise
($0.0173$ versus $0.0002/0$). This does not reverse the graded-overlap
ordering: a dense stream can retain most local assignments while changing a
few symbols, whereas exact match asks whether the entire generated string is
unchanged.

PairAlign's higher exact agreement is also non-degenerate: it shows zero
measured low-diversity collapse, broad vocabulary use, and input-dependent
variable-length termination. Exact equality should therefore be read as a
complementary sequence-level diagnostic, not a substitute for Jaccard or edit
similarity.

\paragraph{Native-rate retrieval strongly favors pretrained SSL streams.}
HuBERT and WavLM achieve R@1/MRR $0.943/0.961$ and $0.977/0.985$, far above
PairAlign's $0.710/0.740$. However, they emit $28.99$ and $32.06$ tok/s,
versus PairAlign's $8.28$ tok/s---approximately $3.5$--$3.9\times$ greater
symbolic density. Native retrieval therefore combines stronger pretrained
acoustic representations with much finer symbolic resolution, both of which
favor normalized edit-distance ranking.

\paragraph{Stress-test interpretation.}
The native-rate comparison establishes a clear boundary on the PairAlign claim.
HuBERT and WavLM preserve stronger graded paired-view structure, substantially
greater external-noise consistency, and markedly stronger retrieval at their
native rates. PairAlign does not match their fine-grained local acoustic
resolution.

At the same time, PairAlign operates at a much lower symbolic rate and attains
higher complete-string agreement without measured collapse. The native-rate
results therefore define different operating points rather than a universal
ordering: pretrained SSL streams provide high-rate, locally detailed symbolic
traces, whereas PairAlign learns a much coarser input-dependent sequence-level
representation.

Because symbolic rate is a major confound in this comparison, these results do
not determine whether the SSL advantage persists after shortening their dense
streams. \S~\ref{sec:rate_controlled_bpe} addresses that question directly
by applying the same post-hoc rate-control principle to Stage~I, HuBERT, and
WavLM and comparing the resulting low-rate sequence geometry with PairAlign.

\section{Rate-Controlled Post-hoc Compression of Geometric and Pretrained SSL Token Streams}
\label{sec:rate_controlled_bpe}

\paragraph{Rate-controlled construction.}
The preceding experiments show that native HuBERT and WavLM streams preserve
stronger fine-grained structure than PairAlign, but at substantially higher
symbolic rates. We therefore test a narrower alternative explanation: whether
PairAlign's compactness--consistency operating point can be reproduced by
post-hoc shortening of either its controlled Stage~I representation or much
stronger pretrained SSL token streams.

For the controlled baseline, we fit deterministic integer BPE to frozen
Stage~I Geometric tokens using $25{,}000$ sequences from $25{,}000$ distinct
$3$ s LibriSpeech training segments ($20.83$ h): $12{,}500$ clean
($10.42$ h) and $12{,}500$ mildly augmented ($10.42$ h) segments drawn from
disjoint sets. Thus, augmented examples are not second views of clean fitting
examples and expose the merge statistics to content-preserving nuisance
variation. PairAlign's LibriSpeech-100 rate of $11.85$ tok/s
($35.55$ tokens/$3$ s) is used as a fixed target, while the number of BPE
merges is determined entirely on the LibriSpeech training corpus, yielding
$122{,}820$ merges.

We apply the same rate-targeting principle independently to the pretrained
HuBERT+KMeans and WavLM+KMeans streams, leaving their encoders, $K=512$ KMeans
quantizers, and consecutive deduplication unchanged. Separate BPE inventories
are fitted on LibriSpeech training data; using the same fixed LibriSpeech-100
PairAlign target rate yields $4{,}190$ HuBERT and $9{,}895$ WavLM merges.
Every BPE transform is then frozen and evaluated unchanged on LibriSpeech-100
and TIMIT. No LibriSpeech-100 or TIMIT token streams enter BPE fitting or
determine the number of merges; the LibriSpeech-100 result contributes only
the fixed PairAlign target rate. Stage~I and PairAlign use training seed
$240$. Paired-view evaluation uses fixed validation-dataloader seed $1234$
and evaluation seed $13$ throughout, ensuring identical anchor--positive
views. BPE controls temporal rate, not vocabulary capacity, because merged
symbols expand the output inventory. The resulting LibriSpeech-100 and TIMIT
paired-view comparisons are reported in
Table~\ref{tab:rate_controlled_bpe_consistency}.

\begin{table*}[t]
\centering
\scriptsize
\setlength{\tabcolsep}{3.5pt}

\begin{subtable}[t]{\textwidth}
\centering
\begin{tabular}{lcccccc}
\toprule
\textbf{LibriSpeech-100 model}
& \textbf{Jaccard}
& \textbf{Edit}
& \textbf{Exact}
& \textbf{Mean Length}
& \textbf{Tok/s}
& \textbf{Active Vocab.} \tabularnewline
\midrule

Stage~I Geometric
& $0.718$
& $0.609$
& $0.264$
& $92.07$
& $30.69$
& $512$ \tabularnewline

Stage~I Geometric + BPE
& $0.421$
& $0.412$
& $0.264$
& $43.54$
& $14.51$
& $95{,}259$ \tabularnewline

PairAlign
& $\mathbf{0.719}$
& $\mathbf{0.629}$
& $\mathbf{0.291}$
& $\mathbf{35.55}$
& $\mathbf{11.85}$
& $512$ \tabularnewline

\bottomrule
\end{tabular}
\caption{
In-distribution evaluation using Stage~I Geometric tokens.
}
\label{tab:rate_controlled_bpe_librispeech}
\end{subtable}

\vspace{0.8em}

\begin{subtable}[t]{\textwidth}
\centering
\begin{tabular}{lcccccc}
\toprule
\textbf{TIMIT model}
& \textbf{Jaccard}
& \textbf{Edit}
& \textbf{Exact}
& \textbf{Mean Length}
& \textbf{Tok/s}
& \textbf{Active Vocab.} \tabularnewline
\midrule

Stage~I Geometric + BPE
& $0.422$
& $0.407$
& $0.267$
& $36.86$
& $12.29$
& $20{,}955$ \tabularnewline

HuBERT + KMeans + BPE
& $0.462$
& $0.496$
& $0.263$
& $39.24$
& $13.08$
& $4{,}532$ \tabularnewline

WavLM + KMeans + BPE
& $0.429$
& $0.465$
& $0.263$
& $38.30$
& $12.77$
& $9{,}273$ \tabularnewline

PairAlign
& $\mathbf{0.747}$
& $\mathbf{0.687}$
& $\mathbf{0.302}$
& $\mathbf{26.10}$
& $\mathbf{8.70}$
& $440$ \tabularnewline

\bottomrule
\end{tabular}
\caption{
Frozen-transfer comparison across geometric and pretrained SSL source streams.
}
\label{tab:rate_controlled_bpe_timit}
\end{subtable}

\caption{
Rate-controlled paired-view comparison. Stage~I BPE is fitted on $25{,}000$
distinct LibriSpeech training segments ($20.83$ h), split equally between
clean and mildly augmented disjoint subsets, yielding $122{,}820$ merges.
HuBERT and WavLM use independently fitted LibriSpeech training inventories of
$4{,}190$ and $9{,}895$ merges, respectively. PairAlign's LibriSpeech-100
operating rate of $11.85$ tok/s is used as the fixed target; the BPE
inventories and number of merges are determined on LibriSpeech training data
and then frozen for evaluation on LibriSpeech-100 and TIMIT. No TIMIT data
enter fitting or rate selection. For both datasets, validation uses fixed
deterministic seed $1234$ and evaluation seed $13$. Token rate is mean length
divided by $3$ s. BPE active-vocabulary cardinality is not directly comparable
with PairAlign's fixed neural alphabet.
}
\label{tab:rate_controlled_bpe_consistency}
\end{table*}

\paragraph{Post-hoc shortening does not reproduce PairAlign's paired-view geometry.}
As shown in Table~\ref{tab:rate_controlled_bpe_consistency}, the controlled
Stage~I experiment first isolates sequence length. BPE reduces Stage~I from
$92.07$ to $43.54$ tokens on LibriSpeech-100, but Jaccard/edit similarity
falls from $0.718/0.609$ to $0.421/0.412$; PairAlign is shorter ($35.55$)
yet reaches $0.719/0.629$. The control is conservative in rate: the number of
Stage~I BPE merges is determined on LibriSpeech training data using $11.85$
tok/s as the fixed target, while frozen evaluation on LibriSpeech-100 yields
$14.51$ tok/s, remaining denser than PairAlign.

The same separation holds under TIMIT transfer. Stage~I+BPE, HuBERT+BPE, and
WavLM+BPE produce $36.86/39.24/38.30$ tokens per segment
($12.29/13.08/12.77$ tok/s), all denser than PairAlign's $26.10$
($8.70$ tok/s), yet achieve Jaccard/edit similarities of
$0.422/0.407$, $0.462/0.496$, and $0.429/0.465$, respectively, versus
$0.747/0.687$ for PairAlign. Thus, fewer symbols alone do not explain
PairAlign's compactness--consistency operating point.

\paragraph{Strong pretrained source geometry does not remove the post-hoc compression gap.}
The SSL controls in Table~\ref{tab:rate_controlled_bpe_consistency} address
whether the Stage~I result is merely a consequence of a weak source
representation. Before compression, the preceding SSL stress test gives
HuBERT and WavLM TIMIT Jaccard/edit similarity $0.767/0.726$ and
$0.773/0.730$, substantially stronger than Stage~I. After BPE, these fall to
$0.462/0.496$ and $0.429/0.465$, respectively. Hence strong dense frame-level
geometry and stable low-rate sequence geometry are empirically distinct
properties.

The replication across Stage~I, HuBERT, and WavLM supports a deliberately
narrow conclusion: \emph{the tested deterministic corpus-level BPE mechanism
does not reproduce PairAlign's combination of compactness and graded
cross-view alignment}. It does not rule out learned, task-specific, or other
post-hoc compression mechanisms.

\paragraph{Frozen BPE and PairAlign realize rate reduction differently.}
Table~\ref{tab:rate_controlled_bpe_consistency} also shows that Stage~I+BPE
changes from $14.51$ tok/s on LibriSpeech-100 to $12.29$ tok/s on TIMIT,
while PairAlign changes from $11.85$ to $8.70$ tok/s. The frozen BPE systems
can therefore exhibit corpus-dependent output rates, but their merge
inventories remain fixed corpus-level transformations of already generated
dense strings. PairAlign instead jointly predicts sequence content and
termination from each acoustic input.

This observation alone is not causal evidence for why their rates differ.
The paired-view geometry supplies the stronger distinction: PairAlign becomes
more compact under TIMIT transfer while retaining edit similarity $0.687$,
whereas Stage~I+BPE obtains $0.407$ and the two SSL+BPE systems obtain
$0.496/0.465$.

\paragraph{Exact equality is preserved by lossless BPE and is therefore not the rate-control discriminator.}
As reported in Table~\ref{tab:rate_controlled_bpe_consistency}, Stage~I exact
match remains $0.264$ on LibriSpeech-100 and $0.267$ on TIMIT after BPE;
HuBERT+BPE and WavLM+BPE likewise retain approximately $0.263$ on TIMIT.
This follows from deterministic lossless encoding: identical source strings
remain identical, while distinct strings do not become identical source
sequences after decoding.

The informative quantities are therefore graded similarities among
non-identical views. BPE can leave exact equality unchanged while substantially
altering Jaccard and edit geometry. PairAlign instead combines stronger graded
correspondence with higher TIMIT exact agreement ($0.302$).


\begin{table*}[t]
\centering
\scriptsize
\setlength{\tabcolsep}{3.6pt}

\begin{tabular}{lcccccc}
\toprule
\textbf{Model}
& \textbf{R@1}
& \textbf{R@5}
& \textbf{R@10}
& \textbf{R@20}
& \textbf{MRR}
& \textbf{Mean FRR} \tabularnewline
\midrule

HuBERT + KMeans + BPE
& $\mathbf{0.807}$
& $\mathbf{0.833}$
& $\mathbf{0.847}$
& $\mathbf{0.863}$
& $\mathbf{0.823}$
& $138.47$ \tabularnewline

WavLM + KMeans + BPE
& $0.760$
& $0.787$
& $0.803$
& $0.843$
& $0.777$
& $241.77$ \tabularnewline

PairAlign
& $0.710$
& $0.780$
& $0.810$
& $0.830$
& $0.740$
& $\mathbf{44.67}$ \tabularnewline

\bottomrule
\end{tabular}

\caption{
TIMIT segment-overlap retrieval after LibriSpeech-trained BPE compression of
the pretrained SSL streams. Frozen transforms are applied identically to
archive and augmented queries before normalized edit-distance ranking.
Both the dataloader and evaluation script use fixed seed $13$, fixing archive
construction, query selection, and augmented query views. All systems have
HitRate $1.0$; both SSL+BPE systems have median FRR $1$.
}
\label{tab:ssl_bpe_retrieval}
\end{table*}

\paragraph{Retrieval survives post-hoc compression more strongly than paired-view geometry.}
Table~\ref{tab:ssl_bpe_retrieval} provides an important qualification.
Despite their large paired-view degradation, HuBERT+BPE retains R@1/MRR
$0.807/0.823$, above PairAlign's $0.710/0.740$, while WavLM+BPE gives
$0.760/0.777$. PairAlign slightly exceeds WavLM+BPE at R@10
($0.810$ versus $0.803$), whereas WavLM+BPE remains higher at R@1, R@5,
R@20, and MRR; HuBERT+BPE is strongest on all reported Recall/MRR metrics.
Thus, cross-view sequence geometry and retrieval discrimination are related
but not interchangeable: BPE can strongly alter the former while preserving
substantial relative-distance information.

Rate control nevertheless sharply reduces the native SSL retrieval margin.
Relative to the preceding native-rate stress test, HuBERT falls from R@1/MRR
$0.943/0.961$ to $0.807/0.823$, and WavLM from $0.977/0.985$ to
$0.760/0.777$. WavLM, the strongest native system, therefore becomes much
closer to PairAlign after compression and is slightly below it at R@10.
Symbolic granularity consequently contributes materially alongside pretrained
encoder quality. HuBERT still retains a meaningful retrieval advantage, so
these results do not support universal PairAlign superiority at low rate.

\paragraph{Compressed SSL retrieval has stronger heads but heavier tails.}
As further shown in Table~\ref{tab:ssl_bpe_retrieval}, HuBERT+BPE and
WavLM+BPE have Mean FRR $138.47$ and $241.77$, versus $44.67$ for
PairAlign, despite median FRR $1$ and HitRate $1.0$ for both SSL+BPE systems.
Their distributions are therefore highly skewed: many queries retrieve a
relevant item immediately, but a minority fail at very deep ranks. PairAlign
has weaker head-of-list accuracy but substantially smaller mean
first-relevant-rank displacement. No system uniformly dominates this
rate-controlled retrieval comparison.


\begin{table*}[t]
\centering
\scriptsize
\setlength{\tabcolsep}{3.4pt}

\begin{tabular}{lccccc}
\toprule
\textbf{Model}
& \textbf{Throughput}
& \textbf{Wall time}
& \textbf{Model latency}
& \textbf{End-to-end}
& \textbf{Peak GPU memory} \tabularnewline
&
\textbf{(segments/s)}
& \textbf{(s)}
& \textbf{(ms/segment)}
& \textbf{(ms/segment)}
& \textbf{(MB)} \tabularnewline
\midrule

Stage~I Geometric + BPE
& $1.26$
& $7509.62$
& $1.49$
& $793.73$
& $126.93$ \tabularnewline

HuBERT + KMeans + BPE
& $29.61$
& $319.50$
& $2.95$
& $33.75$
& $1129.17$ \tabularnewline

WavLM + KMeans + BPE
& $11.20$
& $845.05$
& $20.02$
& $89.30$
& $2272.30$ \tabularnewline

PairAlign
& $5.50$
& $1719.47$
& $194.27$
& $206.09$
& $570.58$ \tabularnewline

\bottomrule
\end{tabular}

\caption{
Measured tokenization efficiency on the same $9{,}461$-segment TIMIT archive,
using batch size $16$ on an NVIDIA L4 GPU and evaluation seed $13$. Model
latency is CUDA-synchronized and batch-amortized. For SSL systems it includes
encoder inference, KMeans assignment, and consecutive deduplication; for
PairAlign it includes autoregressive beam search. End-to-end measurements also
include loading, feature extraction, transfer, and BPE/post-processing.
The BPE implementations apply learned merges sequentially, so these are
implementation measurements rather than intrinsic BPE complexity bounds.
}
\label{tab:rate_controlled_bpe_timing}
\end{table*}

\paragraph{Compactness and representation-construction cost are distinct.}
Table~\ref{tab:rate_controlled_bpe_timing} separates neural tokenization cost
from measured end-to-end construction cost. The neural stage strongly favors
non-autoregressive tokenization: Stage~I, HuBERT, and WavLM require $1.49$,
$2.95$, and $20.02$ ms/segment, respectively, versus $194.27$ ms for
PairAlign's autoregressive decoder. However, measured end-to-end behavior
depends strongly on BPE implementation. HuBERT+BPE and WavLM+BPE remain
faster than PairAlign ($33.75/89.30$ versus $206.09$ ms/segment), whereas
Stage~I+BPE is much slower ($793.73$ ms/segment) because its measured
post-processing dominates. Corresponding throughputs are $29.61$, $11.20$,
$5.50$, and $1.26$ segments/s for HuBERT+BPE, WavLM+BPE, PairAlign, and
Stage~I+BPE.

Memory exhibits a different ordering: Stage~I+BPE, PairAlign, HuBERT+BPE, and
WavLM+BPE use $126.93$, $570.58$, $1129.17$, and $2272.30$ MB,
respectively. For WavLM, sequential BPE accounts for approximately $632.25$ s
of the $845.05$ s archive wall time. These numbers characterize the evaluated
pipelines, not a general complexity ordering between post-hoc merging and
autoregressive generation.

\paragraph{Overall interpretation.}
The combined controls separate three questions. First, Stage~I+BPE provides
the controlled test of whether shortening the representation from which
PairAlign is built recovers PairAlign's graded cross-view geometry; it does
not. Second, HuBERT+BPE and WavLM+BPE show that the same conclusion holds
when post-hoc compression starts from substantially stronger pretrained
frame-level representations. Third, Table~\ref{tab:ssl_bpe_retrieval} shows
that degradation in paired-view geometry does not imply loss of every
discriminative cue: HuBERT in particular retains stronger early-rank retrieval
after compression.

The resulting claim is therefore intentionally bounded. PairAlign learns a
lower-rate, input-conditioned, variable-length sequence with substantially
stronger cross-view organization than the tested Stage~I, HuBERT, and WavLM
BPE controls, despite those compressed streams remaining denser on TIMIT.
This does not imply that PairAlign dominates dense SSL tokenizers, nor that
post-hoc compression is intrinsically inferior: native SSL representations
retain stronger fine-grained local and retrieval information, and alternative
learned compressors remain possible.

\paragraph{Scaling path.}
These results suggest a complementary scaling direction. Pretrained SSL
representations provide strong local acoustic--phonetic geometry, while
PairAlign provides adaptive low-rate sequence learning. Replacing or
initializing PairAlign's Stage~I representation with HuBERT/WavLM features,
while retaining the autoregressive bridge and EMA self-alignment stages, would
test whether stronger pretrained local geometry can be transferred into a
compact adaptive sequence without the graded-geometry loss observed under
fixed BPE.

The present experiments motivate but do not establish this hypothesis. They
instead show that pretrained frame-level quality, symbolic rate, cross-view
sequence geometry, retrieval discrimination, and token-generation cost are
distinct axes of the tokenizer operating point.

\section{Phonetic Association and Boundary Structure of the Learned Tokens}
\label{sec:phonetic_boundary_analysis}


\begin{table*}[t]
\centering
\scriptsize
\setlength{\tabcolsep}{4.2pt}
\renewcommand{\arraystretch}{1.08}

\begin{tabularx}{\textwidth}{
    >{\raggedright\arraybackslash}X
    c
    c
    c
    c
    c
}
\toprule

\textbf{Metric}
&
\makecell{\textbf{Preferred}\\\textbf{direction}}
&
\makecell{\textbf{Stage~I}\\\textbf{Geometric}}
&
\makecell{\textbf{PairAlign}\\\textbf{+ Beta}}
&
\makecell{\textbf{PairAlign gain}\\
$\boldsymbol{\Delta_{\mathrm{PA}}}$}
&
\makecell{\textbf{Relative}\\\textbf{gain}}
\\

\midrule

\multicolumn{6}{l}{\textbf{Ground-truth phone-boundary precision}} \\
\addlinespace[2pt]

Boundary precision @ 10 ms
    & $\uparrow$ & $0.2452$ & $\mathbf{0.3563}$
    & $\mathbf{+0.1111}$ & $\mathbf{+45.3\%}$ \\

Boundary precision @ 20 ms
    & $\uparrow$ & $0.3192$ & $\mathbf{0.4579}$
    & $\mathbf{+0.1388}$ & $\mathbf{+43.5\%}$ \\

Boundary precision @ 30 ms
    & $\uparrow$ & $0.3528$ & $\mathbf{0.5142}$
    & $\mathbf{+0.1613}$ & $\mathbf{+45.7\%}$ \\

\midrule

\multicolumn{6}{l}{\textbf{Ground-truth phone-boundary recall and F1}} \\
\addlinespace[2pt]

Boundary recall @ 10 ms
    & $\uparrow$ & $\mathbf{0.5476}$ & $0.2741$
    & $-0.2734$ & $-49.9\%$ \\

Boundary recall @ 20 ms
    & $\uparrow$ & $\mathbf{0.7052}$ & $0.3468$
    & $-0.3584$ & $-50.8\%$ \\

Boundary recall @ 30 ms
    & $\uparrow$ & $\mathbf{0.7751}$ & $0.3859$
    & $-0.3892$ & $-50.2\%$ \\

Boundary F1 @ 10 ms
    & $\uparrow$ & $\mathbf{0.3321}$ & $0.2952$
    & $-0.0369$ & $-11.1\%$ \\

Boundary F1 @ 20 ms
    & $\uparrow$ & $\mathbf{0.4306}$ & $0.3758$
    & $-0.0548$ & $-12.7\%$ \\

Boundary F1 @ 30 ms
    & $\uparrow$ & $\mathbf{0.4750}$ & $0.4195$
    & $-0.0556$ & $-11.7\%$ \\

Boundary MAE @ 20-ms tolerance
    & $\downarrow$ & $\mathbf{6.556}$ ms & $8.626$ ms
    & $-2.070$ ms & $-31.6\%$ \\

\midrule

\multicolumn{6}{l}{\textbf{Segmentation and local token--phone correspondence}} \\
\addlinespace[2pt]

Boundary-count deviation
    & $\downarrow$ & $0.7948$ & $\mathbf{0.4520}$
    & $\mathbf{+0.3429}$ & $\mathbf{+43.1\%}$ \\

Token-segment phone purity
    & $\uparrow$ & $\mathbf{0.8115}$ & $0.5268$
    & $-0.2847$ & $-35.1\%$ \\

Phone-segment token purity
    & $\uparrow$ & $0.7046$ & $\mathbf{0.8342}$
    & $\mathbf{+0.1296}$ & $\mathbf{+18.4\%}$ \\

Mean tokens per phone segment
    & $\downarrow$ & $2.2580$ & $\mathbf{1.6560}$
    & $\mathbf{+0.6020}$ & $\mathbf{+26.7\%}$ \\

\bottomrule
\end{tabularx}

\caption{
Ground-truth temporal comparison on $4{,}620$ clean $3$ s TIMIT segments.
Stage~I is a dense frame-derived stream; PairAlign positions are mapped to the
$301$ conditioning frames using Beta-reweighted cross-attention and monotone
Viterbi decoding. PairAlign has higher boundary precision, lower boundary-count
deviation, fewer within-phone token fragments, and higher phone-segment token
purity; Stage~I has higher recall/F1, lower matched-boundary MAE, and higher
token-segment phone purity. For $\uparrow$ metrics,
$\Delta_{\mathrm{PA}}=\mathrm{PairAlign}-\mathrm{Geometric}$; for $\downarrow$
metrics, the order is reversed, so positive values always favor PairAlign.
Relative gain is against Stage~I. Boundary-count deviation is the absolute log
predicted/reference boundary-count ratio. Both validation-dataloader and
evaluation-script seeds are fixed to $13$.
}
\label{tab:pairalign_beta_vs_geometric_boundaries}
\end{table*}


\begin{table*}[t]
\centering
\scriptsize
\setlength{\tabcolsep}{4.2pt}
\renewcommand{\arraystretch}{1.08}

\begin{tabularx}{\textwidth}{
    >{\raggedright\arraybackslash}X
    c
    c
    c
    c
    c
}
\toprule

\textbf{Metric}
&
\makecell{\textbf{Preferred}\\\textbf{direction}}
&
\makecell{\textbf{Stage~I}\\\textbf{Geometric}}
&
\makecell{\textbf{PairAlign}\\\textbf{+ Beta}}
&
\makecell{\textbf{PairAlign gain}\\
$\boldsymbol{\Delta_{\mathrm{PA}}}$}
&
\makecell{\textbf{Relative}\\\textbf{gain}}
\\

\midrule

\multicolumn{6}{l}{\textbf{HuBERT-style phonetic association}} \\
\addlinespace[2pt]

Phone purity
    & $\uparrow$ & $\mathbf{0.2915}$ & $0.2465$
    & $-0.0450$ & $-15.4\%$ \\

Cluster purity
    & $\uparrow$ & $0.0809$ & $\mathbf{0.0906}$
    & $\mathbf{+0.0096}$ & $\mathbf{+11.9\%}$ \\

Phone-normalized mutual information
    & $\uparrow$ & $\mathbf{0.1467}$ & $0.0802$
    & $-0.0666$ & $-45.4\%$ \\

LOSO token-to-phone accuracy
    & $\uparrow$ & $\mathbf{0.2894}$ & $0.2428$
    & $-0.0466$ & $-16.1\%$ \\

\midrule

\multicolumn{6}{l}{\textbf{Silence and speaker association}} \\
\addlinespace[2pt]

Silence/non-silence purity
    & $\uparrow$ & $\mathbf{0.8251}$ & $0.8157$
    & $-0.0094$ & $-1.1\%$ \\

Speaker NMI
    & $\downarrow$ & $0.2486$ & $\mathbf{0.2329}$
    & $\mathbf{+0.0157}$ & $\mathbf{+6.3\%}$ \\

Conditional speaker NMI given phone
    & $\downarrow$ & $\mathbf{0.3507}$ & $0.3615$
    & $-0.0109$ & $-3.1\%$ \\

\midrule

\multicolumn{6}{l}{\textbf{Discrete minimal-pair discrimination}} \\
\addlinespace[2pt]

Within-speaker ABX error
    & $\downarrow$ & $\mathbf{0.4550}$ & $0.5103$
    & $-0.0553$ & $-12.2\%$ \\

Across-speaker ABX error
    & $\downarrow$ & $\mathbf{0.4691}$ & $0.4812$
    & $-0.0122$ & $-2.6\%$ \\

\bottomrule
\end{tabularx}

\caption{
Phonetic, nuisance, and discrete ABX comparison of Stage~I and Beta-aligned
PairAlign. HuBERT-style/nuisance metrics use the same $1{,}228{,}504$ valid
phone-aligned frames; ABX uses the same $1{,}563$ within-speaker and $20{,}559$
across-speaker triplets. PairAlign has higher cluster purity and lower
unconditional speaker NMI; Stage~I has higher phone purity, PNMI, LOSO phone
predictability, and lower ABX error. PairAlign's higher conditional speaker NMI
precludes interpreting its lower unconditional NMI as speaker removal. Both
validation-dataloader and evaluation-script seeds are fixed to $13$.
$\Delta_{\mathrm{PA}}$ follows the preferred metric direction as above.
}
\label{tab:pairalign_beta_vs_geometric_phonetics}
\end{table*}


\begin{table*}[t]
\centering
\scriptsize
\setlength{\tabcolsep}{4.4pt}
\renewcommand{\arraystretch}{1.08}

\begin{tabular}{llccc}
\toprule

\textbf{Probe}
&
\textbf{Context}
&
\textbf{Stage~I Geometric}
&
\textbf{PairAlign Raw}
&
\textbf{PairAlign + Beta}
\\

\midrule

\multicolumn{5}{l}{\textbf{Short-context phonetic association}} \\
\addlinespace[2pt]

Phone-normalized MI $\uparrow$
& Bigram
& $\mathbf{0.2072}$
& $0.1084$
& $0.1082$
\\

Phone-normalized MI $\uparrow$
& Trigram
& $\mathbf{0.2987}$
& $0.1474$
& $0.1473$
\\

LOSO token-context $\rightarrow$ phone accuracy $\uparrow$
& Bigram
& $\mathbf{0.2894}$
& $0.2418$
& $0.2403$
\\

LOSO token-context $\rightarrow$ phone accuracy $\uparrow$
& Trigram
& $\mathbf{0.2875}$
& $0.2444$
& $0.2430$
\\

\midrule

\multicolumn{5}{l}{\textbf{Sequence-level speaker recoverability}} \\
\addlinespace[2pt]

Speaker-ID accuracy $\downarrow$
& Unigram
& $0.2015$
& $\mathbf{0.1240}$
& $\mathbf{0.1240}$
\\

Speaker-ID accuracy $\downarrow$
& Bigram
& $0.1814$
& $\mathbf{0.0684}$
& $\mathbf{0.0684}$
\\

Speaker-ID accuracy $\downarrow$
& Trigram
& $0.1307$
& $\mathbf{0.0355}$
& $\mathbf{0.0355}$
\\

Speaker-verification ROC-AUC $\downarrow$
& Unigram
& $0.8679$
& $\mathbf{0.7946}$
& $\mathbf{0.7946}$
\\

Speaker-verification ROC-AUC $\downarrow$
& Bigram
& $0.8122$
& $\mathbf{0.6036}$
& $\mathbf{0.6036}$
\\

Speaker-verification ROC-AUC $\downarrow$
& Trigram
& $0.7298$
& $\mathbf{0.5451}$
& $\mathbf{0.5451}$
\\

Speaker-verification EER $\uparrow$
& Unigram
& $0.2119$
& $\mathbf{0.2766}$
& $\mathbf{0.2766}$
\\

Speaker-verification EER $\uparrow$
& Bigram
& $0.2513$
& $\mathbf{0.4378}$
& $\mathbf{0.4378}$
\\

Speaker-verification EER $\uparrow$
& Trigram
& $0.3312$
& $\mathbf{0.4722}$
& $\mathbf{0.4722}$
\\

\bottomrule
\end{tabular}

\caption{
Short-context phonetic and sequence-level speaker probes on the same $4{,}620$
clean $3$ s TIMIT segments from $462$ speakers; both validation-dataloader and
evaluation-script seeds are fixed to $13$. Phonetic probes use aligned
bigrams/trigrams to predict the central 39-phone label with LOSO evaluation.
Speaker-ID uses five-fold utterance-level logistic regression over
$\ell_2$-normalized token $n$-gram counts; chance is
$1/462\approx0.0022$. Verification uses cosine similarity on the same features
over $20{,}790$ same-speaker and $20{,}790$ different-speaker pairs per order.
Lower ID accuracy/ROC-AUC and higher EER indicate less information recoverable
by these specific lightweight probes, not privacy guarantees. Raw and
PairAlign+Beta are identical on sequence-level speaker probes because Beta
changes only token-to-frame alignment.
}
\label{tab:pairalign_context_privacy_probes}
\end{table*}

The preceding results establish a granularity trade-off: PairAlign is much
shorter while retaining paired-view and relational structure; Stage~I preserves
more local overlap and retrieval discrimination. We now anchor this trade-off
to TIMIT phone annotations and sequence-level speaker probes, asking what
phone-, boundary-, and speaker-associated structure survives compression. Both
the validation dataloader and evaluation script use fixed seed $13$ throughout.

Neither model uses phone labels or an explicit cluster-derived speech-unit
prediction objective. Stage~I combines SimCLR-style acoustic consistency with
vector quantization; PairAlign reorganizes this representation through
autoregressive prediction and sequence-level self-alignment. Unlike
HuBERT/WavLM-style pretraining, neither receives frame-aligned clustered targets
that explicitly bias learning toward segmental, phone-correlated structure.

Phonetic association may nevertheless emerge because phones induce recurring
spectral, voicing, temporal, and transition regularities. We therefore interpret
the following measurements as probes of \emph{emergent phonetic association},
not recovery of a phoneme inventory.

PairAlign positions are mapped to the $301$ conditioning frames using
Beta-reweighted cross-attention and monotone Viterbi decoding.
Tables~\ref{tab:pairalign_beta_vs_geometric_boundaries} and
\ref{tab:pairalign_beta_vs_geometric_phonetics} report boundary, phonetic,
silence, and first-order speaker association;
Table~\ref{tab:pairalign_context_privacy_probes} adds short-context phonetic
and attacker-style speaker-recoverability probes.

\paragraph{PairAlign produces fewer but more selective phone-aligned boundaries.}
PairAlign raises boundary precision from $0.245/0.319/0.353$ to
$0.356/0.458/0.514$ at $10/20/30$ ms, relative gains of roughly
$43$--$46\%$, while reducing boundary-count deviation from $0.795$ to $0.452$.
However, recall is about half of Stage~I across tolerances, yielding lower F1
and higher matched-boundary MAE. PairAlign is therefore not a better boundary
detector: its lower-rate trajectory emits fewer transitions, but those emitted
are more selectively concentrated near reference boundaries. This matches the
broader granularity trade-off: dense Stage~I covers more transitions; PairAlign
compresses them into fewer, more selective changes.

\paragraph{PairAlign reduces within-phone fragmentation but does not induce one-token--one-phone units.}
Mean tokens per reference phone fall from $2.258$ to $1.656$ ($26.7\%$), while
phone-segment token purity rises from $0.705$ to $0.834$, showing fewer token
changes within phones. Conversely, token-segment phone purity falls from
$0.812$ to $0.527$, so PairAlign segments more often span multiple phones.
PairAlign therefore reduces within-phone fragmentation by learning units
coarser than phonemes, not a one-token--one-phone segmentation.

\paragraph{Both tokenizers show emergent phonetic association, but Stage~I retains finer phone identity.}
Without phone supervision, both representations retain measurable phone
structure. Stage~I versus PairAlign gives phone purity
$0.292$ versus $0.247$, PNMI $0.147$ versus $0.080$, and LOSO token-to-phone
accuracy $0.289$ versus $0.243$. PairAlign alone improves cluster purity
($0.081\rightarrow0.091$), but this does not reverse the broader ordering:
Stage~I also has lower within-/across-speaker ABX error
($0.455/0.469$ versus $0.510/0.481$). Thus, PairAlign does not improve phonetic
discriminability; lower-rate self-alignment discards part of Stage~I's
fine-grained phone-correlated structure.

\paragraph{Short token context does not recover the phonetic resolution lost under compression.}
Bigrams and trigrams test whether PairAlign relocates phone information from
single symbols into short compositions. PNMI increases with context for both:
Stage~I rises from $0.147$ to $0.207/0.299$, and PairAlign from about $0.080$
to $0.108/0.147$ for bigrams/trigrams. Yet Stage~I remains substantially
stronger, with LOSO context-to-phone accuracy near $0.29$ versus PairAlign near
$0.24$. PairAlign therefore retains short-context phonetic structure, but its
unigram deficit is not merely redistributed into short $n$-grams.
\paragraph{The phonetic association is emergent rather than imposed by a speech-unit prior.}
Stage~I and PairAlign must discover phone-correlated structure directly from raw
speech, without frame-aligned pseudo-labels or cluster-derived speech-unit
targets. HuBERT instead begins from KMeans targets constructed from MFCCs and
then refines them using learned intermediate representations; WavLM inherits
this HuBERT-style target construction. These objectives therefore provide a
substantially stronger prior toward segmental and phone-correlated structure
than is available to Stage~I or PairAlign.

Accordingly, the phone purity, PNMI, predictability, ABX structure, boundary
alignment, and short-context association observed for Stage~I and PairAlign are
emergent properties of acoustic regularities learned through contrastive
representation learning, quantization, and subsequent sequence-level
self-alignment. PairAlign preserves only part of this fine-grained phonetic
structure while reorganizing it into a substantially lower-rate sequence. Its
tokens should therefore be interpreted as acoustic--symbolic units correlated
with phonetic structure, not as learned phonemes or categories explicitly
supervised toward a speech-unit inventory.

\paragraph{Boundary selectivity and phone discriminability capture different effects.}
PairAlign is more boundary-selective in \emph{precision}: when its compact
trajectory changes, the transition is more likely near a phone boundary.
Stage~I retains greater \emph{coverage and phone identity}: higher recall,
phone association, and minimal-pair discrimination. PairAlign therefore does
not become ``more phonetic''; it reorganizes the dense stream into fewer,
more selective transitions with less phone fragmentation but weaker fine-grained
phone identity. The rate-controlled BPE result further shows that this coarse
organization is not reproduced by post-hoc shortening alone.

\paragraph{The same loss of phonetic resolution explains the retrieval trade-off.}
Stage~I's dense stream retains stronger phone-correlated structure and lower ABX
error, providing more local cues for edit-distance ranking and explaining its
stronger top-rank retrieval and normalized local persistence. PairAlign retains
fewer local distinctions because each symbol plays a coarser role. Yet its
non-trivial phone association, higher boundary precision, reduced within-phone
fragmentation, paired-view consistency, and positive--negative geometry show
that compression does not make the representation acoustically arbitrary.
Useful retrieval therefore requires structured acoustic variation, not
one-token--one-phone correspondence.

\paragraph{First-order speaker association gives a mixed privacy signal.}
Silence/non-silence purity changes little ($0.825\rightarrow0.816$).
Unconditional speaker NMI falls from $0.249$ to $0.233$, whereas conditional
speaker NMI given phone rises from $0.351$ to $0.362$. These measures therefore
do not support speaker removal: conditioning on phone shows that
speaker-associated variation remains within phonetic classes, so compactness
must not be interpreted as anonymization.

\paragraph{Sequence-level probes show reduced speaker recoverability, but not speaker removal.}
Lightweight attacker probes show consistently lower speaker recoverability from
PairAlign. Speaker-ID accuracy falls from $0.202/0.181/0.131$ to
$0.124/0.068/0.036$ for unigram/bigram/trigram features, but remains above the
$462$-speaker chance level $1/462\approx0.0022$. Verification follows the same
ordering: ROC-AUC decreases from $0.868/0.812/0.730$ to
$0.795/0.604/0.545$, while EER rises from $0.212/0.251/0.331$ to
$0.277/0.438/0.472$. PairAlign thus approaches the non-discriminative
$0.5$ AUC/$0.5$ EER point more closely for trigram counts, but speaker
information remains measurable. These results establish reduced
\emph{measured recoverability} under short-order count-based probes, not speaker
invariance; stronger or longer-context attackers may recover more.

\paragraph{Beta improves temporal grounding, not speaker privacy.}
PairAlign Raw and PairAlign+Beta have identical sequence-level speaker-probe
results because Beta changes only token-to-frame alignment, not token identity
or order. It can therefore affect boundary and frame-aligned phonetic metrics
but not bag-of-$n$-gram speaker probes. Speaker-relevant differences arise from
Stage~I versus PairAlign itself; Beta is a temporal-grounding mechanism, not a
speaker-suppression mechanism.

\paragraph{Reduced leakage is an empirical operating-point effect, not a privacy guarantee.}
PairAlign has no speaker-adversarial loss, differential-privacy mechanism,
anonymization objective, or explicit speaker-information constraint. Its lower
speaker recoverability may simply reflect the coarse symbolic bottleneck that
also discards fine-grained acoustic detail. This experiment does not identify
which factors are removed or imply that low-rate tokenization is inherently
privacy-preserving. It shows only that, on this TIMIT protocol and under the
evaluated unigram--trigram identification/verification probes, speaker identity
is harder---but still possible---to recover from PairAlign than Stage~I.

\paragraph{Overall interpretation.}
The results anchor the paper's rate--granularity trade-off to external linguistic
structure. Both tokenizers exhibit emergent phone association despite no
explicit phone-oriented or cluster-derived speech-unit objective. Stage~I
preserves finer structure: higher phone purity, PNMI, short-context association,
LOSO phone prediction, ABX discrimination, boundary recall, and matched-boundary
accuracy, while also exposing more speaker information to the evaluated probes.

PairAlign moves to coarser sequence organization. Its shorter trajectory has
weaker phone identity but fewer within-phone changes, a boundary count closer
to reference segmentation, and substantially higher precision for emitted
boundaries. Short $n$-gram context does not recover the lost phonetic
resolution, indicating genuine loss of local phone-correlated detail rather
than redistribution across neighboring symbols.

The same coarse operating point yields lower measured speaker recoverability,
but not privacy or speaker removal. PairAlign should therefore be viewed neither
as a frame-synchronous phonetic code nor as an anonymized representation, but
as a lower-rate acoustic--symbolic interface that retains sufficient structure
for paired-view consistency, relational geometry, boundary selectivity, and
retrieval while discarding part of Stage~I's fine-grained phonetic and
speaker-discriminative detail.


\section{Local Objective Geometry of Stage~III}
\label{sec:pairalign-local-theory}

\paragraph{Scope.}
Stage~III combines two mechanisms. First, \emph{shared-target fitting} trains
the anchor and its content-preserving positive view toward the same
anchor-derived EMA sequence. Second, \emph{likelihood contrast} separates that
paired target from mismatched EMA candidates in conditional score space. We
ask when these two effects are sufficient for the desired local edit-distance
ordering.

The analysis is deliberately local. We assume neither a dataset-wide
likelihood--edit relationship nor a parametric score--edit model. Instead,
likelihood contrast first yields an explicit row-wise score gap, and a
leave-one-candidate-out (LOO) monotone audit tests when that score separation
transfers to edit space.

On held-out TIMIT, the desired ordering is observed directly in
$99.76$--$99.90\%$ of hard-candidate comparisons. At the primary $k=15$
setting, at least one conditioning view provides a validated LOO calibration
for $71.76\pm0.07\%$ of comparisons across three independently generated EMA
target banks. The realized shared-target contraction condition then certifies
$59.21\pm0.08\%$ of all comparisons from the observed candidate score gap and
$52.92\pm0.20\%$ using only the score gap implied directly by the NCE
objective.


\subsection{Setup and Notation}

Consider a content-preserving pair $(x_i,x_i^{+})$, where $x_i$ is the anchor
and $x_i^{+}$ is its augmented positive view. The student encoder produces
\begin{equation}
    Z_i = Enc_{\theta}(x_i),
    \qquad
    Z_i^{+} = Enc_{\theta}(x_i^{+}).
    \label{eq:local-two-conditions}
\end{equation}

The EMA teacher generates one EOS-terminated target from the \emph{anchor
only},
\begin{equation}
    \widehat{\mathcal Y}_i
    =
    [\widehat{\mathcal T}_i,\mathrm{EOS}],
    \label{eq:local-canonical-target}
\end{equation}
and Stage~III uses this same target for both conditioning paths:
\begin{equation}
\boxed{
    Z_i
    \longrightarrow
    \widehat{\mathcal Y}_i,
    \qquad
    Z_i^{+}
    \longrightarrow
    \widehat{\mathcal Y}_i.
}
\label{eq:local-shared-target}
\end{equation}
Thus, Stage~III does not generate a separate positive-view target
$\widehat{\mathcal Y}_i^{+}$.

Other minibatch examples $j\neq i$ provide mismatched EMA targets
$\widehat{\mathcal Y}_j$. We use $V\in\{A,P\}$ for the anchor- and
positive-conditioned score rows,
\begin{equation}
    Z_i^A=Z_i,
    \qquad
    Z_i^P=Z_i^{+}.
\end{equation}

For one realization of Stage~III positive fitting, let
$\widetilde{\mathcal Y}_i^A$ and
$\widetilde{\mathcal Y}_i^P$ denote the position-wise teacher-forced student
argmax predictions under $Z_i$ and $Z_i^{+}$, respectively, both evaluated
against $\widehat{\mathcal Y}_i$. These are teacher-forced student predictions,
not independently free-running EMA tokenizations. Throughout,
$ED(\cdot,\cdot)$ denotes Levenshtein edit distance.

The local relational property of interest is
\begin{equation}
\boxed{
    ED(
        \widetilde{\mathcal Y}_i^A,
        \widetilde{\mathcal Y}_i^P
    )
    <
    \min
    \left\{
        ED(
            \widetilde{\mathcal Y}_i^A,
            \widehat{\mathcal Y}_j
        ),
        ED(
            \widetilde{\mathcal Y}_i^P,
            \widehat{\mathcal Y}_j
        )
    \right\}.
}
\label{eq:local-relational-ordering}
\end{equation}
That is, the two predictions of the same underlying content should be closer
to one another than either is to a locally competitive mismatched target.

The argument has three steps:
\begin{equation}
\boxed{
\begin{aligned}
    \text{likelihood contrast}
    &\Longrightarrow
    \text{row-local score gap},
    \\
    \text{LOO monotone calibration}
    &\Longrightarrow
    \text{validated edit lower bound},
    \\
    \text{shared-target contraction}
    +\text{edit separation}
    &\Longrightarrow
    \text{Eq.~\ref{eq:local-relational-ordering}}.
\end{aligned}
}
\label{eq:local-theory-roadmap}
\end{equation}


\subsection{Step 1: Likelihood Contrast Gives a Row-Local Score Gap}

Write
$\widehat{\mathcal Y}_j=[\hat y_{j,1},\ldots,\hat y_{j,L_j}]$.
Its clean, length-normalized student log-likelihood under conditioning view
$V$ is
\begin{equation}
    S_{ij,V} = \frac{1}{L_j}\sum_{l=1}^{L_j}\log p_{\theta}\left(\hat y_{j,l}\mid\hat y_{j,<l},Z_i^V\right).
    \label{eq:local-clean-score}
\end{equation}
Here $S_{ii,V}$ scores the paired canonical target
$\widehat{\mathcal Y}_i$, while $S_{ij,V}$ scores mismatched candidate
$\widehat{\mathcal Y}_j$ under the same acoustic condition.

\paragraph{Candidate-specific and row-wise gaps.}
Define the paired-versus-candidate score gap
\begin{equation}
\boxed{\delta_{ij,V} = S_{ii,V}-S_{ij,V}.}
\label{eq:local-delta}
\end{equation}
Thus, $\delta_{ij,V}>0$ means that the paired target scores above candidate
$j$. The hardest realized comparison in the row is summarized by
\begin{equation}
\boxed{G_{i,V} = \min_{j\neq i}\delta_{ij,V} = S_{ii,V}-\max_{j\neq i}S_{ij,V}.}
\label{eq:minimum-row-score-gap}
\end{equation}
Hence $G_{i,V}>0$ means that the paired target outranks every available
mismatch.

For local calibration, let
\begin{equation}
    \mathcal H_k^V(i) = \operatorname{TopKIndices}_{j\neq i}S_{ij,V}
    \label{eq:hard-neighborhood}
\end{equation}
contain the $k$ highest-scoring mismatches. The neighborhood size $k$ affects
only the post-hoc calibration below; the NCE margin uses the complete
available in-batch mismatch set.

\paragraph{NCE-implied gap.}
With likelihood-contrast temperature $\tau_{\ell}>0$, the realized row loss is
\begin{equation}
    \ell_i^V = \log\left[1+\sum_{j\neq i}\exp\left(\frac{S_{ij,V}-S_{ii,V}}{\tau_{\ell}}\right)\right].
    \label{eq:local-row-nce-loss}
\end{equation}
Define
\begin{equation}
\boxed{\Delta_{i,V}^{\mathrm{NCE}} = -\tau_{\ell}\log\sum_{j\neq i}\exp\left(\frac{S_{ij,V}-S_{ii,V}}{\tau_{\ell}}\right).}
\label{eq:local-row-nce-margin}
\end{equation}
Equivalently,
\begin{equation}
    \Delta_{i,V}^{\mathrm{NCE}} = \tau_{\ell}\log\left(\frac{1}{e^{\ell_i^V}-1}\right).
\end{equation}

Because the log-sum-exp contains every individual negative term,
\begin{equation}
\boxed{\delta_{ij,V}\geq G_{i,V}\geq\Delta_{i,V}^{\mathrm{NCE}},\qquad\forall j\neq i.}
\label{eq:score-gap-hierarchy}
\end{equation}

\paragraph{Derivation of the score-gap hierarchy.}
The first inequality follows immediately from the definition of the minimum
row gap:
\begin{equation}
    G_{i,V}=\min_{m\neq i}\delta_{im,V}\leq\delta_{ij,V},\qquad\forall j\neq i.
    \label{eq:delta-geq-row-gap}
\end{equation}
For the second inequality, using
$\delta_{im,V}=S_{ii,V}-S_{im,V}$ gives
\begin{equation}
    \Delta_{i,V}^{\mathrm{NCE}}=-\tau_{\ell}\log\sum_{m\neq i}\exp\left(-\frac{\delta_{im,V}}{\tau_{\ell}}\right).
    \label{eq:nce-gap-delta-form}
\end{equation}
Since $G_{i,V}=\min_{m\neq i}\delta_{im,V}$, there exists a hardest mismatch
$m^\star$ satisfying $\delta_{im^\star,V}=G_{i,V}$. Therefore,
\begin{equation}
    \sum_{m\neq i}\exp\left(-\frac{\delta_{im,V}}{\tau_{\ell}}\right)\geq\exp\left(-\frac{G_{i,V}}{\tau_{\ell}}\right).
\end{equation}
Because $\log(\cdot)$ is increasing and $-\tau_{\ell}<0$, multiplication by
$-\tau_{\ell}$ reverses the inequality:
\begin{equation}
    \Delta_{i,V}^{\mathrm{NCE}}\leq-\tau_{\ell}\log\exp\left(-\frac{G_{i,V}}{\tau_{\ell}}\right)=G_{i,V}.
    \label{eq:nce-gap-leq-row-gap}
\end{equation}
Combining Eqs.~\ref{eq:delta-geq-row-gap} and
\ref{eq:nce-gap-leq-row-gap} yields
\begin{equation}
    \delta_{ij,V}\geq G_{i,V}\geq\Delta_{i,V}^{\mathrm{NCE}},\qquad\forall j\neq i.
\end{equation}
Thus, $\delta_{ij,V}$ is the candidate-specific realized gap,
$G_{i,V}$ is the worst-case realized row gap, and
$\Delta_{i,V}^{\mathrm{NCE}}$ is the conservative row gap implied directly by
the NCE loss. Moreover,
$\Delta_{i,V}^{\mathrm{NCE}}>0$ whenever $\ell_i^V<\log 2$, which also implies
$G_{i,V}>0$.

This result is entirely in score space; relating it to edit distance requires
an additional empirical calibration step.


\subsection{Step 2: LOO Monotone Calibration Transfers Score Gaps to Edit Space}

Define the edit distance between the paired EMA target
$\widehat{\mathcal Y}_i$ and mismatched EMA candidate
$\widehat{\mathcal Y}_j$ by
\begin{equation}
    D_{ij}=ED(\widehat{\mathcal Y}_i,\widehat{\mathcal Y}_j),
\end{equation}
and its normalized form by
\begin{equation}
    d_{ij}=\frac{D_{ij}}{L_{ij}},\qquad L_{ij}=\max\left\{|\widehat{\mathcal Y}_i|,|\widehat{\mathcal Y}_j|\right\}.
    \label{eq:local-normalized-edit}
\end{equation}
Thus, $d_{ij}$ measures the normalized edit separation between the canonical
EMA target for example $i$ and candidate target $j$.

We impose no affine or other parametric relationship between the score gap
$\delta_{ij,V}$ and the normalized edit distance $d_{ij}$. Instead, for each
candidate $j$, we exclude that candidate from the calibration set and use the
remaining locally competitive candidates to construct a monotone lower
envelope relating score gaps to edit distance.

For $j\in\mathcal H_k^V(i)$, define the leave-one-out calibration set
\begin{equation}
    \mathcal T_{i,V}^{(k,-j)}=\mathcal H_k^V(i)\setminus\{j\}.
    \label{eq:loo-training-set}
\end{equation}
Candidate $j$ contributes neither its score gap $\delta_{ij,V}$ nor its edit
distance $d_{ij}$ to the envelope used to evaluate it.

For a query score gap $u$, define
\begin{equation}
\boxed{
    \widehat{\kappa}_{i,V}^{(k,-j)}(u)
    =
    \min_{\substack{m\in\mathcal T_{i,V}^{(k,-j)}\\ \delta_{im,V}\geq u}}
    d_{im}.
}
\label{eq:loo-monotone-envelope}
\end{equation}
Thus, for a required score gap $u$, the envelope returns the smallest
normalized target--candidate edit distance observed among the remaining
calibration candidates whose score gap is at least $u$.

The query is \emph{supported} only when at least one calibration candidate has
a score gap at least as large as $u$, equivalently
\begin{equation}
    u\leq\max_{m\in\mathcal T_{i,V}^{(k,-j)}}\delta_{im,V}.
    \label{eq:loo-support}
\end{equation}
We do not extrapolate beyond this observed score-gap range.

\paragraph{Monotonicity and maximality of the envelope.}
The envelope is non-decreasing in $u$. If $u_2\geq u_1$, then
\begin{equation}
    \left\{m:\delta_{im,V}\geq u_2\right\}\subseteq\left\{m:\delta_{im,V}\geq u_1\right\},
\end{equation}
so increasing the required score gap can only reduce the set over which the
minimum is taken. Therefore,
\begin{equation}
    u_2\geq u_1\quad\Longrightarrow\quad\widehat{\kappa}_{i,V}^{(k,-j)}(u_2)\geq\widehat{\kappa}_{i,V}^{(k,-j)}(u_1).
\end{equation}

The envelope is also the maximal non-decreasing lower envelope consistent with
the $k-1$ calibration candidates. To see this, let $g$ be any non-decreasing
function satisfying
\begin{equation}
    g(\delta_{im,V})\leq d_{im}\qquad\forall m\in\mathcal T_{i,V}^{(k,-j)}.
\end{equation}
For any supported query $u$ and any calibration candidate satisfying
$\delta_{im,V}\geq u$, monotonicity gives
\begin{equation}
    g(u)\leq g(\delta_{im,V})\leq d_{im}.
\end{equation}
Taking the minimum over all calibration candidates with
$\delta_{im,V}\geq u$ therefore yields
\begin{equation}
    g(u)\leq\widehat{\kappa}_{i,V}^{(k,-j)}(u).
\end{equation}
Hence no larger non-decreasing lower envelope can remain consistent with all
$k-1$ calibration candidates.

\paragraph{Held-out calibration.}
Consistency with the $k-1$ calibration candidates does not guarantee that the
excluded candidate $j$ also lies above the envelope. We therefore evaluate
candidate $j$ out of fit and define the held-out calibration event
\begin{equation}
\boxed{
    \mathcal C_{ij,V}^{(k)}
    :=
    \left\{\delta_{ij,V}\text{ is supported}\right\}
    \cap
    \left\{d_{ij}\geq\widehat{\kappa}_{i,V}^{(k,-j)}(\delta_{ij,V})\right\}.
}
\label{eq:loo-calibration-event}
\end{equation}
This is the substantive empirical condition in Step~2: the excluded candidate
is tested against an envelope constructed without using either its score gap
or its edit distance.

When $\mathcal C_{ij,V}^{(k)}$ holds, define the candidate-specific observed
certificate
\begin{equation}
    q_{ij,V}^{\mathrm{obs}}=\widehat{\kappa}_{i,V}^{(k,-j)}(\delta_{ij,V}).
    \label{eq:q-obs}
\end{equation}
By definition of the held-out calibration event,
\begin{equation}
    d_{ij}\geq q_{ij,V}^{\mathrm{obs}}.
\end{equation}
Thus, once candidate $j$ passes the LOO calibration test,
$q_{ij,V}^{\mathrm{obs}}$ is a validated lower bound on its normalized EMA
target--candidate edit distance.

The same LOO envelope, still constructed without candidate $j$, can be queried
using only the NCE-implied row gap:
\begin{equation}
    q_{ij,V}^{\mathrm{NCE}}=\widehat{\kappa}_{i,V}^{(k,-j)}(\Delta_{i,V}^{\mathrm{NCE}}).
    \label{eq:q-nce}
\end{equation}
From Step~1,
\begin{equation}
    \Delta_{i,V}^{\mathrm{NCE}}\leq\delta_{ij,V}.
\end{equation}
Therefore, whenever the observed query $\delta_{ij,V}$ is supported, the NCE
query $\Delta_{i,V}^{\mathrm{NCE}}$ is also supported: any calibration
candidate whose gap is at least $\delta_{ij,V}$ necessarily also has gap at
least $\Delta_{i,V}^{\mathrm{NCE}}$.

Because the envelope is non-decreasing and
$\Delta_{i,V}^{\mathrm{NCE}}\leq\delta_{ij,V}$,
\begin{equation}
    q_{ij,V}^{\mathrm{NCE}}\leq q_{ij,V}^{\mathrm{obs}}.
\end{equation}
Combining this with the held-out calibration inequality gives
\begin{equation}
\boxed{
    d_{ij}\geq q_{ij,V}^{\mathrm{obs}}\geq q_{ij,V}^{\mathrm{NCE}}.
}
\label{eq:local-certificate-hierarchy}
\end{equation}

The two bounds have different levels of information. The observed certificate
$q_{ij,V}^{\mathrm{obs}}$ uses the realized candidate-specific gap
$\delta_{ij,V}$, whereas $q_{ij,V}^{\mathrm{NCE}}$ uses only the weaker
row-level gap $\Delta_{i,V}^{\mathrm{NCE}}$ implied directly by the
likelihood-contrast objective. The latter is therefore generally more
conservative.

Crucially, both $q_{ij,V}^{\mathrm{obs}}$ and
$q_{ij,V}^{\mathrm{NCE}}$ lower-bound the same normalized EMA
target--candidate distance
\begin{equation}
    d_{ij}=\frac{ED(\widehat{\mathcal Y}_i,\widehat{\mathcal Y}_j)}{L_{ij}}.
\end{equation}
They do not directly lower-bound the teacher-forced
prediction--candidate distances
$ED(\widetilde{\mathcal Y}_i^A,\widehat{\mathcal Y}_j)$ or
$ED(\widetilde{\mathcal Y}_i^P,\widehat{\mathcal Y}_j)$. Step~3 transfers this
validated EMA target--candidate lower bound to those prediction--candidate
distances through the reverse triangle inequality.


\subsection{Step 3: Shared-Target Contraction and Local Separation}

\paragraph{Shared-target contraction.}
The contraction argument follows specifically from the shared canonical target
in Eq.~\ref{eq:local-shared-target}. Stage~III does \emph{not} directly
minimize
\begin{equation}
    ED(\widetilde{\mathcal Y}_i^A,\widetilde{\mathcal Y}_i^P).
\end{equation}
Instead, both predictions are fitted to the same
$\widehat{\mathcal Y}_i$.

Define their target-relative errors
\begin{equation}
    e_{i,A}=ED(\widetilde{\mathcal Y}_i^A,\widehat{\mathcal Y}_i),\qquad e_{i,P}=ED(\widetilde{\mathcal Y}_i^P,\widehat{\mathcal Y}_i).
    \label{eq:positive-target-errors}
\end{equation}
and their realized mutual distance
\begin{equation}
    d_i^{AP}=ED(\widetilde{\mathcal Y}_i^A,\widetilde{\mathcal Y}_i^P).
    \label{eq:realized-positive-distance}
\end{equation}
Because both predictions are compared to the same symbolic reference, the
triangle inequality gives
\begin{equation}
\boxed{d_i^{AP}\leq e_{i,A}+e_{i,P}.}
\label{eq:positive-triangle}
\end{equation}
Thus, shared-target fitting induces contraction indirectly by reducing the two
errors to a common target.

\paragraph{Combining valid views.}
For candidate $j$, the LOO calibration may be valid under the
anchor-conditioned row, the positive-conditioned row, or both. Let
\begin{equation}
    \mathcal V_{ij}^{(k)}=\left\{V\in\{A,P\}:\mathcal C_{ij,V}^{(k)}\text{ holds}\right\}.
\end{equation}
For every valid view $V\in\mathcal V_{ij}^{(k)}$, Step~2 provides a lower bound
on the same normalized EMA target--candidate distance $d_{ij}$:
\begin{equation}
    d_{ij}\geq q_{ij,V}^{\mathrm{obs}}\geq q_{ij,V}^{\mathrm{NCE}}.
\end{equation}
If both views are valid, we may therefore retain the stronger of their two
lower bounds; if only one view is valid, this simply reduces to the bound from
that view. Accordingly, define
\begin{equation}
    q_{ij}^{\mathrm{obs}}=\max_{V\in\mathcal V_{ij}^{(k)}}q_{ij,V}^{\mathrm{obs}},\qquad q_{ij}^{\mathrm{NCE}}=\max_{V\in\mathcal V_{ij}^{(k)}}q_{ij,V}^{\mathrm{NCE}}.
    \label{eq:combined-q}
\end{equation}
Because every quantity inside either maximum is at most $d_{ij}$, their
maximum is also at most $d_{ij}$. Hence, for either
$q_{ij}\in\{q_{ij}^{\mathrm{obs}},q_{ij}^{\mathrm{NCE}}\}$,
\begin{equation}
    d_{ij}\geq q_{ij}.
\end{equation}
Recalling that $D_{ij}=L_{ij}d_{ij}$ and $L_{ij}>0$, multiplying both sides by
$L_{ij}$ gives
\begin{equation}
\boxed{D_{ij}\geq L_{ij}q_{ij}.}
\label{eq:target-negative-lower-bound}
\end{equation}

\paragraph{Reverse-triangle lower bound.}
Because Levenshtein edit distance is a metric, the triangle inequality applied
through the anchor-conditioned prediction gives
\begin{equation}
    D_{ij}=ED(\widehat{\mathcal Y}_i,\widehat{\mathcal Y}_j)\leq ED(\widehat{\mathcal Y}_i,\widetilde{\mathcal Y}_i^A)+ED(\widetilde{\mathcal Y}_i^A,\widehat{\mathcal Y}_j)=e_{i,A}+ED(\widetilde{\mathcal Y}_i^A,\widehat{\mathcal Y}_j).
    \label{eq:reverse-triangle-anchor-proof}
\end{equation}
Rearranging yields
\begin{equation}
    ED(\widetilde{\mathcal Y}_i^A,\widehat{\mathcal Y}_j)\geq D_{ij}-e_{i,A}.
    \label{eq:reverse-triangle-anchor}
\end{equation}
Applying the same argument through the positive-conditioned prediction gives
\begin{equation}
    ED(\widetilde{\mathcal Y}_i^P,\widehat{\mathcal Y}_j)\geq D_{ij}-e_{i,P}.
    \label{eq:reverse-triangle-positive}
\end{equation}
Taking the minimum of the two lower bounds gives
\begin{equation}
    \min\left\{ED(\widetilde{\mathcal Y}_i^A,\widehat{\mathcal Y}_j),ED(\widetilde{\mathcal Y}_i^P,\widehat{\mathcal Y}_j)\right\}\geq D_{ij}-\max\{e_{i,A},e_{i,P}\}.
    \label{eq:reverse-triangle-combined}
\end{equation}
Using Eq.~\ref{eq:target-negative-lower-bound},
\begin{equation}
\boxed{\min\left\{ED(\widetilde{\mathcal Y}_i^A,\widehat{\mathcal Y}_j),ED(\widetilde{\mathcal Y}_i^P,\widehat{\mathcal Y}_j)\right\}\geq L_{ij}q_{ij}-\max\{e_{i,A},e_{i,P}\}.}
\label{eq:negative-distance-lower-bound}
\end{equation}

To certify the desired local ordering using
Eq.~\ref{eq:negative-distance-lower-bound}, it is sufficient for its
right-hand side to exceed the realized positive-pair distance $d_i^{AP}$.
That is, we require
\begin{equation}
    L_{ij}q_{ij}-\max\{e_{i,A},e_{i,P}\}>d_i^{AP}.
\end{equation}
Equivalently,
\begin{equation}
\boxed{L_{ij}q_{ij}>d_i^{AP}+\max\{e_{i,A},e_{i,P}\}.}
\label{eq:realized-local-condition}
\end{equation}
Whenever Eq.~\ref{eq:realized-local-condition} holds,
Eq.~\ref{eq:negative-distance-lower-bound} gives
\begin{equation}
    \min\left\{ED(\widetilde{\mathcal Y}_i^A,\widehat{\mathcal Y}_j),ED(\widetilde{\mathcal Y}_i^P,\widehat{\mathcal Y}_j)\right\}>d_i^{AP}.
\end{equation}
Since
\begin{equation}
    d_i^{AP}=ED(\widetilde{\mathcal Y}_i^A,\widetilde{\mathcal Y}_i^P),
\end{equation}
we obtain
\begin{equation}
    ED(\widetilde{\mathcal Y}_i^A,\widetilde{\mathcal Y}_i^P)<\min\left\{ED(\widetilde{\mathcal Y}_i^A,\widehat{\mathcal Y}_j),ED(\widetilde{\mathcal Y}_i^P,\widehat{\mathcal Y}_j)\right\},
\end{equation}
which is exactly the desired local relational ordering in
Eq.~\ref{eq:local-relational-ordering}.

Equation~\ref{eq:realized-local-condition} is therefore not asserted to hold
for every comparison. Rather, it is a sufficient certificate: whenever it is
satisfied, the desired ordering follows from the calibrated edit lower bound
and the reverse-triangle argument.

\paragraph{A conservative sufficient condition.}
The realized certificate in Eq.~\ref{eq:realized-local-condition} uses the
observed positive-pair distance $d_i^{AP}$. Shared-target contraction provides
the upper bound
\begin{equation}
    d_i^{AP}\leq e_{i,A}+e_{i,P}
\end{equation}
from Eq.~\ref{eq:positive-triangle}. Hence,
\begin{equation}
    d_i^{AP}+\max\{e_{i,A},e_{i,P}\}\leq e_{i,A}+e_{i,P}+\max\{e_{i,A},e_{i,P}\}.
\end{equation}
It follows that the stronger condition
\begin{equation}
\boxed{L_{ij}q_{ij}>e_{i,A}+e_{i,P}+\max\{e_{i,A},e_{i,P}\}}
\label{eq:conservative-local-condition}
\end{equation}
implies Eq.~\ref{eq:realized-local-condition}, and therefore also certifies
Eq.~\ref{eq:local-relational-ordering}.

The condition in Eq.~\ref{eq:conservative-local-condition} is more
conservative because it replaces the realized distance $d_i^{AP}$ by its
possibly larger triangle-inequality upper bound $e_{i,A}+e_{i,P}$.



\subsection{Held-Out TIMIT Audit}
\label{sec:local-monotone-audit}

\paragraph{Protocol.}
We evaluate all $4{,}620$ TIMIT validation examples using the trained PairAlign
checkpoint. The validation dataloader uses seed $1234$ and the evaluation
script uses seed $13$, fixing the anchor--positive population.

We independently generate three EMA target banks, each containing one
anchor-derived target per example. This gives $13{,}860$ conditioning rows per
view in total. Batch size is $16$, yielding $15$ available mismatches per full
row. We report $k\in\{5,8,15\}$, with $k=15$ as the primary setting because it
uses the complete in-batch mismatch set. The NCE margin always uses all
available mismatches; $k$ affects only the post-hoc calibration neighborhood.

EMA targets use the Stage~III teacher sampler with maximum target length $60$,
length-cap ratio $0.15$, and minimum cap $2$. The exploration/final sampling
settings are top-$p$ $0.97/0.88$, temperature $1.30/1.00$, and repetition
penalty $1.03/1.30$.

Clean score matrices $S_{ij,V}$ are evaluated with prefix corruption and
structured self-attention dropout disabled, using $\tau_{\ell}=0.5$.
Equation~\ref{eq:local-row-nce-margin} is evaluated in float64 with shifted
log-sum-exp.

For the shared-target contraction audit, we restore the stochastic Stage~III
positive-fitting path: prefix masking $0.1$, structured self-attention dropout
$0.1$, and $16$ stochastic realizations per fixed EMA target. Within each
realization, the same corrupted canonical prefix is supplied to both
conditioning branches, matching the implemented shared-target training path.

Unless otherwise noted, uncertainty below denotes mean $\pm$ standard
deviation across the three independently generated EMA target banks for the
same trained checkpoint; it is therefore target-sampling variability, not
training-run variability.

\paragraph{Score-gap audit.}
Equation~\ref{eq:score-gap-hierarchy} has zero numerical violations. The
NCE-implied margin is strictly positive for
$99.827\pm0.043\%$ of anchor-conditioned rows and
$98.759\pm0.066\%$ of positive-conditioned rows.

The gap is also substantial in magnitude. Across banks, the median
NCE-implied and realized minimum row gaps are
\begin{equation}
\begin{aligned}
    \operatorname{median}
    [\Delta_{i,A}^{\mathrm{NCE}}]
    &=
    4.520\pm0.023,
    &
    \operatorname{median}[G_{i,A}]
    &=
    4.725\pm0.019,
    \\
    \operatorname{median}
    [\Delta_{i,P}^{\mathrm{NCE}}]
    &=
    3.689\pm0.026,
    &
    \operatorname{median}[G_{i,P}]
    &=
    3.892\pm0.035.
\end{aligned}
\label{eq:empirical-score-gaps}
\end{equation}
The mean difference
$G_{i,V}-\Delta_{i,V}^{\mathrm{NCE}}$ is
$0.197\pm0.002$ for anchor-conditioned rows and
$0.209\pm0.002$ for positive-conditioned rows. Thus, the objective-derived
margin is conservative but closely tracks the realized worst-case row
separation.

\paragraph{LOO calibration and joint geometry.}
Table~\ref{tab:local-monotone-joint-audit} reports bank-wise mean $\pm$
standard deviation. For the Spearman columns, each bank first contributes the
median row-wise Spearman correlation.

\begin{table}[t]
\centering
\scriptsize
\setlength{\tabcolsep}{3.2pt}
\begin{tabular}{c cc cc c c cc}
\toprule
&
\multicolumn{2}{c}{\textbf{Median Spearman $\rho$}}
&
\multicolumn{2}{c}{\textbf{LOO coverage (\%)}}
&
\textbf{Valid}
&
\textbf{Actual}
&
\multicolumn{2}{c}{\textbf{Realized certificate (\%)}}
\\
\cmidrule(lr){2-3}
\cmidrule(lr){4-5}
\cmidrule(lr){8-9}
$k$
& A & P
& A & P
& \textbf{A/P}
& \textbf{order}
& \textbf{Obs.}
& \textbf{NCE}
\\
\midrule
5
& $0.359\!\pm\!0.000$
& $0.362\!\pm\!0.006$
& $61.11\!\pm\!0.29$
& $61.33\!\pm\!0.40$
& $52.71\!\pm\!0.21$
& $99.765\!\pm\!0.017$
& $40.29\!\pm\!0.24$
& $37.99\!\pm\!0.33$
\\
8
& $0.426\!\pm\!0.006$
& $0.421\!\pm\!0.007$
& $67.77\!\pm\!0.20$
& $67.98\!\pm\!0.24$
& $62.97\!\pm\!0.29$
& $99.828\!\pm\!0.015$
& $49.40\!\pm\!0.33$
& $45.45\!\pm\!0.44$
\\
15
& $0.591\!\pm\!0.003$
& $0.589\!\pm\!0.003$
& $73.05\!\pm\!0.08$
& $73.13\!\pm\!0.05$
& $71.76\!\pm\!0.07$
& $99.898\!\pm\!0.009$
& $59.21\!\pm\!0.08$
& $52.92\!\pm\!0.20$
\\
\bottomrule
\end{tabular}
\caption{
Held-out TIMIT local-geometry audit across three independently generated EMA
target banks. Values are mean $\pm$ standard deviation across banks.
A/P denote anchor- and positive-conditioned score rows. LOO coverage is
computed only for non-extrapolated candidate-specific queries. ``Valid A/P''
is the fraction of all candidate comparisons for which at least one view
satisfies $\mathcal C_{ij,V}^{(k)}$. ``Actual order'' is the empirical
frequency of Eq.~\ref{eq:local-relational-ordering}. ``Obs.'' uses
$q_{ij}^{\mathrm{obs}}$, whereas ``NCE'' uses the more conservative
objective-derived $q_{ij}^{\mathrm{NCE}}$.
}
\label{tab:local-monotone-joint-audit}
\end{table}

The score gap exhibits a consistent positive local rank association with
normalized edit distance. At $k=15$, the median correlation is
$0.591\pm0.003$ for anchor conditioning and
$0.589\pm0.003$ for positive conditioning. More importantly, the relation is
tested out-of-fit: among supported held-out candidates, LOO coverage reaches
$73.05\pm0.08\%$ and $73.13\pm0.05\%$, respectively.

Candidate-specific support is $80.0\%$, $87.5\%$, and approximately $93.3\%$
for $k=5,8,15$. This is largely structural: if the excluded point has the
largest score gap in its neighborhood, no remaining calibration candidate has
an equal or larger gap. We therefore do not interpret support rate itself as
calibration evidence. The NCE query is supported in every evaluated case.

The desired ordering in Eq.~\ref{eq:local-relational-ordering} is substantially
more prevalent than the sufficient certificate. At $k=15$, it holds in
$99.898\pm0.009\%$ of comparisons, while at least one conditioning view
provides validated LOO calibration for $71.76\pm0.07\%$.

Under the realized shared-target contraction condition,
the candidate-specific score gap certifies
$59.21\pm0.08\%$ of all $k=15$ comparisons. Using only the weaker
NCE-implied gap still certifies $52.92\pm0.20\%$. Conditional on successful
local calibration, these rates are
$82.51\pm0.05\%$ and $73.74\pm0.21\%$, respectively.

Across the three banks and $221{,}760$ total stochastic positive-fitting
realizations,
\begin{equation}
\begin{aligned}
    \mathbb E[e_{i,A}]
    &=
    8.636\pm0.012,
    &
    \mathbb E[e_{i,P}]
    &=
    10.574\pm0.012,
    \\
    \mathbb E[d_i^{AP}]
    &=
    8.478\pm0.001.
\end{aligned}
\end{equation}
The mean conservative threshold in
Eq.~\ref{eq:conservative-local-condition} is
$30.219\pm0.028$, compared with
$19.487\pm0.011$ for the realized threshold in
Eq.~\ref{eq:realized-local-condition}. No violation of
Eq.~\ref{eq:positive-triangle} is observed.

As implementation consistency checks, we additionally observe zero violations
of the NCE score-gap inequality, zero violations of
\begin{equation}
    q_{ij,V}^{\mathrm{NCE}}
    \leq
    q_{ij,V}^{\mathrm{obs}}
    \leq
    d_{ij}
\end{equation}
among validated candidates, and zero violations of the relational ordering
among positively certified comparisons.


\paragraph{Interpretation and limitations.}
The analysis isolates the roles of the two Stage~III mechanisms.
Shared-target fitting does not directly minimize positive-pair edit distance;
rather, both conditioning paths are fitted to one anchor-derived canonical EMA
sequence, yielding the contraction relation in
Eq.~\ref{eq:positive-triangle}. Likelihood contrast independently provides
score separation through
\begin{equation}
    \delta_{ij,V}
    \geq
    G_{i,V}
    \geq
    \Delta_{i,V}^{\mathrm{NCE}},
\end{equation}
and the held-out monotone audit tests when that separation transfers to edit
space. When the validated edit lower bound exceeds the realized
shared-target contraction radius, the desired local relational ordering
follows.

The certificate rate is a sufficient-condition rate, not a failure rate for
uncertified comparisons. A comparison may remain uncertified because its LOO
query is unsupported or because the empirical lower envelope is conservative
even when Eq.~\ref{eq:local-relational-ordering} holds directly.

Finally, the result is local to the finite candidate neighborhoods considered
by Stage~III and is not a global likelihood--Levenshtein embedding theorem.
The analysis also concerns training-time student quantities:
EMA-generated candidate targets, clean-prefix student likelihoods, and
teacher-forced student predictions. Final PairAlign tokenizations are produced
independently by the EMA teacher under free-running beam search; their
paired-view and negative-pair edit geometry is therefore evaluated directly
elsewhere in the paper.

\end{document}